\documentclass[twoside,11pt]{article}

\usepackage{blindtext}

\usepackage[preprint]{jmlr2e}

\usepackage{setspace}
\usepackage[utf8]{inputenc} 
\usepackage[T1]{fontenc}    
\usepackage{hyperref}       
\usepackage{url}            
\usepackage{booktabs}       
\usepackage{amsfonts}       
\usepackage{nicefrac}       
\usepackage{microtype}      
\usepackage[pdftex,dvipsnames,table]{xcolor}         
\usepackage{caption} 
\usepackage{subcaption} 

\usepackage[
  color=blue!20!white,
  textsize=tiny,
  colorinlistoftodos,
  prependcaption,
]{todonotes}

\usepackage{marginnote}
\newcommand{\cmark}{\textcolor{Green}{\ding{51}}}
\newcommand{\xmark}{\textcolor{BrickRed}{\ding{55}}}
\usepackage{cite}
\usepackage{amsmath,amssymb,amsfonts}
\usepackage{graphicx}
\usepackage{textcomp}

\def\BibTeX{{\rm B\kern-.05em{\sc i\kern-.025em b}\kern-.08em
    T\kern-.1667em\lower.7ex\hbox{E}\kern-.125emX}}

\usepackage{dsfont,empheq}
\usepackage{mathrsfs}
\usepackage{algorithm}
\usepackage{algpseudocode}
\usepackage{algorithmicx}
\usepackage[textsize=tiny]{todonotes}
\usepackage{mathtools}
\usepackage{syntonly,bm,wrapfig}
\usepackage{cases,balance}

\usepackage{makecell}

\usepackage{pifont}
\newcommand{\uparrowcolor}[1]{%
  \begin{tikzpicture}
    \draw[->, line width=1.5pt, #1] (0,0) -- (0,0.3);
  \end{tikzpicture}%
}

\newcommand{\downarrowcolor}[1]{%
  \begin{tikzpicture}
    \draw[->, line width=1.5pt, #1] (0,0) -- (0,-0.3);
  \end{tikzpicture}%
}

\newcommand{\hfdownarrowcolor}[1]{%
  \begin{tikzpicture}
    \draw[->, line width=1.5pt, #1] (0,0) -- (0.2,-0.3);
  \end{tikzpicture}%
}

\usepackage{romannum}
\AtBeginDocument{\pagenumbering{arabic}}

\newcommand\numberthis{\addtocounter{equation}{1}\tag{\theequation}}

\newcommand{\E}{\mathbb{E}}
\newcommand{\lambdain}{\lambda \in \Delta^{M}}
\newcommand{\xin}{x \in \mathbb{R}^d}
\newcommand{\tdz}{\tilde{z}}

\newcommand{\red}[1]{{\color{Red} #1}}

\newcommand{\green}[1]{{\color{Green} #1}}
\newcommand{\gray}[1]{{\color{gray} #1}}
\newcommand{\yellow}[1]{{\color{Goldenrod} #1}}

\newcommand{\info}[1]{%
 \let\marginpar\marginnote
 \reversemarginpar
 \renewcommand{\baselinestretch}{0.8}
\todo[linecolor=OliveGreen!25!White,backgroundcolor=OliveGreen!25,bordercolor=OliveGreen,]{\textcolor{OliveGreen}{Info:} #1}}

\usepackage[most]{tcolorbox}
\usepackage{pgf}

\newcommand{\emphblockoption}{drop shadow,
    colframe=black!60,
    colback=RoyalBlue!10,
    coltitle=white!, 
    left=.0pt,
    right=.0pt,
    boxrule=0pt,
    arc=0pt}
\tcbset{emphblock/.style={code={\pgfkeysalsofrom{\emphblockoption}}}}

\usepackage{thmtools,thm-restate}

\newtheorem{thm}{Theorem}[section]
\newtheorem{prop}{Proposition}[section]
\newtheorem{lmm}{Lemma}[section]

\newtheorem{dfn}{Definition}[section]

\newtheorem{example}{Example}
\newtheorem{remark}{Remark}

\newtheorem{assumption}{Assumption}

\usepackage{float}
\usepackage{placeins}
\usepackage{multirow}
\usepackage{siunitx}

\usepackage[toc,page,header]{appendix}
\usepackage{minitoc}

\numberwithin{equation}{section}

\newcommand\blfootnote[1]{%
     \begingroup
     \renewcommand\thefootnote{}\footnote{#1}%
     \addtocounter{footnote}{-1}%
      \endgroup
    }

\usepackage{lastpage}
\jmlrheading{23}{2023}{1-\pageref{LastPage}}{1/21; Revised 5/22}{9/22}{21-0000}{Lisha Chen and Heshan Fernando and Yiming Ying and Tianyi Chen}

\ShortHeadings{Three-Way Trade-Off in Multi-Objective Learning}{Optimization, Generalization and Conflict-Avoidance}
\firstpageno{1}

\mtcsetdepth{secttoc}{3}

\begin{document}
\doparttoc 
\faketableofcontents 


\title{Three-Way Trade-Off in Multi-Objective Learning: Optimization, Generalization and Conflict-Avoidance}

\author{\name Lisha Chen$^{\dagger}$ \email chenl21@rpi.edu \\
\addr Department of Electrical, Computer \& Systems Engineering\\
Rensselaer Polytechnic Institute, United States
\AND
\name Heshan Fernando$^{\dagger}$ \email fernah@rpi.edu \\
\addr Department of Electrical, Computer \&  Systems Engineering\\
Rensselaer Polytechnic Institute, United States
\AND
\name Yiming Ying \email yying@albany.edu \\
\addr Department of Mathematics and Statistics\\
University at Albany, State University of New York, United States
\AND
\name Tianyi Chen \email chentianyi19@gmail.com \\
\addr Department of Electrical, Computer \&  Systems Engineering\\
Rensselaer Polytechnic Institute, United States
}
\blfootnote{$^{\dagger}$ Equal contribution.}
\editor{My editor}

\maketitle

\allowdisplaybreaks

\begin{abstract}
Multi-objective learning (MOL) problems often arise in emerging machine learning problems when there are multiple learning criteria, data modalities, or learning tasks. Different from single-objective learning, one of the critical challenges in MOL is the potential conflict among different objectives during the iterative optimization process. Recent works have developed various \emph{dynamic weighting} algorithms for MOL such as MGDA and its variants, where the central idea is to find an update direction that \emph{avoids conflicts} among objectives. Albeit its appealing intuition,  empirical studies show that dynamic weighting methods may not always outperform static ones. To understand this theory-practical gap, we focus on a new stochastic variant of  MGDA - the \textsf{M}ulti-\textsf{o}bjective gradient with \textsf{Do}uble sampling (\textsf{MoDo}) algorithm, and study the generalization performance of the dynamic weighting-based MoDo and its interplay with optimization through the lens of algorithm stability. 
Perhaps surprisingly, we find that the key rationale behind MGDA -- updating along conflict-avoidant direction - may \emph{hinder} dynamic weighting algorithms from achieving the optimal ${\cal O}(1/\sqrt{n})$ population risk, where $n$ is the number of training samples.
We further demonstrate the impact of the variability of dynamic weights on the three-way trade-off among optimization, generalization, and conflict avoidance that is unique in MOL.
We showcase the generality of our theoretical framework by analyzing other existing stochastic MOL algorithms under the framework.
Experiments on various multi-task learning benchmarks are performed to demonstrate the practical applicability. Code is available at \url{https://github.com/heshandevaka/Trade-Off-MOL}.
\footnote{
Preliminary results in this paper were presented in part at the 2023 Advances in Neural Information Processing Systems~\citep{chen2023threeway_moo}.}

\end{abstract}

\vspace{0.2cm}
\begin{keywords}
  Multi-objective optimization, statistical learning theory, algorithm stability, Pareto stationarity, gradient conflict
\end{keywords}

\section{Introduction}

Multi-objective learning (MOL) emerges frequently as a new unified learning paradigm from recent machine learning problems such as 
learning under fairness and safety constraints \citep{zafar2017fairness}; learning across multiple tasks including multi-task learning \citep{sener2018multi} and meta-learning \citep{ye2021MOML}; and, learning across multiple agents that may not share a global utility including federated learning~\citep{federeated_MTL_smith2017} and multi-agent reinforcement learning~\citep{vanmoffaert14a_MORL}. 

This work considers solving the empirical version of MOL  defined on the training dataset as $S=\{z_1, \ldots, z_n\}$. 
The performance of a model   $x \in \mathbb{R}^d$ on a datum  $z$ for the $m$-th objective is denoted as $f_{z,m}:\mathbb{R}^d\mapsto\mathbb{R}$, and its performance on the entire training dataset $S$ is measured by the $m$-th empirical objective $f_{S, m}(x)$ for $m \in [M]$. MOL optimizes the vector-valued objective, given by
\begin{equation}\label{eq.opt1}
  \min\limits_{x \in \mathbb{R}^d}  ~~F_S(x)\coloneqq   [f_{S,1}(x), \dots, f_{S,M}(x)].
\end{equation}
One natural method for solving \eqref{eq.opt1} is to optimize the (weighted) average of the multiple objectives, also known as {\em static or unitary weighting}~\citep{kurin2022defense,xin2022current}. 
However, this method may face challenges due to \emph{potential conflicts} among multiple objectives during the optimization process; e.g., conflicting gradient directions $\langle \nabla f_{S,m}(x), \nabla f_{S,m'}(x)\rangle<0$, if choosing the gradient-based optimizer. 
A popular alternative is thus to {\em dynamically weight} gradients from different objectives to avoid conflicts and   obtain a direction $d(x)$ that   optimizes all objective functions jointly that we call a \emph{conflict-avoidant} (CA) direction. 
Algorithms in this category include the multi-gradient descent algorithm (MGDA)~\citep{Desideri2012mgda}, its stochastic variants~\citep{liu2021stochastic,fernando2022mitigating,zhou2022_SMOO}.  
While the idea of finding CA direction in dynamic weighting-based approaches is very appealing, recent empirical studies reveal that dynamic weighting methods may not outperform static weighting in some MOL benchmarks~\citep{kurin2022defense,xin2022current}, especially when it involves stochastic updates and deep  models. 
Specifically, observed by~\citep{kurin2022defense}, the vanilla stochastic MGDA can be under-optimized, leading to larger optimization error than static weighting. The reason behind this optimization performance degradation has been studied in~\citep{zhou2022_SMOO,fernando2022mitigating}, which suggest the vanilla stochastic MGDA has biased update, and propose momentum-based methods to address this issue.
Nevertheless, in~\citep{xin2022current}, it has been demonstrated that the training errors of MGDA and static weighting are similar, while their main difference lies in the \emph{generalization performance}.
Unfortunately, the reason behind this testing performance degradation is not fully understood and remains an open question.

To gain a deeper understanding of the dynamic weighting methods, a natural question is
\begin{center}
  \textit{{\bf Q1:} What are the major sources of errors in dynamic weighting-based MOL methods?}  
\end{center}

To answer this question theoretically, we first introduce a proper measure of testing performance in MOL -- the \emph{Pareto stationary measure} in terms of the population objectives, which will immediately imply stronger measures such as  Pareto optimality under strongly convex objectives. We then decompose this measure into \emph{generalization} error and \emph{optimization} error and further introduce a new metric termed \emph{CA distance} that reflects the algorithm's ability to update along CA direction and is unique to MOL; see Sections \ref{sec.prel-mol} and \ref{sub:measure_decompose}. 

To characterize the performance of MOL methods in a unified manner, we introduce a generic dynamic weighting-based MOL method that we term stochastic Multi-Objective gradient with DOuble sampling algorithm (\textbf{MoDo}), which uses a step size $\gamma$ to control the change of dynamic weights. Roughly speaking, by controlling $\gamma$, MoDo includes MGDA (large $\gamma$) and static weighting algorithm ($\gamma=0$) as two special cases; see Section \ref{sec.modo}. We first analyze the generalization error of the model learned by MoDo through the lens of algorithmic stability \citep{bousquet2002stability,hardt2016train,lei2020fine} in the framework of statistical learning theory.  To our best knowledge, this is the \emph{first-ever-known}   stability analysis for  MOL algorithms. Here the key contributions lie in defining a new notion of stability - MOL uniform stability and then establishing a tight upper bound (matching lower bound) on the MOL uniform stability for the MoDo algorithm that involves two coupled sequences; see Section \ref{sub:MOL_gen_err}. 
Note that the only other existing stability analysis for two coupled sequences is for minmax problems, where the two sequences are optimizing the opposite objective functions w.r.t. two variables, and the variables are stacked into one to derive similar properties as single objective learning. In contrast, our analysis for two sequences with different objectives is of self-interest and generalizes to other settings such as bilevel and compositional optimization. 
We then analyze the optimization error of MoDo and its distance to CA directions, where the key contributions lie in relaxing \emph{the bounded function value/gradient assumptions} and significantly improving the convergence rate of state-of-the-art dynamic weighting-based  method~\citep{fernando2022mitigating}; see Section \ref{sub:MOL_opt_err}. 

Different from the stability analysis for single-objective learning~\citep{hardt2016train}, the techniques used in our generalization and optimization analysis allow to remove conflicting assumptions such as bounded gradient and bounded function value assumptions in the strongly convex and unconstrained setting, as well as use step sizes larger than $\mathcal{O}(1/t)$ in \citep{hardt2016train} to ensure both small generalization and optimization errors, which are of independent interest.

\begin{figure}[t]
\hspace{-0.15cm}
 \begin{subfigure}[b]{0.215\textwidth}
 \centering
 \includegraphics[width=0.95\textwidth]{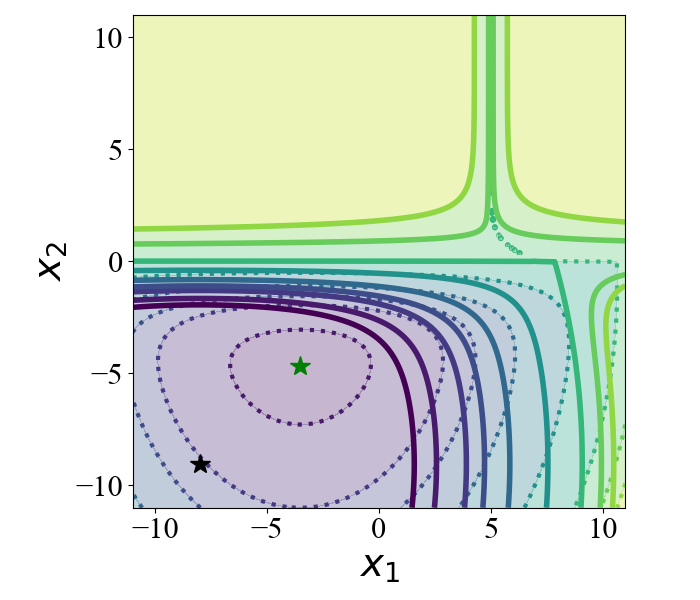}
 \caption{Objective 1}
 \label{fig:toy-task-1}
\end{subfigure}
  \hspace{-0.15cm}
\begin{subfigure}[b]{0.187\textwidth}
 \centering
 \includegraphics[width=0.95\textwidth]{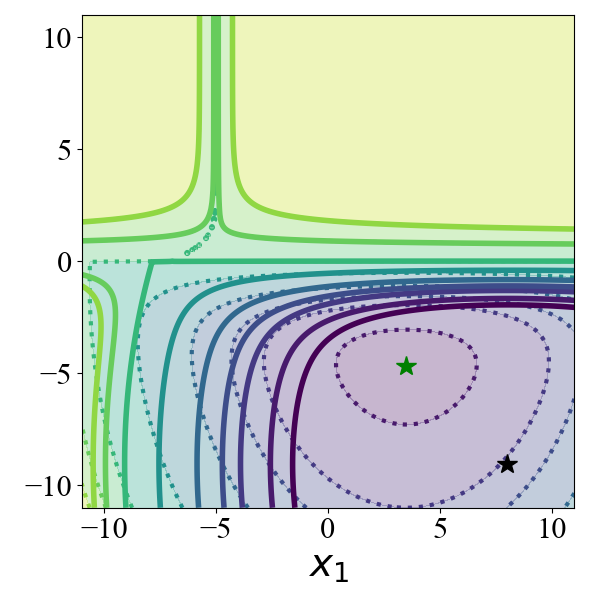}
 \caption{Objective 2}
 \label{fig:toy-task-2}
\end{subfigure} 
  \hspace{-0.15cm}
\begin{subfigure}[b]{0.187\textwidth}
 \centering
 \includegraphics[width=0.95\textwidth]{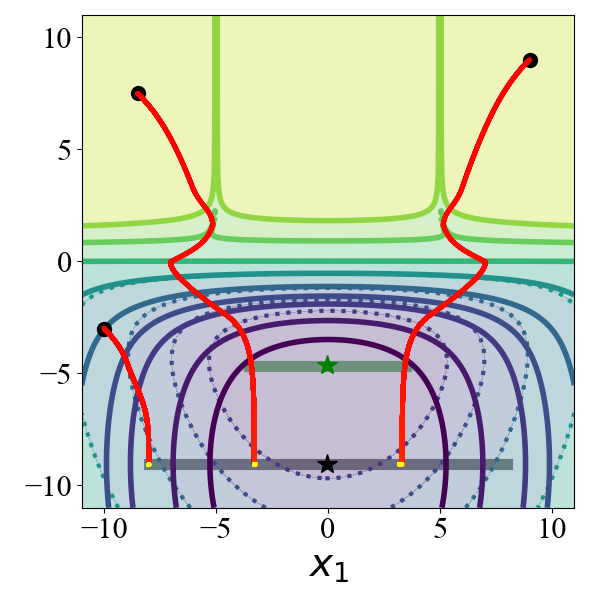}
 \caption{MGDA}
 \label{fig:toy-mgda}
\end{subfigure}
  \hspace{-0.15cm}
\begin{subfigure}[b]{0.187\textwidth}
 \centering
 \includegraphics[width=0.95\textwidth]{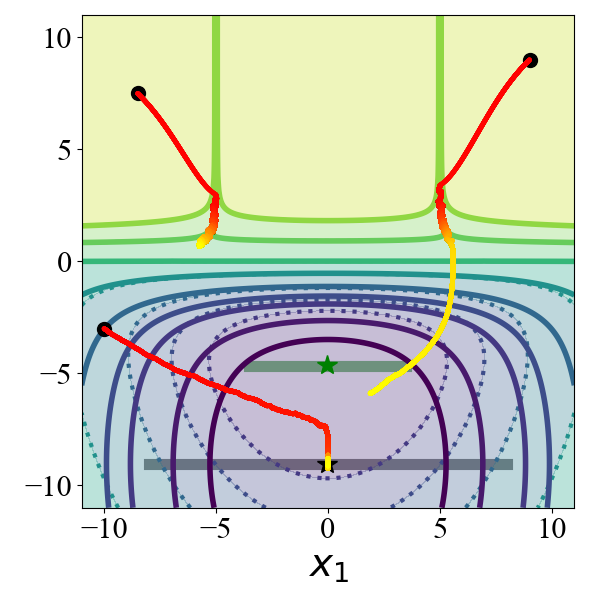}
 \caption{Static}
 \label{fig:toy-sgd}
\end{subfigure}
  \hspace{-0.15cm}
\begin{subfigure}[b]{0.221\textwidth}
 \centering
 \includegraphics[width=0.95\textwidth]{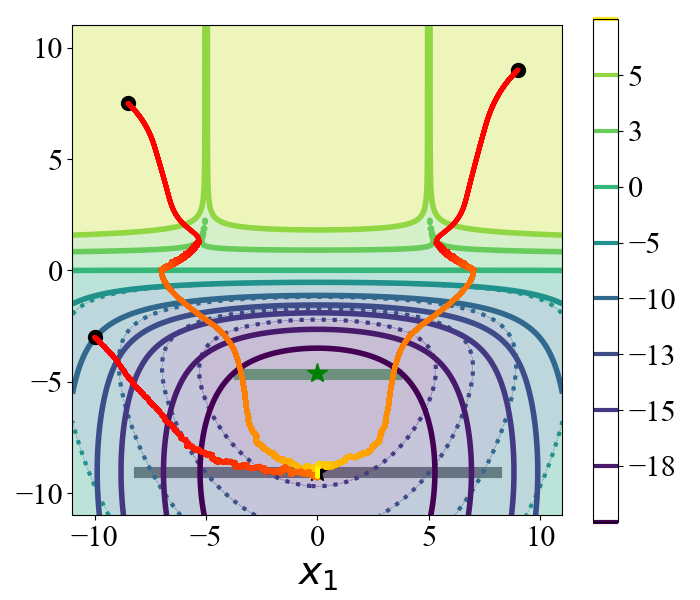}
 \caption{MoDo (dynamic)}
 \label{fig:toy-modo}
\end{subfigure}
\caption{An example from~\citep{liu2021conflict} with two objectives (\ref{fig:toy-task-1} and \ref{fig:toy-task-2}) to show the three-way trade-off in MOL. 
Figures~\ref{fig:toy-mgda}-\ref{fig:toy-modo} show the optimization trajectories,
where the {\bf black} $\bullet$ marks initializations of the trajectories, colored from \red{\bf red} (start) to \yellow{\bf yellow} (end).
The background solid/dotted contours display the landscape of the average empirical/population objectives. 
The \gray{\bf gray}/\green{\bf green} bar marks empirical/population Pareto front, and the {\bf black $\star$}/\green{\bf green $\star$}  marks solution to the  average  objectives. 
}
\label{fig:toy-comp}
\end{figure}

Given the test performance degradation of dynamic weighting methods in MOL and the holistic analysis of dynamic weighting methods provided in {\bf Q1}, a  follow-up question is
\begin{center}
  \textit{{\bf Q2:} What may cause the empirical performance degradation of dynamic weighting methods?}  
\end{center}

\emph{Visualizing MOL solution concepts.} 
To reason the root cause for this, we first compare different MOL algorithms in a toy example shown in Figure~\ref{fig:toy-comp}. We find MGDA can navigate along CA directions and converge to the empirical Pareto front under all initializations, while static weighting gets stuck in some initializations; at the same time, the empirical Pareto solution obtained by MGDA may incur a larger population risk than the suboptimal empirical solution obtained by the static weighting method; finally, if the step size $\gamma$ of dynamic weights is carefully tuned, MoDo can converge along CA directions to the empirical Pareto optimal solution that also generalizes well. 

\begin{figure}[t]
\centering
\includegraphics[width=0.3\textwidth]{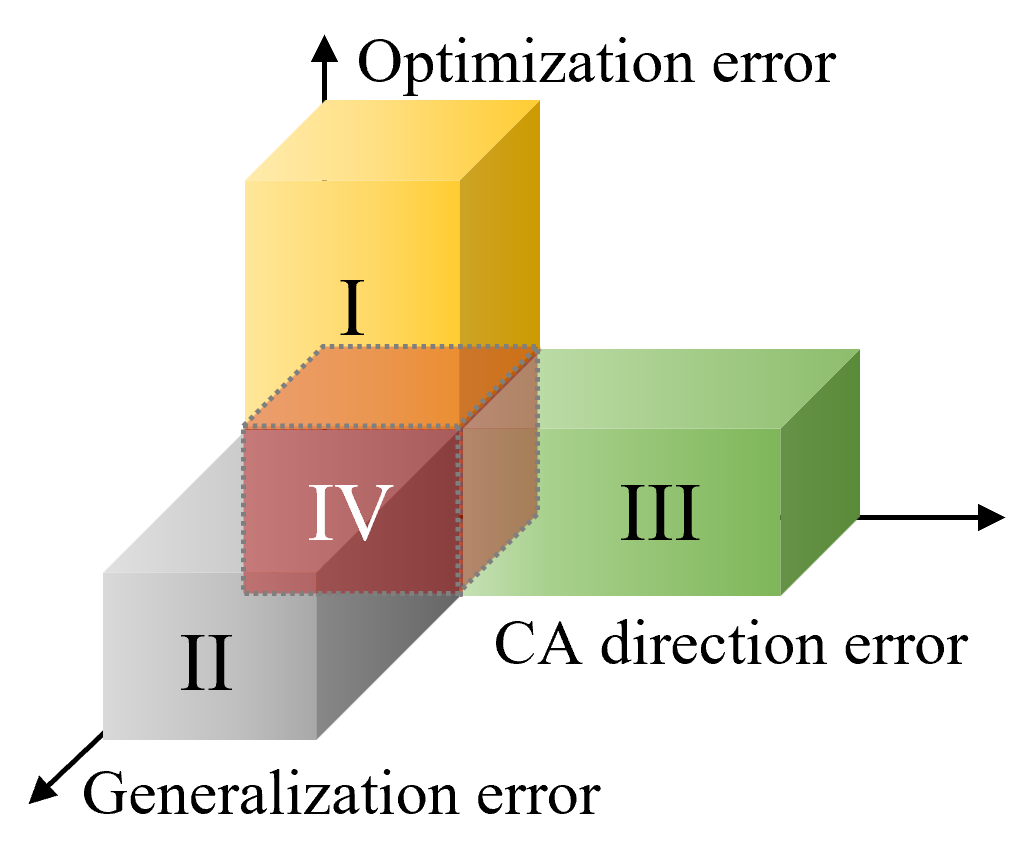}
\hspace{-0.2cm}
\small
\begin{tabular}[b]{cccccc}
    \hline\hline
  \multirow{2}{*}{Region}\!\! & \!\!\multirow{2}{*}{Opt.}\!\! & \!\!\multirow{2}{*}{Gen.}\!\! & \!\!\multirow{2}{*}{Conflict} \!\!
  & \makecell{Step sizes \\ $\alpha$ \& $\gamma$} 
  & \makecell{Iterations $T$ \\ \& data size   $n$}   \\ 
  \midrule
  \Romannum{1} & \uparrowcolor{BrickRed}  & \downarrowcolor{Green} & \downarrowcolor{Green} 
  & $\gamma = \omega( \alpha )$ 
  & $T = o(n)$  \\
  \Romannum{2} & \downarrowcolor{Green} & \uparrowcolor{BrickRed} & \downarrowcolor{Green} 
  & $\gamma = \omega( \alpha )$
  & $T = \omega(n)$ \\
  \Romannum{3} & \downarrowcolor{Green} & \downarrowcolor{Green} & \uparrowcolor{BrickRed} 
  & $\gamma = o( T^{-1} )$ 
  & $T = \omega(n)$  \\
    \Romannum{4} & \hfdownarrowcolor{Green} & \hfdownarrowcolor{Green} & \hfdownarrowcolor{Green}
  &\!\!\! \!\!\! {\small $\gamma =\Theta(\alpha^{\frac{1}{2}})= \Theta(T^{-\frac{1}{4}})$}\!\!\! \!\!\! 
  & {\small$T = \Theta(n^{\frac{4}{5}})$}  \\
    \hline\hline
    \vspace{-0.1cm}
\end{tabular}
\caption{An illustration of three-way trade-off among optimization, generalization, and conflict avoidance in the strongly convex case; $\alpha$ is the step size for $x$, $\gamma$ is the step size for weights $\lambda$, where $o(\cdot)$ denotes a strictly slower growth rate, $\omega(\cdot)$ denotes a strictly faster growth rate, and $\Theta(\cdot)$ denotes the same growth rate. Arrows {\color{Green} $\mathbf\downarrow$} and {\color{BrickRed} $\mathbf\uparrow$} respectively represent diminishing in an optimal rate and growing in a fast rate w.r.t. $n$, while {\color{Green} $\mathbf\searrow$} represents diminishing w.r.t. $n$, but not in an optimal rate.}
\label{fig:tradeoff_3way}
\end{figure}

Aligned with this toy example, our theoretical results suggest a novel \emph{three-way trade-off} in the performance of dynamic weighting-based MOL algorithm; see Section \ref{sub:opt_gen_ca_tradeoff}. Specifically, it suggests that the step size for dynamic weighting $\gamma$ plays a central role in the trade-off among convergence to the CA direction, convergence to empirical Pareto stationarity, and generalization error; see Figure \ref{fig:tradeoff_3way}. In this sense, MGDA has an edge in convergence to the CA direction but it could sacrifice generalization; the static weighting method cannot converge to the CA direction but guarantees convergence to empirical Pareto solutions and their generalization. 
Our analysis also suggests that MoDo achieves a small population risk under a proper combination of step sizes and the number of iterations.

\vspace{0.2cm}
\noindent\textbf{Technical challenges.} 
Our technical contributions are highly non-trivial, and the technical challenges are summarized below. 

\begin{enumerate}
  \item [C1)] 
  The definition of PS testing risk~\eqref{eq:excess_risk_decompose} is unique in MOL, and the introduction of sampling-determined algorithms overcomes a key challenge brought by the classical function value-based risk measures -- the unnecessarily small step size choice.
Specifically, prior stability analysis in function values for single objective learning~\citep{hardt2016train} requires $1/t$ step size decay in the nonconvex case, otherwise, the generalization error bound will depend exponentially on the number of iterations. However, such step size choice leads to a very slow convergence of the optimization error.
This is addressed by the definitions of gradient-based measures and sampling-determined MOL algorithms, which yield stability bound in $\mathcal{O}(T/n)$
without any step size decay. See more discussions below Theorem~\ref{crlr:bound_stab_grad_MGDA}. 
\item [C2)] The stability of the dynamic weighting algorithm in the strongly convex (SC) case is non-trivial compared to single objective learning~\citep{hardt2016train} because it involves two coupled sequences during the update. As a result, the classical contraction property for the update of model parameters that is often used to derive stability does not hold. This is addressed by controlling the change of $\lambda_t$ by the step size $\gamma$, and using mathematical induction to derive a tighter bound. See Appendix~\ref{sub_app:proof_gen_sc_app}. 
\item [C3)] In the SC case with an unbounded domain, the function is not Lipschitz or the gradients are generally unbounded, which violates the commonly used bounded gradient assumption for proving the stability bound and optimization convergence. We relax this assumption by proving that the iterates generated by dynamic weighting algorithms in the SC case are bounded on the trajectory in Lemma~\ref{lemma:x_t_bounded_sc_smooth}. 
\end{enumerate}

\section{Problem Formulation and Target of Analysis}
In this section, we first introduce the problem formulation of MOL, the target of analysis, the metric to measure its generalization, and then present the    MGDA algorithm  and its stochastic variant.

\subsection{Preliminaries of MOL}\label{sec.prel-mol}

Denote the vector-valued objective
function on datum $z$ as $F_z(x) = [f_{z,1}(x), \dots, f_{z,M}(x)]$.
The training and testing performance  of $x$  can then be measured by the empirical objective $F_S(x)$ and the population objective $F(x)$ which are, respectively, defined as $F_S(x)\coloneqq \frac{1}{n} \sum_{i=1}^n F_{z_i}(x)$ and $F(x)\coloneqq \mathbb{E}_{z\sim \cal D}[F_z(x)]$.
Their gradients are denoted as $\nabla F_S(x)$ and $\nabla F(x) \in \mathbb{R}^{d\times M}$.

Analogous to the stationary and optimal solutions in single-objective learning,
we define Pareto stationary and Pareto optimal solutions for MOL problem $\min_{x \in \mathbb{R}^d}F(x)$ as follows.

\begin{dfn}[Pareto stationary and Pareto optimal solutions]
If there exists a convex combination of the gradient vectors that equals to zero, i.e., there exists $\lambdain$ such that $\nabla F(x) \lambda = 0$, then $x \in \mathbb{R}^d$ is  Pareto stationary. 
If there is no  $x\in\mathbb{R}^d$ and $x \neq x^*$ such that, for all $m\in[M]$ $f_{m}(x) \leq  f_{m}(x^*)$, with $f_{m'}(x) <  f_{m'}(x^*)$ for at least one $m'\in [M]$, then $x^*$ is Pareto optimal.
If there is no  $x\in\mathbb{R}^d$ such that for all $m\in[M]$, $f_{m}(x) < f_{m}(x^*)$, then $x^*$ is weakly Pareto optimal. 
\end{dfn}

By definition, at a Pareto stationary solution, there is no common descent direction for all objectives. A necessary and sufficient condition for $x$ being  Pareto stationary for smooth objectives is that $\min_{\lambdain}\|\nabla F(x) \lambda\| = 0$~\citep{tanabe2018proximal}.
Therefore, $\min_{\lambdain}\|\nabla F(x) \lambda\|$ can be used as a measure of Pareto stationarity (PS)~\citep{Desideri2012mgda,fliege2019complexity,tanabe2018proximal,liu2021stochastic,fernando2022mitigating}. We will refer to the aforementioned quantity as the \emph{PS population risk} henceforth and its empirical version as \emph{PS empirical risk} or \emph{PS optimization error}.
We next introduce the target of our analysis based on the above definitions.

\subsection{Target of analysis and error decomposition}\label{sub:measure_decompose}
In the existing generalization analysis for MOL,   measures based on function values have been used to derive generalization guarantees derived from Pareto optimality \citep{cortes2020agnostic, sukenik2022generalization}. However, for general nonconvex smooth MOL problems,  it can only be guaranteed for an algorithm to converge to Pareto stationarity of the empirical objective, i.e.,  a small $\min_{\lambdain}\|\nabla F_S(x) \lambda\|$ ~\citep{Desideri2012mgda,fliege2019complexity}. Thus, it is not reasonable to measure population risk in terms of Pareto optimality in this case.
Furthermore, when all the objectives are convex or strongly convex, Pareto stationarity is a sufficient condition for weak Pareto optimality or Pareto optimality, respectively, as stated in Proposition~\ref{prop:PS_imply_PO}.

\begin{prop}{\citep[Lemma~2.2]{tanabe2018proximal}}
\label{prop:PS_imply_PO}
If $f_{m}(x)$ are convex or strongly-convex for all $m\in [M]$,
and $x \in \mathbb{R}^d$ is a Pareto stationary point of $F(x)$, then $x$ is weakly Pareto optimal or Pareto optimal. 
\end{prop}

Next we proceed to decompose the PS population risk in~\eqref{eq:excess_risk_decompose}.

\noindent\textbf{Error Decomposition.}
Given a model ${x}$, the PS population risk  can be decomposed into
\begin{equation}
\underbrace{\min _{{\lambda}\in \Delta^{M}}\|\nabla F({x}) \lambda\| }_{\text {PS population risk }R_{\rm pop}(x)} 
= \underbrace{\min_{\lambda\in \Delta^{M}} \|\nabla F({x}) \lambda\|-\min _{\lambda\in \Delta^{M}} \|\nabla F_S({x}) \lambda \|}_{\text{PS generalization error }R_{\rm gen}(x) }+\underbrace{\min _{\lambda\in \Delta^{M}} \|\nabla F_S({x}) \lambda \|
 }_{\text {PS optimization error }R_{\rm opt}(x)}  
\label{eq:excess_risk_decompose}
\end{equation}
where the optimization error quantifies the training performance, i.e., how well does model $x$ perform on the training data; and
the generalization error (gap) quantifies the difference of the testing performance on new data sampled from $\cal D$ and the training performance, i.e., how well does the model $x$ perform on unseen testing data compared to the training data. 

Let $A: \mathcal{Z}^n \mapsto \mathbb{R}^d$ denote a randomized MOL algorithm. Given training data $S$, we are interested in the expected  performance of the output model $x = A(S)$, which is measured by $\E_{A,S} [R_{\rm pop}(A(S))]$.
From~\eqref{eq:excess_risk_decompose} and linearity of expectation, it holds that
\begin{align}\label{eq:pop_risk_decompose_AS}
 \E_{A,S} [R_{\rm pop}(A(S))] = \E_{A,S} [R_{\rm gen}(A(S))] + \E_{A,S} [R_{\rm opt}(A(S))].   
\end{align}

\noindent\textbf{Distance to CA direction.}
As demonstrated in Figure~\ref{fig:toy-comp}, the key merit of dynamic weighting    over static weighting algorithms lies in its ability to navigate through conflicting gradients. 
Consider an update direction $d = -\nabla F_S(x)\lambda$, where $\lambda$ is the dynamic weights from a simplex $\lambdain \coloneqq \{\lambda \in \mathbb{R}^M~|~ \mathbf{1}^\top \lambda = 1,~\lambda \geq 0\}$.
To obtain such a steepest CA direction in unconstrained learning, we can reformulate the problem at each iteration~\citep{fliege2019complexity},
with the goal of maximizing the minimum descent (among all objectives) along the update direction $d$, where the minimum descent given direction $d$ and step size $\alpha$ can be computed by
\begin{align*}
  \frac{1}{\alpha}\min_{m\in [M]} {f_{S,m}(x) - f_{S,m}(x+\alpha d)} \approx \min_{m\in [M]} -\langle \nabla f_{S,m}(x), d\rangle
   = \min_{\lambda\in \Delta^M} -\langle \nabla F_S(x)\lambda, d\rangle.
\end{align*}
Since the solutions to $\lambda$ and $d$ may not necessarily be singletons, we further explicitly regularize the $\ell$-2 norm of $\lambda$ and $d$ so as to put more emphasis on all the objectives instead of focusing on the worst one, and to ensure $d$ does not go to infinity.
With the above measurement, the algorithm aims to find an update direction $d$ that maximizes the following
\begin{align}
  \max_{d\in \mathbb{R}^d} \min_{\lambda\in \Delta^M} -\langle \nabla F_S(x)\lambda, d\rangle + \frac{\rho}{2}  \|\lambda\|^2 - \frac{1}{2}\|d\|^2 
\end{align}
where $\rho \geq 0$ is a regularization constant.
By the min-max theorem, it can be further reformulated as
\begin{align}
  \max_{\lambda\in \Delta^M} \min_{d\in \mathbb{R}^d}  \langle \nabla F_S(x)\lambda, d\rangle - \frac{\rho}{2} \|\lambda\|^2 + \frac{1}{2}\|d\|^2
\end{align}
where the optimal solution is $d = -\nabla F_S(x)\lambda_\rho^*(x)$, with $\lambda_\rho^*(x) \in \arg\min_{\lambda\in \Delta^M} \frac{1}{2}\|\nabla F_S(x)\lambda\|^2 + \frac{\rho}{2} \|\lambda\|^2 $.
Then the CA direction is calculated as
\begin{align}
\label{eq:ca_direction_problem}
{\rm CA~direction}~~~~~ 
 d(x) = -\nabla F_S(x)\lambda^*_\rho(x)~~\mathrm{s.t.}
 ~~
 \lambda^*_\rho(x) \in \mathop{\arg\min}_{\lambdain} \|\nabla F_S(x)\lambda\|^2  + \rho \|\lambda\|^2 .
\end{align}
The regularized MGDA adopts $d(x)$ as the update direction at each iteration, as summarized in Algorithm~\ref{alg:MGDA}.
Let $d_\lambda(x) = - \nabla F_Z(x) \lambda $ denote the stochastic update direction with random mini-batch data $Z$, and $x\in \mathbb{R}^d$, $\lambdain$ generated by the stochastic algorithm $A$. We measure the so-termed CA distance via
\begin{align}
{\rm CA~direction~distance}\qquad \mathcal{E}_{\rm ca}(x, \lambda) \coloneqq &
 \|\E_A[d_\lambda(x) - d(x)]\| ^2,
 \\
{\rm CA~weight~distance}\qquad \mathcal{E}_{\rm caw}(x, \lambda) \coloneqq &
 \|\E_A[\lambda - \lambda^*_\rho (x)]\| ^2.  
\end{align}

With the above definitions of measures that quantify the performance in different aspects, we then introduce a stochastic gradient algorithm for MOL studied in this work.

\begin{figure}[t]
\begin{minipage}{0.45\textwidth}
\begin{algorithm}[H]
\small
\caption{Regularized MGDA}\label{alg:MGDA} 
\begin{algorithmic}[1]
\State \textbf{input} Training data $S$, initial model $x_0$, and the learning rates $\{\alpha_t\}_{t=0}^{T}$.
\For {$t=0, \dots, T-1$}
\State Compute gradients $\nabla F_S(x_t)$
\State Compute CA direction $d(x_t)$  by~\eqref{eq:ca_direction_problem}
\State Update $x_{t+1}$ via $x_{t+1} = x_t + \alpha_t d(x_t)$
\EndFor
\State \textbf{output} $x_T$
\end{algorithmic} 
\end{algorithm}
\vspace{-4mm}
\end{minipage}
\hspace{3mm}
\begin{minipage}{0.5\textwidth} 
\begin{algorithm}[H]
\small
\caption{MoDo - Stochastic MGDA}\label{alg:modo} 
\begin{algorithmic}[1]
\State \textbf{input} Training data $S$, initial model $x_0$, initial weight  $\lambda_0$, and learning rates $\{\alpha_t\}_{t=0}^{T}$, $\{\gamma_t\}_{t=0}^{T}$.
\For {$t=0, \dots, T-1$}
\State Compute $\nabla F_{ z_{t+1,1}}(x_t)$ and $\nabla F_{ z_{t+1,2}}(x_t)$
\State Update $\lambda_{t+1}$ by~\eqref{eq:lambda-h1h2-update} 
\State Update $x_{t+1}$  by~\eqref{eq:x-modo-update}
\EndFor
\State \textbf{output} $x_T$
\end{algorithmic} 
\end{algorithm}
\end{minipage}
\end{figure}
\subsection{A stochastic algorithm for MOL}\label{sec.modo}
 MGDA finds  $\lambda^*(x)$ in~\eqref{eq:ca_direction_problem} using the full-batch gradient $\nabla F_S(x)$, and then constructs $d(x) = -\nabla F_S(x) \lambda^*(x)$, a CA direction for all empirical objectives $f_{S,m}(x)$; see details in Algorithm~\ref{alg:MGDA}.
However, in practical statistical learning settings, the full-batch gradient $\nabla F_S(x)$ may be costly to obtain, and thus one may resort to a stochastic estimate of $\nabla F_S(x)$ instead.
The direct  stochastic counterpart of MGDA, referred to as the stochastic multi-gradient algorithm in \citep{liu2021stochastic},  replaces the full-batch gradients $\nabla f_{S, m}(x)$ in (\ref{eq:ca_direction_problem}) with their stochastic approximations $\nabla f_{z, m}(x)$ for $z\in S$,
which, however, introduces a biased stochastic estimate of $\lambda_{t+1}^*$, thus a biased CA direction; see~\citep[Section~4]{liu2022convergence} and~\citep[Section~2.3]{fernando2022mitigating}.

To provide a tight analysis, we introduce a simple yet theoretically grounded stochastic variant of MGDA - stochastic Multi-Objective gradient with DOuble sampling algorithm  (MoDo).
MoDo obtains an unbiased stochastic estimate of the gradient of problem~\eqref{eq:ca_direction_problem} through double (independent) sampling because
\begin{align*}
  \E_{z_{t,1},z_{t,2}}[\nabla F_{z_{t,1}}(x_t)^\top\nabla F_{z_{t,2}}(x_t)\lambda_t] 
  =& \nabla F_S(x_t)^\top\nabla F_S(x_t)\lambda_t.\numberthis
\end{align*}
At each iteration $t$, denote $z_{t,s}$ as an independent sample from $S$ with $s \in [2]$, and $\nabla F_{z_{t,s}}(x_t)$ as a stochastic estimate of $\nabla F_{S}(x_t)$.
 MoDo updates  $x_t$ and $\lambda_t$ by
\begin{subequations}
\begin{align}
\lambda_{t+1} &= \Pi_{\Delta^{M}}\left(\lambda_t - \gamma_t \big(\nabla F_{z_{t,1}}(x_t)^\top \nabla F_{z_{t,2}}(x_t) + \rho \mathrm{I}\big)  \lambda_t\right) \label{eq:lambda-h1h2-update}\\
x_{t+1} &= 
x_t -  {\alpha_t} \nabla F_{{Z}_{t+1}}(x_t)  \lambda_{t+1}  \label{eq:x-modo-update}
\end{align}
\end{subequations}
where $\alpha_t, \gamma_t$ are step sizes, $\Pi_{\Delta^M}(\cdot)$ denotes Euclidean projection to the simplex $\Delta^M$, ${Z}_{t+1} = \{z_{t+1,1}, z_{t+1,2}\}$, and $\nabla F_{Z_{t+1}}(x_t) = \frac{1}{|Z_{t+1}|} \sum_{z\in Z_{t+1}} \nabla F_z(x_t)$. We summarize the MoDo algorithm in Algorithm~\ref{alg:modo} and will focus on its theoretical analysis subsequently.

\section{Optimization, Generalization and Three-Way Trade-Off} 
\label{sec:generalization_and_stability_in_gradients}

This section presents the theoretical analysis of the PS population risk associated with the   MoDo algorithm, where the analysis of generalization error is in Section~\ref{sub:MOL_gen_err} and that of optimization error is in Section~\ref{sub:MOL_opt_err}.  
A summary of our main results is given in Table~\ref{tab:risk_compare_refined}.

\subsection{Multi-objective generalization and uniform stability}\label{sub:MOL_gen_err}

We first bound the expected PS generalization error by the generalization in gradients in Proposition~\ref{prop:gen_err_minnorm_bound_grad}, then introduce the MOL uniform stability and establish its connection to the generalization in gradients. Finally, we bound the MOL uniform stability.
\begin{prop}\label{prop:gen_err_minnorm_bound_grad}
With $\|\cdot\|_{\rm F}$ denoting the Frobenious norm, the PS generalization error
$R_{\rm gen}(A(S))$ in~\eqref{eq:pop_risk_decompose_AS} is bounded by
\begin{equation}
\E_{A,S} [R_{\rm gen}(A(S))]
\leq 
\E_{A,S} [\|\nabla F(A(S))- \nabla F_S(A(S)) \|_{\rm F} ] .
\label{eq:gen_err_minnorm_bound_grad}
\end{equation}
\end{prop}

\begin{proof}
\label{proof:gen_err_minnorm_bound_grad}
For a given model $x$, it holds that
\begin{align*}
  R_{\rm gen}(x)
  =& \min_{\lambdain} \|\nabla F({x}) \lambda\|-\min _{\lambdain} \left\|\nabla F_S({x}) \lambda\right\| 
  = - \max_{\lambdain} - \|\nabla F({x}) \lambda\| + \max _{\lambdain} 
 - \left\|\nabla F_S({x}) \lambda\right\| \\
  \stackrel{(a)}{\leq} & 
  \max_{\lambdain} (\|\nabla F({x}) \lambda\| -\|\nabla F_S({x}) \lambda \|)
  \stackrel{(b)}{\leq} 
  \max_{\lambdain} (\|(\nabla F({x})- \nabla F_S({x})) {\lambda} \|) \\
  \stackrel{(c)}{\leq} &
  {\max_{\lambdain}(\|\nabla F(x)- \nabla F_S(x) \|_{\rm F} \|\lambda\|_{\rm F})} 
  \leq \|\nabla F(x)- \nabla F_S(x) \|_{\rm F} 
\numberthis
\end{align*}
where $(a)$ follows from the subadditivity of max operator, $(b)$ follows from triangle inequality, $(c)$ follows from Cauchy-Schwartz inequality.
Setting $x = A(S)$,
and taking expectation over $A,S$ on both sides of the above inequality proves the result.
\end{proof}

With Proposition~\ref{prop:gen_err_minnorm_bound_grad}, next, we introduce the concept of MOL uniform stability tailored to MOL problems and show that PS generalization error in MOL can be bounded by the  MOL uniform stability. Then we analyze their bound in general nonconvex case and strongly convex case, respectively.

\begin{table}[tb]
\centering
\fontsize{8}{9}\selectfont
\caption{Comparison of optimization error, generalization error and population risk under different assumptions for static and dynamic weighting. ``NC'', ``C'', ``SC'' represent nonconvex and strongly convex, and ``Lip-C'', ``S'' represent Lipschitz continuous and smooth, respectively. Non-dominant terms are omitted.
The results are given in $\mathcal{O}(\cdot)$ if not otherwise specified.}
\begin{tabular}{c|c|cccc}
\hline\hline
{Assmp} 
& {Method} & 
Optimization  &  \!\!\!\!Generalization  \!\!  & Risk 
& CA weight distance \\
\midrule
\!\!\multirow{2}{*}{\makecell{NC, \\Lip-C, S}} \!\!
& Static
& $(\alpha T)^{-\frac{1}{2}} + \alpha^{\frac{1}{2}} $ 
& $ T^{\frac{1}{2}} n^{-\frac{1}{2}}$ 
& $ n^{-\frac{1}{6}}$
& $\Theta(1)$\\
& Dynamic
& $ (\alpha T)^{-\frac{1}{2}} + \alpha^{\frac{1}{2}} + \gamma^{\frac{1}{2}}$ 
& $ T^{\frac{1}{2}} n^{-\frac{1}{2}}$
& $ n^{-\frac{1}{6}}$
& $ \gamma\rho^{-1} + \alpha\gamma^{-1}\rho^{-2}$ \\
\hline
\!\!\multirow{2}{*}{\makecell{\\SC, S}} \!\!
& Static
& $ (1 - \alpha)^{\frac{T}{2}} + \alpha^{\frac{1}{2}} $ 
& $ n^{-\frac{1}{2}}$ 
& $ n^{-\frac{1}{2}}$ 
& $\Theta(1)$ \\
& Dynamic
& \makecell[c]{\footnotesize$
\min\{ 
(\alpha T)^{-\frac{1}{2}} + \alpha^{\frac{1}{2}} + \gamma^{\frac{1}{2}} + \rho^{\frac{1}{2}}$, \\
$(1 - \alpha)^{\frac{T}{2}} + \alpha^{\frac{1}{2}} + \gamma T
\}$} 
&\!\!\!\! {\footnotesize$\begin{cases}
n^{-\frac{1}{2}}, ~{\gamma = \mathcal{O}(T^{-1})}\\ T^{\frac{1}{2}} n^{-\frac{1}{2}},~ {\rm o.w.}  
\end{cases}$}
\hspace{-5mm}
& \!\!\!\!{\footnotesize$\begin{cases} 
n^{-\frac{1}{2}}\\
n^{-\frac{1}{6}}
\end{cases}$}
\hspace{-5mm}
& $ \gamma\rho^{-1} + \alpha\gamma^{-1}\rho^{-2}$ \\
\hline\hline
\end{tabular}
\label{tab:risk_compare_refined}
\vspace{-0.3cm}
\end{table}

\begin{dfn}[MOL uniform stability]
\label{def:stab_grad}
A randomized algorithm $A: \mathcal{Z}^n \mapsto \mathbb{R}^d$, is MOL-uniformly stable with $\epsilon_{\rm F}$   if for all neighboring  datasets $S, S^{\prime}$ that differ in at most one sample, we have 
\begin{equation}\label{eq:stab_grad_p2norm}
  \sup _z ~
  \mathbb{E}_A\big[\|\nabla F_z(A(S) )-\nabla F_z(A(S^{\prime}) )\|_{\rm F}^2 \big] 
  = \epsilon_{\rm F}^2.
\end{equation}
\end{dfn}
Next, we show the relation between the upper bound of PS generalization error in~\eqref{eq:gen_err_minnorm_bound_grad} and MOL uniform stability in~\eqref{eq:stab_grad_p2norm}.
\begin{prop}[MOL uniform stability and generalization]
\label{prop:stability_gen_grad_F}
Assume for any $z$, the function $F_z(x)$ is differentiable.
If a randomized MOL algorithm $A: \mathcal{Z}^n \mapsto \mathbb{R}^d$ is MOL-uniformly stable with $\epsilon_{\rm F}$,
then
\begin{equation}\label{eq:stability_gen_grad_F}
 \mathbb{E}_{A, S}[\|\nabla F(A(S))-\nabla F_S(A(S)) \|_{\rm F}] 
\leq 4\epsilon_{\rm F} 
+ \sqrt{n^{-1} \mathbb{E}\left[\mathbb{V}_{z\sim \cal D}(\nabla F_z(A(S) ))\right]}
\end{equation}
where the variance is defined as
{\small $ \mathbb{V}_{z\sim \cal D}(\nabla F_z(A(S) ))=\mathbb{E}_{z\sim \cal D}\big[\|\nabla F_z(A(S) )-\mathbb{E}_{z\sim \cal D}[\nabla F_z(A(S) )]\|_{\rm F}^2\big]$}.
\end{prop}

Proposition~\ref{prop:stability_gen_grad_F} establishes the connection between the upper bound of the  PS generalization error and the MOL uniform stability, where the former can be bounded above by the latter plus the variance of the stochastic gradient over the population data distribution.
It is worth noting that the standard arguments of bounding the generalization error measured in function values by the uniform stability measured in function values~\citep[Theorem~2.2]{hardt2016train} is not applicable here as the summation and norm operators are not exchangeable.
More explanations are provided in the proof 
in Appendix~\ref{sub_app:proof_stability_gen_grad_F}.
\begin{tcolorbox}[emphblock]
\begin{thm}[PS generalization error of MoDo in the NC case]
\label{crlr:bound_stab_grad_MGDA}
Let $A$ be the MoDo algorithm.
If $\sup _z \mathbb{E}_A\left[\|\nabla F_z(A(S) ) \|_{\rm F}^2 \right] \leq G^2  $ for any $S$, then the MOL uniform stability  $\epsilon_{\rm F}^2$  in Definition~\ref{def:stab_grad} for MoDo algorithm satisfies 

~~~a) the MOL uniform stability  $\epsilon_{\rm F}^2$ for $A_t(S)$ with $t\in [T]$ is upper bounded by $ \epsilon_{\rm F}^2 = \mathcal{O} (T n^{-1}) $;

~~~b) there exist functions $F_z(x)$, neighboring datasets $S$, $S'$, and $t\in [T]$ such that the MOL uniform stability $\epsilon_F$ for $A_t(S)$ is lower bounded by
  $\epsilon_{\rm F}^2 = \Omega (T n^{-1}). $ \\
And the PS generalization error at iteration $t\in [T]$ is $\E_{A,S}[R_{\rm gen}(A_t(S))] = \mathcal{O}(T^{\frac{1}{2}} n^{-\frac{1}{2}} ).$
\end{thm}
\end{tcolorbox}

Proof of Theorem~\ref{crlr:bound_stab_grad_MGDA} is provided in Appendix~\ref{sub_app:proof_PS_gen_NC}.
Compared to the function value uniform stability upper bound  in~\citep[Theorem~3.12]{hardt2016train}  for nonconvex single-objective learning, Theorem~\ref{crlr:bound_stab_grad_MGDA} does not require a step size decay $\alpha_t =\mathcal{O}(1/t)$, thus can enjoy at least a polynomial convergence rate of optimization errors w.r.t. $T$. 
The tightness of the stability upper bound is verified by providing a matching lower bound of the stability.
Combining Theorem~\ref{crlr:bound_stab_grad_MGDA} with Proposition~\ref{prop:stability_gen_grad_F}, to ensure the generalization error is diminishing with $n$, one needs to choose $T = o(n)$, which lies in the ``early stopping'' regime and results in potentially large optimization error.
While the ``early stopping'' phenomenon has been indeed observed in practice for general nonconvex settings, we next provide a tighter bound in the strongly convex (SC) case that allows a larger choice of $T$.
Below we list the standard assumptions used to derive the  MOL uniform stability in the SC case.

\begin{assumption}
\label{assmp:lip_cont_grad_f} The gradient
$\nabla f_{z,m}(x)$ is $\ell_{f, 1}$-Lipschitz continuous for all $m \in [M]$ for all $z$, then
$\nabla F_z(x)$ is $\ell_{F,1}$-Lipschitz continuous in Frobenius norm with $\ell_{F,1} = \sqrt{M}\ell_{f,1}$.
\end{assumption}

\begin{assumption}
\label{assmp:sconvex}
For all $m \in [M]$, $z \in \cal Z$,
$f_{z,m}(x)$ is $\mu$-strongly convex w.r.t. $x$, with $\mu > 0$.
\end{assumption}

Note that Assumptions~\ref{assmp:lip_cont_grad_f} and~\ref{assmp:sconvex} are only used to derive the MOL uniform stability in the SC case but not in the general NC case. 
In the SC case, the gradient norm $\|\nabla F_z(x)\|_{\rm F}$ is generally unbounded in  $\mathbb{R}^d$. Therefore, one cannot assume Lipschitz continuity of $f_{z,m}(x)$. We address this challenge by showing that $\{x_t\}_{t=1}^T$ generated by the MoDo algorithm are bounded as stated in Lemma~\ref{lemma:x_t_bounded_sc_smooth}.
Combining with Assumption~\ref{assmp:lip_cont_grad_f},  the gradient norm $\|\nabla F_z(x_t)\|_{\rm F}$ is also bounded, which serves as a stepping stone to derive the MOL stability bound.
\begin{lmm}[$x_t$ bounded for SC and smooth objectives]
\label{lemma:x_t_bounded_sc_smooth}
Suppose Assumptions~\ref{assmp:lip_cont_grad_f} and \ref{assmp:sconvex} hold. 
For $\{x_t\}, t \in [T]$  generated by the dynamic weighting algorithms such as MoDo and SMG with weight $\lambdain$, step size $\alpha_t = \alpha$, and $0 \leq  \alpha \leq  \ell_{f,1}^{-1} $, then 

~~~a) there exists a finite positive constant $c_x$ such that $\| x_t \| \leq c_x$; 

~~~b) there exists finite positive constants $\ell_f $, $\ell_F = \sqrt{M} \ell_f$, such that for all $\lambdain$, we have 
$\|\nabla F(x_t)\lambda\| \leq \ell_f$, and 
$\|\nabla F(x_t)\|_{\rm F} \leq \ell_F $.
\end{lmm}

Proof of Lemma~\ref{lemma:x_t_bounded_sc_smooth} is deferred to Appendix~\ref{sub_app:proof_xt_bounded_SC}.
With Lemma~\ref{lemma:x_t_bounded_sc_smooth}, the MOL uniform stability and the PS generalization error of MoDo are provided below.
\begin{tcolorbox}[emphblock]
\begin{thm}[PS generalization error of MoDo in SC case]\label{thm:grad_stability_modo_sc}
Suppose Assumptions~\ref{assmp:lip_cont_grad_f} and~\ref{assmp:sconvex} hold. 
The MOL uniform stability $\epsilon_F$ in Definition~\ref{def:stab_grad} for MoDo algorithm with $\gamma = \mathcal{O}(T^{-1})$ satisfies

~~~a) $\epsilon_F$ for $A_t(S)$ with $t\in [T]$ is upper bounded by
\begin{equation}\label{eq:modo_SC_stab_gen}
\epsilon_{\rm F}^2
= \mathcal{O} \big({{M}{n}^{-1}
(\alpha  + {M \gamma} + {M}{n}^{-1} ) } \big) ;
\end{equation}   
~~~b) there exist functions $F_z(x)$ that satisfy Assumptions~\ref{assmp:lip_cont_grad_f} and~\ref{assmp:sconvex}, neighboring datasets $S$, $S'$, and $t\in [T]$ such that $\epsilon_F$ for $A_t(S)$ is lower bounded by
\begin{equation}
\epsilon_{\rm F}^2
= \Omega ({M}{n^{-2}} ).
\end{equation}   
The PS generalization error at iteration $t\in [T]$ is 
\begin{align}\label{eq:modo_SC_gen}
\E_{A,S}[R_{\rm gen}(A_t(S))] = \mathcal{O}(n^{-\frac{1}{2}}).
\end{align}
\end{thm}
\end{tcolorbox}

See the proof of Theorem~\ref{thm:grad_stability_modo_sc} in Appendix~\ref{sub_app:proof_gen_sc_app}.
Theorem~\ref{thm:grad_stability_modo_sc} provides both the upper and lower bounds for the MOL uniform stability of MoDo in terms of the step sizes $\alpha,\gamma$, and training data size $n$. Below we provide a remark on how the choice of these parameters affects the stability. 
\begin{remark}
\label{rmk:grad_stability_modo_sc}
If we choose $\alpha = \Theta(T^{-\frac{1}{2}})$, $\gamma = \mathcal{O}(T^{-1})$, and $T = \Theta(n^2)$, the MOL uniform stability upper bound matches the lower bound in an order of $n^{-2}$, suggesting that our bound is tight.
The generalization error bound in~\eqref{eq:modo_SC_gen} is a direct implication from the MOL uniform stability bound in~\eqref{eq:modo_SC_stab_gen}, Propositions~\ref{prop:gen_err_minnorm_bound_grad} and \ref{prop:stability_gen_grad_F}.
It states that the PS generalization error of MoDo is ${\cal O}(n^{-\frac{1}{2}})$, which matches the generalization error of static weighting up to a constant~\citep{lei2022stability_NC_nonsmooth}.
Our result also indicates that when all the objectives are strongly convex, choosing small step sizes $\alpha$ and $\gamma$ leads to a smaller MOL uniform stability and thus can benefit the generalization error.
\end{remark}

\subsection{Multi-objective CA distance and optimization error}\label{sub:MOL_opt_err}
In this section, we bound the multi-objective PS optimization error $\min_{\lambda\in \Delta^{M}}\|\nabla F_S(x) \lambda\|$ which has been the main metric in the recent MOL optimization literature such as~\citep{Desideri2012mgda,liu2021stochastic,fernando2022mitigating}.
As discussed in Section~\ref{sub:measure_decompose}, this measure being zero implies the model $x$ achieves a Pareto stationarity for the empirical problem.

Below we list an additional standard assumption used to derive the theoretical results.
\begin{assumption}[Lipschitz $F_z(x)$]
\label{assmp:lip_cont_f}
For all $m \in [M]$,
$f_{z,m}(x)$ are $\ell_{f}$-Lipschitz continuous 
for all $z$, then
$F_z(x)$ are $\ell_{F}$-Lipschitz continuous in Frobenius norm  with $\ell_F = \sqrt{M}\ell_f$.
\end{assumption}

We first introduce the theoretical results on the CA direction and CA weight distances, given in Theorems~\ref{lemma:converge_MGDA_bounded_grad} and~\ref{thm:dist_CA_weight}.

\begin{tcolorbox}[emphblock]
\begin{thm}[CA direction distance of MoDo]
\label{lemma:converge_MGDA_bounded_grad}
Suppose either: 1) Assumptions~\ref{assmp:lip_cont_grad_f} and~\ref{assmp:lip_cont_f} hold; or, 2) Assumptions~\ref{assmp:lip_cont_grad_f} and~\ref{assmp:sconvex}
hold, with $\ell_f $ and $\ell_F$ defined in  Lemma~\ref{lemma:x_t_bounded_sc_smooth}. 
For $\{x_t\}, \{\lambda_t\}$ generated by MoDo with step size $\alpha_t = \alpha$, $\gamma_t = \gamma$, and  regularization $\rho \geq 0$, 
given training data $S$, it holds that
\begin{equation}\label{eq:converge_MGDA_bounded_grad}
\frac{1}{T} \sum_{t=1}^T \mathcal{E}_{\rm ca}(x_t, \lambda_{t+1})
= \mathcal{O} 
\big({\gamma^{-1} T^{-1}} + {M^{\frac{1}{2}} {\alpha}^{\frac{1}{2}} {\gamma^{-\frac{1}{2}}} } + \gamma M  + {\rho}{\gamma^{-1}} + \rho 
\big). 
\end{equation}
\end{thm}
\end{tcolorbox}
Theorem~\ref{lemma:converge_MGDA_bounded_grad} establishes the convergence to the CA direction using the measure introduced in Section~\ref{sub:measure_decompose}, with proof is provided in Appendix~\ref{sub_app:proof_CA_dist_MoDo}.
For example, one can choose $\alpha = \Theta(T^{-\frac{3}{4}})$,  $\gamma = \Theta(T^{-\frac{1}{4}})$, and $\rho = \mathcal{O}(T^{-\frac{1}{2}})$, then the RHS of~\eqref{eq:converge_MGDA_bounded_grad} is $\mathcal{O}(T^{-\frac{1}{4}})$. 

Below we provide a stronger result on the distance to the CA weight when the regularization is strictly enforced, i.e., $\rho > 0$, for both static weighting and MoDo algorithms.

\begin{prop}[CA weight distance of static weighting]
  Suppose Assumption~\ref{assmp:lip_cont_grad_f} holds. Then there exists $\lambdain$ for static weighting such that
\begin{align}
  \mathcal{E}_{\rm caw}(x_{T}, \lambda) =\Theta(1).
\end{align}
\end{prop}

\begin{proof}
First, for the upper bound, since both $\lambda, \lambda_{\rho}^*(x_T)\in \Delta^M$, it holds that
\begin{align}
  \mathcal{E}_{\rm caw}(x_{T}, \lambda)\leq \E_A  [\|\lambda - \lambda_\rho^*(x_T)\|^2] \leq 4.
\end{align}
Second, for the lower bound, for a given $\lambda_\rho^*(x_T)\in \Delta^M$ with $M\geq 2$, 
let $\lambda_{\rho,m}^*(x_T)$ denote the $m$-th element of $\lambda_\rho^*(x_T)$ with $m\in [M]$.
Define $m^* \coloneqq \mathop{\arg\min}_{m\in [M]} \E_A[\lambda_{\rho,m}^*(x_T)]$.
Then $\E_A[\lambda_{\rho,m^*}^*(x_T)]\leq \frac{1}{M}\leq \frac{1}{2}$.
Then there exists $\lambdain$ with $\lambda_{m^*} = 1$ such that
\begin{align}
\mathcal{E}_{\rm caw}(x_{T}, \lambda)
= \|\lambda - \E_A[\lambda_\rho^*(x_T)]\|^2
  \geq \Big(1 - \frac{1}{M} \Big)^2
  \geq \frac{1}{4}.  
\end{align}
Combining the upper and lower bounds yields the result.
\end{proof}

\begin{tcolorbox}[emphblock]
\begin{thm}[CA weight distance of MoDo]
  \label{thm:dist_CA_weight}
Suppose either: 1) Assumptions~\ref{assmp:lip_cont_grad_f} and~\ref{assmp:lip_cont_f} hold; or, 2) Assumptions~\ref{assmp:lip_cont_grad_f} and~\ref{assmp:sconvex}
hold, with $\ell_f $ and $\ell_F$ defined in  Lemma~\ref{lemma:x_t_bounded_sc_smooth}. 
For $\{x_t\}, \{\lambda_t\}$ generated by MoDo with step size $\alpha_t = \alpha$, $\gamma_t = \gamma$, and  regularization $\rho > 0$, 
given training data $S$,
it holds that
\begin{align}
  \mathcal{E}_{\rm caw}(x_{T}, \lambda_{T+1}) & 
  = \mathcal{O} 
  \big( (1 - \rho \gamma)^T 
    + \rho^{-1}\gamma (M^{\frac{1}{2}} + \rho)^2
    + \rho^{-2} \gamma^{-1} \alpha M 
\big).
\end{align}
\end{thm}
\end{tcolorbox}

Proof of Theorem~\ref{thm:dist_CA_weight} is provided in Appendix~\ref{sub_app:proof_CA_weight_dist_MoDo}.
Below we provide a remark on the difference between the two measures: CA direction distance, and CA weight distance.
\begin{remark}
\label{rmk:dist_CA_weight}
  Compared to the convergence to CA direction, the convergence to CA weight is stronger because it involves last-iterate point convergence instead of average-iterate function value convergence, and can only be guaranteed with $\rho = \omega(\max\{\gamma, \alpha^{\frac{1}{2}}\gamma^{-\frac{1}{2}}, \gamma^{-1} T^{-1} \})$, resulting in a trade-off in convergence to PS stationarity and convergence to the CA weight.
  Using the distance to the CA weight as a measure, the lower bound of the static weighting method in this measure can be derived, which is a constant, strictly greater than the upper bound of MoDo, further justifying the benefit of MoDo over static weighting in this measure. 
\end{remark}

Next, we introduce the PS optimization error of MoDo in Theorem~\ref{thm:opt_err_modo_nonconvex_bounded_grad}.
\begin{tcolorbox}[emphblock]
\begin{thm}[PS optimization error of MoDo]
\label{thm:opt_err_modo_nonconvex_bounded_grad}
Given training data $S$,
define $c_F$ such that $\E_A [F_S (x_{1}) \lambda_1] - \min_{x\in \mathbb{R}^d} \E_A [F_S (x) \lambda_1] \leq c_F$.
Considering $\{x_t\} $ generated by MoDo (Algorithm~\ref{alg:modo}), with $\alpha_t = \alpha \leq 1/(2\ell_{f,1})$, $\gamma_t = \gamma$.
Suppose

~~~1)  Assumptions~\ref{assmp:lip_cont_grad_f} and~ \ref{assmp:lip_cont_f} hold (NC case), then 
\begin{equation}
\E_A \Big[\min_{t\in [T]}~ R_{\rm opt}(x_t) \Big]
= \mathcal{O} 
\big( \alpha^{-\frac{1}{2}} T^{-\frac{1}{2}}  + {\gamma^{\frac{1}{2}} M^{\frac{1}{2}} }  + { \alpha^{\frac{1}{2}} } + {\rho^{\frac{1}{2}}} 
\big);
\end{equation}
~~~2)  Assumptions~\ref{assmp:lip_cont_grad_f}, \ref{assmp:sconvex}
hold (SC case), with $\ell_f $   defined in  Lemma~\ref{lemma:x_t_bounded_sc_smooth}, then
\begin{equation}
\E_A \Big[\min_{t\in [T]}~ R_{\rm opt}(x_t) \Big]
= \mathcal{O}
\big(  \min\{ \alpha^{-\frac{1}{2}} T^{-\frac{1}{2}}  + {\gamma^{\frac{1}{2}} M^{\frac{1}{2}} }  + { \alpha^{\frac{1}{2}} } + {\rho^{\frac{1}{2}}}, (1 - \alpha)^{\frac{T}{2}} + \alpha^{\frac{1}{2}} + M^{\frac{1}{2}}\gamma T \}
\big).
\end{equation}
\end{thm}
\end{tcolorbox}

Proof of Theorem~\ref{thm:opt_err_modo_nonconvex_bounded_grad} is provided in Appendix~\ref{sub_app:bound_opt_err_MoDo}.
Below we provide a remark on Theorem~\ref{thm:opt_err_modo_nonconvex_bounded_grad} under different choices of step sizes.
\begin{remark}
\label{remark:PS_opt_err}
Note that the original result with the squared PS optimization error in the general nonconvex case is
\begin{align}
\frac{1}{T}\sum_{t=1}^T 
\E_A \Big[\min_{\lambdain}\| \nabla F_S(x_t) \lambda \|^2 \Big] = \mathcal{O}\big( {\alpha^{-1} T^{-1}} + \alpha + \gamma
+ \rho \big) .
\end{align}
And it holds for any choice of $\alpha, \gamma, T$ as long as $\alpha \leq 1/(2\ell_{f,1})$.
Based on this result, one optimal choice to ensure the best $\mathcal{O}(\epsilon^{-2})$ sample complexity of the optimization error in square is $\alpha = \Theta(T^{-\frac{1}{2}})$, $\gamma = \Theta(T^{-\frac{1}{2}})$, $\rho = \mathcal{O}(T^{-\frac{1}{2}})$. However, this results in a constant error bound of the CA distance according to Theorem~\ref{lemma:converge_MGDA_bounded_grad}.
To ensure better convergence to CA direction, one possible choice, $\alpha = \Theta(T^{-\frac{3}{4}})$,  $\gamma = \Theta(T^{-\frac{1}{4}})$, and $\rho = \mathcal{O}(T^{-\frac{1}{2}})$,  is suboptimal with regard to the convergence to Pareto stationarity, as evidenced by Theorem~\ref{thm:opt_err_modo_nonconvex_bounded_grad}.  This exhibits a trade-off between convergence to the CA direction and convergence to Pareto stationarity.
\end{remark}

\subsection{Optimization, generalization and conflict avoidance trade-off}\label{sub:opt_gen_ca_tradeoff}

Combining the results in Sections~\ref{sub:MOL_gen_err} and~\ref{sub:MOL_opt_err}, we are ready to analyze and summarize the three-way trade-off of MoDo.
With $A_t(S) = x_t$ denoting the output of algorithm $A$ at the $t$-th iteration, we can decompose the PS population risk  $R_{\rm pop}(A_t(S))$ as   (cf.~\eqref{eq:excess_risk_decompose} and \eqref{eq:gen_err_minnorm_bound_grad})
\begin{align*}
& \E_{A,S}\big[R_{\rm pop}(A_t(S)) \big]  
\leq 
\E_{A,S}\Big[\min_{\lambdain}\|\nabla F_S(A_t(S)) \lambda\| \Big] 
+ \E_{A,S}\Big[ \| \nabla F(A_t(S)) -\nabla F_S(A_t(S)) \|_{\rm F} \Big] .
\end{align*}

\textbf{The general nonconvex  case.}
Suppose Assumptions~\ref{assmp:lip_cont_grad_f} and \ref{assmp:lip_cont_f} hold.
By the generalization error bound in Theorem~\ref{crlr:bound_stab_grad_MGDA}, and the optimization error bound in Theorem~\ref{thm:opt_err_modo_nonconvex_bounded_grad},  denote $\hat{t} \in \arg\min_{t\in [T]}R_{\rm opt}(x_t)$, the PS population risk of the output of MoDo can be bounded by 
\begin{equation}\label{eq:bound_nc_pop}
\E_{A,S}\Big[ R_{\rm pop}(A_{\hat t}(S)) \Big] =  
\mathcal{O}\left(\alpha^{-\frac{1}{2}} T^{-\frac{1}{2}} + \alpha^{\frac{1}{2}} + \gamma^{\frac{1}{2}}
+ T^{\frac{1}{2}}n^{-\frac{1}{2}}\right) .
\end{equation}
\emph{Discussion of trade-off.}
Choosing step sizes {$\alpha = \Theta(T^{-\frac{1}{2}})$, $\gamma =\Theta(T^{-\frac{1}{2}})$,  
and number of steps $T = \Theta(n^{\frac{2}{3}})$, then  the expected PS population risk is 
$ \mathcal{O}(n^{-\frac{1}{6}} )$, which matches the PS population risk upper bound of a general nonconvex single objective in~\citep{lei2022stability_NC_nonsmooth}.
A clear trade-off in this case is between the optimization error and generalization error, controlled by $T$. Indeed, increasing $T$ leads to smaller optimization errors but larger generalization errors, and vice versa.
To satisfy convergence to  CA direction, it requires $\gamma = \omega (\alpha)$ based on Lemma~\ref{lemma:converge_MGDA_bounded_grad}, and the optimization error in turn becomes worse, so does the PS population risk.
Specifically, choosing $\alpha = \Theta(T^{-\frac{1}{2}})$, $\gamma = \Theta(T^{-\frac{1}{4}})$, and $T = \Theta(n^{\frac{4}{5}})$ leads to the expected PS population risk in $\mathcal{O}(n^{-\frac{1}{10}})$, and the distance to CA direction in $\mathcal{O}(n^{-\frac{1}{10}})$.
This shows another trade-off between conflict avoidance and optimization error.

\textbf{The strongly convex case.}
Suppose Assumptions~\ref{assmp:lip_cont_grad_f} and \ref{assmp:sconvex} hold.
By the generalization error and the optimization error given in Theorems~\ref{thm:grad_stability_modo_sc} and \ref{thm:opt_err_modo_nonconvex_bounded_grad}, the
  PS population risk of MoDo can be bounded by 
\begin{equation}
\E_{A,S}\Big[ R_{\rm pop}(A_{\hat t}(S)) \Big]
= \mathcal{O}\left( \alpha^{-\frac{1}{2}} T^{-\frac{1}{2}} + \alpha^{\frac{1}{2}} + \gamma^{\frac{1}{2}}
+ n^{-\frac{1}{2}}\right).   
\end{equation}
\emph{Discussion of trade-off.}
Choosing $\alpha  = \Theta(T^{-\frac{1}{2}} )$,  $\gamma = o(T^{-1})$, and 
$ T = \Theta(n^2)$, the expected PS population risk in gradients is  
$ \mathcal{O}( n^{-\frac{1}{2}} )$. 
However, choosing $\gamma = o(T^{-1})$ leads to large distance to the CA direction according to Lemma~\ref{lemma:converge_MGDA_bounded_grad}  because the term ${4}/{(\gamma T)}$ in \eqref{eq:converge_MGDA_bounded_grad} increases with $T$.
To ensure convergence to the CA direction, it requires $\gamma = \omega(T^{-1})$, under which the tighter bound in Theorem~\ref{thm:grad_stability_modo_sc} does not hold but the bound in Theorem~\ref{crlr:bound_stab_grad_MGDA} still holds. In this case, the PS population risk under the proper choice of $\alpha, \gamma$, and $T$ is  
$\mathcal{O}(n^{-\frac{1}{6}} )$ as discussed in the previous paragraph.
Therefore, to avoid conflict among gradients, MoDo needs to sacrifice the sample complexity of PS population risk, demonstrating a trade-off between conflict avoidance and PS  population risk, as demonstrated in Figure~\ref{fig:tradeoff_3way}.

\section{Application to Other MOL Algorithms}

Our theoretical framework is general and can be applied to other stochastic MOL algorithms to analyze the three errors.
To demonstrate this, we apply our framework to analyze other stochastic MOL algorithms including SMG~\citep{liu2021stochastic} and MoCo~\citep{fernando2022mitigating} in this section. 
Both SMG and MoCo mitigate the bias in the CA direction during optimization. 
To achieve this, SMG~\citep{liu2021stochastic} increases the batch size during optimization,  MoCo~\citep{fernando2022mitigating} uses momentum-based methods.
We then describe in detail the updates of SMG and MoCo.

SMG substitutes the full-batch gradient $\nabla F_S(x_t)$ used in MGDA in~\eqref{eq:ca_direction_problem}  with its stochastic estimate $\nabla F_{Z_t}(x_t)$, where $Z_t$ is a randomly sampled mini-batch of data at the $t$-th iteration, and its batch size $|Z_t|$ is increasing with the iteration $t$ to mitigate the CA direction bias. The SMG algorithm is summarized in Algorithm~\ref{alg:SMG}.

Different from SMG, MoCo~\citep{fernando2022mitigating} adopts a momentum-based method to mitigate the CA direction bias. It uses the moving average of the stochastic gradient to compute the CA direction. The MoCo algorithm uses the update functions given by
\begin{subequations}\label{eq:update_moco}
\begin{align}
&Y_{t+1} = Y_t - \beta_t(Y_t - \nabla F_{z_{t+1}}(x_t)) \label{eq:update_Y_moco}\\
&\lambda_{t+1} \in {\arg\min}_{\lambdain} \|Y_t \lambda\|^2 \label{eq:update_lam_moco}\\
&x_{t+1} = x_t - \alpha_t Y_t\lambda_t
\label{eq:update_x_moco}
\end{align}    
\end{subequations}
where $Y_t$ is the moving average of the stochastic gradients, and $Y_t\lambda_t$ is the estimated CA direction.
The MoCo algorithm is summarized in Algorithm~\ref{alg:moco}.

\begin{figure}[t]
\begin{minipage}{0.49\textwidth} 
\begin{algorithm}[H]
\small
\caption{SMG~\citep{liu2021stochastic}}
\label{alg:SMG} 
\begin{algorithmic}[1]
\State \textbf{Input:} Initial model $x_0$, the learning rates $\{\alpha_t\}_{t=0}^{T}$, and the regularization $\rho=0$.
\For {$t=0, \dots, T-1$}
\State Compute $\nabla F_{Z_t}(x_t)$ with $|Z_t| = \mathcal{O}(t)$
\State Compute CA direction at $x_t$ with~\eqref{eq:ca_direction_problem} 
\indent by replacing $\nabla F_S(x_t)$ with $\nabla F_{Z_t}(x_t)$;
\State Update $x_{t+1}$ via $x_{t+1} = x_t + \alpha_t d(x_t)$;
\EndFor
\State \textbf{output} $x_T$
\end{algorithmic} 
\end{algorithm}
\end{minipage}
\hspace{3mm}
\begin{minipage}{0.49\textwidth} 
\vspace{4mm}
\begin{algorithm}[H]
\small
\caption{MoCo~\citep{fernando2022mitigating}}
\label{alg:moco} 
\begin{algorithmic}[1]
\State \textbf{Input:} Initial model $x_0$,  the learning rates $\{\alpha_t\}_{t=0}^{T}$, and the regularization $\rho=0$.
\State Set $Y_0 = \nabla F_{z_{0}}(x_t)$;
\For {$t=0, \dots, T-1$}
\State Sample gradients $\nabla F_{z_{t+1}}(x_t)$;
\State Update $Y_{t+1}$ by~\eqref{eq:update_Y_moco};
\State Update $\lambda_{t+1}$ by~\eqref{eq:update_lam_moco} 
\State Update $x_{t+1}$  by~\eqref{eq:update_x_moco}
\EndFor
\State \textbf{output} $x_T$
\end{algorithmic} 
\end{algorithm}
\end{minipage}
\end{figure}

Next, we proceed to formally introduce our theoretical results on the three errors for SMG and MoCo algorithms in the general nonconvex case.

\subsection{Multi-objective generalization}

We summarize the PS generalization error bounds of SMG (Algorithm~\ref{alg:SMG}) and MoCo (Algorithm~\ref{alg:moco}) in the general nonconvex case in Theorem~\ref{thm:PS_gen_SMG_moco}.
The proof of this theorem follows similar steps as the proof for the PS generalization error of MoDo in Theorem~\ref{crlr:bound_stab_grad_MGDA}, by first deriving their MOL uniform stability bounds, and applying Proposition~\ref{prop:stability_gen_grad_F} to connect their PS generalization errors with their MOL uniform stability bounds.
\begin{tcolorbox}[emphblock]
\begin{thm}[Generalization errors of SMG and MoCo]
\label{thm:PS_gen_SMG_moco}
Let $A$ be the SMG or MoCo algorithm.
If $\mathbb{E}_A\!\left[\|\nabla F_z(A(S) ) \|_{\rm F}^2 \right] \!\!\leq\! G^2$ for any $z$ and $S$, then
the PS generalization errors  of SMG and MoCo satisfy
\begin{align}\label{eq:smg_moco_NC_gen}
\E_{A,S}[R_{\rm gen}(A(S))] = \mathcal{O}(T^{\frac{1}{2}}n^{-\frac{1}{2}}).
\end{align}
\end{thm}
\end{tcolorbox}
The proof is deferred to Appendix~\ref{sub_app:proof_PS_gen_SMG_MoCo_NC}.
Theorem~\ref{thm:PS_gen_SMG_moco} indicates that the PS generalization errors of SMG and MoCo in the general nonconvex case are in the same rates as MoDo, with $\mathcal{O}(T^{\frac{1}{2}}n^{-\frac{1}{2}})$. This bound also matches the generalization error bound of static weighting or single objective learning in the general nonconvex case. 
The result further demonstrates the generality of our proposed theoretical framework to analyze the MOL uniform stability and PS generalization errors of various stochastic MOL algorithms.

In the following sections, we further demonstrate how to apply our proposed theoretical framework to analyze the CA distances and PS optimization errors of the stochastic MOL algorithms, SMG and MoCo.

\subsection{Multi-objective CA distance and optimization error}

\begin{table*}
\caption{Comparison with prior stochastic MOL algorithms in terms of assumptions and the guarantees of the three errors, where logarithmic dependence is omitted, and Opt., CA dist., and Gen. are short for optimization error, CA distance, and generalization error, respectively.}
\label{tab:opt_err_compare_detail}
\centering
\fontsize{8}{9}\selectfont
\begin{tabular}{c|c|c|c|c|c|c|c}
\toprule
Algorithm 
& \makecell[c]{Batch\\size} & NC 
& \makecell[c]{Lipschitz \\ $\lambda^*(x)$}
& \makecell[c]{Bounded \\
function} 
& \makecell[c]{Opt.} 
& \makecell[c]{CA  \\
dist.}
& \makecell[c]{Gen.}\\
\midrule
{SMG~\citep[Thm~5.3]{liu2021stochastic}} 
& $\mathcal{O}(t)$ 
& \xmark & \cmark  & \xmark  & $T^{-\frac{1}{8}}$ & - & -\\
{CR-MOGM~\citep[Thm~3]{zhou2022_SMOO}} 
& $\mathcal{O}(1)$ & \cmark & \xmark  & \cmark & $T^{-\frac{1}{4}}$ 
& - & - \\
{MoCo~\citep[Thm~2]{fernando2022mitigating}}
& $\mathcal{O}(1)$ 
& \cmark & \xmark  & \xmark  & $T^{-\frac{1}{20}}$ & $T^{-\frac{1}{5}}$ & -\\
{MoCo~\citep[Thm~4]{fernando2022mitigating}} 
 & $\mathcal{O}(1)$ & 
 \cmark & \xmark   & \cmark   & $T^{-\frac{1}{4}}$ & - & - \\
\rowcolor{RoyalBlue!10}
{SMG (Ours, Thms~\ref{thm:PS_gen_SMG_moco}-\ref{thm:opt_err_smg_moco})} 
& $\mathcal{O}(t)$ & 
\cmark & \xmark & \xmark   & $T^{-\frac{1}{8}}$ 
& $T^{-\frac{1}{2}}$ & $T^{\frac{1}{2}}n^{-\frac{1}{2}}$ \\
\rowcolor{RoyalBlue!10}
{MoCo (Ours, Thms~\ref{thm:PS_gen_SMG_moco}-\ref{thm:opt_err_smg_moco})} 
& $\mathcal{O}(1)$ & 
\cmark & \xmark   & \xmark & $ T^{-\frac{1}{16}}$
& $ T^{-\frac{1}{4}}$
& $ T^{\frac{1}{2}}n^{-\frac{1}{2}}$ \\
\rowcolor{RoyalBlue!10}
{\bf MoDo (Ours, Thms~\ref{crlr:bound_stab_grad_MGDA},\ref{lemma:converge_MGDA_bounded_grad},\ref{thm:opt_err_modo_nonconvex_bounded_grad})} 
& $\mathcal{O}(1)$ 
& \cmark & \xmark & \xmark & $ T^{-\frac{1}{4}}$ 
& - & $T^{\frac{1}{2}}n^{-\frac{1}{2}}$ \\
\rowcolor{RoyalBlue!10}
{\bf MoDo (Ours, Thms~\ref{crlr:bound_stab_grad_MGDA},\ref{lemma:converge_MGDA_bounded_grad},\ref{thm:opt_err_modo_nonconvex_bounded_grad})} 
& $\mathcal{O}(1)$ 
& \cmark & \xmark & \xmark & $ T^{-\frac{1}{8}}$ 
& $ T^{-\frac{1}{4}}$ & $T^{\frac{1}{2}}n^{-\frac{1}{2}}$ \\
\bottomrule
\end{tabular}
\end{table*}
Notably, in the CA distance and {optimization error} analysis, we have also developed new techniques to relax the assumptions and/or improve the final convergence rates of different algorithms; see a detailed comparison in Table~\ref{tab:opt_err_compare_detail}.
To obtain the improved analysis, one critical property is that the CA direction is unique and H\"{o}lder continuous, despite that $\lambda^*(x)$ is not Lipschitz continuous in general, as proved in Lemmas~\ref{lemma:uniqueness_CA} and~\ref{lemma:bound_diff_update_holder}.

For notation simplicity, we use  $Q \in \mathbb{R}^{d\times M}$ to denote either full-batch gradient matrix $\nabla F_S(x)$ or its stochastic estimate. 
Then the subproblem without regularization, i.e., $\rho=0$, can be defined as
\begin{align}
  \min_{\lambdain}& ~~\|Q \lambda\|^2
\end{align}
which is a constrained quadratic programming problem. The estimate of the CA direction used in either SMG with $Q = \nabla F_{Z_t}(x_t)$, or MoCo with $Q = Y_t$, can be computed by
$d_{Q} = Q \lambda_{Q}^*$, with $\lambda_{Q}^* \in \mathop{\arg\min}_{\lambdain} ~\|Q \lambda\|^2 $.

We then proceed to prove the H\"{o}lder continuity of $d_{Q}$ w.r.t. $Q$ in Lemma~\ref{lemma:bound_diff_update_holder}, which is essential for deriving the CA direction distance and PS optimization error of SMG and MoCo. This result also generalizes to constrained quadratic programming problems with general compact and convex set constraints.
\begin{lmm}[H\"{o}lder continuity of $d_{Q}$ w.r.t. $Q$]
\label{lemma:bound_diff_update_holder}
For all $Q, Q' \in \mathbb{R}^{d\times M}$, define $\lambda^* \in \mathop{\arg\min}_{\lambdain} \|Q \lambda\|^2 $, 
and $\lambda^{*\prime} \in \mathop{\arg\min}_{\lambdain} \|Q' \lambda\|^2 $, and
$d_{Q} = Q\lambda^*$, $d_{Q}' = Q'\lambda^{*\prime}$,
then 
\begin{align}
\|d_{Q}  - d_{Q'}\|^2
\leq 4 \max \Bigg\{\sup_{\lambdain}\|Q\lambda\|, \sup_{\lambdain}\|Q'\lambda\| \Bigg\} \cdot \sup_{\lambdain}\|(Q-Q^{\prime})\lambda\| .
\end{align}
\end{lmm}

\begin{proof}
With uniqueness of $d_{Q} $ given in Lemma~\ref{lemma:uniqueness_CA},
we can then rewrite $ \|d_{Q}  - d_{Q'}\|^2 = \|Q \lambda^* - Q' \lambda^{*\prime}\|^2$ as
  \begin{align*}
    \| Q \lambda^* -Q^{\prime} \lambda^{*\prime} \|^2 
    =&\|Q \lambda^*\|^2+\|Q^{\prime} \lambda^{* \prime}\|^2-2\langle Q \lambda^*, Q^{\prime} \lambda^{* \prime}\rangle \\
    =&\|Q \lambda^*\|^2-\|Q^{\prime} \lambda^{* \prime}\|^2+2\langle Q^{\prime} \lambda^{* \prime}, Q^{\prime} \lambda^{* \prime}-Q \lambda^*\rangle \\
    =& \| Q \lambda^*\|^2 - \| Q^{\prime} \lambda^{* \prime}\|^2
    + \underbrace{2\langle Q^{\prime} \lambda^{*\prime},  Q^{\prime} \lambda^{* \prime}- Q^{\prime} \lambda^*\rangle}_{\leq 0}  
    +2\langle Q^{\prime} \lambda^{* \prime},  Q^{\prime} \lambda^*- Q \lambda^*\rangle 
    \end{align*}
    where $\langle Q^{\prime} \lambda^{* \prime},  Q^{\prime} \lambda^{* \prime}- Q^{\prime} \lambda^*\rangle  \leq 0$ by Lemma~\ref{lemma:descent direction}, \eqref{eq:convex_inner_prod_property}.
Then it can be further bounded by
    \begin{align*}
    \| Q \lambda^* -Q^{\prime} \lambda^{*\prime} \|^2
    \stackrel{(a)}{\leq} & \min _{\lambda \in \Delta^{M}}\|Q \lambda\|^2
    -\min_{\lambda \in \Delta^{M}} \|Q^{\prime} \lambda\|^2
    +2\|Q^{\prime} \lambda^{* \prime}\| \|(Q^{\prime}-Q) \lambda^*\|\\
    = & -\max _{\lambda \in \Delta^{M}} -\|Q \lambda\|^2
    +\max_{\lambda \in \Delta^{M}} - \|Q^{\prime} \lambda\|^2
    +2\|Q^{\prime} \lambda^{* \prime}\|\|(Q^{\prime}-Q) \lambda^*\|\\
    \stackrel{(b)}{\leq} & 
    \max _{\lambda \in \Delta^{M}} \big(\|Q \lambda\|^2
    - \|Q' \lambda\|^2 \big)
    +2\|Q^{\prime} \lambda^{* \prime}\|\|(Q^{\prime}-Q) \lambda^*\|\\
    \stackrel{(c)}{\leq} & \max _{\lambda \in \Delta^{M}}\|(Q-Q^{\prime}) \lambda\|
    \big(\|Q\lambda\| + \|Q'\lambda\| \big)
    +2\|Q^{\prime} \lambda^{* \prime}\|\|(Q^{\prime}-Q) \lambda^*\| \\
    {\leq} & 4 \max \Big\{\sup_{\lambdain}\|Q\lambda\|, \sup_{\lambdain}\|Q'\lambda\| \Big\} \cdot \sup_{\lambdain}\|(Q-Q^{\prime})\lambda\|
    \end{align*}
    where 
    $(a)$ follows from Cauchy-Schwarz inequality;
    $(b)$ follows from subadditivity of maximum operator;
    $(c)$ follows from triangle inequality.
    The proof is complete.
\end{proof}
 
With the property in Lemma~\ref{lemma:bound_diff_update_holder}, we are able to derive the CA direction distances of SMG (Algorithm~\ref{alg:SMG}) and MoCo (Algorithm~\ref{alg:moco}), as summarized in Theorem~\ref{crlr:bias_CA_mitigate}.
\begin{tcolorbox}[emphblock]
\begin{thm}[CA direction distances of SMG and MoCo]
\label{crlr:bias_CA_mitigate}
Suppose either:
1)  Assumptions~\ref{assmp:lip_cont_grad_f}, \ref{assmp:lip_cont_f} hold; 
or,
2)  Assumptions~\ref{assmp:lip_cont_grad_f}, \ref{assmp:sconvex}
hold, with $\ell_f $  defined in  Lemma~\ref{lemma:x_t_bounded_sc_smooth}.
Considering $\{x_t\}$ and $\{\lambda_t\}$ generated by SMG or MoCo, with $\alpha_t = \alpha \leq 1/(2\ell_{f,1})$,  then under either condition~1) or 2), their CA direction distances can be bounded by
\begin{align}
\text{(SMG) }\quad
&\frac{1}{T} \sum_{t=1}^T \mathcal{E}_{\rm ca}(x_t, \lambda_{t+1}) = {\mathcal{O}}(T^{-\frac{1}{2}}) \\
\text{(MoCo) }\quad
&\frac{1}{T} \sum_{t=1}^T \mathcal{E}_{\rm ca}(x_t, \lambda_{t+1}) = \mathcal{O}(T^{-\frac{1}{4}}) .
\end{align}
\end{thm}
\end{tcolorbox}

Proof of Theorem~\ref{crlr:bias_CA_mitigate} is deferred to Appendix~\ref{sub_app:proof_CA_direction_SMG_MoCo}.
Theorem~\ref{crlr:bias_CA_mitigate} indicates that increasing the batch size during optimization as in SMG or using momentum-based methods for gradient estimation as in MoCo can both mitigate the bias in the CA direction, and lead to convergence of the CA direction distances for stochastic MOL algorithms. 

Based on Lemma~\ref{lemma:bound_diff_update_holder}, we derive improved PS optimization error bounds for SMG and MoCo.
In addition, we remove the assumption of bounded function values on the trajectory, by deriving a tighter bound on the inner product term. 
The PS optimization error bounds of SMG (Algorithm~\ref{alg:SMG}) and MoCo (Algorithm~\ref{alg:moco}) are summarized in Theorem~\ref{thm:opt_err_smg_moco}.
\begin{tcolorbox}[emphblock]
\begin{thm}[PS optimization errors of SMG and MoCo]
\label{thm:opt_err_smg_moco}
Suppose Assumptions~\ref{assmp:lip_cont_grad_f} and~\ref{assmp:lip_cont_f} hold. 
Define $c_F$ such that $\E_A [F_S (x_{1}) \lambda_1] - \min_{x\in \mathbb{R}^d} \E_A [F_S (x) \lambda_1] \leq c_F$.
Considering $\{x_t\}$ generated by SMG or MoCo, with $\alpha_t = \alpha \leq 1/(2\ell_{f,1})$, and proper choices of $\alpha, \beta$ depending on $T$,  then 
their PS optimization errors can be bounded by
\begin{align}
\text{(SMG) }\quad
& 
\E_A \Big[\min_{t\in [T]}~ R_{\rm opt}(x_t)\Big]
= \tilde{\cal O}\Big( T^{-\frac{1}{8}} \Big);\\
\text{(MoCo) }\quad
& 
 \E_A \Big[\min_{t\in [T]}~ R_{\rm opt}(x_t) \Big]
= {\cal O}\Big( T^{-\frac{1}{16}} \Big).
\end{align}
\end{thm}
\end{tcolorbox}
Proof of Theorem~\ref{thm:opt_err_smg_moco} is deferred to Appendix~\ref{sub_app:proof_PS_opt_SMG_MoCo}.
Theorem~\ref{thm:opt_err_smg_moco} provides the PS optimization error guarantees of SMG and MoCo under the same assumptions as Theorem~\ref{thm:opt_err_modo_nonconvex_bounded_grad}, which relaxes the assumption of bounded function values on the optimization trajectory as used in~\citep{fernando2022mitigating,zhou2022_SMOO}. It also improves the convergence rate of MoCo in PS optimization error without such an assumption, see the comparison in Table~\ref{tab:opt_err_compare_detail}.

\section{Related Works}
We review related work from the following three aspects -- multi-task learning, theory of MOL, and generalization based on algorithm stability.

\paragraph{Multi-task learning (MTL).}
MTL, as one application of MOL, leverages shared information among different tasks to train a model that can perform multiple tasks.
MTL has been widely applied to natural language processing, computer vision, and robotics \citep{hashimoto2016joint}, \citep{ruder2017overview}, \citep{zhang2021survey}, \citep{vandenhende2021multi}.
From the optimization perspective, a simple method for MTL is to take the weighted average of the per-task losses as the objective. 
The weights can be static or dynamic during optimization.
And weighting for different tasks can be chosen based on different criteria such as uncertainty \citep{kendall2017multi}, gradient norms \citep{chen2018gradnorm} or task difficulty \citep{guo2018dynamic}. 
These methods are often heuristic and designed for specific applications.
Another line of work tackles MTL through MOL~\citep{sener2018multi,yu2020gradient,liu2021conflict}. A foundational algorithm in this regard is MGDA~\citep{Desideri2012mgda}, which takes dynamic weighting of gradients to obtain a CA direction for all objectives. 
Stochastic variants of MGDA with optimization convergence guarantees have been proposed in~\citep{liu2021stochastic,zhou2022_SMOO,fernando2022mitigating}.
Algorithms for finding a set of Pareto optimal models rather than one have been proposed in~\citep{navon2020learning, liu2021profiling, mahapatra2021exact,  yang2021pareto, kyriakis2021pareto, zhao2021multi,  lin2022pareto, momma2022multi}.

\paragraph{Theory of MOL.}
Optimization convergence analysis for the deterministic MGDA algorithm has been provided in \citep{fliege2019complexity}. 
Later on, stochastic variants of MGDA were introduced~\citep{liu2021stochastic,zhou2022_SMOO,fernando2022mitigating}.
However, the vanilla stochastic MGDA introduces a biased estimate of the dynamic weight, which results in the biased estimate of CA direction during optimization. 
To address this issue, 
Liu et al.~\citep{liu2021stochastic} proposes to increase the batch size during optimization.
However, they employ an assumption, Lipschitz continuity of $\lambda^*(x)$ w.r.t. the gradients $\nabla F_S(x)$ to establish convergence, which does not hold in general, as first proved in~\citep[Proposition~2]{zhou2022_SMOO}.
To address this, momentum-based bias reduction algorithms~\citep{zhou2022_SMOO,fernando2022mitigating} were proposed.
{These works provide a theoretical guarantee of convergence in PS optimization error, which can be achieved by the simplest static weighting method.}
Therefore, although
the community has a rich history of investigating the optimization of MOL algorithms, their theoretical benefits over static weighting, and their generalization guarantees remain open.
Not until recently, generalization guarantees for MOL were theoretically analyzed. In~\citep{cortes2020agnostic}, a min-max formulation to solve the MOL problem is analyzed, where the weights are chosen based on the maximum function values, rather than the CA direction.  More recently,
\citep{sukenik2022generalization} provides generalization guarantees for MOL  for a more general class of weighting. These two works analyze generalization based on Rademacher complexity of the hypothesis class of the learner, with generalization error bound independent of the training process. 
Different from these works, we use the algorithm stability framework to derive the first algorithm-dependent generalization error bounds, highlighting the effect of the training dynamics. 
In addition, in contrast to prior works for MOL theory, which focus solely on either optimization~\citep{fernando2022mitigating,zhou2022_SMOO} or generalization~\citep{cortes2020agnostic,sukenik2022generalization}, we propose a holistic framework to analyze the three types of errors, namely, optimization, generalization, and CA distance in MOL with an instantiation of the proposed MoDo algorithm.
This allows us to study how the theoretical test performance depends on hyperparameters such as the number of iterations and step sizes, and how to choose hyperparameters to achieve the best trade-off among the errors.

\paragraph{Algorithm stability and generalization.} 
{Stability analysis  dates back to the work \citep{devroye1979distribution} in 1970s.}
Uniform stability and its relationship with generalization were studied  in~\citep{bousquet2002stability} for the exact minimizer of the ERM problem with strongly convex objectives. 
The work \citep{hardt2016train} pioneered the stability analysis for stochastic gradient descent (SGD) algorithms with convex and smooth objectives. The results were extended and refined in \citep{kuzborskij2018data} with data-dependent bounds, in \citep{charles2018stability,richards2021stability,lei2022stability} for non-convex objectives, and in \citep{bassily2020stability,lei2020fine} for SGD with non-smooth and convex losses. However, all these studies mainly focus on single-objective learning problems.
To our best knowledge, there is no existing work on the stability and generalization analysis for multi-objective learning problems and our results on its stability and generalization are the \emph{first-ever-known} ones.

\begin{figure*}[t]
\centering
\begin{subfigure}[b]{0.3\textwidth}
 \centering
 \includegraphics[width=\textwidth]{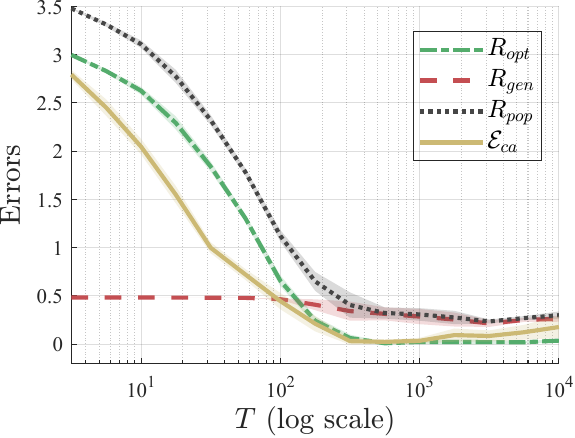}
  \vspace{-0.2cm}
 \caption{Number of iterations $T$.}
 \label{fig:T_opt_gen}
\end{subfigure}
\begin{subfigure}[b]{0.3\textwidth}
 \centering
 \includegraphics[width=\textwidth]{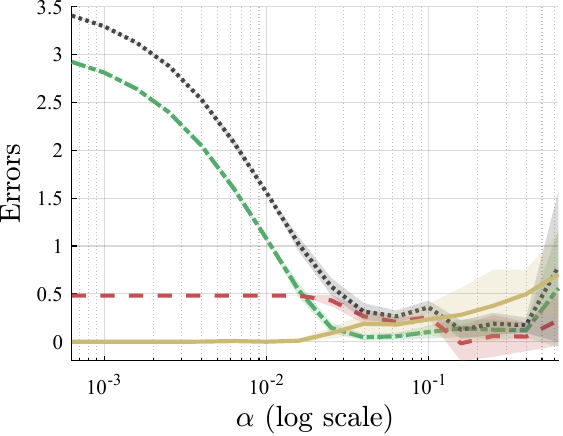}
  \vspace{-0.2cm}
 \caption{Different step size $\alpha$.}
 \label{fig:alpha_opt_gen}
\end{subfigure}
\begin{subfigure}[b]{0.3\textwidth}
 \centering
 \includegraphics[width=\textwidth]{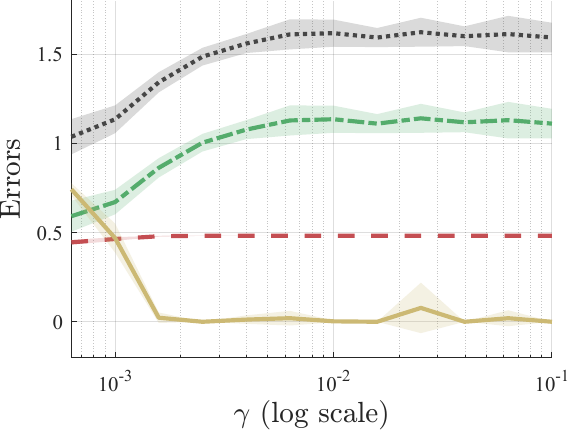}
  \vspace{-0.2cm}
 \caption{Different step size $\gamma$.}
 \label{fig:gamma_opt_gen}
\end{subfigure}
\caption{Optimization, generalization, and CA direction distances of MoDo in the strongly convex case under different $T,\alpha,\gamma$. The default  parameters are $T = 100$, $\alpha = 0.01$, $\gamma=0.001$.}
\label{fig:alpha_gamma_T_SC}
\end{figure*}

\begin{figure*}[t]
\centering
\begin{subfigure}[b]{0.31\textwidth}
 \centering
 \includegraphics[width=\textwidth]{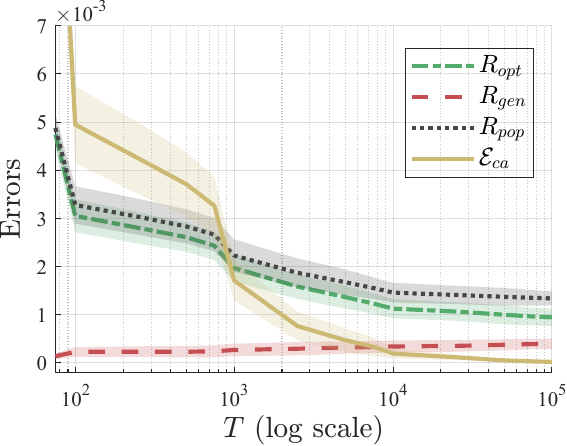}
  \vspace{-0.4cm}
 \caption{Number of iterations $T$}
 \label{fig:T-ablation}
\end{subfigure}
\begin{subfigure}[b]{0.3\textwidth}
 \centering
 \includegraphics[width=\textwidth]{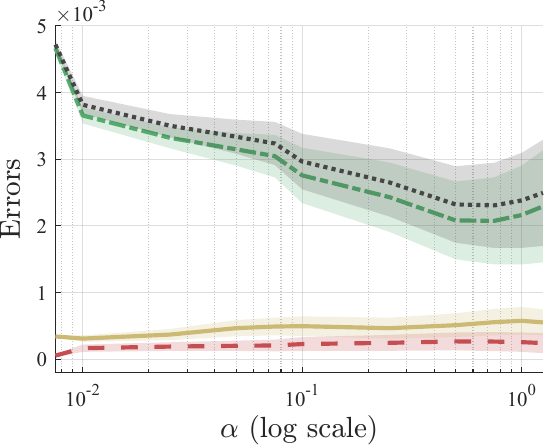}
  \vspace{-0.4cm}
 \caption{Different step size $\alpha$}
 \label{fig:lr-ablation}
\end{subfigure}
\begin{subfigure}[b]{0.3\textwidth}
 \centering
 \includegraphics[width=\textwidth]{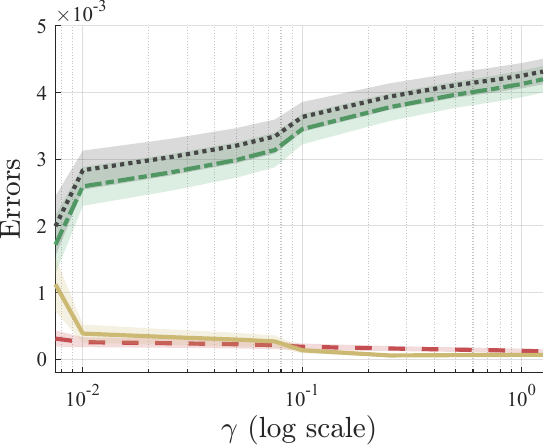}
  \vspace{-0.4cm}
 \caption{Different step size $\gamma$}
 \label{fig:gamma-ablation}
\end{subfigure}
\caption{Optimization, generalization, and CA direction distances of MoDo for MNIST image classification under different  $T$, $\alpha$, and $\gamma$.
The default parameters are $T=1000$, $\alpha=0.1$, and $\gamma=0.01$. }
\label{fig:T-alpha-gamma-ablation}
\end{figure*}

\section{Experiments}
 
In this section, we conduct experiments to further demonstrate the three-way trade-off among the optimization, generalization, and conflict avoidance of the MoDo algorithm on synthetic and various real MOL tasks.

\subsection{Synthetic experiments}

\subsubsection{Experiments on toy strongly-convex objectives}
 
Our theory in the strongly convex case is first verified through a synthetic experiment; see the details in Appendix~\ref{sub_app:synthetic_experiment}. Figure~\ref{fig:T_opt_gen} shows the PS optimization error and PS population risk, as well as the distance to CA direction, decreases as $T$ increases, which corroborates Lemma~\ref{lemma:converge_MGDA_bounded_grad}, and Theorem~\ref{thm:opt_err_modo_nonconvex_bounded_grad}. In addition, the generalization error, in this case, does not vary much with $T$, verifying Theorems~\ref{thm:grad_stability_modo_sc}.
In Figure~\ref{fig:alpha_opt_gen}, the optimization error first decreases and then increases as $\alpha$ increases, which is consistent with Theorem~\ref{thm:opt_err_modo_nonconvex_bounded_grad}.  Notably, we observe a threshold for $\alpha$ below which the distance to the CA direction converges even when the optimization error does not converge, while beyond which the distance to the CA direction becomes larger, verifying Lemma~\ref{lemma:converge_MGDA_bounded_grad}.
Additionally,
Figure~\ref{fig:gamma_opt_gen} demonstrates that increasing $\gamma$ enlarges the PS optimization error, PS generalization error, and thus the PS population risk, but decreases the distance to  CA direction, which supports Lemma~\ref{lemma:converge_MGDA_bounded_grad}.

\subsubsection{Experiments on toy nonconvex objectives}

\begin{figure*} 
\begin{subfigure}[b]{0.32\textwidth}
 \centering
 \includegraphics[width=0.9\textwidth]{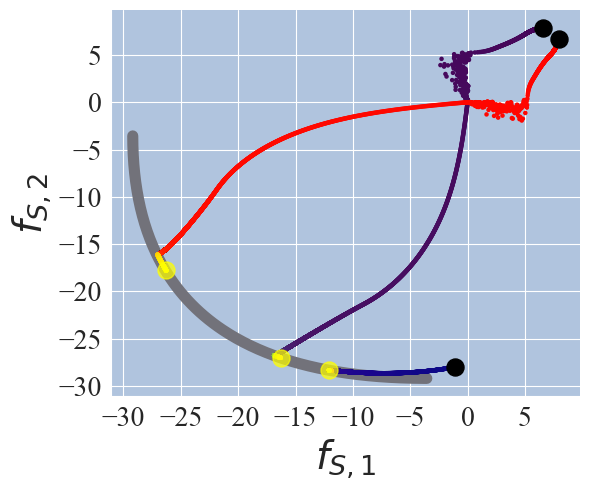}
\end{subfigure}
\begin{subfigure}[b]{0.32\textwidth}
 \centering
 \includegraphics[width=0.9\textwidth]{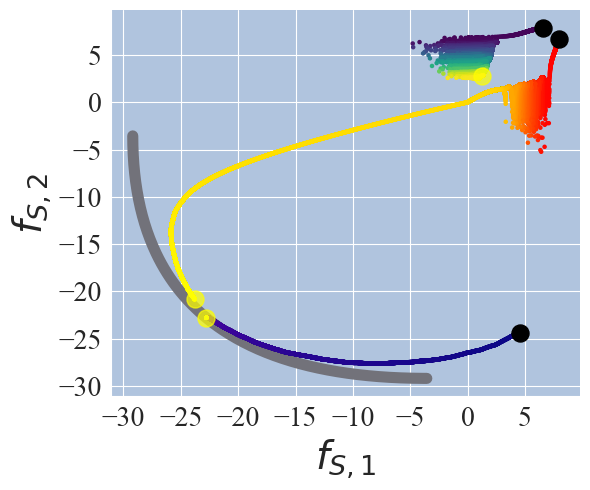}
\end{subfigure} 
\begin{subfigure}[b]{0.32\textwidth}
 \centering
 \includegraphics[width=0.9\textwidth]{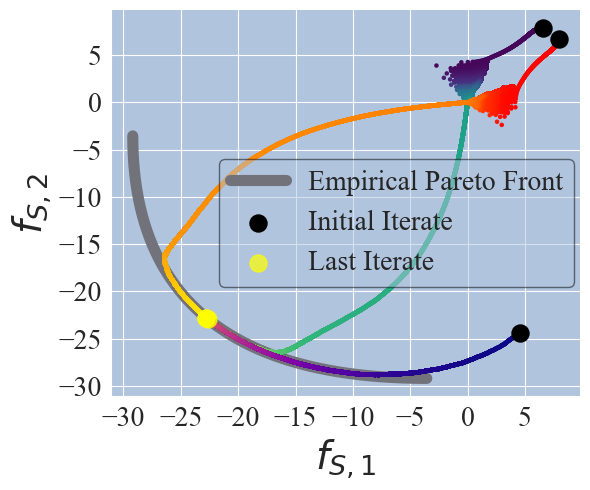}
\end{subfigure}
\begin{subfigure}[b]{0.32\textwidth}
 \centering
 \includegraphics[width=0.9\textwidth]{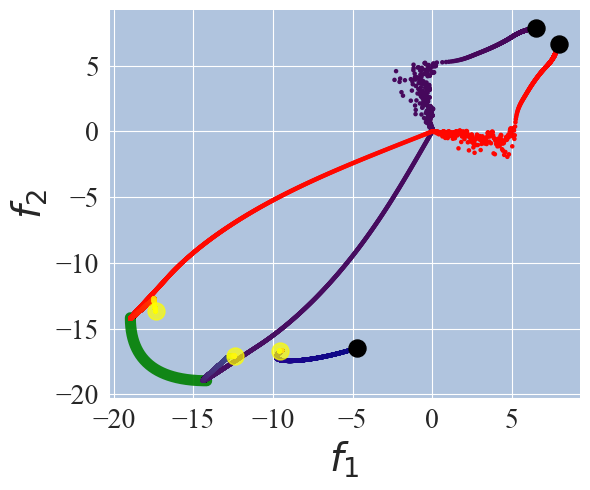}
 \caption{MGDA}
 \label{sfig:mgd-pop-os}
\end{subfigure}
\begin{subfigure}[b]{0.32\textwidth}
 \centering
 \includegraphics[width=0.9\textwidth]{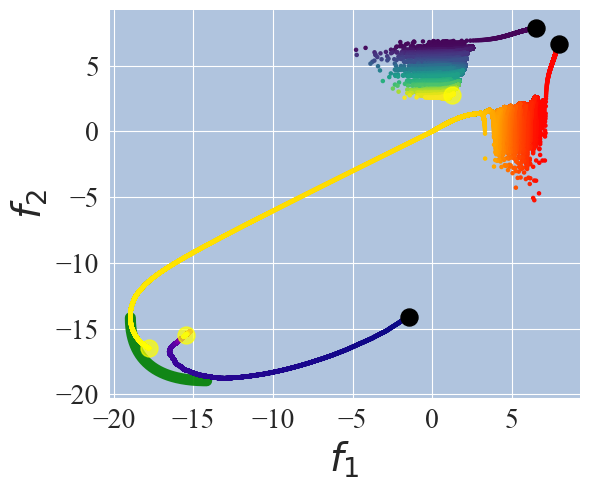}
 \caption{Static}
 \label{sfig:sgd-pop-os}
\end{subfigure}
\begin{subfigure}[b]{0.32\textwidth}
 \centering
 \includegraphics[width=0.9\textwidth]{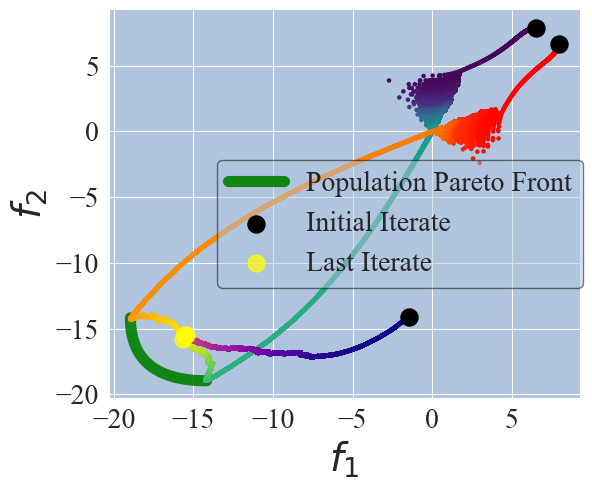}
 \caption{MoDo}
 \label{sfig:modo-pop-os}
\end{subfigure}
\caption{Convergence of MGDA, static weighting and MoDo to the \gray{\bf empirical (gray, upper)} and \green{\bf population (green, lower)} Pareto fronts.
The horizontal and vertical axes in the first/second row are the values of the two empirical/population objectives. 
Three colormaps are used for the trajectories from three initializations, respectively, where the same colormaps represent the trajectories of the same initializations, darker colors in one colormap indicate earlier iterations and lighter colors indicate later iterations.}
\label{fig:toy-pf-methods}
\vspace{-0.4cm}
\end{figure*}

Figure~\ref{fig:toy-pf-methods}, in addition to Figure~\ref{fig:toy-comp}, illustrates the trajectories of different methods to the empirical and population Pareto fronts (PF).
Figure~\ref{sfig:mgd-pop-os} shows that MGDA converges to the empirical PF regardless of the initializations, yet not all solutions exhibit low population risk due to the gap between empirical and population PFs, caused by finite stochastic training data.
In Figure~\ref{sfig:sgd-pop-os}, static weighting yields mixed results, with one trajectory converging to the center of the empirical PF and others getting stuck, oscillating around suboptimal parameters for a long time.
Despite lacking convergence to the CA direction to quickly go through the suboptimal valley, one static weighting solution achieves low population risk, cf. the trajectory with red to yellow colormap in the second row of Figure~\ref{sfig:sgd-pop-os}.
In Figure~\ref{sfig:modo-pop-os}, MoDo is able to converge to the empirical PF slightly slower than MGDA, and demonstrates good generalization with solutions attaining low population risks.

To explore the impact of different values of $\gamma$ on the performance of MoDo, we conducted experiments across various $\gamma$, observing its behavior in recovering static weighting ($\gamma=0$) and MGDA ($\gamma=10^{-2}$).
Results are visualized in Figure~\ref{fig:toy_modo_gamma_contour} with contours of objectives, and Figure~\ref{fig:toy_modo_gamma_PF} with empirical and population PFs.
Figures~\ref{sfig:modo_gamma_0} and~\ref{sfig:modo_PF_gamma0} verify that MoDo recovers static weighting with $\gamma=0$, where for two initializations, they get stuck in the valley. When convergence occurs, its solution aligns with the optimal solution of average empirical objectives.
In contrast, Figures~\ref{sfig:modo_gamma_1e-2} and~\ref{sfig:modo_PF_gamma1E-2} show that MoDo with $\gamma=10^{-2}$ resembles MGDA, navigating the valley without extended stagnation and converging to the empirical PF for all initializations. However, similar to MGDA, it ceases updates upon reaching the empirical PF, leading to suboptimal population risk for one solution with a trajectory colored from purple to pink to yellow.
We also conduct experiments on $0<\gamma< 10^{-2}$.
With $\gamma = 10^{-12}$, as shown in Figures~\ref{sfig:modo_gamma_1E-12}, and~\ref{sfig:modo_PF_gamma1E-12}, MoDo slowly traverses the valley, and ultimately converging to the optimal solution of the average of objectives.
While with $\gamma = 10^{-4}$, as shown in Figures~\ref{sfig:modo_gamma_1E-4}, and~\ref{sfig:modo_PF_gamma1E-4}, MoDo swiftly navigates the valley, stopping parameter updates upon reaching the empirical PF. 
The resulting solutions are closer to the average objective optimum compared to the $\gamma=10^{-2}$ scenario, owing to the smaller variations in the dynamic weight $\lambda$.

\begin{figure}[t]
\vspace{-0.15cm}
\begin{subfigure}[b]{0.245\textwidth}
 \centering
 \includegraphics[width=0.99\textwidth]{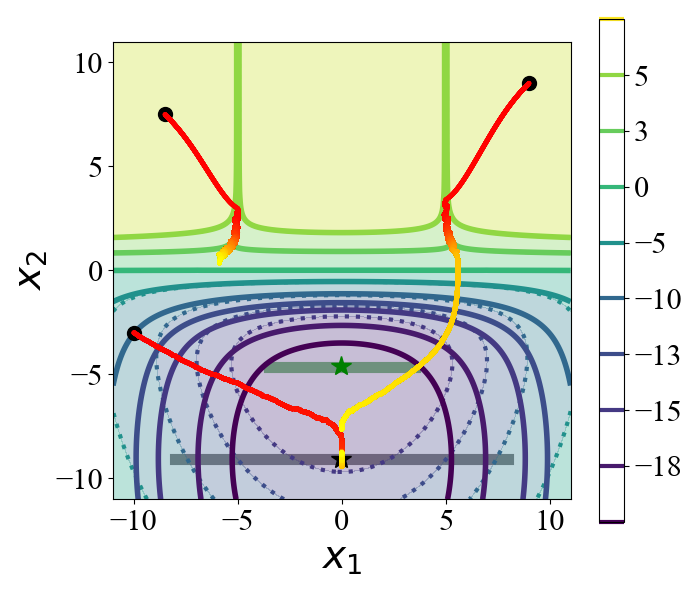}
 \caption{$\gamma=0$}
 \label{sfig:modo_gamma_0}
\end{subfigure}
\begin{subfigure}[b]{0.245\textwidth}
 \centering
 \includegraphics[width=0.99\textwidth]{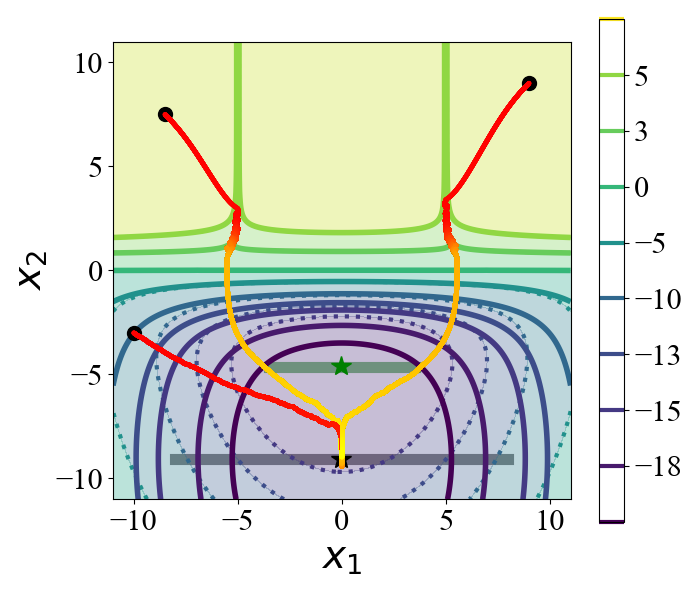}
 \caption{$\gamma=10^{-12}$}
 \label{sfig:modo_gamma_1E-12}
\end{subfigure} 
\begin{subfigure}[b]{0.245\textwidth}
 \centering
 \includegraphics[width=0.99\textwidth]{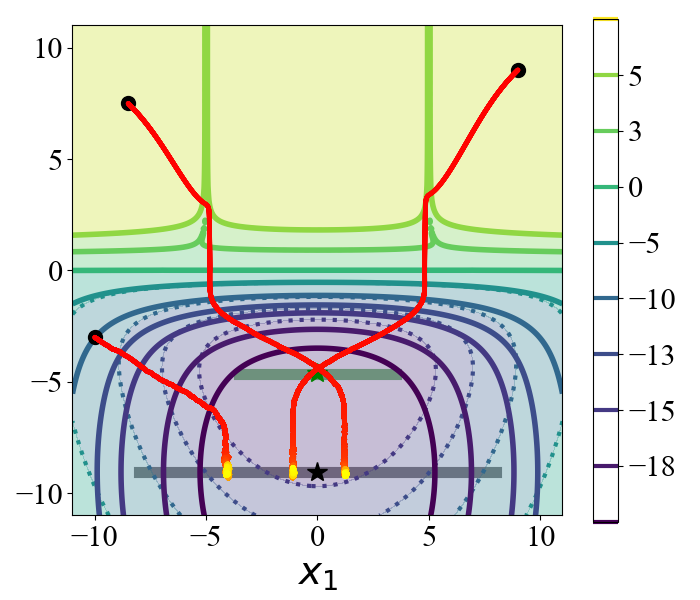}
 \caption{$\gamma=10^{-4}$}
 \label{sfig:modo_gamma_1E-4}
\end{subfigure} 
\begin{subfigure}[b]{0.245\textwidth}
 \centering
 \includegraphics[width=0.99\textwidth]{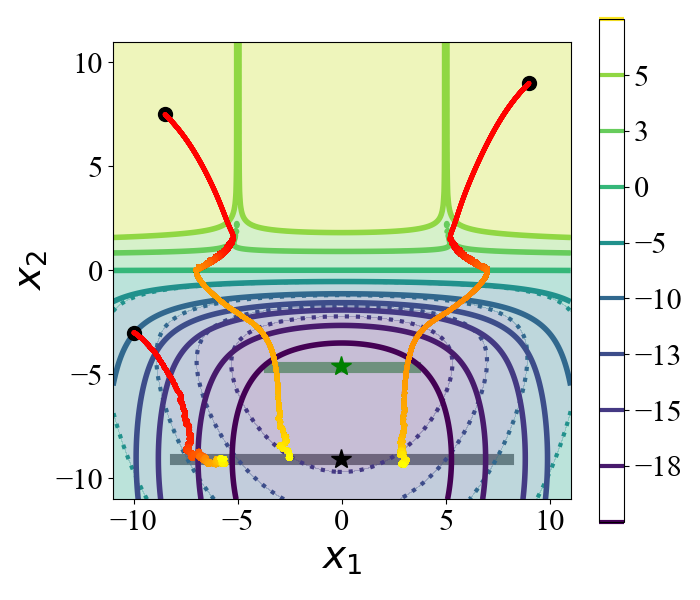}
 \caption{$\gamma=10^{-2}$}
 \label{sfig:modo_gamma_1e-2}
\end{subfigure}
\caption{Trajectories of MoDo under different $\gamma$ on the contour of the average of objectives.
The {\bf black} $\bullet$ marks initializations of the trajectories, colored from \red{\bf red} (start) to \yellow{\bf yellow} (end).
The background solid/dotted contours display the landscape of the average empirical/population objectives. 
The \gray{\bf gray}/\green{\bf green} bar marks empirical/population Pareto front, and the {\bf black $\star$}/\green{\bf green $\star$}  marks solution to the average objectives. }
\label{fig:toy_modo_gamma_contour}
\vspace{-0.4cm}
\end{figure}

\begin{figure}[ht]
\begin{subfigure}[b]{0.245\textwidth}
 \centering
 \includegraphics[width=0.99\textwidth]{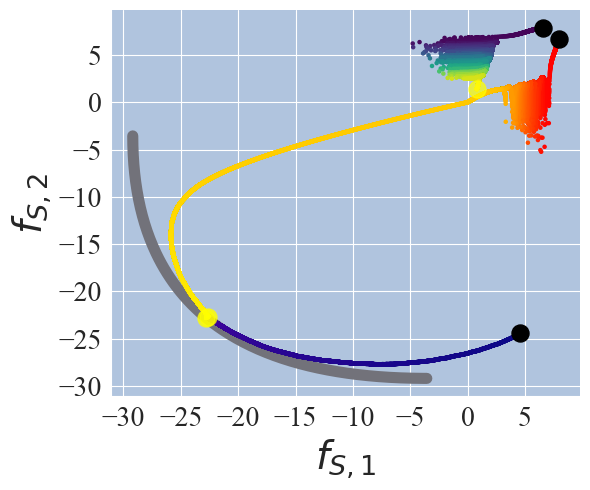}
\end{subfigure}
\begin{subfigure}[b]{0.245\textwidth}
 \centering
 \includegraphics[width=0.99\textwidth]{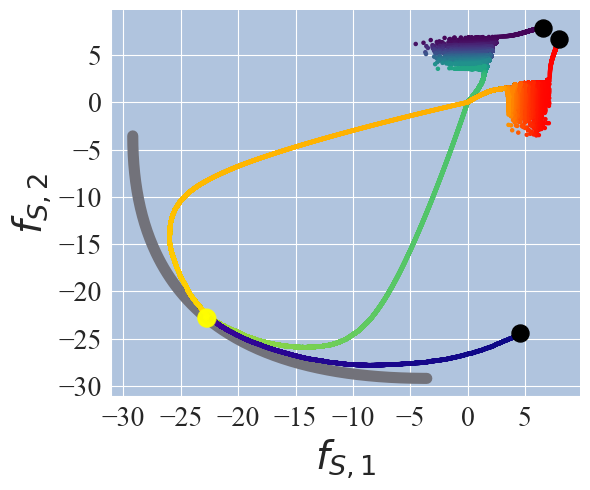}
\end{subfigure} 
\begin{subfigure}[b]{0.245\textwidth}
 \centering
 \includegraphics[width=0.99\textwidth]{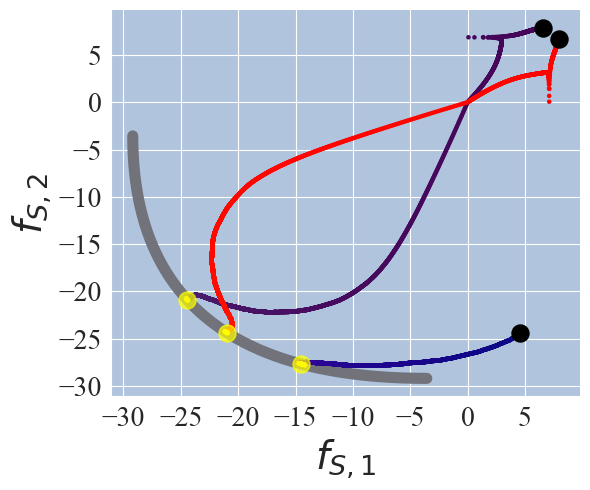}
\end{subfigure}
\begin{subfigure}[b]{0.245\textwidth}
 \centering
 \includegraphics[width=0.99\textwidth]{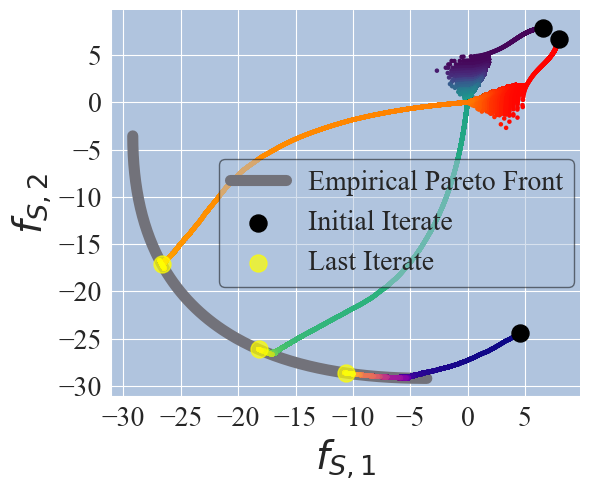}
\end{subfigure}
\begin{subfigure}[b]{0.245\textwidth}
 \centering
 \includegraphics[width=0.99\textwidth]{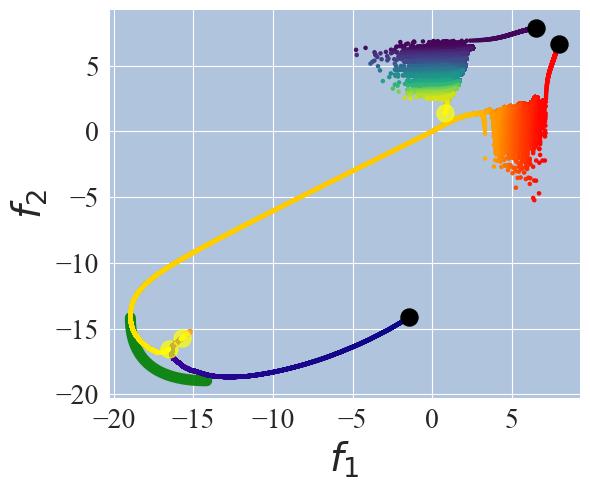}
 \caption{$\gamma=0$}
 \label{sfig:modo_PF_gamma0}
\end{subfigure}
\begin{subfigure}[b]{0.245\textwidth}
 \centering
 \includegraphics[width=0.99\textwidth]{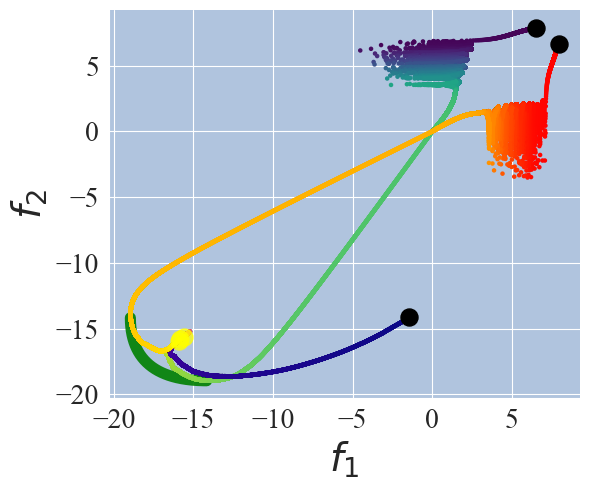}
 \caption{$\gamma=10^{-12}$}
 \label{sfig:modo_PF_gamma1E-12}
\end{subfigure}
\begin{subfigure}[b]{0.245\textwidth}
 \centering
 \includegraphics[width=0.99\textwidth]{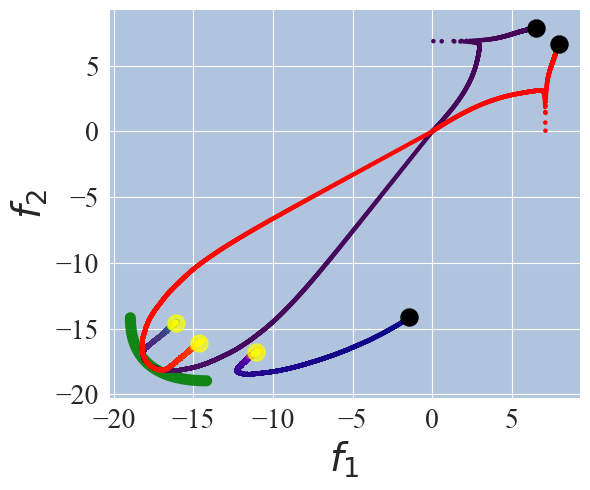}
 \caption{$\gamma=10^{-4}$}
 \label{sfig:modo_PF_gamma1E-4}
\end{subfigure}
\begin{subfigure}[b]{0.245\textwidth}
 \centering
 \includegraphics[width=0.99\textwidth]{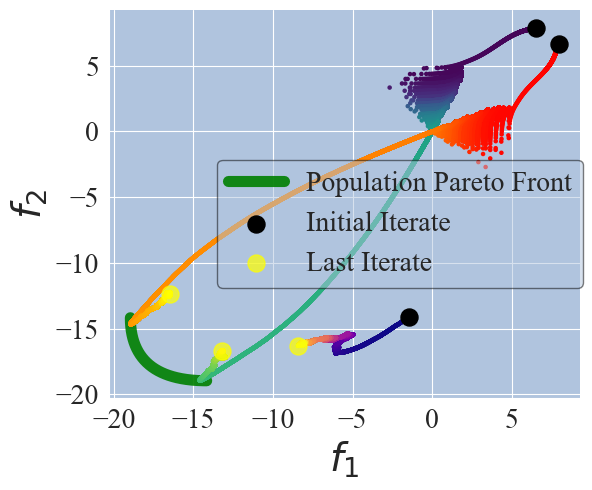}
 \caption{$\gamma=10^{-2}$}
 \label{sfig:modo_PF_gamma1E-2}
\end{subfigure}
\caption{Convergence of MoDo to the \gray{\bf empirical (gray, upper)} and \green{\bf population (green, lower)} Pareto fronts under different $\gamma$.
The horizontal and vertical axes in the first/second row are the values of the two empirical/population objectives. 
Three colormaps are used for the trajectories from three initializations, respectively, where the same colormaps represent the trajectories of the same initializations, darker colors in one colormap indicate earlier iterations and lighter colors indicate later iterations.}
\label{fig:toy_modo_gamma_PF}
\end{figure} 

\subsubsection{Multi-objective MNIST Experiments}

We further verify our theory in the NC case on MNIST image classification~\citep{lecun1998mnist} using a multi-layer perceptron (MLP) and three objectives: cross-entropy, mean squared error (MSE), and Huber loss. Implementation details are provided in Appendix~\ref{sub_app:image_classification_experiment}.
Following Section~\ref{sub:measure_decompose}, we evaluate the performance in terms of $R_{\rm pop}(x)$, $R_{\rm opt}(x)$, $R_{\rm gen}(x)$, and $\mathcal{E}_{\rm ca}(x,\lambda)$.
The exact PS population risk $R_{\text{pop}}(x)$ is not accessible without the true data distribution.
To estimate the PS  population risk, we evaluate $\min _{{\lambda}\in \Delta^{M}}\|\nabla F_{S_{\text{te}}}({x}) \lambda\|$ on the testing data set $S_{\text{te}}$ that is independent of training data set $S$. 
The PS optimization error $R_{\text{opt}}(x)$ is obtained by $\min _{{\lambda}\in \Delta^{M}}\|\nabla F_{S}({x}) \lambda\|$, and the {PS generalization error $R_{\text{gen}}(x)$} is estimated by $\min _{{\lambda}\in \Delta^{M}}\|\nabla F_{S_{\text{te}}}({x}) \lambda\| - R_{\text{opt}}(x)$.

We examine the impact of different choices of $T$, $\alpha$, and $\gamma$ on the three-way trade-off in Figure~\ref{fig:T-alpha-gamma-ablation} (averaged over 10 random seeds with 0.5 standard deviations as error bars). 
Figure~\ref{fig:T-ablation} shows that
increasing $T$ reduces optimization error and CA direction distance but increases generalization error, aligning with optimization error bound in Theorem~\ref{thm:opt_err_modo_nonconvex_bounded_grad}, the CA direction distance bound in Theorem~\ref{lemma:converge_MGDA_bounded_grad}, and the generalization bound in Theorem~\ref{crlr:bound_stab_grad_MGDA}. 
Figure~\ref{fig:lr-ablation} shows that increasing $\alpha$ leads to an initial decrease and subsequent increase in PS optimization error and population risk. which aligns with Theorem~\ref{thm:opt_err_modo_nonconvex_bounded_grad} and~\eqref{eq:bound_nc_pop}. On the other hand, there is an overall increase in CA direction distance with $\alpha$, which aligns with Theorem~\ref{lemma:converge_MGDA_bounded_grad}.
Figure~\ref{fig:gamma-ablation} shows that increasing $\gamma$ increases both the PS population and optimization errors but decreases CA direction distance.
This matches our bounds for PS optimization error in Theorem~\ref{thm:opt_err_modo_nonconvex_bounded_grad}, PS population risk in~\eqref{eq:bound_nc_pop}, and CA direction distance in Theorem~\ref{lemma:converge_MGDA_bounded_grad}. 

\begin{table}[ht]
\caption{MNIST classification with cross-entropy, MSE, and Huber loss as objectives.}
\label{tab:toy-mnist-results}
\fontsize{9}{10}\selectfont
\centering
\setlength{\tabcolsep}{0.3em} 
{\renewcommand{\arraystretch}{1.4}
\begin{tabular}{c c c c c c c}
\toprule
 \multirow{2}{*}{Method} & Cross-entropy & MSE loss & Huber loss & \multirow{2}{*}{$R_{\text{pop}}$ ($\num{e-3}$) $\downarrow$} & \multirow{2}{*}{$R_{\text{opt} }$ ($\num{e-3}$) $\downarrow$}& \multirow{2}{*}{$|R_{\text{gen}}|$ ($\num{e-3}$) $\downarrow$}\\
 \cline{2-4}
 & Loss ($\num{e-3}$) $\downarrow$ & Loss ($\num{e-3}$) $\downarrow$ & Loss ($\num{e-3}$) $\downarrow$  \\ 
 \midrule
 Static & \textbf{306.9$\pm$3.9} & 13.2$\pm$0.14 & 2.2$\pm$0.03 
 & 2.1$\pm$0.56 & 1.9$\pm$0.5 & 0.2$\pm$0.19\\
 MGDA & 363.6$\pm$4.1 & 13.5$\pm$0.13 & \textbf{1.9$\pm$0.01} 
 & \textbf{1.3$\pm$0.24} & \textbf{1.1$\pm$0.2} & 0.2$\pm$0.13\\
 \textbf{MoDo}  & 317.9$\pm$3.4 & \textbf{13.1$\pm$0.13} & 2.1$\pm$0.05
 & 2.1$\pm$0.38 & 1.9$\pm$0.4 & \textbf{0.1$\pm$0.09}
 \\ \bottomrule
\end{tabular}}
\end{table}
We then discuss the results in Table~\ref{tab:toy-mnist-results}.
Experiments are repeated 10 times with average performance and standard deviations reported.
Observed from Table~\ref{tab:toy-mnist-results}, 
for cross-entropy loss, MGDA performs the worst while static weighting performs the best.
On the other hand, for Huber loss, MGDA performs the best while static weighting performs the worst. 
This result aligns with the scales of the loss functions, with static weighting with equal weights favoring the largest-scale cross-entropy loss and MGDA favoring the smallest-scale Huber loss.
Compared to MGDA, MoDo achieves much better performance on cross-entropy loss, and comparable performance on Huber loss.
GDA has the lowest PS population risk and PS optimization error, likely due to its strong performance on Huber loss and similarly small generalization errors across methods in this setting. 
Overall, MoDo demonstrates a balance by combining static weighting and MGDA's advantages to perform well across objectives with varying scales.

\subsection{Multi-task supervised learning experiments}

In this section, we conduct a comparative analysis between MoDo, MGDA, and static weighting on real-world supervised multi-task learning benchmarks. In these experiments, we go beyond individual task (test) performance and introduce an overarching performance metric that takes into account all tasks. This metric is denoted as $\Delta \mathcal{A}_{\rm id} \%$, which captures the average performance degradation of an MOL algorithm compared to dedicated single-task learners. Formally, this metric is expressed as follows
 \begin{equation}
 \Delta \mathcal{A}_{\rm id} \%\!=\! \frac{1}{M}\sum_{m=1}^{M}(-1)^{\ell_m}\left(S_{\mathcal{A},m} - S_{\mathcal{B}, m}\right)\!/\!S_{\mathcal{B}, m} \times 100,
 \end{equation}
 where $M$ is the number of tasks, $S_{\mathcal{A}, m}$ is the performance of method $\mathcal{A}$ for task $m$, and $S_{\mathcal{B}, m}$ is the performance of the independent task learner dedicated to task $m$. Here, $\ell_m\!=\!1$ if higher values for $S_{\mathcal{A}, m}$ are better and $0$ otherwise. Additional experiments and details are provided in Appendix~\ref{sub_app:image_classification_experiment}.
Hyperparameters of all methods are chosen based on the best validation performance. 

\subsubsection{Office-31 dataset}

The experiments of multi-task image classification
on Office-31 multi-domain dataset~\citep{saenko2010adapting} with 3 objectives are summarized in Table~\ref{tab:office-31-results}.
Observed from Tables~\ref{tab:office-31-results}, Static weighting outperforms MGDA  in some datasets as well as in terms of $\Delta \mathcal{A}_{\rm id}  \%$. As an interpolation of the two methods, by the correct choice of hyper-parameters, MoDo performs on par or better compared to both Static and MGDA, and hence achieves the best overall performance in terms of $\Delta \mathcal{A}_{\rm id} \%$.

\begin{table}[H]
\fontsize{9}{10}\selectfont
\caption{Classification results on Office-31 dataset. }
\centering
\setlength{\tabcolsep}{0.25em} 
{\renewcommand{\arraystretch}{1.3}
\begin{tabular}{c c c c c}
\toprule
 \multirow{2}{*}{Method} & Amazon & DSLR & Webcam
 & \multirow{2}{*}{$\Delta \mathcal{A}_{\rm id} \% \downarrow$} \\ 
 \cline{2-4}
 & Test Acc $\uparrow$ &  Test Acc $\uparrow$ & Test Acc $\uparrow$  \\ 
 \midrule
 Static & 84.62 {$\pm$ 0.71}  & 94.43 $\pm$ 0.96 & 97.44 $\pm$ 1.20 & 2.56 $\pm$ 0.37\\
 MGDA & 79.45 $\pm$ 0.11 & \textbf{96.56 $\pm$ 1.20} & \textbf{97.89 $\pm$ 0.74} & 3.65 $\pm$ 0.64\\
\textbf{MoDo}  & \textbf{85.13 $\pm$ 0.58} & 95.41 $\pm$ 0.98 & 96.78 $\pm$ 0.65 & \textbf{2.26 $\pm$ 0.31}\\ 
 \bottomrule
\end{tabular}}
\vspace{5 pt}
\label{tab:office-31-results}
\end{table}

\subsubsection{Office-home dataset}

Next, we report the experimental results in multi-task image classification on Office-home multi-domain dataset~\citep{venkateswara2017deep}  with 3 objectives in Table~\ref{tab:office-home-results}.
Similar to the results on Office-31, MoDo performs comparable or better compared to MGDA and static weighting considering per task metrics, and achieves the best performance in  $\Delta \mathcal{A}_{\rm id} \%$ out of the three methods.
 
\begin{table}[H]
\caption{Classification results on Office-home dataset.}
\label{tab:office-home-results}
\fontsize{9}{10}\selectfont
\centering
\setlength{\tabcolsep}{0.5em} 
{\renewcommand{\arraystretch}{1.3}
\begin{tabular}{c c c c c c}
\toprule
 \multirow{2}{*}{Method} & Art & Clipart & Product & Real-world
 & \multirow{2}{*}{$\Delta \mathcal{A}_{\rm id} \% \downarrow$} \\ 
 \cline{2-5}
 & Test Acc $\uparrow$ &  Test Acc $\uparrow$ & Test Acc $\uparrow$ & Test Acc $\uparrow$ \\ 
 \midrule
 Static &  64.14 $\pm$ 1.40  & \textbf{79.57} $\pm$ 1.09 & 90.00 $\pm$ 0.50 & 78.94 $\pm$ 0.87 & 2.85 $\pm$ 1.08 \\
 MGDA &  61.71 $\pm$ 1.33 & 73.95 $\pm$ 0.43 & \textbf{90.17 $\pm$ 0.27} & 79.35 $\pm$ 1.15  & 5.29  $\pm$ 0.47 \\
\textbf{MoDo}  & \textbf{65.50 $\pm$ 0.55} & 79.44 $\pm$ 0.29 & 89.72 $\pm$ 0.94  & \textbf{79.65 $\pm$ 0.67} & \textbf{2.24 $\pm$ 0.48} \\ 
 \bottomrule
\end{tabular}}
\end{table}

\subsubsection{NYU-v2 dataset}

Finally, we present the experiments on NYU-v2 dataset~\citep{silberman2012indoor}, which consists of image segmentation, depth estimation, and surface normal estimation tasks. Unlike image classification on Office-31 and Office-home datasets, this is a single-input multi-output problem. The dataset consists of images from video sequences of indoors. It can be seen from Table~\ref{tab:nyu-v2-results} that MoDo outperforms both MGDA and Static methods in some tasks and also in terms of $\Delta \mathcal{A}_{\rm id} \%$. The poor performance of MGDA in terms of $\Delta \mathcal{A}_{\rm id} \%$ is due to the large preference towards the Surface normal estimation task, which negatively affects its performance on other tasks.

\begin{table}[H]
\tiny
\caption{Segmentation, depth, and surface normal estimation results on NYU-v2 dataset.}
\label{tab:nyu-v2-results}
\hspace{-5mm}
\setlength{\tabcolsep}{0.3em} 
{\renewcommand{\arraystretch}{1.4}
\begin{tabular}{c c c c c  c  c c c c c}
\toprule
\multirow{3}{*}{Method} & \multicolumn{2}{c}{Segmentation} & \multicolumn{2}{c}{Depth ($\times 10^{-1}$)} & \multicolumn{5}{c}{Surface Normal} & \multirow{3}{*}{$\Delta \mathcal{A}_{\rm id}  \% \downarrow$} \\ \cline{2-3}\cline{4-5}\cline{6-10}
& \multicolumn{2}{c}{(Higher Better)} & \multicolumn{2}{c}{(Lower Better)} & \multicolumn{2}{c}{\makecell{Angle Distance \\ (Lower Better)}} & \multicolumn{3}{c}{\makecell{Within $t^\circ$ \\ (Higher better)}}\\
& mIoU & Pix Acc & Abs Err & Rel Err & Mean & Median & 11.25 & 22.5 & 30\\ \hline
Static & 52.02{$\pm$0.69} & 74.21{$\pm$0.57}   & \textbf{3.98$\pm$0.03} & \textbf{1.64$\pm$0.01} & 23.79$\pm$0.10 & 17.44$\pm$0.15 & 34.07$\pm$0.17 & 60.17$\pm$0.31  & 71.48$\pm$0.29 &  3.98$\pm$0.70  \\
MGDA   & 46.39$\pm$0.17 & 70.27$\pm$0.24 & 4.27$\pm$0.02 & 1.74$\pm$0.01 & \textbf{22.34$\pm$0.03} & \textbf{15.70$\pm$0.08} & \textbf{37.71$\pm$0.21} & \textbf{63.96$\pm$0.11} & \textbf{74.50$\pm$0.06} & 6.25$\pm$0.38\\
\textbf{MoDo} & \textbf{52.64$\pm$0.19}  & \textbf{74.67$\pm$0.08}  & \textbf{3.98$\pm$0.02} & 1.65$\pm$0.02 & 23.45$\pm$0.06  & 17.09$\pm$0.05  & 34.79$\pm$0.11  & 60.90$\pm$0.13 & 72.12$\pm$0.11 & \textbf{3.21$\pm$0.34 }\\ 
\bottomrule
\end{tabular}}
\end{table}

\section{Conclusions, Limitations, and Future Work}
 
This work studied the three-way trade-off in MOL -- among optimization, generalization, and conflict avoidance.
Our results showed that, in the general nonconvex setting, the well-known trade-off between optimization and generalization controlled by the number of iterations also exists in MOL. Moreover, dynamic weighting algorithms like MoDo introduced a new dimension of trade-off in terms of conflict avoidance compared to static weighting. 
We demonstrated that this three-way trade-off can be controlled by the step size $\gamma$ for updating the dynamic weighting parameter $\lambda$ and the number of iterations $T$. Proper choice of these parameters led to decent performances on all three metrics. 
We also demonstrated the power of this new analytical framework by applying it to analyze SMG and MoCo. 

\paragraph{Limitations.}
One limitation of the proposed theoretical framework is that it focuses on specific algorithms, and stochastic variants of MGDA, including MoDo, MoCo, and SMG, for smooth objectives in unconstrained learning. 
Future research could explore the theory of other MOL algorithms for non-smooth objectives or constrained learning.
Another limitation of the proposed MoDo algorithm is that it requires computing two stochastic gradients per iteration, resulting in a higher per-iteration complexity.
Nevertheless, there is still room for improvement in its complexity and the corresponding trade-off by adopting variance reduction techniques or implementing objective sampling, 
which we leave for future work.

\paragraph{Future work.}
Our work has broad implications in advancing both the theory and practice of multi-objective optimization, with potential future applications as follows.
\emph{1) Theoretical Applications:}
Our theoretical framework extends its utility beyond our proposed MoDo algorithm, allowing analysis of various  MOL algorithms like CAGrad and PCGrad. 
Additionally, it aids in the investigation of the advantages of MOL algorithms over static weighting in reducing CA distance. This validates their use when CA distance reduction is crucial.
\emph{2) Practical Applications:}
Our theory is crucial for optimizing hyperparameters (e.g., step size, iterations) to minimize testing risks effectively. It also enables informed algorithm selection based on the trade-off among three errors. Lastly, our theory may inspire the development of MOL algorithms that balance these errors more effectively.


\acks{The work of L. Chen, H. Fernando, and T. Chen was supported by the National Science Foundation (NSF) MoDL-SCALE project   2134168 and the RPI-IBM Artificial Intelligence Research Collaboration (AIRC).
The work of Y. Ying was partially supported by NSF (DMS-2110836, IIS-2103450, and IIS-2110546).}

\newpage

\setcounter{secnumdepth}{3}
\setcounter{tocdepth}{3}
\mtcsetdepth{secttoc}{3}

\appendix
\begin{center}
\setstretch{1.5}
{\Large \bf Appendix}
\end{center}

\vspace{-1cm}
\addcontentsline{toc}{section}{} 
\part{} 
\parttoc 
\secttoc

\allowdisplaybreaks

\newpage
\section{Notations}


A summary of notations used in this work is listed in Table~\ref{tab:notations} for ease of reference.
Throughout the paper, we use $\|\cdot\|$ to denote the spectral norm, and $\|\cdot\|_{\rm F}$ to denote the Frobenius norm.
\begin{table}[ht]
\caption{Notations and their descriptions.}
  \label{tab:notations}
  \small
  \centering
  \begin{tabular}{l|l l }
  \toprule
  Notations & Descriptions   \\
  \midrule
  $x \in \mathbb{R}^{d}$ &Model parameter, or decision variable &   \\
  $z\in \mathcal{Z}$ &Data point for training or testing &   \\
  $S\in \mathcal{Z}^n$ &Dataset such that $S = \{z_1,\dots, z_n\}$&   \\
  \makecell[l]{$f_{z,m}(x)$, $f_{S,m}(x)$ \\
  } &
  \makecell[l]{A scalar-valued objective function evaluated on data point $z$,\\ with $f_{z,m}: \mathbb{R}^d \mapsto \mathbb{R}$, 
  or on dataset $S$, $f_{S,m}$, 
  with $f_{S,m} \coloneqq \frac{1}{|S|} \sum_{z\in S} f_{z,m}(x) $} \\
  $f_{m}(x)$ &A scalar-valued population objective function, $f_{m}(x) \coloneqq \E_z [f_{z,m}(x)] $ \\
  $\nabla f_{m}(x)$ &Gradient of $f_{m}(x)$, with 
  $\nabla f_{m}(x): \mathbb{R}^d \mapsto \mathbb{R}^d$ \\
  \makecell[l]{$F_{z}(x)$, $F_{S}(x)$ \\
  } &
  \makecell[l]{A vector-valued objective function evaluated on data point $z$,\\ with $F_{z}: \mathbb{R}^d \mapsto \mathbb{R}^M$,
  or on dataset $S$, with
  $F_{S} \coloneqq \frac{1}{|S|} \sum_{z\in S} F_{z}(x) $ 
  } \\
  $F(x)$ & A vector-valued population objective, $F(x) \coloneqq \E_z [f_{z,m}(x)] $ \\
  $\nabla F(x)$ &Gradient of $F(x)$, with 
  $\nabla F(x): \mathbb{R}^d \mapsto \mathbb{R}^{d \times M}$ \\
  $\lambda \in \Delta^{M} $ &Weighting parameter in an $(M-1)$-simplex  &\\
  $\lambda^*_\rho(x) \in \Delta^{M} $ &CA weight, optimal solution to~\eqref{eq:ca_direction_problem}, when $\rho=0$, it is  simplified as $\lambda^*(x) $  \\
  $\mathbf{1} \in \mathbb{R}^{M} $ &All-one vector with dimension $M$  &\\
  \hline 
  $\alpha$ &Step size to update model parameter $x$\\
  $\beta$ & Moving average coefficient for MoCo\\
  $\gamma$ &Step size to update weight $\lambda$\\
  $\rho$ &Regularization parameter in~\eqref{eq:ca_direction_problem} \\
  \bottomrule
  \end{tabular}
  \vspace{0.2cm}
\end{table}



\section{Bounding the PS Generalization Error}
\label{sec:appdx_gen_err}

\subsection{Proof of Proposition~\ref{prop:stability_gen_grad_F}} 
\label{sub_app:proof_stability_gen_grad_F}

\begin{proof}
\label{proof:stability_gen_grad_F}
The proof extends that of~\citep{lei2022stability_NC_nonsmooth} for single objective learning to our MOL setting.
Recall that $S=\{z_1, \ldots, z_n \}$,  which are drawn i.i.d. from the data distribution $\cal D$. 
Define the perturbed dataset $S^{(i)}=\{z_1, \ldots, z_i', \ldots, z_n \}$ sampled i.i.d. from $\cal D$ with $z_i'$ independent of $z_j$, for all $i,j\in [n]$. 
Let $\tdz$ be an independent sample of $z_j, z_j'$ , for all $j\in [n]$, and  from the same distribution $\cal D$.
We first decompose the difference of population gradient and empirical gradient on the algorithm output $n(\nabla F(A(S))-\nabla F_S(A(S)))$ as follows using the gradient on $A(S^{(i)})$.
Since $\mathbb{E}_{\tdz} [\nabla F_{\tdz} (A(S) )]=\nabla F(A(S))$, it holds that
\begin{align*}
  & n\big(\nabla F(A(S))-\nabla F_S(A(S))\big) 
  = n\E_{\tdz} [\nabla F_{\tdz} (A(S) )]- n\nabla F_S(A(S)) \\
  =& n \E_{\tdz} [\nabla F_{\tdz} (A(S) )]- \Bigg(\sum_{i=1}^n \nabla F_{z_i}(A(S)) \Bigg) 
  + \sum_{i=1}^n \Big(\E_{z_i'}[\nabla F(A(S^{(i)}))] 
  - \E_{z_i'}[\nabla F(A(S^{(i)}))] \Big) \\
  &+ \sum_{i=1}^n \Big(\E_{z_i'}[\nabla F_{z_i}(A(S^{(i)}))] 
  - \E_{z_i'}[\nabla F_{z_i}(A(S^{(i)}))] \Big) \\
  = & \sum_{i=1}^n \mathbb{E}_{\tdz, z_i^{\prime}}
  [\nabla F_{\tdz} (A(S) )- 
  \nabla F_{\tdz}(A(S^{(i)}) ) ] 
  +
   \sum_{i=1}^{n}
   \underbrace{\mathbb{E}_{z_i^{\prime}}[\mathbb{E}_{\tdz}[\nabla F_{\tdz}(A(S^{(i)}) )]-\nabla F_{z_i}(A(S^{(i)}) )]}_{\xi_i(S)} \\
  &+
   \sum_{i=1}^n \mathbb{E}_{z_i^{\prime}}\big[\nabla F_{z_i}(A(S^{(i)}) )-\nabla F_{z_i}(A(S) )\big]
\numberthis
\label{eq:stab_gen_grad_decompose}
\end{align*}
where the last equality follows from rearranging and that $z_i,z_i',\tdz$ are mutually independent.
Applying triangle inequality to~\eqref{eq:stab_gen_grad_decompose}, it then follows that
\begin{align*}
 n\|\nabla F(A(S))-\nabla F_S(A(S))\|_{\rm F} 
\leq 
& \sum_{i=1}^n \mathbb{E}_{\tdz, z_i^{\prime}} {[\|\nabla F_{\tdz}(A(S) )-\nabla F_{\tdz}(A(S^{(i)}) )\|_{\rm F}]+ \Big\|\sum_{i=1}^n \xi_i(S)\Big\|_{\rm F} } \\
&+\sum_{i=1}^n \mathbb{E}_{z_i^{\prime}}[\|\nabla F_{z_i}(A(S^{(i)}) )-\nabla F_{z_i}(A(S) )\|_{\rm F}] .
\numberthis
\end{align*}
Note $S$ and $S^{(i)}$ differ by a single sample. By Definition~\ref{def:stab_grad}, the MOL uniform stability $\epsilon_{\rm F}$, and Jensen's inequality, we further get
\begin{align}\label{eq:grad_diff_norm_xi}
  n \mathbb{E}\left[\left\|\nabla F(A(S))-\nabla F_S(A(S))\right\|_{\rm F}\right] \leq 2 n \epsilon_{\rm F}
  +\mathbb{E}\Big[\Big\|\sum_{i=1}^n \xi_i(S)\Big\|_{\rm F} \Big].
\end{align}
We then proceed to bound $\mathbb{E}\big[\big\|\sum_{i=1}^n \xi_i(S)\big\|_{\rm F} \big]$, which satisfies
\begin{align}
  \Bigg( \mathbb{E}\Big[\Big\|\sum_{i=1}^n \xi_i(S)\Big\|_{\rm F} \Big] \Bigg)^2
  \leq \mathbb{E}\Big[\Big\|\sum_{i=1}^n \xi_i(S)\Big\|_{\rm F}^2 \Big]
  = \sum_{i=1}^n \underbrace{\mathbb{E}\big[\|\xi_i(S)\|_{\rm F}^2 \big]}_{J_{1,i}}
  +\sum_{i, j \in[n]: i \neq j} 
  \underbrace{\mathbb{E}[\langle\xi_i(S), \xi_j(S)\rangle]}_{J_{2,i,j}}.
  \label{eq:bound_xi_J1_J2}
\end{align}
For $J_{1,i}$, according to the definition of $\xi_i(S)$ in~\eqref{eq:stab_gen_grad_decompose} and Jensen inequality, it holds that
\begin{align*}
 J_{1,i} = \mathbb{E}[\|\xi_i(S)\|_{\rm F}^2] 
& =\mathbb{E}\left[\big\|\mathbb{E}_{z_i^{\prime}}\big[\mathbb{E}_{\tdz}[\nabla F_{\tdz}(A(S^{(i)}) )]-\nabla F_{z_i}(A(S^{(i)}) )\big]\big\|_{\rm F}^2\right] \\
& \stackrel{(a)}{\leq} \mathbb{E}\left[\big\|\mathbb{E}_{\tdz}[\nabla F_{\tdz}(A(S^{(i)}) )]-\nabla F_{z_i}(A(S^{(i)}) )\big\|_{\rm F}^2\right] \\
& \stackrel{(b)}{=} \mathbb{E}\left[\big\|\mathbb{E}_{\tdz}[\nabla F_{\tdz}(A(S) )]-\nabla F_{z_i^{\prime}}(A(S) )\big\|_{\rm F}^2\right] \\
& =\mathbb{E} \left[\mathbb{V}_{\tdz}(\nabla F_{\tdz}(A(S) ))\right],
\numberthis\label{eq:stab_gen_xi_i_square_bound_J1i}
\end{align*}
where $(a)$ follows from Jensen's inequality, $(b)$ follows from the symmetry between $z_i$ and $z_i^{\prime}$.
To bound $J_{2,i,j}$ with $i \neq j$, further introduce
 $S^{\prime \prime}=\left\{z_1^{\prime \prime}, \ldots, z_n^{\prime \prime}\right\}$ which are drawn  i.i.d. from the data distribution $\cal D$. Then for each $i, j \in[n]$ with $i \neq j$, introduce $S_j$ as a neighboring dataset of $S$ by replacing its $z_j$ with $z_j^{\prime \prime}$, and $S_j^{(i)}$ as a neighboring dataset of $S^{(i)}$ by replacing its $z_j$ with $z_j^{\prime \prime}$, i.e.,
\begin{subequations}
\begin{align}
S_j 
&=\{z_1, \ldots, z_{j-1}, z_j^{\prime \prime}, z_{j+1}, \ldots, z_n\}, \\
S_j^{(i)} 
&=\{z_1, \ldots, z_{i-1}, z_i^{\prime}, z_{i+1}, \ldots, z_{j-1}, z_j^{\prime \prime}, z_{j+1}, \ldots, z_n \} .
\end{align}    
\end{subequations}
Then the idea is to bound $J_{2,i,j}$ using the newly introduced neighboring datasets $S_j$ and $S_j^{(i)}$, so as to connect to the definition of the stability $\epsilon_{\rm F}$.
We first show that $\mathbb{E}\left[\left\langle\xi_i(S), \xi_j(S)\right\rangle\right]
= \mathbb{E}\left[\left\langle\xi_i(S)-\xi_i(S_j), \xi_j(S)-\xi_j(S_i)\right\rangle\right]$ because for $i \neq j$,
\begin{align}
\mathbb{E}\left[\left\langle\xi_i(S_j), \xi_j(S)\right\rangle\right] & \stackrel{(c)}{=} 0,  
~~
\mathbb{E}\left[\left\langle\xi_i(S_j), \xi_j(S_i)\right\rangle\right] \stackrel{(d)}{=} 0, 
~~
\mathbb{E}\left[\left\langle\xi_i(S_j), \xi_j(S_i)\right\rangle\right] \stackrel{(e)}{=} 0.
\label{eq:inner_xi_i_xi_j_Sj_0}
\end{align}  
For $i \neq j$, $(c)$ follows from
\begin{align}
\mathbb{E}\left[\left\langle\xi_i(S_j), \xi_j(S)\right\rangle\right] & =\mathbb{E E}_{z_j}\left[\left\langle\xi_i(S_j), \xi_j(S)\right\rangle\right] 
=\mathbb{E}\left[\left\langle\xi_i(S_j), \mathbb{E}_{z_j}\left[\xi_j(S)\right]\right\rangle\right]=0,
\end{align}
where the second identity holds since $\xi_i(S_j)$ is independent of $z_j$ and the last identity follows from $\mathbb{E}_{z_j}\left[\xi_j(S)\right]=0$ due to  the symmetry between $\tdz$ and $z_i$, and their independence with $S^{(i)}$, derived as
\begin{align}\label{eq:E_xi_0}
  \mathbb{E}_{z_i}\left[\xi_i(S)\right]
  = \E_{z_i} \Big[\mathbb{E}_{z_i^{\prime}}[\mathbb{E}_{\tdz}[\nabla F_{\tdz}(A(S^{(i)}) )]-\nabla F_{z_i}(A(S^{(i)}) )] \Big] =0, \quad \forall i \in[n] .
\end{align}
In a similar way, for $i \neq j$, $(d)$  and $(e)$ follow from
\begin{align}
\mathbb{E}\left[\left\langle\xi_i(S), \xi_j(S_i)\right\rangle\right] & =\mathbb{E E}_{z_i}\left[\left\langle\xi_i(S), \xi_j(S_i)\right\rangle\right] 
 =\mathbb{E}\left[\left\langle\xi_j(S_i), \mathbb{E}_{z_i}\left[\xi_i(S)\right]\right\rangle\right]=0, \\
\mathbb{E}\left[\left\langle\xi_i(S_j), \xi_j(S_i)\right\rangle\right] & =\mathbb{E} \mathbb{E}_{z_i}\left[\left\langle\xi_i(S_j), \xi_j(S_i)\right\rangle\right] 
=\mathbb{E}\left[\left\langle\xi_j(S_i), \mathbb{E}_{z_i}\left[\xi_i(S_j)\right]\right\rangle\right]=0 .
\end{align}
Based on~\eqref{eq:inner_xi_i_xi_j_Sj_0}, for $i \neq j$ we have 
\begin{align*}
J_{2,i,j} =& \mathbb{E}\left[\left\langle\xi_i(S), \xi_j(S)\right\rangle\right]
= \mathbb{E}\left[\left\langle\xi_i(S)-\xi_i(S_j), \xi_j(S)-\xi_j(S_i)\right\rangle\right] \\
\leq & \mathbb{E}\left[\left\|\xi_i(S)-\xi_i(S_j)\right\|_{\rm F} \left\|\xi_j(S)-\xi_j(S_i)\right\|_{\rm F} \right] \\
\leq & \frac{1}{2} \mathbb{E}\left[\left\|\xi_i(S)-\xi_i(S_j)\right\|_{\rm F}^2\right]
+\frac{1}{2} \mathbb{E}\left[\left\|\xi_j(S)-\xi_j(S_i)\right\|_{\rm F}^2\right] \numberthis
\label{eq:xi_i_xi_j}
\end{align*}
where we have used $a b \leq \frac{1}{2}\left(a^2+b^2\right)$. According to the definition of $\xi_i(S)$ and $\xi_i(S_j)$ we know the following identity for $i \neq j$
\begin{align*}
\mathbb{E}\left[\left\|\xi_i(S)-\xi_i(S_j)\right\|_{\rm F}^2\right] 
=& \mathbb{E}\Big[\Big\| \mathbb{E}_{z_i^{\prime}} \mathbb{E}_{\tdz}
\big[\nabla F_{\tdz}(A(S^{(i)}) )-\nabla F_{\tdz}(A(S_j^{(i)}) )\big] \\
& +\mathbb{E}_{z_i^{\prime}}\big[\nabla F_{z_i}(A(S_j^{(i)}) )-\nabla F_{z_i}(A(S^{(i)}) )\big] \Big\|_{\rm F}^2\Big].
\numberthis
\end{align*}
It then follows from the  inequality $(a+b)^2 \leq$ $2\left(a^2+b^2\right)$ and the Jensen's inequality that
\begin{align*}
\mathbb{E}[\|\xi_i(S)-\xi_i(S_j)\|_{\rm F}^2] 
\leq & 2 \mathbb{E}[\| \nabla F_{\tdz}(A(S^{(i)}) )-
\nabla F_{\tdz}(A(S_j^{(i)}) ) \|_{\rm F}^2] \\
&+ 2 \mathbb{E}[\|\nabla F_{z_i}(A(S_j^{(i)}) )-\nabla F_{z_i}(A(S^{(i)}) )\|_{\rm F}^2]. 
\numberthis
\end{align*}
Since $S^{(i)}$, $S_j^{(i)}$ and $S^{(j)}$, $S_i^{(j)}$ are two pairs of neighboring datasets, it follows from the definition of stability that
\begin{align}
  \mathbb{E}\left[\left\|\xi_i(S)-\xi_i(S_j)\right\|_{\rm F}^2\right] \leq 4 \epsilon_{\rm F}^2, ~~\text{and}~~\mathbb{E}\left[\left\|\xi_j(S)-\xi_j(S_i)\right\|_{\rm F}^2\right] \leq 4 \epsilon_{\rm F}^2, \quad \forall i \neq j .
\end{align}
We can plug the above inequalities back into~\eqref{eq:xi_i_xi_j} and bound $J_{2,i,j}$ by
\begin{align}
  J_{2,i,j} = \mathbb{E}\left[\left\langle\xi_i(S), \xi_j(S)\right\rangle\right] \leq 4 \epsilon_{\rm F}^2 , \quad \forall i \neq j .
  \label{eq:stab_gen_xi_i_square_bound_J2ij}
\end{align}
Combining the bound for $J_{1,i}$ in~\eqref{eq:stab_gen_xi_i_square_bound_J1i} and $J_{2,i,j}$ in~\eqref{eq:stab_gen_xi_i_square_bound_J2ij} and substituting them back into~\eqref{eq:bound_xi_J1_J2}, it then follows that
\begin{align*}
 \mathbb{E}\Big[\Big\|\sum_{i=1}^n \xi_i(S) \Big\|_{\rm F}^2 \Big] 
 &=\mathbb{E}\Big[\sum_{i=1}^n\|\xi_i(S)\|_{\rm F}^2 \Big]+\sum_{i, j \in[n]: i \neq j} \mathbb{E}[\langle\xi_i(S), \xi_j(S)\rangle] \\
 & \leq n \mathbb{E}\left[\mathbb{V}_{\tdz}(\nabla F_{\tdz}(A(S) ))\right]+4 n(n-1) \epsilon_{\rm F}^2.\numberthis
\end{align*}
We can plug the above inequality back into \eqref{eq:grad_diff_norm_xi}, use the subadditivity of square root function, and get
\begin{align*}
n \mathbb{E}[\| \nabla F(A(S)) & -\nabla F_S(A(S)) \|_{\rm F} ] \leq 4 n \epsilon_{\rm F} 
+\sqrt{n \mathbb{E}\left[\mathbb{V}_{\tdz}(\nabla F_{\tdz}(A(S) ))\right]}. \numberthis
\end{align*}
The proof is complete.
\end{proof}

\subsection{Proof of Theorem~\ref{crlr:bound_stab_grad_MGDA} 
 -- Generalization of MoDo in the NC case}
 \label{sub_app:proof_PS_gen_NC}

In this subsection, we prove Theorem~\ref{crlr:bound_stab_grad_MGDA}, which establishes the PS generalization error of MoDo, SMG, and MoCo in the nonconvex case.

\paragraph{Organization of proof.}
We first define the concept of sampling-determined MOL algorithms in Definition~\ref{def:sample-determined alg}, which significantly generalizes the concept defined in~\citep{lei2022stability_NC_nonsmooth}   in single-objective learning algorithms.
Then we show that MoDo is sampling-determined in Proposition~\ref{prop:SMG_smp_dtm}.
Combining Propositions~\ref{prop:gen_err_minnorm_bound_grad} and~\ref{prop:SMG_smp_dtm}, we are able to prove the upper bound of the MOL uniform stability in Theorem~\ref{crlr:bound_stab_grad_MGDA}. A matching lower bound of the MOL uniform stability is provided in Lemma~\ref{lemma:lower_bound_stability_NC}. Combining the upper and lower bounds, the proof for Theorem~\ref{crlr:bound_stab_grad_MGDA} is complete.

\begin{dfn}[Sampling-determined   algorithm~\citep{lei2022stability_NC_nonsmooth}]
\label{def:sample-determined alg}
Let $A$ be a randomized algorithm that randomly chooses an index sequence $I(A)=\{i_{t,s}\}$ to compute stochastic gradients. We say a symmetric algorithm $A$ is sampling-determined if the output model is fully determined by $\{z_i: i \in I(A)\}$. 
\end{dfn}

\begin{prop}[MoDo, SMG, MoCo, are sampling-determined]
\label{prop:SMG_smp_dtm}
MoDo (Algorithms~\ref{alg:modo},~\ref{alg:SMG}, and~\ref{alg:moco}) are sampling determined.
In other words, 
Let $I(A)=\{i_t\}$ be the sequence of index chosen by these algorithms from training set $S = \{z_1,\dots, z_n\}$, and $z_i \stackrel{\rm i.i.d.}{\sim} \mathcal{P}$ for all $i\in [n]$  to build stochastic gradients, the output $A(S)$ is determined by $\{z_j\mid j \in I(A)\}$. To be precise, $A(S)$ is independent of $z_j$ if $j \notin I(A)$.
\end{prop}

\begin{proof}[Proposition~\ref{prop:SMG_smp_dtm}]
\label{proof:SMG_smp_dtm}
Let $I(A) = \{I_{1}, \dots, I_{T}\}$, $I_t = \{i_{t,s}\}_{s=1}^2$ and $i_{t,s} \in [n]$ for all $1 \leq t \leq T$.
Let $Z_{I(A)} = \{z_{i_{t,s}}\mid t\in [T], s\in [2] \}$.
By the description in Algorithm~\ref{alg:modo}, 
$A(S) = G_{Z_{I_{T}}} \circ\dots \circ   G_{Z_{I_{1}}} (x_0)$, 
where $G_{Z}(\cdot)$ is the stochastic update function of the model parameter given random mini-batch $Z$.
Therefore, for all possible random mini-batch $Z$ selected by $A$, we have
\begin{align*}
  \mathbb{P}(A(S) = x\mid z_j=z, j\notin I(A))
  =&\mathbb{P}(G_{Z_{I_{T}}} \circ\dots \circ   G_{Z_{I_{1}}}(x_0) = x\mid z_j=z, j\notin I(A)) 
  \\
  =&\mathbb{P}(G_{Z_{I_T}}  \circ\dots \circ   G_{Z_{I_1}} (x_0) = x\mid j\notin I(A)) \\
  =& \mathbb{P}(A(S) = x \mid j\notin I(A))
\numberthis\label{eq:AS_independent_zj}
\end{align*}
where the last equality holds because $z_j \notin S_{I(A)}$, and $z_j$ is independent of all elements in $S_{I(A)}$ by $\rm i.i.d.$ sampling.
Therefore, $A(S)$ is independent of $z_j$ if $j \notin I(A)$, MoDo is sampling-determined.

Similarly, for SMG, let $I_t = \{i_{t,s}\}_{s=1}^{|Z_t|}$, and $i_{t,s}\in [n]$ for all $t \in [T]$, then \eqref{eq:AS_independent_zj} still holds for $A$ being the SMG algorithm. Therefore, SMG is also sampling-determined.

Finally, for MoCo, its update at iteration $t$ depends on the stochastic sample selected at iteration $t$, as well as all the stochastic samples at previous iterations. Denote the update function at each iteration as $G_{Z_{I_{1:t}}}(x_t)$, where $Z_{I_{1:t}} = \{z_{I_1}, z_{I_2},\ldots, z_{I_t}\}$, then we have
\begin{align*}
  \mathbb{P}(A(S) = x\mid z_j=z, j\notin I(A))
  =&\mathbb{P}(G_{Z_{I_{1:T}}} \circ\dots \circ   G_{Z_{I_{1:1}}}(x_0) = x\mid z_j=z, j\notin I(A)) 
  \\
  =&\mathbb{P}(G_{Z_{I_{1:T}}}  \circ\dots \circ   G_{Z_{I_{1:1}}} (x_0) = x\mid j\notin I(A)) \\
  =& \mathbb{P}(A(S) = x \mid j\notin I(A)).
\numberthis\label{eq:AS_independent_zj_moco}
\end{align*}
which proves MoCo is sampling-determined.
\end{proof}

\begin{lmm}{\citep[Theorem~5~(b)]{lei2022stability_NC_nonsmooth}}
\label{lemma:stab_grad_prob_bound}
Let $A$ be a sampling-determined random algorithm (Definition~\ref{def:sample-determined alg}) and $S, S^{\prime}$ be neighboring datasets with $n$ data points that differ only in the $i$-th data point. 
If $\sup _z \mathbb{E}_A\left[\|\nabla F_z(A(S) )\|_{\rm F}^2 \mid i \in I(A)\right] \leq G^2$ for any $S$, then
\begin{align}
\sup _z \mathbb{E}_A[\| \nabla F_z(A(S) )-\nabla F_z(A(S^{\prime}) ) \|_{\rm F}^2 ] 
 \leq 4 G^2 \cdot \mathbb{P}\{i \in I(A)\} .
\end{align}
\end{lmm}

\begin{lmm}[Lower bound of MOL uniform stability in the NC case]
\label{lemma:lower_bound_stability_NC}
There exists a vector-valued objective function $F_z(x)$, where  for each $m\in [M]$, $z\in \mathcal{Z}$, the scalar-valued function $f_{z,m}(x)$ is  nonconvex and smooth, and there exists neighboring datasets $S$ and $S'$ with $|S|=|S'|=n$, which differ with at most one sample, and a randomized algorithm MoDo, denoted as $A$, such that the MOL uniform stability of the $t$-th iteration output with $t\in [T]$ is lower bounded by
\begin{align*}
\sup _z \mathbb{E}_A[\| \nabla F_z(A(S) )-\nabla F_z(A(S^{\prime}) ) \|_{\rm F}^2 ] 
= \Omega \Big(\frac{t}{n} \Big).
\end{align*} 
\end{lmm}

\begin{proof}
  \label{proof:lower_bound_NC}
From the definition of the sampling-determined algorithms, 
and that MoDo selects two samples at each iteration,
we can compute the probability of $i^* \in I(A)$ as
\begin{align}
  \mathbb{P}(i^* \in I(A) ) = 1 - \Big(\frac{n-1}{n} \Big)^{2T}
\end{align}
whose lower bound can be computed by
\begin{align}
  \mathbb{P}(i^* \in I(A) ) = 1 - \Big(\frac{n-1}{n} \Big)^{2T}
  \stackrel{(a)}{\geq} 
  1 - 1 + \frac{1}{1 + \frac{2T-1}{n}} \cdot \frac{2T}{n}
  =  
  \frac{T}{ n(1 + \frac{2T-1}{n})}
  \stackrel{(b)}{\geq} 
  \frac{2T}{ 3n}
\end{align}
where $(a)$ follows from the inequality that 
$(1+c)^r \leq 1+\frac{r c}{1-(r-1) c}$ for $c \in [-1, \frac{1}{r-1} ), r>1$, plugging in  $r = 2T > 1$, $c = -1/n < 0 < 1/ (r - 1)$;
and $(b)$ follows from $T \leq n$.

The MOL uniform stability of a sampling-determined algorithm in the general non-convex case can then be lower bounded by
\begin{align*}
  & \sup _z \mathbb{E}_A[\| \nabla F_z(A(S) )-\nabla F_z(A(S^{\prime}) ) \|_{\rm F}^2 ] \\
  =& \sup _z
  \Big ( \mathbb{E}_A[\| \nabla F_z(A(S) )-\nabla F_z(A(S^{\prime}) ) \|_{\rm F}^2 \mid i^* \in I(A)] 
  \cdot \mathbb{P}(i^* \in I(A)) \\
  &\qquad  + \mathbb{E}_A[\| \nabla F_z(A(S) )-\nabla F_z(A(S^{\prime}) ) \|_{\rm F}^2 \mid i^* \notin I(A)] 
  \cdot \mathbb{P}(i^* \notin I(A))
  \Big ) \\
  =& \sup _z
  \mathbb{E}_A[\| \nabla F_z(A(S) )-\nabla F_z(A(S^{\prime}) ) \|_{\rm F}^2 \mid i^* \in I(A)] 
  \cdot \mathbb{P}(i^* \in I(A)) \\
  \geq &
  \sup _z
  \mathbb{E}_A[\| \nabla F_z(A(S) )-\nabla F_z(A(S^{\prime}) ) \|_{\rm F}^2 \mid i^* \in I(A)] 
  \cdot \frac{2T}{3n} .
  \numberthis
  \label{eq:NC_stability_prob_lower_bound}
\end{align*}
We proceed to bound the term 
$\sup _z \mathbb{E}_A[\| \nabla F_z(A(S) )-\nabla F_z(A(S^{\prime}) ) \|_{\rm F}^2 \mid i^* \in I(A)] $ 
in the above inequality by constructing the following simple example with $M=2$, $|S| = |S'| = n > 10$, $S = \{0,\dots, 0 \}$, $S' = \{0,\dots, 0, -\frac{1}{8 }\pi\}$.
\begin{align*}
  f_{z,1}(x) = f_{z,2}(x) = \sin (x + z) \\
  \nabla f_{z,1}(x) = \nabla f_{z,2}(x) = \cos (x+z)
\end{align*}
For algorithm $A$, choose $2 \leq T\leq 10 < n $, step size $\alpha_t = \alpha = \frac{\pi}{80}$, initialization $x_0 = x_0' = \frac{1}{4}\pi$. 
Let $x_t = A_t(S)$, and $x_t' = A_t (S')$.
Since $|\nabla f_{z,1}(x)| \leq 1$, we have
\begin{align}
  |x_0 - x_T|
  \leq \alpha \Bigg|\sum_{t=0}^{T-1} \nabla F_{z}(x_t) \lambda_t \Bigg|
  \leq \alpha \sum_{t=0}^{T-1} |\nabla F_{z}(x_t) \lambda_t|
  \leq \alpha T \leq \frac{1}{8} \pi .
\end{align}
Similarly, we have
\begin{align}
  |x_0' - x_T'| \leq \frac{1}{8} \pi .
\end{align}
Therefore, for all $t \in [T]$, it holds that
\begin{align}\label{eq:bound_x_t_lower_stab}
 \frac{1}{8} \pi \leq x_t \leq \frac{3}{8} \pi, ~~
 \frac{1}{8} \pi \leq x_t' \leq \frac{3}{8} \pi.
\end{align}
We need to bound $\E_A[\| \nabla F_z(A(S) )-\nabla F_z(A(S^{\prime}) ) \|_{\rm F}^2 \mid i^* \in I(A)]$ in \eqref{eq:NC_stability_prob_lower_bound}. Considering the case $i^* \in I(A)$,
let $t_0 \in [T-1]$ denote the first iteration to select $i^*$, then
\begin{align*}
   \nabla f_{z'}(x_{t_0}') - \nabla f_z(x_{t_0}) 
   =& \cos (x_{t_0}' + z') - \cos (x_{t_0} + z) 
   = \cos \Big(x_{t_0}' - \frac{1}{8} \pi \Big) - \cos (x_{t_0} ) \\
   =& -2 \sin \Big(x_{t_0} - \frac{1}{16}\pi \Big) \sin \Big( - \frac{1}{16}\pi \Big) 
   \geq 0.076
\end{align*}
which implies 
\begin{align}
  x_{t_0 + 1} - x_{t_0 + 1}'
  =& x_{t_0} - x_{t_0 }' + \alpha(\nabla f_{z'}(x_{t_0}') - \nabla f_z(x_{t_0})) \nonumber\\
  =& \alpha(\nabla f_{z'}(x_{t_0}') - \nabla f_z(x_{t_0}))
  \geq 0.076 \alpha > 0 .
  \label{eq:bound_diff_x_t0_lower_stab}
\end{align}

We then prove by mathematical induction that $x_T - x_T' \geq 0.076 \alpha$ using the following statements:\\
1) $x_{t_0+1} - x_{t_0+1}' \geq 0.076 \alpha$;\\
2) $x_{t+1} - x_{t+1}' \geq x_{t} - x_{t}'\geq 0.076 \alpha$ if $x_{t} - x_{t}'\geq 0.076 \alpha > 0$.

The first statement is proved in~\eqref{eq:bound_diff_x_t0_lower_stab}. The second statement can be proved by
\begin{align*}
  x_{t+1} - x_{t+1}'
  =& x_{t} - x_{t}'
  + \alpha(\nabla f_{z'}(x_{t}') - \nabla f_z(x_{t}))\\
  =& x_{t} - x_{t}'
  + \alpha(\cos(x_{t}' + z') - \cos(x_{t} + z)) \geq 0 .
\end{align*}
The last inequality follows from that for $t_0 < t \leq T$, $\frac{1}{8} \pi \leq x_t' < x_t  \leq \frac{3}{8} \pi$ as~\eqref{eq:bound_x_t_lower_stab}, 
where $\nabla f_z(x) = \cos(x+z)$ is monotonically decreasing with $x+z \in [0, \frac{1}{2}\pi]$. And since $x_t - x_t'>0$, $z'\leq z $, $x'_t+z'\leq x_t + z$, therefore $\cos(x_{t}' + z') - \cos(x_{t} + z) \geq 0$.
Then we arrive at the following result
\begin{align}
  x_T - x_T' \geq 0.076 \alpha \geq \frac{7\pi}{800} .
\end{align}
By the Mean value theorem, there exists $\bar{x} \in [x_T', x_T] \subset [\frac{1}{8}\pi, \frac{3}{8}\pi]$ such that
\begin{align}
  |\nabla f_z(A(S) )-\nabla f_z(A(S^{\prime}))|
  = | \nabla f_z(x_T )-\nabla f_z(x_T') |
  = |\nabla^2 f_z(\bar{x} )| |x_T - x_T'|
  \geq \sin \Big(\frac{1}{8}\pi \Big) \frac{7\pi}{800}.
  \label{eq:mean_value_grad_diff_lower_bound}
\end{align}
Therefore, combining~\eqref{eq:NC_stability_prob_lower_bound} and \eqref{eq:mean_value_grad_diff_lower_bound} yields
\begin{align}
  \sup _z \mathbb{E}_A[\| \nabla F_z(A(S) )-\nabla F_z(A(S^{\prime}) ) \|_{\rm F}^2 ] 
  = \Omega \Big(\frac{T}{n} \Big).
\end{align}
The proof of the lower bound in the nonconvex case is complete.
\end{proof}  

The following remark discusses the application of the above MOL uniform stability lower bound to single-objective learning.
\begin{remark}
\label{rmk:lower_bound_stability_NC}
  Our lower bound in the NC case can be easily reduced to the single objective learning problem with sampling-determined algorithms since our proof is based on the construction of a special case with identical multi-objectives where the update of $\lambda$ does not affect the update of $x$.
  The reduction to the single-objective learning setting is also the first lower bound with sampling-determined algorithms for single-objective learning in the NC case that matches the upper bound in~\citep{lei2022stability_NC_nonsmooth}.
\end{remark}

\subsubsection{Proof of Theorem~\ref{crlr:bound_stab_grad_MGDA}} 
\label{ssub_app:proof_PS_gen_MoDo}

\begin{proof}
\label{proof:bound_stab_grad_MGDA}
From Proposition~\ref{prop:SMG_smp_dtm},  MoDo algorithm is sampling-determined.
Then based on Lemma~\ref{lemma:stab_grad_prob_bound}, its MOL uniform stability in Definition~\ref{def:stab_grad} can be bounded by
\begin{align}\label{eq:epsilon_F_bound}
\epsilon_{\rm F}^2 
 \leq 4 G^2 \cdot \mathbb{P}\{i \in I(A)\} .
\end{align}
Let $i_t$ be the index of the sample selected by $A$ at $t$-th iteration, $I_t(A)$ be the indices of the samples selected by $A$ up to the $t$-th iteration, with $t\in [T]$, and $i^*$ be the index of the data point that is different in $S$ and $S'$.
Then
\begin{align}\label{eq:prob_IA}
\mathbb{P}\{i^* \in I_t(A)\} \leq \sum_{k=1}^t \mathbb{P}\left\{i_k=i^*\right\} \leq \frac{t}{n} .
\end{align}
Combining \eqref{eq:epsilon_F_bound} and \eqref{eq:prob_IA} gives the MOL uniform stability of the $t$-th iterate with $t\in [T]$ is upper bounded by
\begin{align}\label{eq:ep_F_Tn}
 \epsilon_{\rm F}^2 \leq 
 \frac{4 G^2 t}{n} \leq 
 \frac{4 G^2 T}{n}  .   
\end{align}
This proves the upper bound in a). The proof of the lower bound in b) is given in Lemma~\ref{lemma:lower_bound_stability_NC} in Appendix~\ref{sub_app:proof_PS_gen_NC}.
Then based on Propositions~\ref{prop:gen_err_minnorm_bound_grad}-\ref{prop:stability_gen_grad_F}, the PS generalization error is upper bounded by
\begin{align*}
 \E_{A,S}[R_{\rm gen}(A_t(S))]
 {\leq} &
 \E_{A,S} [ \|\nabla F(A_t(S))- \nabla F_S(A_t(S)) \|_{\rm F}] 
 \tag*{by Proposition~\ref{prop:gen_err_minnorm_bound_grad}}\\
 {\leq} &
 4\epsilon_{\rm F} 
+ \sqrt{n^{-1} \mathbb{E}_S\left[\mathbb{V}_{z\sim \cal D}(\nabla F_z(A_t(S) ))\right]} 
\tag*{by Proposition~\ref{prop:stability_gen_grad_F}}\\
{=}& 
\mathcal{O}(T^{\frac{1}{2}}n^{-\frac{1}{2}}).
\tag*{by \eqref{eq:ep_F_Tn}}
\end{align*}
The proof of the upper bound is complete.
Lemma~\ref{lemma:lower_bound_stability_NC} provides the lower bound.
Combining both completes the proof.
\end{proof}

\subsection{Proof of Theorem~\ref{thm:PS_gen_SMG_moco} -- Generalization of SMG and MoCo in the NC case} 
\label{sub_app:proof_PS_gen_SMG_MoCo_NC}

\begin{proof}
  The proof follows similar steps as the proof for Theorem~\ref{crlr:bound_stab_grad_MGDA}. 
  First, Proposition~\ref{prop:SMG_smp_dtm} states that SMG and MoCo are sampling-determined, and thus their MOL uniform stability $\epsilon_{\rm F}^2 = \mathcal{O}(Tn^{-1})$ by Lemma~\ref{lemma:stab_grad_prob_bound}. 
  Then combining with Proposition~\ref{prop:stability_gen_grad_F} which connects the MOL uniform stability and PS generalization error, it yields that their PS generalization errors are $\E_{A,S}[R_{\rm gen}(A_t(S))] = \mathcal{O}(T^{\frac{1}{2}}n^{-\frac{1}{2}})$ for all $t\in [T]$.
\end{proof}

\subsection{Proof of Lemma~\ref{lemma:x_t_bounded_sc_smooth} -- $x_t$ bounded in the SC case}
\label{sub_app:proof_xt_bounded_SC}

\paragraph{Technical challenges.}
In this work, we focus on analyzing stochastic MGDA-based MOL algorithms in the unconstrained setting. This is because in the constrained setting, MGDA with projected gradient descent on $x$ has no guarantee to find the CA direction, and a new algorithm needs to be developed to achieve this~\citep{tanabe2018proximal}.
However, a fundamental challenge in the unconstrained strongly convex setting is that a strongly convex function is not Lipschitz continuous on $\mathbb{R}^d$.
We overcome this challenge by showing that $\{x_t\}_{t=1}^T$ generated by the MoDo algorithm is bounded on the trajectory, so is the gradient $\|\nabla f_{z,m}(x_t)\|$ for all $m \in [M]$, and $z \in \cal Z$.
Thereby, we are able to derive the upper bound of PS optimization and generalization errors without the Lipschitz continuity assumption for strongly convex objectives.

\paragraph{Organization of proof.}
Without loss of generality, we assume $\inf_{x\in \mathbb{R}^d} f_{z,m}(x) < \infty$ for all $m \in [M]$ and $z \in \cal Z$ in the strongly convex case.
In Lemma~\ref{lemma:x_star_bounded}, we  show that the optimal solution of $F_z(x) \lambda$ given any stochastic sample $z\in \cal Z$, and weighting parameter $\lambdain$, is bounded.
In Lemma~\ref{lemma:bound_x_update_sc_smooth}, we show that if the argument parameter is bounded, then the updated parameter by MoDo at each iteration is also bounded by exploiting the co-coerciveness of strongly convex and smooth objectives.
Finally,
based on Lemma~\ref{lemma:x_star_bounded} and Lemma~\ref{lemma:bound_x_update_sc_smooth}, we  first prove that with a bounded initialization $x_0$, the model parameter $\{x_t\}_{t=1}^T$ generated by MoDo algorithm is bounded on the trajectory. Then by the  smoothness assumption of $f_{z,m}(x)$, we immediately have that $\|\nabla f_{z,m}(x)\|$ is bounded for $x\in \{x_t\}_{t=1}^T$ generated by MoDo algorithm, which completes the proof of Lemma~\ref{lemma:x_t_bounded_sc_smooth}.
Lemma~\ref{lemma:x_t_bounded_sc_smooth} paves the way for deriving the MOL uniform stability and PS generalization error of MoDo in the strongly convex and unconstrained setting.

\subsubsection{Auxiliary Lemmas} 
\label{ssub_app:auxiliary_proof_xt_bounded_SC}

\begin{lmm}\label{lemma:x_star_bounded}
Suppose Assumptions~\ref{assmp:lip_cont_grad_f}, \ref{assmp:sconvex} hold. 
W.l.o.g., assume $\inf_{x\in \mathbb{R}^d} f_{z,m}(x) < \infty$ for all $m \in [M]$ and $z \in \cal Z$.
For any given $\lambdain$, and stochastic sample $z\in \cal Z$,
define $x^*_{\lambda, z} = \arg\min_{x\in \mathbb{R}^d} F_z(x) \lambda$, 
 then   $\inf_{x\in \mathbb{R}^d} F_z(x) \lambda < \infty $ and $\|x^*_{\lambda, z}\| < \infty$, i.e., there exist finite positive constants $c_{F^*}$ and $c_{x^*}$ such that 
\begin{equation}
    \inf_{x\in \mathbb{R}^d} F_z(x) \lambda \leq c_{F^*} ~~~{\rm and}~~~ \|x^*_{\lambda, z}\| \leq c_{x^*}.
\end{equation}  
\end{lmm}

\begin{proof}
Under Assumption~\ref{assmp:sconvex}, for all $m \in [M]$, $f_{z,m}(x)$ is strongly convex w.r.t. $x$, thus has a unique minimizer.
Define the minimizer $x^*_{m,z} = \arg\min_{x\in \mathbb{R}^d} f_{z,m}(x)$.
Since a strongly convex function is coercive,    $\inf_{x\in \mathbb{R}^d} f_{z,m}(x) < \infty$, i.e., $f_{z,m}(x^*_{m,z}) < \infty$, implies that $\|x^*_{m,z}\|< \infty$.

By Assumption~\ref{assmp:lip_cont_grad_f}, the $\ell_{f,1}$-smoothness of $ f_{z,m}(x) $,  for $x$ such that $\|x\| < \infty$
\begin{align*}
f_{z,m} (x)  
\leq &
f_{z,m} (x^*_{m,z})  +
\langle \nabla f_{z,m}(x^*_{m,z}) , x - x^*_{m,z} \rangle
+ \frac{\ell_{f,1}}{2} \|x - x^*_{m,z}\|^2  \\
\leq & f_{z,m} (x^*_{m,z})  +\frac{\ell_{f,1}}{2} \|x - x^*_{m,z}\|^2 
< \infty .
\numberthis
\end{align*}
Since $F_z(x)\lambda$ is convex w.r.t. $x$,  for all $\lambdain$, with $\lambda = [\lambda_1, \dots, \lambda_M]^{\top}$, we have
\begin{align}
  F_z \Big(\frac{1}{M}\sum_{m=1}^M x_{m,z}^*  \Big)\lambda   
  \leq 
  \frac{1}{M}\sum_{m=1}^M  F_z(x_{m,z}^*  )\lambda   
  = \frac{1}{M}\sum_{m=1}^M \sum_{m'=1}^M f_{m',z}(x_{m,z}^* )\lambda_{m'}
  < \infty .
\end{align}
Therefore, for all $\lambdain$, we have
\begin{align}
 \inf_{x\in \mathbb{R}^d} F_z(x) \lambda 
 \leq  F_z \Big(\frac{1}{M}\sum_{m=1}^M x_{m,z}^*  \Big)\lambda 
 < \infty .
\end{align}
Since $F_z(x) \lambda$ is strongly convex, thus coercive, then $\|x^*_{\lambda, z}\| < \infty$, which proves the result.
\end{proof}

\begin{lmm}\label{lemma:bound_x_update_sc_smooth}
Suppose Assumptions~\ref{assmp:lip_cont_grad_f}, \ref{assmp:sconvex} hold, and
define   $\kappa = 3 \ell_{f,1}/\mu \geq 3$.
For any given $\lambdain$, and a stochastic sample $z\in \cal Z$,
define $x^*_{\lambda,z} = \arg\min_{x} F_z(x) \lambda$. 
Then by Lemma~\ref{lemma:x_star_bounded}, there exists a positive finite constant $c_{x,1} \geq c_{x^*}$ such that $\|x^*_{\lambda, z}\| \leq c_{x^*} \leq c_{x,1}$.
Recall the multi-objective gradient update is  
\begin{equation}
    G_{\lambda,z}(x) = x - \alpha \nabla F_z(x) \lambda
\end{equation}
with step size
$0 \leq  \alpha \leq  \ell_{f,1}^{-1} $.
Defining $c_{x,2} = (1 + \sqrt{2 \kappa}) c_{x,1}$, we have that
\begin{equation}
\text{ if}~~~\|x\| \leq c_{x,2},~\text{ then}~~~\| G_{\lambda,z}(x) \| \leq c_{x,2}.
\end{equation}
\end{lmm}

\begin{proof}
We divide the proof into two cases: 
1) when $ \|x\| < c_{x,1} $; and, 
2) when $c_{x,1} \leq \|x\| \leq c_{x,2} $.

1) For the first case,  $\|x\| < c_{x,1} \leq c_{x, 2}$, then we have
\begin{align*}
\|G_{\lambda,z}(x)\|  
\leq & \|G_{\lambda,z}(x) - x^*\| + \| x^*\| \\
\stackrel{(a)}{=} &
\|G_{\lambda,z}(x) - G_{\lambda,z}(x^*)\| + \| x^*\| 
\stackrel{(b)}{\leq}  \|x - x^*\| + \| x^*\| \\
\leq & \|x \| + 2\| x^*\| 
\leq  3 c_{x,1} \leq (1 + \sqrt{6}  )c_{x,1} 
 \leq (1 + \sqrt{2 \kappa}  )c_{x,1} \leq c_{x,2}
\numberthis
\end{align*}
where $(a)$ follows from $\nabla F_z(x^*) \lambda = 0$, and $(b)$ follows from the non-expansiveness of the gradient update for strongly convex and smooth function.

2) For the second case,  $c_{x,1} \leq \|x\| \leq c_{x,2} $,   we first consider $\alpha = \ell_{f,1}^{-1}$.
Let $\mu' = \mu / 3$.
Note that since $F_z(x) \lambda$ is $\mu$-strongly convex, it is also $\mu'$-strongly convex.
By strong convexity and  smoothness of $F_z(x) \lambda$, the gradients are co-coercive~\citep[Theorem 2.1.12]{nest_intro_lecture}, i.e., for any $x$ we have
\begin{align}
( \nabla F_z(x)\lambda) ^{\top} (x-x^*) \geq \frac{ \ell_{f,1}^{-1}\|\nabla F_z(x)\lambda\|^2}{1+  {\kappa}^{-1}}+\frac{ {\mu'} \|x-x^*\|^2}{1+ {\kappa}^{-1}}   .
\end{align}
Rearranging and applying Cauchy-Schwartz inequality, we have
\begin{align*}
(\nabla {F}_z(x) \lambda )^{\top} x 
& \geq 
(\nabla {F}_z(x) \lambda)^{\top} x^*+
\frac{ \ell_{f,1}^{-1}\|\nabla F_z(x)\lambda\|^2}{1+  {\kappa}^{-1}}+\frac{ {\mu'} \|x-x^*\|^2}{1+ {\kappa}^{-1}}   \\
& \geq 
-c_{x,1} \| \nabla {F}_z(x) \lambda\|
+ \frac{ \ell_{f,1}^{-1}\|\nabla F_z(x)\lambda\|^2}{1+  {\kappa}^{-1}}+\frac{ {\mu'} \|x-x^*\|^2}{1+ {\kappa}^{-1}} .  
\numberthis\label{eq:innerprod_grad_x}
\end{align*}
By the definition of $G_{\lambda,z}(x)$, 
\begin{align}\label{eq:udpate_lam_z_equation}
\|G_{\lambda,z}(x)\|^2 
& = \left\|x-\frac{1}{\ell_{f,1}} \nabla F_z(x) \lambda \right\|^2 
= \|x\|^2+\frac{1}{\ell_{f,1}^2}\|\nabla F_z(x) \lambda \|^2- \frac{2}{\ell_{f,1}} (\nabla F_z(x) \lambda)^{\top} x .
\end{align}
Substituting \eqref{eq:innerprod_grad_x} into \eqref{eq:udpate_lam_z_equation} yields
\begin{align*}
\hspace{-0.4cm} \|G_{\lambda,z}(x)\|^2  
\leq &
\|x\|^2+\frac{1}{\ell_{f,1}^2}\|\nabla F_z(x)\lambda \|^2 
+\frac{2}{\ell_{f,1}} \Big( c_{x,1} \| \nabla {F}_z(x) \lambda\|
- \frac{ \ell_{f,1}^{-1}\|\nabla F_z(x)\lambda\|^2}{1+  {\kappa}^{-1}}
- \frac{ {\mu'} \|x-x^*\|^2}{1+ {\kappa}^{-1}} \Big) \\
= & \|x\|^2+\frac{2}{\ell_{f,1}} 
\Big( c_{x,1} \| \nabla {F}_z(x) \lambda\|
-\frac{1}{2 \ell_{f,1}} (\frac{1- {\kappa}^{-1}}{1+  {\kappa}^{-1}} )\|\nabla F_z(x) \lambda \|^2
-\frac{\mu'}{1+ {\kappa}^{-1}} \|x-x^* \|^2 \Big) \\
\leq & \|x\|^2+\frac{2}{\ell_{f,1}} 
\sup_{\tau \in \mathbb{R}} \Big( \underbrace{c_{x,1} \cdot\tau
-\frac{1}{2 \ell_{f,1}} (\frac{1- {\kappa}^{-1}}{1+  {\kappa}^{-1}} )\tau^2}_{I_1} -\frac{\mu'\|x-x^* \|^2}{1+ {\kappa}^{-1}} \Big) . 
\numberthis\label{eq:bound_update_x_lam_z}
\end{align*}
Since $\kappa \geq 3$, thus $\frac{1- {\kappa}^{-1}}{1+  {\kappa}^{-1}} > 0$,
then $I_1$ is a quadratic function w.r.t. $\tau$, and is strictly concave, thus can be bounded above by
\begin{align}
\sup_{\tau \in \mathbb{R}}~~ c_{x,1} \cdot\tau
-\frac{1}{2 \ell_{f,1}} (\frac{1- {\kappa}^{-1}}{1+  {\kappa}^{-1}} )\tau^2
\leq \frac{c_{x,1}^2 \ell_{f,1}}{2} \frac{1+ {\kappa}^{-1}}{1-  {\kappa}^{-1}}.
\end{align} 
Substituting this back into \eqref{eq:bound_update_x_lam_z} gives that
\begin{align*}
\|G_{\lambda,z}(x)\|^2  
\leq &  
\|x\|^2+\frac{2}{\ell_{f,1}} 
\Big( \frac{c_{x,1}^2 \ell_{f,1}}{2} \frac{1+ {\kappa}^{-1}}{1-  {\kappa}^{-1}}
-\frac{\mu'}{1+ {\kappa}^{-1}} \|x-x^* \|^2 \Big) \\
= &  
\|x\|^2+ 
{c_{x,1}^2 } \frac{1+ {\kappa}^{-1}}{1-  {\kappa}^{-1}}
- {2}  \frac{{\kappa}^{-1} }{1+ {\kappa}^{-1}} \|x-x^* \|^2  \\
\leq &
\|x\|^2+ 
{c_{x,1}^2 } \frac{1+ {\kappa}^{-1}}{1-  {\kappa}^{-1}}
- {2}  \frac{{\kappa}^{-1} }{1+ {\kappa}^{-1}} (\|x \| - \| x^* \| )^2 \\
\leq & 
\underbrace{\|x\|^2+ 
2 {c_{x,1}^2 } - {{\kappa}^{-1} }  (\|x \| - c_{x,1} )^2}_{I_2}
\numberthis
\end{align*}
where the last inequality follows from $\kappa \geq 3$,  thus $\frac{1+ {\kappa}^{-1}}{1-  {\kappa}^{-1}} \leq 2$, $- {2}  \frac{{\kappa}^{-1} }{1+ {\kappa}^{-1}} \leq -\kappa^{-1}$,   and $\|x^*\|\leq c_{x,1} \leq \|x\|$ by assumption. 

For $c_{x,1} \leq \|x\| \leq c_{x,2}$, $I_2$ is a strictly convex quadratic function of $\|x\|$, which achieves its maximum at $\|x\| = c_{x,1}$ or $\|x\| = c_{x,2}$. Therefore,
\begin{align*}
 \|G_{\lambda,z}(x)\|^2  
\leq &  
\max\{ 
3 {c_{x,1}^2 }  , 
c_{x,2}^2+ 
2 {c_{x,1}^2 } - {{\kappa}^{-1} }  (c_{x,2} - c_{x,1} )^2\} \\
\stackrel{(c)}{=}  &  
\max\{ 
3 c_{x,1}^2, 
c_{x,2}^2 \} 
\stackrel{(d)}{<} c_{x,2}^2
\numberthis
\end{align*}
where $(c)$ follows from the definition that $c_{x,2} = (1 + \sqrt{2 \kappa}) c_{x,1}$;
$(d)$ follows from $\kappa \geq 3$, and thus $3 c_{x,1}^2  < (1 + \sqrt{2 \kappa})^2 c_{x,1}^2 = c_{x,2}^2$.
We have proved the case for $\alpha = \ell_{f,1}^{-1}$.
The result for $ 0 \leq  \alpha < \ell_{f,1}^{-1}$ follows by observing that,
\begin{align*}
  \|G_{\lambda,z}(x)\|
  =& \|x - \alpha \nabla F_z(x) \lambda \| \\
  =& \| (1 - \alpha \ell_{f,1} ) x + \alpha \ell_{f,1} (x - \ell_{f,1}^{-1} \nabla F_z(x) \lambda) \| \\
  \leq & (1 - \alpha \ell_{f,1} ) \|x\| + \alpha \ell_{f,1} \|x - \ell_{f,1}^{-1} \nabla F_z(x) \lambda\| 
  \leq c_{x,2}.
\numberthis
\end{align*}
The proof is complete.
\end{proof}

\subsubsection{Proof of Lemma~\ref{lemma:x_t_bounded_sc_smooth}} 
\label{ssub_app:proof_xt_bounded_SC}

\begin{proof}
We first prove (a), i.e., $\{x_t\}$ generated by the MoDo algorithm are bounded.
Define $\kappa = 3\ell_{f,1}/\mu$ and $x^*_{\lambda, z} = \arg\min_{x} F_z(x) \lambda$ with $\lambdain$. 
Under Assumptions~\ref{assmp:lip_cont_grad_f}, \ref{assmp:sconvex}, 
by Lemma~\ref{lemma:x_star_bounded}, $\|x^*_{\lambda, z}\| < \infty $,
i.e., there exists a finite positive constant $c_{x^*}$ such that $\|x^*_{\lambda, z}\| \leq c_{x^*}$.
Choose the initial iterate to be bounded, i.e., there exists a finite positive constant $c_{x_0}$ such that  $\|x_0\| \leq c_{x_0}$.
Then we will prove that for $\{x_t\}$  generated by MoDo algorithm with $\alpha_t = \alpha$ and $0 \leq  \alpha \leq  \ell_{f,1}^{-1} $, we have 
\begin{equation}\label{eq:x_t_bound_cx}
\| x_t \| \leq c_x,~~~{\rm with}~~~c_x = \max\{ (1 + \sqrt{2\kappa} ) c_{x^*}, c_{x_0} \}.
\end{equation}
To prove~\eqref{eq:x_t_bound_cx}, 
we rely on Lemma~\ref{lemma:bound_x_update_sc_smooth}, which states that if the current iterate $x_t$ is bounded, then with MoDo update, the next iterate $x_{t+1}$ is also bounded.
Let $c_{x,1} = \max\{(1 + \sqrt{2\kappa} )^{-1} c_{x_0}, c_{x^*}\}$, and $c_{x,2} = (1 + \sqrt{2\kappa} ) c_{x,1} = \max\{ c_{x_0},(1 + \sqrt{2\kappa} ) c_{x^*}\}$
in Lemma~\ref{lemma:bound_x_update_sc_smooth}.
We then consider the following two cases:

1) If $(1 + \sqrt{2\kappa} ) c_{x^*} \leq c_{x_0}$, 
then $\|x^*_{\lambda, z}\| \leq c_{x^*} \leq (1 + \sqrt{2\kappa} )^{-1} c_{x_0}$.
Then it satisfies the condition in Lemma~\ref{lemma:bound_x_update_sc_smooth} that $\|x^*_{\lambda, z}\| \leq c_{x,1}$ and $ \|x_0\| \leq c_{x,2}$. Applying Lemma~\ref{lemma:bound_x_update_sc_smooth} yields $\|x_1\| \leq c_{x,2}$.

2) If $(1 + \sqrt{2\kappa} ) c_{x^*} > c_{x_0}$, 
then $\|x_0\| \leq c_{x_0} < (1 + \sqrt{2\kappa} ) c_{x^*}$.
Then it satisfies the condition in Lemma~\ref{lemma:bound_x_update_sc_smooth} that $\|x^*_{\lambda, z}\| \leq c_{x,1}$ and $ \|x_0\| \leq c_{x,2}$. Applying Lemma~\ref{lemma:bound_x_update_sc_smooth} yields $\|x_1\| \leq c_{x,2}$.

Therefore, \eqref{eq:x_t_bound_cx} holds for $t=1$.
We then prove by induction that \eqref{eq:x_t_bound_cx} also holds for $t \in [T]$.
Assume~\eqref{eq:x_t_bound_cx} holds at $1\leq k \leq T-1$, i.e.,
\begin{align}
 \|x_k\| \leq c_x = c_{x,2}   
\end{align}
Then by Lemma~\ref{lemma:bound_x_update_sc_smooth}, at $k+1$,
\begin{align}
 \|x_{k+1}\| = \|G_{\lambda_{k+1},Z_{k+1}}(x_k)\| \leq c_{x,2}.   
\end{align}
Since $\|x_1\| \leq c_{x,2}$, for $t = 0, \dots, T-1$, we have
\begin{align}
\|x_{t+1} \| 
= \|G_{\lambda_{t+1}, Z_{t+1}}(x_t)\|
\leq   c_{x,2}. 
\end{align}
Therefore, by mathematical induction,
$\|x_t\| \leq c_{x,2} = c_x, \text{for all } t \in [T]$.
The proof of (a) is thus complete.

We then prove (b).
This result follows directly from (a), Assumption~\ref{assmp:lip_cont_grad_f}, i.e., the $\ell_{f,1}$-smoothness assumption for all objectives, and boundedness of the Pareto optimal solutions given in Lemma~\ref{lemma:x_star_bounded}.
Specifically, 
by Lemma~\ref{lemma:x_star_bounded},  there exist finite positive constant $c_{x^*}$ such that   
$\|x^*_{\lambda, z}\| \leq c_{x^*}$.
Then by Assumption~\ref{assmp:lip_cont_grad_f}, the $\ell_{f,1}$-Lipschitz continuity of the gradient $\nabla F_z(x) \lambda$, we have
\begin{align*}
\|\nabla F_z(x) \lambda\| 
=& \|\nabla F_z(x) \lambda
- \nabla F_z(x^*_{\lambda, z}) \lambda\| \\ 
\leq &
\ell_{f,1} \|x - x^*_{\lambda, z}\|
\leq 
\ell_{f,1} (\|x \| + \| x^*_{\lambda, z}\|)
\leq \ell_{f,1} (c_x + c_{x^*})
\numberthis
\end{align*}
where the first equality uses the fact that $\nabla F_z(x^*_{\lambda, z}) \lambda = 0$.
Define $\ell_f \coloneqq \ell_{f,1} (c_x + c_{x^*})$, and $\ell_F \coloneqq \sqrt{M} \ell_{f} $, and then it holds for all $\lambdain$ that
\begin{align}
 \|\nabla F(x_t)\lambda\| \leq \ell_f
 \quad \text{and} \quad
 \|\nabla F(x_t)\| \leq \|\nabla F(x_t)\|_{\rm F} \leq \sqrt{M} \ell_{f} = \ell_F.
\end{align}
The proof of (b) is thus complete.
\end{proof}

\subsection{Proof of Theorem~\ref{thm:grad_stability_modo_sc} -- Generalization of MoDo in the SC case}
\label{sub_app:proof_gen_sc_app}

\paragraph{Technical challenges.}
One challenge that the strongly convex objectives are not Lipschitz for $\xin$ is addressed by Lemma~\ref{lemma:x_t_bounded_sc_smooth}.
Another challenge compared to static weighting or single-objective learning is that the MoDo algorithm involves the update of two coupled sequences $\{x_t\}$ and $\{\lambda_t\}$.
Consequently, the commonly used argument that the SGD update for strongly convex objectives has the classical contraction property for model parameter $x$~\citep{hardt2016train} does not necessarily hold in our case since the weighting parameter $\lambda$ is changing, as detailed in Section~\ref{subs:property_modo_update}.
Nevertheless, we manage to derive a tight stability bound when $\gamma = \mathcal{O}(T^{-1})$, as detailed in Section~\ref{subs:mol_stability_upper}.

\paragraph{Organization of proof.}
In Section~\ref{subs:property_modo_update}, we prove the properties of the MoDo update, including approximate expansiveness or non-expansiveness and boundedness.
Building upon these properties, in Section~\ref{subs:mol_stability_upper}, we prove the upper bound of argument stability in Theorem~\ref{thm:bound_E_diff_x_lam_warm_modo}, and the upper bound of MOL uniform stability.
To show the tightness of the upper bound, in Section~\ref{app_subs:lower_bound}, Theorem~\ref{thm:lower_bound}, we derive a matching lower bound of MOL uniform stability.
Combining the upper bound in Section~\ref{subs:mol_stability_upper} and the lower bound in Section~\ref{app_subs:lower_bound} leads to the results in Theorem~\ref{thm:grad_stability_modo_sc}, whose proof is in Section~\ref{app_subs:proof_stab_SC_modo}.

\subsubsection{Expansiveness and boundedness of MoDo update}\label{subs:property_modo_update}
In this section, we prove the properties of the update function of MoDo at each iteration, including boundedness and approximate expansiveness, which is then used to derive the algorithm stability.
For $z, z_1, z_2 \in S$, $\lambdain$, recall that the update functions of MoDo  is
\begin{align*}
G_{x, z_1,z_2}(\lambda) &= \Pi_{\Delta^{M}}\left(\lambda - \gamma (\nabla F_{z_1} (x)^\top \nabla F_{z_2}(x) + \rho \mathrm{I} )  \lambda \right) \\ 
G_{\lambda, z} (x) &= x - \alpha \nabla F_{z}(x) \lambda .
\end{align*}

\begin{lmm}[Boundedness of  MoDo update]
\label{lemma:bounded_agn_update}  
Let $\ell_f$ be a positive constant.
If $\|\nabla F_z(x) \lambda\| \leq \ell_f$  for all $\lambdain$, $z \in S$ and $x \in \{x_t\}_{t=1}^T$ generated by the MoDo algorithm with step size $\alpha_t \leq \alpha$,
then  $G_{\lambda, z}(x)$ is $(\alpha \ell_f)$-bounded on the trajectory of MoDo, i.e.,
\begin{align}
\sup_{x\in \{x_t\}_{t=1}^T}\|G_{\lambda,z}(x) - x\|
\leq \alpha \ell_f .
\end{align}
\end{lmm}

\begin{proof}
\label{proof:bounded_agn_SMGDA_update} 
For all $x \in \{x_t\}_{t=1}^T$,  $\lambda \in \Delta^{M}$, and  $z \in S$, since $\|\nabla F_z(x) \lambda\| \leq \ell_f$,
we have
\begin{align}
  \|G_{\lambda, z}(x) - x\|
  \leq &\| \alpha \nabla F_{z}(x) \lambda \|
  \leq \alpha \ell_f
\end{align}
which proves the boundedness.  
\end{proof}

\begin{lmm}[Properties of  MoDo update  in  SC case]
\label{lemma:modo_sconvex_pgd_lam_update_warm}
Suppose Assumptions~\ref{assmp:lip_cont_grad_f}, \ref{assmp:sconvex}
hold. 
Let $\ell_f$ be a positive constant.
If for all $\lambda, \lambda' \in \Delta^M$, $z \in S$, and $x \in \{x_t\}_{t=1}^T$, $x' \in \{x_t'\}_{t=1}^T$ generated by the MoDo algorithm on datasets $S$ and $S'$, respectively, we have $\|\nabla F_z(x) \lambda\| \leq \ell_f$, $\|\nabla F_z(x') \lambda'\| \leq \ell_f$, and $\|\nabla F_z(x)\| \leq \ell_F$, $\|\nabla F_z(x')\| \leq \ell_F$, and step sizes of MoDo satisfy $\alpha_t \leq \alpha$, $\gamma_t \leq \gamma$,
 it holds that
\begin{align*}
\|G_{\lambda,z}(x) - G_{\lambda',z}(x')\|^2
\leq &
(1 - 2\alpha \mu + 2\alpha^2 \ell_{f,1}^2)\|x - x'\|^2  \\
& + 2\alpha \ell_F \|x - x'\| \| \lambda - \lambda' \|
+ 2\alpha^2 \ell_F^2 \|\lambda - \lambda'\|^2  
\numberthis\\
\|G_{x,z_1,z_2}(\lambda) - G_{x',z_1,z_2}(\lambda')\|^2
\leq &
\Big((1+ \ell_{F}^2 \gamma)^2 + (1+ \ell_{F}^2 \gamma)\ell_{g,1}\gamma \Big)\|\lambda - \lambda' \|^2 \\
&+ \Big((1+ \ell_{F}^2 \gamma)\ell_{g,1}\gamma  + \ell_{g,1}^2\gamma^2\Big) \|x - x'\|^2  .
\numberthis
\end{align*}
\end{lmm}

\begin{proof}
\label{proof:modo_sconvex_pgd_lam_update_warm}
The squared norm of the difference of  $G_{\lambda,z}(x)$ and $G_{\lambda',z}(x')$ can be bounded by
\begin{align*}
&\|G_{\lambda,z}(x) - G_{\lambda',z}(x')\|^2 \\
{=}  &
\|x - x'\|^2 - 2\alpha \langle x-x', \nabla F_z(x)\lambda - \nabla F_z(x') \lambda' \rangle 
+ \alpha^2 \|\nabla F_z(x)\lambda - \nabla F_z(x') \lambda'\|^2 \\
\stackrel{(a)}{\leq}  &
\|x - x'\|^2 - 2\alpha \langle x-x', (\nabla F_z(x) - \nabla F_z(x')) \lambda \rangle 
+ 2 \alpha^2 \|(\nabla F_z(x) - \nabla F_z(x')) \lambda \|^2  \\
&+ 2\alpha \langle x-x', \nabla F_z(x') (\lambda' - \lambda) \rangle 
+ 2 \alpha^2 \|\nabla F_z(x') ( \lambda - \lambda') \|^2 \\
\stackrel{(b)}{\leq} &
(1 - 2\alpha \mu + 2\alpha^2 \ell_{f,1}^2)\|x - x'\|^2 
+ 2\alpha \langle x-x', \nabla F_z(x') (\lambda' - \lambda) \rangle
+ 2\alpha^2 \ell_F^2 \|\lambda' - \lambda\|^2 \\
\stackrel{(c)}{\leq} &
(1 - 2\alpha \mu + 2\alpha^2 \ell_{f,1}^2)\|x - x'\|^2 
+ 2\alpha \ell_F \|x - x'\| \| \lambda' - \lambda \|
+ 2\alpha^2 \ell_F^2 \|\lambda' - \lambda\|^2 
\numberthis
\label{eq:bound_decompose_I1I2_pgd_lam_update_warm_modo}
\end{align*}
where $(a)$ follows  from rearranging and that $\|a+b\|^2 \leq 2 \|a\|^2 + 2 \|b\|^2$;  
$(b)$ follows from the $\mu$-strong convexity of $F_z(x)\lambda$, $\ell_{f,1}$-Lipschitz continuity of $\nabla F_z(x)\lambda$, and that $\|\nabla F_z(x')\| \leq \ell_F$ for $x' \in \{x_t'\}_{t=1}^T$;
and, $(c)$ follows from Cauchy-Schwartz inequality.

And $\| G_{x,z_1,z_2}(\lambda) - G_{x',z_1,z_2}(\lambda') \|$ can be bounded by
\begin{align*}
 &\| G_{x,z_1,z_2}(\lambda) - G_{x',z_1,z_2}(\lambda') \| \\
=& \|\Pi_{\Delta^{M}}(\lambda - \gamma \big(\nabla F_{z_1}(x)^\top \nabla F_{z_2}(x) + \rho \mathrm{I} \big)\lambda )
- \Pi_{\Delta^{M}}(\lambda' - \gamma \big(\nabla F_{z_1}(x')^\top \nabla F_{z_2}(x') + \rho \mathrm{I} \big) \lambda' )\| \\
\stackrel{(d)}{\leq} &
\|(1 - \gamma \rho) (\lambda - \lambda') - \gamma (\nabla F_{z_1}(x)^\top \nabla F_{z_2}(x) \lambda 
- \nabla F_{z_1}(x')^\top \nabla F_{z_2}(x') \lambda') \| \\
\stackrel{(e)}{\leq} &
\|\lambda - \lambda'\|
+ \gamma \| \nabla F_{z_1}(x)^\top \nabla F_{z_2}(x) (\lambda - \lambda') \|
+ \gamma \| (\nabla F_{z_1}(x)^\top \nabla F_{z_2}(x) - \nabla F_{z_1}(x')^\top \nabla F_{z_2}(x')) \lambda' \|\\
\stackrel{(f)}{\leq} &
\|\lambda - \lambda' \| + \gamma \ell_F^2  \|\lambda - \lambda' \|
+ \gamma \|( \nabla F_{z_1}(x)^\top \nabla F_{z_2}(x) - \nabla F_{z_1}(x')^\top \nabla F_{z_2}(x') )\lambda'\|\\
\stackrel{(g)}{\leq} &
\|\lambda - \lambda' \| + \gamma \ell_F^2 \|\lambda - \lambda' \| 
+ \gamma \big(\| (\nabla F_{z_1}(x)- \nabla F_{z_1}(x'))^\top \nabla F_{z_2}(x) \lambda'\| \\
&+\| \nabla F_{z_1}(x')^\top (\nabla F_{z_2}(x) - \nabla F_{z_2}(x')) \lambda'\| \big) \\
\stackrel{(h)}{\leq}  &
(1+ \ell_F^2  \gamma)\|\lambda - \lambda' \| + 
 (\ell_f\ell_{F,1}+\ell_F\ell_{f,1}) \gamma \|x - x'\|
\numberthis
\label{eq:lam_update_bound_modo}
\end{align*}
where $(d)$ follows from non-expansiveness of projection; $(e)$ follows from triangle inequality, $(f)$ follows from $\|\nabla F_z(x)\| \leq \ell_F$ for $x \in \{x_t'\}_{t=1}^T$, $(g)$ follows from triangle inequality; and $(h)$ follows from $\ell_{f,1}$-Lipschitz continuity of $\nabla F_z(x)\lambda'$, $\ell_{F,1}$-Lipschitz continuity of $\nabla F_z(x)$, $\|\nabla F_z(x)\| \leq \ell_F$ for $x \in \{x_t'\}_{t=1}^T$, and $\|\nabla F_z(x) \lambda'\| \leq \ell_f$ for $x \in \{x_t\}_{t=1}^T$.

Let $\ell_{g,1} = \ell_f\ell_{F,1}+\ell_F\ell_{f,1} $.
Taking square on both sides of~\eqref{eq:lam_update_bound_modo} yields
\begin{align*}
 &\| G_{x,z_1,z_2}(\lambda) - G_{x',z_1,z_2}(\lambda') \|^2 \\
\leq & \Big((1+ \ell_{F}^2 \gamma)\|\lambda - \lambda' \| + 
 \ell_{g,1}\gamma \|x - x'\| \Big)^2 \\
= & 
(1+ \ell_{F}^2 \gamma)^2\|\lambda - \lambda' \|^2 
+ 2(1+ \ell_{F}^2 \gamma)\ell_{g,1}\gamma \|\lambda - \lambda' \|  \|x - x'\|
+ \ell_{g,1}^2\gamma^2 \|x - x'\|^2 \\
\leq & (1+ \ell_{F}^2 \gamma)^2\|\lambda - \lambda' \|^2 
+ (1+ \ell_{F}^2 \gamma)\ell_{g,1}\gamma (\|\lambda - \lambda' \|^2+  \|x - x'\|^2)
+ \ell_{g,1}^2\gamma^2 \|x - x'\|^2 \\
=& \Big((1+ \ell_{F}^2 \gamma)^2 + (1+ \ell_{F}^2 \gamma)\ell_{g,1}\gamma \Big)\|\lambda - \lambda' \|^2 
+ \Big((1+ \ell_{F}^2 \gamma)\ell_{g,1}\gamma  + \ell_{g,1}^2\gamma^2\Big) \|x - x'\|^2 .
\numberthis
\label{eq:lam_update_square_bound_modo}
\end{align*}
The proof is complete.
\end{proof}

\subsubsection{Growth recursion}

\begin{lmm}[Growth recursion with approximate expansiveness]
\label{lemma:growth_recursion_approximate}
Fix an arbitrary sequence of updates $G_1, \ldots, G_T$ and another sequence $G_1^{\prime}, \ldots, G_T^{\prime}$. Let $x_0=x_0^{\prime}$ be a starting point in $\Omega$ and define $\delta_t=\left\|x_t^{\prime}-x_t\right\|$ where $x_t, x_t^{\prime}$ are defined recursively through
\begin{align*}
  x_{t+1}=G_t(x_t), \quad 
  x_{t+1}^{\prime}
  =G_t^{\prime}(x_t^{\prime}) \quad(t>0) .
\end{align*}
Let $\eta_t > 0, \nu_t \geq 0$, and $\varsigma_t \geq 0$.
Then, for any $p> 0$, and $t \in [T]$, we have the recurrence relation (with $\delta_0=0$)
\begin{align*}
\delta_{t+1}^2 & \leq
\begin{cases}
\eta_t \delta_t^2 + \nu_t,
& G_t=G_t^{\prime} \text { is } (\eta_t, \nu_t) \text {-approximately expansive in square;}\\
(1+p)
\min \{\eta_t \delta_t^2+\nu_t, \delta_t^2 \} 
+ (1+\frac{1}{p}) 4 \varsigma_t^2 
& G_t \text { and } G_t^{\prime} \text { are } \varsigma_t \text {-bounded}, \\
& G_t \text { is } (\eta_t, \nu_t) \text {-approximately expansive in square.} 
\end{cases}
\end{align*}
\end{lmm}

\begin{proof}
When $G_t$  and  $G_t^{\prime}$ are $\varsigma_t$-bounded, we can bound $\delta_{t+1}$ by
 \begin{align*}
     \delta_{t+1}
    = \|x_{t+1} - x_{t+1}'\|
    = &\|G_t(x_{t}) - G_t'(x_{t}')\|\\
    =& \|G_t(x_{t}) -x_t - G_t'(x_{t}') + x_t' + x_t - x_t'\| \\
    \leq & 
    \|G_t(x_{t}) -x_t\| + \|G_t'(x_{t}') - x_t'\| + \|x_t - x_t'\| \\
    \leq &
    2 \varsigma_t + \delta_t.
    \numberthis
    \label{eq:bounded_delta_iter}
\end{align*} 
Alternatively, when $G_t$  and  $G_t^{\prime}$ are $\varsigma_t$-bounded,
 $G_t$  is  $(\eta_t, \nu_t)$-approximately expansive, we have
\begin{align*}
     \delta_{t+1}
    = \|x_{t+1} - x_{t+1}'\|
    = &\|G_t(x_{t}) - G_t'(x_{t}')\|\\
    =& \|G_t(x_{t}) -G_t(x_{t}') + G_t(x_{t}') - G_t'(x_{t}') \| \\
    \leq &
    \|G_t(x_{t}) -G_t(x_{t}')\| + \|G_t(x_{t}') - G_t'(x_{t}') \| \\
    \leq &
    \eta_t \delta_t + \nu_t + \|G_t(x_{t}') -x_t' - G_t'(x_{t}') + x_t' \| \\
    \leq & 
    \eta_t \delta_t + \nu_t +
    \|G_t(x_{t}') -x_t'\| + \|G_t'(x_{t}') - x_t'\|  \\
    \leq &
    \eta_t \delta_t + \nu_t + 2 \varsigma_t .
\numberthis
\end{align*}
When $G_t = G_t'$, is $(\eta_t, \nu_t)$-approximately expansive in square,  given $\delta_t^2$, $\delta_{t+1}^2$ can be bounded by
\begin{align}
    \delta_{t+1}^2
    = \|x_{t+1} - x_{t+1}'\|^2
    = \|G_t(x_{t}) - G_t(x_{t}')\|^2
    \leq \eta_t \|x_t - x_t'\|^2 +\nu_t
    =\eta_t \delta_t^2 +\nu_t.
\end{align}
When $G_t$  and  $G_t^{\prime}$ are $\varsigma_t$-bounded, applying~\eqref{eq:bounded_delta_iter}, we can bound $\delta_{t+1}^2$ by
\begin{align}
  \delta_{t+1}^2 
  \leq (\delta_t + 2\varsigma_t)^2
  \leq (1+p)\delta_t^2 + (1+1/p) 4\varsigma_t^2
\end{align}
where $p> 0$ and the last inequality follows from $(a+b)^2 \leq (1+p)a^2 + (1+1/p)b^2$.

Alternatively, when $G_t$  and  $G_t^{\prime}$ are $\varsigma_t$-bounded,
 $G_t$  is  $(\eta_t, \nu_t)$-approximately expansive in square, the following holds
\begin{align*}
 \delta_{t+1}^2
= \|x_{t+1} - x_{t+1}'\|^2
=& \|G_t(x_{t}) - G_t'(x_{t}')\|^2\\
=& \|G_t(x_{t}) -G_t(x_{t}') + G_t(x_{t}') - G_t'(x_{t}') \|^2 \\
\leq &
(1+p)\|G_t(x_{t}) -G_t(x_{t}')\|^2 + (1+1/p)\|G_t(x_{t}') - G_t'(x_{t}') \|^2 \\
\leq &
(1+p)(\eta_t \delta_t^2 + \nu_t) + (1+1/p)\|G_t(x_{t}') -x_t' - G_t'(x_{t}') + x_t' \|^2 \\
\leq & 
(1+p)(\eta_t \delta_t^2 + \nu_t) +
2(1+1/p)(\|G_t(x_{t}') -x_t'\|^2 + \|G_t'(x_{t}') - x_t'\|^2)  \\
\leq &
(1+p)(\eta_t \delta_t^2 + \nu_t) + (1+1/p)4 \varsigma_t^2 .
\numberthis
\end{align*}
The proof is complete.
\end{proof}

\subsubsection{Upper bound of MOL uniform stability}\label{subs:mol_stability_upper}
In Theorem~\ref{thm:bound_E_diff_x_lam_warm_modo} we bound the argument stability, which is then used to derive the MOL uniform stability and PS generalization error in Theorem~\ref{thm:grad_stability_modo_sc}.
\begin{thm}[Argument stability of MoDo in the SC case]
\label{thm:bound_E_diff_x_lam_warm_modo}
Suppose Assumptions~\ref{assmp:lip_cont_grad_f}, \ref{assmp:sconvex}, 
hold.
Let $A$ be the MoDo algorithm in Algorithm~\ref{alg:modo}.
Choose the step sizes $\alpha_t \leq \alpha \leq \min \{1 /(2\ell_{f,1}), \mu/(2\ell_{f,1}^2) \}$,
and $\gamma_t \leq \gamma \leq {\min\{\frac{\mu^2}{120\ell_f^2\ell_{g,1}}, \frac{1}{8(3\ell_f^2  + 2\ell_{g,1})} \}}/{T} $.
Then it holds for all $t\in [T]$ that
\begin{align}
\E_A[\|A_t(S) - A_t(S')\|^2]
\leq 
\frac{48}{\mu n} \ell_f^2  
\Big(\alpha + \frac{12 + 4M \ell_f^2}{\mu n} + \frac{10 M\ell_f^4 \gamma}{\mu} \Big) .
\end{align}
\end{thm}

\begin{proof}[Theorem~\ref{thm:bound_E_diff_x_lam_warm_modo}]
\label{proof:bound_E_diff_x_lam_warm_modo}
Under  Assumptions~\ref{assmp:lip_cont_grad_f}, \ref{assmp:sconvex}, 
 Lemma~\ref{lemma:x_t_bounded_sc_smooth} implies that for $\{x_t\}$ generated by the MoDo algorithm, and for all $\lambdain$, and for all $m \in [M]$, 
\begin{equation}
\|\nabla F_{z}(x_t) \lambda \| \leq \ell_{f,1} (c_x + c_{x^*}) = \ell_{f} .
~~~{\rm and} ~~~\|\nabla F_{z}(x_t) \|\leq \|\nabla F_{z}(x_t) \|_{\rm F} \leq \sqrt{M} \ell_{f} = \ell_F.
\end{equation}
For notation simplicity, denote $\delta_t = \|x_t - x_t'\|$, $\zeta_t = \|\lambda_t - \lambda_t'\|$, $x_T = A_T(S)$ and $x_T' = A_T(S')$.
Denote the index of the different sample in $S$ and $S'$ as $i^*$, and the set of indices selected at the $t$-th iteration as $I_t$, i.e., $I_t = \{i_{t,s}\}_{s=1}^3$.
When $i^* \notin I_t$, for  any $c_1>0$, based on Lemma~\ref{lemma:modo_sconvex_pgd_lam_update_warm}, we have
\begin{align*}
\delta_{t+1}^2 
\leq & (1 - 2\alpha_t \mu + 2\alpha_t^2 \ell_{f,1}^2) \delta_t^2  
+ 2\alpha_t \ell_F \delta_t \zeta_{t+1}
+ 2\alpha_t^2 \ell_F^2 \zeta_{t+1}^2  \\  
\leq & (1 - 2\alpha_t \mu + 2\alpha_t^2 \ell_{f,1}^2) \delta_t^2 
+ \alpha_t \ell_F (c_1\delta_t^2 + c_1^{-1}\zeta_{t+1}^2)
+ 2\alpha_t^2 \ell_F^2 \zeta_{t+1}^2 \\
\leq & (1 - \alpha_t \mu ) \delta_t^2 
+ \alpha_t \ell_F (c_1\delta_t^2 + c_1^{-1}\zeta_{t+1}^2)
+ 2\alpha_t^2 \ell_F^2 \zeta_{t+1}^2
\numberthis
\end{align*}
where the second last inequality is due to Young's inequality; the last inequality is due to choosing $\alpha_t \leq \mu / (2\ell_{f,1}^2) $.

When $i^* \in I_t$,  from Lemma~\ref{lemma:bounded_agn_update}, the $(\alpha_t\ell_f)$-boundedness of the update at $t$-th iteration, and Lemma~\ref{lemma:growth_recursion_approximate}, the growth recursion, for a given constant $p>0$, we have 
\begin{align}
  \delta_{t+1}^2 
  \leq (1+p)\delta_t^2 + (1+1/p)4\alpha_t^2\ell_f^2.
\end{align}
Taking expectation of $\delta_{t+1}^2$ over $I_t$, we have
\begin{align*}
\!\!\! \E_{I_t}[\delta_{t+1}^2]  
\leq &
\mathbb{P}(i^* \notin I_t)
\Big((1 - \alpha_t \mu ) \delta_t^2 
+ \alpha_t \ell_F c_1\delta_t^2 +
(\alpha_t \ell_F c_1^{-1} + 2\alpha_t^2 \ell_F^2)\E_{I_t}[\zeta_{t+1}^2 \mid i^*\notin I_t]  \Big) \\
&\qquad\qquad +\mathbb{P}(i^* \in I_t)\Big({(1+p)} \delta_t^2 
+ {(1+1/p)} 4\alpha_t^2 \ell_f^2 \Big) \\
\leq  &
\Big(1 - \alpha_t ( \mu - \ell_F c_1 ) \mathbb{P}(i^* \notin I_t)
+ p\mathbb{P}(i^* \in I_t)\Big)\delta_t^2 \\
&+ \alpha_t \underbrace{(\ell_F c_1^{-1} + 2\alpha \ell_F^2)}_{c_{2}} \E_{I_t}[\zeta_{t+1}^2 \mid i^*\notin I_t] \mathbb{P}(i^* \notin I_t)
+ \Big(1 + \frac{1}{p}\Big) \mathbb{P}(i^* \in I_t) 4\alpha_t^2 \ell_f^2 .
\numberthis
\label{eq:delta_square_iter_modo}
\end{align*} 
At each iteration of MoDo, we randomly select three independent samples (instead of one)  from the training set $S$.
Then the probability of selecting the different sample from $S$ and $S'$ at the $t$-th iteration, $\mathbb{P}(i^* \in I_t)$ in the above equation, can be computed as follows
\begin{align*}
  \mathbb{P}(i^* \in I_t) = 1 - \Big(\frac{n-1}{n} \Big)^2 \leq \frac{2}{n}.
\numberthis
\end{align*}
Consequently, the probability of selecting the same sample from $S$ and $S'$ at the $t$-th iteration is $\mathbb{P}(i^* \notin I_t) = 1 - \mathbb{P}(i^* \in I_t) $.

Let $\ell_{g,1} = \ell_f\ell_{F,1}+\ell_F\ell_{f,1} $.
Recalling when $i^* \notin I_{t}$,  $\zeta_{t+1} \leq (1+ \ell_{F}^2 \gamma_t)\zeta_{t} + 2 \gamma_t \ell_{g,1} \delta_{t}$ from Lemma~\ref{lemma:modo_sconvex_pgd_lam_update_warm}, it follows that
\begin{align*}
  \zeta_{t+1}^2 \leq &
  \Big((1+ \ell_{F}^2 \gamma_t)^2 + (1+ \ell_{F}^2 \gamma_t)\ell_{g,1}\gamma_t \Big) \zeta_t^2 
 + \Big((1+ \ell_{F}^2 \gamma_t)\ell_{g,1}\gamma_t  + \ell_{g,1}^2\gamma_t^2\Big) \delta_t^2 \\
 \leq & 
 \big(1 + \underbrace{(3\ell_{F}^2 + 2\ell_{g,1})}_{c_{3}} \gamma_t \big) \zeta_t^2
 + 3 \ell_{g,1}\gamma_t\delta_t^2 
\numberthis
\end{align*}
where the last inequality follows from $\ell_{g,1}\gamma_t \leq 1$, and $\ell_{F}^2 \gamma_t \leq 1$. \\
And since $\zeta_t$ and $\delta_t$ are independent of $I_t$,
it follows that
\begin{align}
  \E_{I_t}[\zeta_{t+1}^2 \mid i^*\notin I_t]
  \leq 
   \big(1 + c_{3} \gamma_t \big) \zeta_t^2
 + 3 \ell_{g,1}\gamma_t\delta_t^2 .
\label{eq:E_zeta_t_plus}
\end{align}
Combining \eqref{eq:delta_square_iter_modo} and \eqref{eq:E_zeta_t_plus}, we have
\begin{align*}
\E_{I_t}[\delta_{t+1}^2]  
\leq  & 
\Big(1 - \alpha_t ( \mu - \ell_F c_1 ) \mathbb{P}(i^* \notin I_t) + p\mathbb{P}(i^* \in I_t) \Big)\delta_t^2 + 
\Big(1 + \frac{1}{p}\Big) \mathbb{P}(i^* \in I_t) 4\alpha_t^2 \ell_f^2
\\
&\quad 
+ \alpha_t c_{2} 
\Big(\big(1 + c_{3} \gamma_t \big) \zeta_t^2
 + 3 \ell_{g,1}\gamma_t\delta_t^2 \Big) \mathbb{P}(i^* \notin I_t) 
\numberthis\label{eq:E_zeta_t_plus_square} \\
=&
\Big(\eta_t + p\mathbb{P}(i^* \in I_t) \Big)\delta_t^2 
+ \alpha_t c_{2} 
\big(1 + c_{3} \gamma_t \big) \zeta_t^2
  \mathbb{P}(i^* \notin I_t) 
+ \Big(1 + \frac{1}{p}\Big) \mathbb{P}(i^* \in I_t) 4\alpha_t^2 \ell_f^2
\end{align*}
where  we define $\eta_t = 1 - \alpha_t ( \mu - \ell_F c_1 - 3 c_{2} \ell_{g,1} \gamma_t) \mathbb{P}(i^* \notin I_t)  $. \\
While when $i^* \in I_{t}$, for a given constant $p_2 > 0$, we have
\begin{align*}
   \zeta_{t+1}
 =& \|\Pi_{\Delta^{M}}(\lambda_t - \gamma_t h_{t,1}(x_t)^\top h_{t,2}(x_t) \lambda_t )
 - \Pi_{\Delta^{M}}(\lambda_t' - \gamma_t h_{t,1}'(x_t')^\top h_{t,2}'(x_t') \lambda_t' )\| \\
 \leq &
 \|\lambda_t - \lambda_t'
 - \gamma_t (h_{t,1}(x_t)^\top h_{t,2}(x_t) \lambda_t 
  - h_{t,1}'(x_t')^\top h_{t,2}'(x_t') \lambda_t' )\| \\
  \leq &
 \|\lambda_t - \lambda_t'\|
  + 2\gamma_t \ell_F\ell_f 
  \leq \zeta_t + 2\gamma_t \sqrt{M}\ell_f^2\\
  \zeta_{t+1}^2 \leq &
(1 + p_2)\zeta_t^2 + (1 + 1/p_2)4\gamma_t^2 M\ell_f^4 .
  \numberthis
\label{eq:lam_update_bound_diff_sample_modo}
\end{align*}
Taking expectation of $\zeta_{t+1}^2$ over $I_t$ gives
\begin{align*}
\E_{I_t}[\zeta_{t+1}^2] 
= &\E_{I_t}[\zeta_{t+1}^2 \mid i^* \in I_t] \mathbb{P}(i^* \in I_t)
+\E_{I_t}[\zeta_{t+1}^2 \mid i^* \notin I_t] \mathbb{P}(i^* \notin I_t)\\
\leq &   
\Big((1 + p_2)\zeta_t^2 + (1 + 1/p_2)4\gamma_t^2 M\ell_f^4\Big) \mathbb{P}(i^* \in I_t)
+ \Big((1 + c_{3} \gamma_t)\zeta_t^2
 + 3 \ell_{g,1}\gamma_t\delta_t^2\Big)\mathbb{P}(i^* \notin I_t) \\
 \leq &
 \Big(1+ c_{3}\gamma_t + \frac{3}{n}p_2 \Big)\zeta_t^2
 +(1 + \frac{1}{p_2})4\gamma_t^2 M\ell_f^4 \frac{3}{n}
+ 3\ell_{g,1}\gamma_t\delta_t^2. 
\numberthis
\label{eq:zeta_square_iter_modo}
\end{align*}
Based on linearity of expectation and applying \eqref{eq:zeta_square_iter_modo} recursively yields
\begin{align*}
  \E[\zeta_{t+1}^2] 
  \leq & \sum_{t' = 0}^{t}\Big((1 + \frac{1}{p_2})4\gamma^2 M\ell_f^4 \frac{3}{n}
+ 3\ell_{g,1}\gamma\E[\delta_{t'}^2] \Big)
\Bigg(\prod_{k=t'+1}^{t} \Big(1+ c_{3}\gamma + \frac{3}{n}p_2 \Big) \Bigg) \\
= &
\sum_{t' = 0}^{t}\Big((1 + \frac{1}{p_2})4\gamma^2 M\ell_f^4 \frac{3}{n}
+ 3\ell_{g,1}\gamma\E[\delta_{t'}^2] \Big)
 \Big(1+ c_{3}\gamma + \frac{3}{n}p_2 \Big)^{t-t'} \\
\stackrel{(a)}{\leq}&
\sum_{t' = 0}^{t}\Big((1 + \frac{8T}{n})4\gamma^2 M\ell_f^4 \frac{3}{n}
+ 3\ell_{g,1}\gamma\E[\delta_{t'}^2] \Big)
 \Big(1+ \frac{1}{2T}  \Big)^{t-t'}  \\
\stackrel{(b)}{\leq} &
 \sum_{t' = 0}^{t}\Big((1 + \frac{8T}{n})4\gamma^2 M\ell_f^4 \frac{3}{n}
+ 3\ell_{g,1}\gamma\E[\delta_{t'}^2] \Big)
 e^{\frac{1}{2}} \\
\stackrel{(c)}{\leq} &
 2\gamma\sum_{t' = 0}^{t}\Big((1 + \frac{8T}{n})4\gamma M\ell_f^4 \frac{3}{n}
+ 3\ell_{g,1}\E[\delta_{t'}^2] \Big) 
 \numberthis
 \label{eq:bound_E_zeta_square_modo}
\end{align*}
where $(a)$ follows from choosing $\gamma_t \leq \gamma \leq  1/(8c_{3}T)$, $p_2 = n /(8T)$, $(b)$ follows from $t - t' \leq T$, and  $(1+\frac{a}{T})^T \leq e^a$, and the inequality $(c)$ follows from $e^{\frac{1}{2}}< 2$.
Note that $\delta_0 =0,  \zeta_1 = 0$.
Applying \eqref{eq:delta_square_iter_modo} at $t=0$ gives
\begin{align*}
  \E[\delta_1^2]
  \leq \frac{3}{n} \left(1+ \frac{1}{p}\right)  4\alpha^2 \ell_f^2
\end{align*}
which together with \eqref{eq:zeta_square_iter_modo}   gives
\begin{align*}
   \E[ \zeta_2^2]  
   \leq  3\ell_{g,1} \gamma_1 \delta_1^2
   + \left(1 +\frac{1}{p_2}\right)4\gamma_1^2 M \ell_f^4\frac{3}{n} .
\end{align*}
Therefore, for $0 \leq t\leq 1$, it satisfies that
\begin{align}
\E[\delta_t^2]
\leq & \Bigg(\frac{3}{n} (1 + \frac{1}{p})4\alpha^2 \ell_f^2
+ 24 M\ell_f^4 c_{2} (\frac{8\gamma T}{n} + \gamma)  \frac{\alpha}{n} \Bigg)
\underbrace{\Big(\sum_{t' = 0}^{t-1}
(1 - \frac{1}{2} \alpha \mu  + \frac{3p}{n})^{t-t'-1} \Big)}_{\nu_t} \nonumber\\
=& \Bigg(\frac{3}{n} (1+ \frac{1}{p})4\alpha^2 \ell_f^2
+ 24 M\ell_f^4 c_{2} \Big( \frac{8\gamma T}{n} + \gamma \Big)  \frac{\alpha}{n}\Bigg) \nu_t .
\label{eq:delta_t_bound_c_t_modo}
\end{align}
Next, we will prove by induction that \eqref{eq:delta_t_bound_c_t_modo} also holds for $t> 1$.
Assuming that \eqref{eq:delta_t_bound_c_t_modo} holds for all $0 \leq t \leq k \leq T-1$, we   apply \eqref{eq:E_zeta_t_plus_square} to the case where $t=k$ to obtain
\begin{align*}
\E[\delta_{k+1}^2] &\leq \Big(\eta_{k} + \frac{3p}{n}\Big) \E[\delta_{k}^2] + \alpha_k c_{2} \bigg(1 + c_{3} \gamma_k \bigg) \E[ \zeta_k^2 ] \mathbb{P}(i^* \notin I_t)  + \frac{3}{n} \Big(1 + \frac{1}{p}\Big) 4\alpha_k^2 \ell_f^2 \\
&\stackrel{(a)}{\leq} \Big(\eta_{k} + \frac{3p}{n}\Big) \E[\delta_{k}^2 ] \\
&\qquad + 2\alpha_k c_{2} \gamma \Bigg(\sum_{t'=1}^{k} \Bigg((1 + \frac{8T}{n} )  \frac{12\gamma M\ell_f^4}{n} + 3\ell_{g,1}\E[\delta_{t'}^2] \Bigg) \Bigg) \mathbb{P}(i^* \notin I_t)  + \frac{3}{n} \Bigg(1 + \frac{1}{p}\Bigg) 4\alpha_{k}^2 \ell_f^2 \\
&\stackrel{(b)}{\leq} \underbrace{\Bigg( \Big( \eta_{k} + \frac{3p}{n} \Big) \nu_k + 1 + 6\alpha_k c_{2} \ell_{g,1} \gamma \Big( \sum_{t'=1}^{k} \nu_{t'} \Big) \mathbb{P}(i^* \notin I_t) \Bigg)}_{J_1} \\
&\quad\times \Bigg( \frac{3}{n} \Bigg(1 + \frac{1}{p}\Bigg) 4\alpha^2 \ell_f^2 + 24 M \ell_f^4 c_2 \Bigg( \frac{8\gamma T}{n} + \gamma \Bigg) \frac{\alpha}{n} \Bigg)
\numberthis
\label{eq:bound_delta_t_plus_modo}
\end{align*}
where $(a)$ follows from~\eqref{eq:bound_E_zeta_square_modo}, and $(b)$ follows from~\eqref{eq:delta_t_bound_c_t_modo} for $0\leq t\leq k$ and that $\gamma k \leq \gamma T \leq 1$.
The coefficient $J_1$ in \eqref{eq:bound_delta_t_plus_modo} can be further bounded by
\begin{align*}
J_1 =&\Big( \eta_{k} + \frac{3p}{n} \Big)
\nu_k + 1
+ 6\alpha_k c_{2} \ell_{g,1}  \gamma
(\sum_{t'=1}^{k} c_{t'} ) \mathbb{P}(i^* \notin I_t)  \\
\stackrel{(c)}{\leq} &
\Big( \eta_{k} + \frac{3p}{n} \Big)
\nu_k + 1
+6\alpha_k c_{2} \ell_{g,1}  k \gamma \nu_k \mathbb{P}(i^* \notin I_t)   \\
\stackrel{(d)}{\leq} &
\Big( 1 - \alpha_k ( \mu - \ell_F c_1 - 3 c_{2} \ell_{g,1} \gamma (1+2 k) ) \mathbb{P}(i^* \notin I_t)  + \frac{3p}{n} \Big)
\nu_k + 1 \\
\stackrel{(e)}{\leq} &
\Big( 1 - \frac{1}{2} \alpha \mu  + \frac{3p}{n} \Big)
\nu_k + 1
\numberthis\label{eq:bound_delta_t_plus_coeff_modo}
\end{align*}
where $(c)$ follows from $\nu_{t}\leq \nu_{t+1}$, $\gamma_t \leq \gamma$ for all $t=0,\dots, T$; 
$(d)$ follows from the definition of $\eta_k $;
$(e)$ is because $\gamma \leq  {\mu^2}/ ({120\ell_F^2\ell_{g,1}}{T}) $, 
$\alpha \leq 1/ (2\ell_{f,1}) \leq 1/ (2\mu)$ and choosing $c_1 = \mu/(4\ell_F)$ leads to
\begin{align*}
\ell_F c_1 + 3 c_{2} \ell_{g,1} \gamma (1+2 k) \gamma 
\leq  & 
\ell_F c_1
+ 6 (\ell_F c_1^{-1} + 2\alpha \ell_F^2)\ell_{g,1} (k+1) \gamma \\
\leq & \frac{1}{4} \mu + 6 (4\mu^{-1} +2\alpha ) \ell_F^2\ell_{g,1}\frac{k+1}{T}\frac{\mu^2}{120\ell_F^2\ell_{g,1}} 
\leq \frac{1}{2} \mu .
\end{align*}
Combining \eqref{eq:bound_delta_t_plus_modo} and \eqref{eq:bound_delta_t_plus_coeff_modo} implies
\begin{align}
  \E[\delta_{k+1}^2]
\leq &  
\Bigg(\Big( 1 - \frac{1}{2} \alpha \mu + \frac{3p}{n} \Big)
\nu_k + 1\Bigg)
\Bigg(\frac{3}{n} (1+ \frac{1}{p})4\alpha^2 \ell_f^2
+ 24M\ell_f^4 c_{2} \big( \frac{8\gamma T}{n} + \gamma \big)  \frac{\alpha}{n}\Bigg) \nonumber  \\
=& c_{k+1}
\Bigg(\frac{3}{n} (1+ \frac{1}{p})4\alpha^2 \ell_f^2
+ 24M\ell_f^4 c_{2} \big( \frac{8\gamma T}{n} + \gamma \big)  \frac{\alpha}{n}\Bigg) 
\label{eq:k_plus_holds_modo}
\end{align}
where the equality follows by the definition of $\nu_t$ given in \eqref{eq:delta_t_bound_c_t_modo}.
The above statements from \eqref{eq:bound_delta_t_plus_modo}-\eqref{eq:k_plus_holds_modo} show that if \eqref{eq:delta_t_bound_c_t_modo} holds for all $t$ such that $0\leq t\leq k \leq T-1$, it also holds for $t = k+1$.
Therefore, we can conclude that for all $ t\in [T]$,  it follows
\begin{align}
\E[\delta_t^2]
\leq & 
\nu_T 
\Bigg(\frac{3}{n} (1+ \frac{1}{p})4\alpha^2 \ell_f^2
+ 24M \ell_f^4 c_{2} \big( \frac{8\gamma T}{n} + \gamma \big)  \frac{\alpha}{n}\Bigg) \nonumber \\
=& \Bigg(\frac{3}{n} (1+ \frac{1}{p})4\alpha^2 \ell_f^2
+ 24M \ell_f^4 c_{2} \big( \frac{8\gamma T}{n} + \gamma \big)  \frac{\alpha}{n}\Bigg)
\Bigg(\sum_{k = 0}^{T-1}
\Big(1 - \frac{1}{2}\alpha \mu  + \frac{3p}{n} \Big)^{T-k-1} \Bigg) \nonumber \\
=& \Bigg(\frac{3}{n} (1+ \frac{12}{\alpha \mu n})4\alpha^2 \ell_f^2
+ 24M \ell_f^4 c_{2} \big( \frac{8\gamma T}{n} + \gamma \big)  \frac{\alpha}{n}\Bigg)
\Big(\frac{1}{4}\alpha \mu \Big)^{-1}
\Bigg(1 - \Big(1 - \frac{1}{4}\alpha \mu \Big)^{T} \Bigg)
\label{eq:delta_t_bound_c_T_geq_3_modo}
\end{align}
where the first inequality follows from $\nu_t \leq \nu_T$ for all $t\in [T]$; the last equality follows from taking $p = \alpha \mu n/12 $, and computing the sum of geometric series.
By  plugging in  $c_1 = \mu /(4\ell_F)$, $c_{2} = \ell_F c_1^{-1} + 2\alpha \ell_F^2$,  $c_{3} = 3\ell_{F}^2 + 2\ell_{g,1}$, for all $ t\in [T]$,we have that
\begin{align*}
\E[\delta_t^2] \leq &
\Big( \frac{3}{n }(1 + \frac{12}{\alpha \mu n} ) 4\alpha^2 \ell_f^2
+ 24 M\ell_f^4 c_{2} c_{3}^{-1} \frac{\alpha}{n^2}+24 M\ell_f^4 c_{2} \frac{ \alpha \gamma}{n }\Big)
(\frac{1}{4}\alpha \mu)^{-1} \\
\leq &
\frac{48}{\mu n} \ell_f^2  
\Big(\alpha + \frac{12}{\mu n} +\frac{2M\ell_f^2 c_{2} c_{3}^{-1}}{n} + {2M\ell_f^2 c_{2}\gamma} \Big) \\
\leq &
\frac{48}{\mu n} \ell_f^2  
\Big(\alpha + \frac{12 + 4M \ell_f^2}{\mu n} + \frac{10 M\ell_f^4 \gamma}{\mu} \Big)
\numberthis
\end{align*}
where the last inequality follows from $c_2 = \ell_F^2 (4\mu^{-1} + 2\alpha)\leq 5 M  \ell_f^2 \mu^{-1}$, and $c_2 c_3^{-1} \leq 5\ell_F^2\mu^{-1}/(3\ell_F^2) \leq 2\mu^{-1}$.
\end{proof}


\subsubsection{Lower bound of MOL uniform stability} 
\label{app_subs:lower_bound}

In this section, we construct Example~\ref{exmp:lower_bound_quadratic_SC} with a lower bound of stability for the MoDo algorithm.
Before proceeding to the example, we first define $\mathbf{1}$ as the all-one vector in $\mathbb{R}^M$, $\widetilde\Delta^{M} \coloneqq \{\lambda \in \mathbb{R}^M \mid \mathbf{1}^\top \lambda = 1 \}$, and 
$P_{\bf 1} \coloneqq \mathrm{I} - \frac{1}{M} \mathbf{1} \mathbf{1}^\top$. Then given any vector $u \in \mathbb{R}^M$, $\Pi_{\widetilde\Delta^{M}}(u) = P_{\bf 1} u + \frac{1}{M}\mathbf{1}$. 

\begin{example}\label{exmp:lower_bound_quadratic_SC}
Recall that $S = \{z_1, z_2, \dots, z_j, \dots z_n\}$, $S' = \{z_1, z_2,\dots, z_j', \dots, z_n\}$, where $S$ and $S'$ differ only in the $j$-th data point.
Define the $m$-th objective function as 
\begin{align}\label{eq:f_mz_SC_exmp1}
 f_{z,m}(x) = \frac{1}{2}  x^\top A x - b_m z^\top x  
\end{align}
where $A$ is a symmetric positive definite matrix, $\mu = 16 n^{-\frac{1}{3}} > 0$ is the smallest eigenvalue of $A$, and $v$ is the corresponding eigenvector of $A$. 
For the datasets $S$, and $S'$,
let  $z_i = c_i v$,  with $\E_{z \in S}[z] =  \mu v$, $\E_{z \in S'}[z] =  \mu' v$, $z_j - z_j' = v$, i.e., $\mu - \mu' = \frac{1}{n}$.
For simplicity, let  $M = 2$, $b = [1, 1+\sqrt{2}]^\top$ such that  $P_\mathbf{1} b = b_P = [-\frac{1}{\sqrt{2}}, \frac{1}{\sqrt{2}}]^\top$, where $b_P$ is the eigenvector of $P_\mathbf{1}$ with  eigenvalue $1$.    
\end{example}

\paragraph{Technical challenges.} 
The multi-objective learning (MOL) problem encompasses single-objective learning (SOL) when all involved objectives are identical. In such cases, the matching stability lower bound in SOL~\citep{zhang2022stability_lower} can be directly applied. 
However, 
unlike the general NC case where the MOL uniform stability for sampling-determined algorithms can be derived independently of $\gamma$, the MOL uniform stability upper bound in the SC case explicitly depends on the step size $\gamma$.
Hence, in the SC case, our goal is to establish a lower bound for MOL uniform stability that explicitly accounts for the impact of $\gamma$. This extends beyond the straightforward application of the stability lower bound in SOL.

To achieve this, one challenge lies in the projection operator when updating $\lambda$.
Unlike deriving the upper bound, where the projection operator can be handled by its non-expansive property,
due to the nature of inequality-constrained quadratic programming, neither a simple closed-form solution can be obtained, nor a  non-trivial tight lower bound can be derived in general.
We overcome this challenge by showing that when  $\gamma = \mathcal{O}(T^{-1})$, the Euclidean projection onto the simplex is equivalent to a linear operator in Lemma~\ref{lemma:proj_lambda_linear}.
Another challenge compared to deriving the lower bound for single-objective learning is that the update of MoDo involves two coupled sequences,  $\{x_t\}$ and $\{\lambda_t\}$.
The update of $x_t$ and $x_t'$ involves different weighting parameters $\lambda_t$ and $\lambda_t'$, where $\{x_t\}$, $\{\lambda_t\}$ and $\{x_t'\}$, $\{\lambda_t'\}$ are generated by the MoDo algorithm on neighboring training data $S$ and $S'$, respectively.
We overcome this challenge by deriving a recursive relation of the vector $[x_t - x_t'; \lambda_t - \lambda_t']$ in Lemma~\ref{lemma:lower_bound_v_bp}.

\paragraph{Organization of proof.} Lemma~\ref{lemma:proj_lambda_linear} proves that under proper choice of initialization of $\lambda$ and step size $\gamma$, the projection of the updated $\lambda$ onto simplex $\Delta^{M}$ is equal to the projection of that onto the set $\widetilde\Delta^{M} \coloneqq \{\lambda \in \mathbb{R}^M \mid \mathbf{1}^\top \lambda = 1 \}$. 
And thus the projection is equivalent to a linear transformation.
Thanks to Lemma~\ref{lemma:proj_lambda_linear}, we are able to derive a recursive relation the vector $[x_t - x_t'; \lambda_t - \lambda_t']$ in Lemma~\ref{lemma:lower_bound_v_bp}.
Finally, relying on the recursive relation, we derive a lower bound for the recursion of $[\E_A\|x_t - x_t'\|; \E_A\|\lambda_t - \lambda_t'\|]$, depending on a $2\times 2$ transition matrix. Based on its eigendecomposition, we could compute the $T$-th power of such a transition matrix, which is used to derive the final lower bound of $\E_A\|A(S) - A(S')\|$ in Theorem~\ref{thm:lower_bound}.

\begin{lmm}\label{lemma:proj_lambda_linear}
Suppose Assumptions~\ref{assmp:lip_cont_grad_f}, \ref{assmp:sconvex}
hold.
For MoDo algorithm,
choose $\lambda_0 = \frac{1}{M} \mathbf{1}$, $\gamma \leq \frac{1}{2M T \ell_F\ell_f }$, and  define 
\begin{equation}
\lambda_{t}^+ \coloneqq \lambda_{t} - \gamma \nabla F_{z_{t,1}}(x_t)^\top \nabla F_{z_{t,2}}(x_t) \lambda_t
\end{equation}
then the update of $\lambda_t$ for MoDo algorithm is $\lambda_{t+1} = \Pi_{\Delta^{M}}(\lambda_{t}^+)$.

Define the  set $\widetilde\Delta^{M} \coloneqq \{\lambda \in \mathbb{R}^M \mid \mathbf{1}^\top \lambda = 1 \}$, 
$P_{\bf 1} \coloneqq I - \frac{1}{M} \mathbf{1} \mathbf{1}^\top$, $\lambda_{P,t} \coloneqq \Pi_{\widetilde\Delta^{M}}(\lambda_t^+) = P_{\bf 1}\lambda_t^+ + \frac{1}{M} \mathbf{1}$.
 Then for $t = 0, \dots, T-1$, it holds that
 \begin{align}
 \lambda_{t+1} 
 = P_{\bf 1} \Big(\lambda_t - \gamma \nabla F_{z_{t,1}}(x_t)^\top \nabla F_{z_{t,2}}(x_t) \lambda_t \Big) + \frac{1}{M} \mathbf{1} =P_{\bf 1}\lambda_t^+ + \frac{1}{M} \mathbf{1}.  
 \end{align}
\end{lmm}

\begin{proof}
By the update of $\lambda_t$, we have
\begin{align*}
\|\lambda_{t+1} - \lambda_t \| 
=& \|\Pi_{\Delta^{M}} \Big(\lambda_t - \gamma \nabla F_{z_{t,1}}(x_t)^\top \nabla F_{z_{t,2}}(x_t) \lambda_t \Big) - \lambda_t\| \\
\leq & \| \lambda_t - \gamma \nabla F_{z_{t,1}}(x_t)^\top \nabla F_{z_{t,2}}(x_t) \lambda_t  - \lambda_t\| \\
\leq & \gamma \|\nabla F_{z_{t,1}}(x_t)^\top \nabla F_{z_{t,2}}(x_t) \lambda_t\| 
\leq \gamma \ell_F\ell_f 
\numberthis
\end{align*}
where the last inequality follows from Lemma~\ref{lemma:x_t_bounded_sc_smooth}, with $\ell_f = \ell_{f,1} (c_x + c_{x^*})$. 

Then for all $t \in [T-1]$, it holds that
\begin{align}
\|\lambda_t - \lambda_0\| = 
\Bigg\|\sum_{k=0}^{t-1} \lambda_{k+1} - \lambda_k \Bigg\|
\leq \sum_{k=0}^{t-1} \|\lambda_{k+1} - \lambda_k \|
\leq \gamma t \ell_F\ell_f  
\leq  \frac{t}{2MT}
\end{align}
where the last inequality follows from $\gamma \leq \frac{1}{2M T \ell_F\ell_f }$.

Then for $t \in [T-1]$, it holds that
\begin{align*}
 \|\lambda_{t}^+ - \lambda_0\| \leq & \|\lambda_{t}^+ -\lambda_t \| + \|\lambda_t - \lambda_0 \|  \\
\leq & \gamma \|\nabla F_{z_{t,1}}(x_t)^\top \nabla F_{z_{t,2}}(x_t) \lambda_t\| + \gamma t \ell_F\ell_f  
\leq \gamma (t+1) \ell_F\ell_f  
\leq \gamma T \ell_F\ell_f  \leq \frac{1}{2M} .
\numberthis \label{eq:bound_diff_lam_t_m_noproj_lam_0}
\end{align*}
By the update of $\lambda_t$, and the definition of projection,
\begin{align}
 \lambda_{t+1} = \Pi_{\Delta^{M}}(\lambda_t^+) = \mathop{\arg\min}_{\lambdain} \|\lambda - \lambda_t^+\|^2 .
\end{align}
Also we have 
\begin{align}
  \lambda_{P,t}  = \Pi_{\widetilde\Delta^{M}}(\lambda_t^+) = \mathop{\arg\min}_{\lambda\in {\widetilde\Delta^{M}}} \|\lambda - \lambda_t^+\|^2.
\end{align}
Let $\lambda_{P,t} = [\lambda_{P,t,1},\dots, \lambda_{P,t,M}]^\top$. Then it holds that
\begin{align*}
 |\lambda_{P,t,m} - \lambda_{0,m}|
 \leq \|\lambda_{P,t} - \lambda_{0}\| = \|\Pi_{\widetilde\Delta^{M}}(\lambda_t^+) - \lambda_0\| 
 \stackrel{(a)}{\leq} \|\lambda_t^+ - \lambda_0\| 
 \stackrel{(b)}{\leq} \frac{1}{2M}
\end{align*}
where $(a)$ follows from non-expansiveness of projection and that $\lambda_0 \in \widetilde\Delta^{M}$; $(b)$ follows from~\eqref{eq:bound_diff_lam_t_m_noproj_lam_0}.

Therefore, each element of $\lambda_{P,t}$ satisfies
\begin{align}
0 \leq \frac{1}{M} - \frac{1}{2M} 
\leq \lambda_{0,m} - |\lambda_{P,t,m} - \lambda_{0,m} | 
\leq \lambda_{P,t,m}
\leq \lambda_{0,m} + |\lambda_{P,t,m} - \lambda_{0,m} |
\leq \frac{3}{2M} \leq 1 
\end{align}
which shows that $\lambda_{P,t} = \Pi_{\widetilde\Delta^{M}}(\lambda_t^+) \in \Delta^{M}$.
Therefore it holds that,
\begin{align}
\|\lambda_{P,t} - \lambda_t^+\|^2 
\stackrel{(c)}{\geq} \min_{\lambdain} \|\lambda - \lambda_t^+\|^2 \stackrel{(d)}{\geq} \min_{\lambda \in {\widetilde\Delta^{M}}} \|\lambda - \lambda_t^+\|^2
\stackrel{(e)}{=} \|\lambda_{P,t} - \lambda_t^+\|^2
\end{align}
where $(c)$ is because $\lambda_{P,t} \in \Delta^{M}$; $(d)$ is because $\Delta^{M} \subset \widetilde\Delta^{M}$ by the definition of the simplex; $(e)$ is because $\lambda_{P,t} = \Pi_{\widetilde\Delta^{M}}(\lambda_t^+)$.
Then the equality holds that
\begin{align}
\|\lambda_{P,t} - \lambda_t^+\|^2  =  \min_{\lambdain} \|\lambda - \lambda_t^+\|^2 
\end{align}
and 
\begin{align}
 P_{\bf 1}\lambda_t^+ + \frac{1}{M} \mathbf{1} = \lambda_{P,t} = {\arg\min}_{\lambdain} \|\lambda - \lambda_t^+\|^2 
= \Pi_{\lambdain}(\lambda_t^+)
= \lambda_{t+1}.
\end{align}
The proof is complete.
\end{proof}

With the help of Lemma~\ref{lemma:proj_lambda_linear}, which simplifies the Euclidean projection operator as a linear operator, we then prove in Lemma~\ref{lemma:lower_bound_v_bp}, the recursive relation of $x_t - x_t'$ and $\lambda_t - \lambda_t'$.

\begin{lmm}\label{lemma:lower_bound_v_bp}
Suppose Assumptions~\ref{assmp:lip_cont_grad_f}, \ref{assmp:sconvex}
hold.
Under Example~\ref{exmp:lower_bound_quadratic_SC}, choose $\lambda_0 = \frac{1}{M} \mathbf{1}$, $\gamma \leq \frac{1}{2M T \ell_F\ell_f }$ for the MoDo algorithm. 
Denote $\{x_t\}$, $\{\lambda_t\}$ and $\{x_t'\}$, $\{\lambda_t'\}$ as the sequences  generated by the MoDo algorithm with dataset $S$ and  $S'$, respectively.
Then it holds that  
\begin{align}
 x_{t} - x_t' = \varphi_{x,t} v,\quad \text{and} \quad 
 \lambda_t - \lambda_t' = \varphi_{\lambda,t}  b_P 
\end{align}
and $\varphi_{x,t} $, $\varphi_{\lambda,t} $ satisfy the following recursion
\begin{align*}
& \varphi_{x,t+1} = (1  - \alpha \mu + \alpha  \gamma c_{t,3} c_{t,1}  \mu ) \varphi_{x,t} 
 +\alpha c_{t,3} (1 -\gamma c_{t,1} c_{t,2} ) \varphi_{\lambda,t} 
 + \mathds{1}(i_{t,3} = j) \alpha 
 (b^\top \lambda_{t+1}') \\
 & \qquad \qquad  
 + \mathds{1}(i_{t,1} = j) 
 \gamma \alpha  c_{t,3} (\mu v^\top x_t' - c_{t,2} b^\top \lambda_t')
 - \mathds{1}(i_{t,2} = j) \gamma \alpha  c_{t,3} c_{t,1}' 
  b^\top \lambda_t' 
\numberthis \\ 
& \varphi_{\lambda,t+1} = ( 1  -\gamma c_{t,1} c_{t,2} ) \varphi_{\lambda,t} 
 +\gamma  c_{t,1} \mu \varphi_{x,t}\\
&\qquad \qquad + \mathds{1}(i_{t,1} = j) 
 \gamma \big(\mu ( v^\top x_t') - c_{t,2} (b^\top \lambda_t') \big)
 -\mathds{1}(i_{t,2} = j) \gamma
 c_{t,1}' (b^\top \lambda_t') 
\numberthis
\end{align*}
where $\mathds{1}(\cdot)$ is the indicator function.
\end{lmm}

\begin{proof}
Denote $z_{t,s}$, and $z_{t,s}'$, $s \in [3]$, as the samples selected in the $t$-th iteration  from $S$ and $S'$, respectively.
According to the MoDo algorithm  update of $x_t$, and the definition of the problem in~\eqref{eq:f_mz_SC_exmp1}, we have
\begin{align*}
x_{t+1} 
=& x_t - \alpha \nabla F_{Z_{t+1}}(x_t) \lambda_{t+1} 
= x_t - \alpha \big[A x_t - b_1 \bar{z}_{t+1}, \dots, A x_t  - b_M \bar{z}_{t+1}\big] \lambda_{t+1} \\ 
=& x_t  - \alpha A x_t 
+ \alpha \bar{z}_{t+1} (b^\top \lambda_{t+1}) .
\numberthis
\end{align*}  
The difference $x_{t+1} - x_{t+1}'$ can be computed by
\begin{align*}
 x_{t+1} - x_{t+1}'  
= &\big(  I  - \alpha A \big) (x_t - x_t')
+ \alpha  ( \bar{z}_{t+1} b^\top \lambda_{t+1}  - \bar{z}_{t+1}' b^\top \lambda_{t+1}' ) \\
=& \big(  I  - \alpha A \big) (x_t - x_t')
+ \alpha \bar{z}_{t+1} b^\top  (\lambda_{t+1} -  \lambda_{t+1}' )  
+ \alpha ( \bar{z}_{t+1} - \bar{z}_{t+1}') b^\top \lambda_{t+1}'  \\
=& \big(  I  - \alpha A \big) (x_t - x_t')
+ \alpha \bar{z}_{t+1}  b^\top (\lambda_{t+1} - \lambda_{t+1}' ) 
+ \mathds{1} (i_{t,3}=j) \alpha v b^\top \lambda_{t+1}' 
\numberthis\label{eq:x_iter_general}
\end{align*}
where $\mathds{1}(\cdot)$ denotes the indicator function,
and the last equation follows from that 
$\bar{z}_{t+1} - \bar{z}_{t+1}' = 0$ if $i_{t,3}\neq j$, and 
$\bar{z}_{t+1} - \bar{z}_{t+1}' = z_j - z_j' = v$ if $j \in I_{t+1} $.

By Lemma~\ref{lemma:proj_lambda_linear}, in Example~\ref{exmp:lower_bound_quadratic_SC}, $\lambda_{t+1} 
 = P_{\bf 1} \big(\lambda_t - \gamma \nabla F_{z_{t,1}}(x_t)^\top \nabla F_{z_{t,2}}(x_t) \lambda_t \big) +\frac{1}{M} \mathbf{1}$, 
which can be further derived as
\begin{align*}
\lambda_{t+1} 
 =& P_{\bf 1} \big(\lambda_t - \gamma \nabla F_{z_{t,1}}(x_t)^\top \nabla F_{z_{t,2}}(x_t) \lambda_t \big) +\frac{1}{M} \mathbf{1}\\
 =& P_{\bf 1} \Big(\lambda_t - \gamma  \big(\mathbf{1}   x_t^\top A - b z_{t,1}^\top \big) \big( A x_t - z_{t,2} b^\top \lambda_{t} \big)  \Big) +\frac{1}{M} \mathbf{1} \\
 \stackrel{(a)}{=}& \lambda_t - \gamma  \big(P_{\bf 1} \mathbf{1}   x_t^\top A - P_{\bf 1} b z_{t,1}^\top \big) \big( A x_t - z_{t,2} b^\top \lambda_{t} \big) \\
\stackrel{(b)}{=}& \lambda_t +\gamma b_P z_{t,1}^\top  \big( A x_t - z_{t,2} b^\top \lambda_{t} \big) 
\numberthis\label{eq:lam_update_lower_bound_P}
\end{align*}
where $(a)$ follows from rearranging the equation and that $P_{\bf 1} \lambda_t + \frac{1}{M} \mathbf{1} = \Pi_{\widetilde\Delta^{M}} (\lambda_t) = \lambda_t$ as $\lambda_t \in \widetilde\Delta^{M}$; $(b)$ follows from that $P_{\bf 1} b = b_P$ and $P_{\bf 1} \mathbf{1} = 0$.

The difference $\lambda_{t+1} - \lambda_{t+1}'$ can be derived as
\begin{align*}
  \lambda_{t+1} - \lambda_{t+1}' 
  {=}&   (\lambda_t - \lambda_t')
  + \gamma b_P \big( z_{t,1}^\top A x_t -  z_{t,1}'^\top A x_t' \big) 
  -\gamma b_P \big( z_{t,1}^\top  z_{t,2} b^\top \lambda_t
 - z_{t,1}'^\top  z_{t,2}'  b^\top \lambda_t' \big) \\
 \stackrel{(c)}{=}&   (\lambda_t - \lambda_t')
  + \gamma b_P z_{t,1}^\top A (x_t - x_t')
  + \gamma b_P (z_{t,1} - z_{t,1}' )^\top A x_t'    
  -\gamma b_P z_{t,1}^\top  z_{t,2} b^\top (\lambda_t - \lambda_t') \\
  &- \gamma b_P (z_{t,1} - z_{t,1}')^\top  z_{t,2} b^\top  \lambda_t'
  - \gamma b_P  z_{t,1}'^\top ( z_{t,2} - z_{t,2}') b^\top  \lambda_t' \\
  \stackrel{(d)}{=}&
    (\lambda_t - \lambda_t')
  + \gamma b_P z_{t,1}^\top A (x_t - x_t')
  - \gamma b_P z_{t,1}^\top  z_{t,2} b^\top (\lambda_t - \lambda_t') \\
  & + \mathds{1}(i_{t,1} = j) \gamma b_P (\mu v^\top x_t'    
   - v^\top  z_{t,2} b^\top  \lambda_t')
  - \mathds{1}(i_{t,2} = j) \gamma b_P  z_{t,1}'^\top v b^\top  \lambda_t'
  \numberthis\label{eq:lam_iter_general}
\end{align*}
where 
$(c)$ follows from rearranging the equation; 
$(d)$ follows from that 
$z_{t,s} - z_{t,s}' = 0$ if $i_{t,s}\neq j$, and 
$z_{t,s} - z_{t,s}' = z_j - z_j' = v$ if $i_{t,s}= j$, and that $A v = \mu v$.

Combining~\eqref{eq:x_iter_general} and \eqref{eq:lam_iter_general} gives
\begin{align*}
 \begin{bmatrix}
 x_{t+1} - x_{t+1}' \\
 \lambda_{t+1} - \lambda_{t+1}' 
 \end{bmatrix}
 \!=\!&
 \begin{bmatrix}
 C_{x,x,t} & C_{x,\lambda,t} \\
 C_{\lambda,x,t} & C_{\lambda,\lambda,t}
 \end{bmatrix}
 \begin{bmatrix}
 x_{t} - x_{t}' \\
 \lambda_{t} - \lambda_{t}' 
 \end{bmatrix}
 \!+ \!\mathds{1}(i_{t,3} = j) \alpha 
 \begin{bmatrix}
  v b^\top \lambda_{t+1}' \\
 0
 \end{bmatrix}  \!-\!\mathds{1}(i_{t,2} = j) 
 \gamma   
  \begin{bmatrix}
  \alpha c_{t,3} c_{t,1}' v 
  b^\top \lambda_t' \\
  b_P c_{t,1}'     
  b^\top \lambda_t' 
  \end{bmatrix}\\
 &  + \mathds{1}(i_{t,1} = j) \gamma 
 \begin{bmatrix}
 \alpha c_{t,3} v (\mu  v^\top x_t' - c_{t,2}  b^\top \lambda_t') \\
 b_P (\mu  v^\top x_t' - c_{t,2}  b^\top \lambda_t')  
 \end{bmatrix}
  \numberthis\label{eq:vec_x_lam_iter_general}
\end{align*}
where the matrices are defined as 
\begin{subequations}
\begin{align}
 \label{eq:C_x_x_t}
 C_{x,x,t} =&   I  - \alpha  A + \alpha  \gamma \bar{z}_{t+1} b^\top b_P z_{t,1}^\top A  
 = I  - \alpha  A + \alpha  \gamma c_{t,3} c_{t,1}  \mu v v^\top \\
 \label{eq:C_x_lam_t}
 C_{x,\lambda,t} =& \alpha \bar{z}_{t+1} b^\top  ( I  -\gamma b_P z_{t,1}^\top z_{t,2} b^\top)
 = \alpha c_{t,3}  v b^\top ( 1 -\gamma c_{t,1} c_{t,2} ) \\
 \label{eq:C_lam_x_t}
C_{\lambda,x,t} =& \gamma b_P z_{t,1}^\top A 
= \gamma  c_{t,1} \mu b_P v^\top  \\
\label{eq:C_lam_lam_t}
C_{\lambda,\lambda,t} =&  ( I  -\gamma b_P z_{t,1}^\top z_{t,2} b^\top)
=  ( I  -\gamma c_{t,1} c_{t,2} b_P  b^\top).
\end{align}
\end{subequations}
Next we show by induction that 
\begin{align}\label{eq:x_diff_lam_diff_v_bp}
 x_{t} - x_t' = \varphi_{x,t} v,\quad \text{and} \quad 
 \lambda_t - \lambda_t' =  \varphi_{\lambda,t}  b_P .  
\end{align}
First, when $t=0$, $x_0 - x_0' = 0 = \varphi_{x,0} v$ and $\lambda_0 - \lambda_0' =  \varphi_{\lambda,0}  b_P$ with $\varphi_{x,0} = 0$ and $\varphi_{\lambda,0} = 0$. Therefore~\eqref{eq:x_diff_lam_diff_v_bp} holds at $t=0$.
Supposing that~\eqref{eq:x_diff_lam_diff_v_bp} holds for $t = k $, next we show that it also holds at $t = k+1$.

At $t = k+1$, applying~\eqref{eq:vec_x_lam_iter_general} for $x_{k+1} - x_{k+1}'$, and substituting~\eqref{eq:C_x_x_t},~\eqref{eq:C_x_lam_t} yields 
\begin{align*}
  x_{k+1} - x_{k+1}' 
 = &\begin{bmatrix}
 C_{x,x,k} & C_{x,\lambda,k} 
 \end{bmatrix}
 \begin{bmatrix}
 x_k - x_k' \\
 \lambda_k - \lambda_k' 
 \end{bmatrix} 
 +\mathds{1}(i_{k,3} = j) \alpha v b^\top \lambda_{k+1}' \\
 & \qquad +\mathds{1}(i_{k,1} = j) 
 \gamma \alpha c_{k,3} v (\mu v^\top x_k' - c_{k,2}  b^\top \lambda_k')
  - \mathds{1}(i_{k,2} = j) \gamma  \alpha c_{k,3} c_{k,1}' v b^\top \lambda_k' \\
 =& (I  - \alpha  A + \alpha \gamma c_{k,3} c_{k,1} \mu v v^\top) 
 \varphi_{x,k} v
 + \alpha c_{k,3} v ( 1 -\gamma c_{k,1} c_{k,2} ) \varphi_{\lambda,k} 
 +\mathds{1}(i_{k,3} = j) \alpha v b^\top \lambda_{k+1}' \\
 & + \mathds{1}(i_{k,1} = j) \gamma  \alpha c_{k,3} v (\mu v^\top x_k' - c_{k,2}  b^\top \lambda_k')
 - \mathds{1}(i_{k,2} = j) \gamma  \alpha c_{k,3} c_{k,1}' v 
  b^\top \lambda_k' \\
 =& 
 (1  - \alpha \mu + \alpha  \gamma c_{k,3} c_{k,1}  \mu ) \varphi_{x,k} v
 + \alpha c_{k,3} (1 -\gamma c_{k,1} c_{k,2} )
 \varphi_{\lambda,k} v 
 +\mathds{1}(i_{k,3} = j) \alpha (b^\top \lambda_{k+1}') v  \\
 &+\mathds{1}(i_{k,1} = j) \gamma  \alpha c_{k,3} v (\mu v^\top x_k' - c_{k,2}  b^\top \lambda_k')
 - \mathds{1}(i_{k,2} = j) \gamma  \alpha c_{k,3} c_{k,1}' v b^\top \lambda_k' \\
 =& \varphi_{x,k+1} v
\numberthis
\end{align*}
where $\varphi_{x,k+1}$ is computed by
\begin{align*}
\varphi_{x,k+1} =&   (1  - \alpha \mu + \alpha  \gamma c_{k,3} c_{k,1}  \mu ) \varphi_{x,k} 
 +\alpha c_{k,3} (1 -\gamma c_{k,1} c_{k,2} ) \varphi_{\lambda,k} 
 +\mathds{1}(i_{k,3} = j) \alpha (b^\top \lambda_{k+1}') \\
 &+\mathds{1}(i_{k,1} = j) \gamma  \alpha c_{k,3} (\mu v^\top x_k' - c_{k,2}  b^\top \lambda_k')
 - \mathds{1}(i_{k,2} = j) \gamma \alpha c_{k,3} c_{k,1}' 
  b^\top \lambda_k'.
\numberthis
\end{align*}
Therefore, for all $t \in [T]$, it holds that $x_{t} - x_t' = \varphi_{x,t} v $.

At $t = k+1$, apply~\eqref{eq:vec_x_lam_iter_general} for $\lambda_{k+1} - \lambda_{k+1}'$, and substitute~\eqref{eq:C_lam_x_t},~\eqref{eq:C_lam_lam_t} yields
\begin{align*}
  \lambda_{k+1} - \lambda_{k+1}' 
 = &\begin{bmatrix}
 C_{\lambda,x,k} & C_{\lambda,\lambda,k} 
 \end{bmatrix}
 \begin{bmatrix}
 x_k - x_k' \\
 \lambda_k - \lambda_k' 
 \end{bmatrix} \\
 &\quad  +\mathds{1}(i_{k,1} = j) 
 \gamma  \big(\mu b_P  v^\top x_k' - c_{k,2}  b_P  b^\top \lambda_k' \big)
 -\mathds{1}(i_{k,2} = j) \gamma c_{k,1}' b_P b^\top \lambda_k' 
 \\
 =& \gamma c_{k,1} \mu b_P v^\top \varphi_{x,k} v
 + ( I -\gamma c_{k,1} c_{k,2} b_P b^\top) b_P 
 \varphi_{\lambda,k}  \\
 &  + \mathds{1}(i_{k,1} = j) \gamma  
 \big( \mu b_P  v^\top x_k' - c_{k,2} b_P  b^\top \lambda_k' \big)
 -\mathds{1}(i_{k,2} = j) 
 \gamma c_{k,1}' b_P b^\top \lambda_k' \\
 =& \gamma  c_{t,1} \mu \varphi_{x,k} b_P   
 +  ( 1 -\gamma c_{k,1} c_{k,2} ) \varphi_{\lambda,k} b_P \\
 &  +\mathds{1}(i_{k,1} = j) \gamma 
 \big(\mu ( v^\top x_k')  - c_{k,2} (b^\top \lambda_k' ) \big) b_P 
 -\mathds{1}(i_{k,2} = j) \gamma 
 c_{k,1}' (b^\top \lambda_k') b_P   \\
 =& \varphi_{\lambda,k+1} b_P
\numberthis
\end{align*}
where $\varphi_{\lambda,k+1}$ is computed by
\begin{align*}
\varphi_{\lambda,k+1} = &
\gamma  c_{k,1} \mu \varphi_{x,k} +  ( 1  -\gamma c_{k,1} c_{k,2} ) \varphi_{\lambda,k} \\
 &+ \mathds{1}(i_{k,1} = j) 
 \gamma \big(\mu ( v^\top x_k') - c_{k,2} (b^\top \lambda_k' ) \big)
 -\mathds{1}(i_{k,2} = j) 
 \gamma c_{k,1}' (b^\top \lambda_k')  .
\numberthis
\end{align*}
Therefore, for all $t \in [T]$, it holds that $\lambda_t - \lambda_t' =  \varphi_{\lambda,t}  b_P $.
\end{proof}

Lemma~\ref{lemma:lower_bound_v_bp} provides the recursive relation of $x_t - x_t'$ and $\lambda_t - \lambda_t'$.
And Lemma~\ref{lemma:grad_geq_0} below provides another property used to derive the lower bound of $\E[\|x_T - x_T'\|]$ in Theorem~\ref{thm:lower_bound}.

\begin{lmm}\label{lemma:grad_geq_0}
Suppose Assumptions~\ref{assmp:lip_cont_grad_f}, \ref{assmp:sconvex}
hold.
Under Example~\ref{exmp:lower_bound_quadratic_SC}, choose $x_0 = x_0' = 7v$, $\alpha = \frac{1}{4\mu T}$ for the MoDo algorithm. 
Denote $\{x_t\}$, $\{\lambda_t\}$ and $\{x_t'\}$, $\{\lambda_t'\}$ as the sequences  generated by the MoDo algorithm with dataset $S$ and  $S'$, respectively.
Then it holds that  
\begin{align}
v^\top \E_A [A x_t' - z_{t,2} b^\top \lambda_t'] \geq 0 
\quad\text{and}\quad
b^\top\E_A[\lambda_{t+1}'] \geq b^\top\E_A[\lambda_{t}']. 
\end{align}  
\end{lmm}

\begin{proof}
From the update of $x_t'$, we have   
\begin{align*}
x_{t+1}' 
=& x_t'  - \alpha A x_t' 
+ \alpha Z_{t+1}' (b^\top \lambda_{t+1}') \\
=& x_t'  - \alpha A x_t' 
+ \alpha c_{t,3}' v (b^\top \lambda_{t+1}') 
\numberthis
\end{align*}
Suppose $x_t' = c_{x,t}'v$, then $x_{t+1}' = c_{x,t+1}'v$ with
\begin{align*}
 c_{x,t+1}' =  (1 - \alpha \mu) c_{x,t}' + \alpha c_{t,3}'  (b^\top \lambda_{t+1}') \\
 \text{and}~~
 \E_A[x_{t+1}' ] =  (1 - \alpha \mu) \E_A[x_{t}' ] + \alpha \mu' v \E_A(b^\top \lambda_{t+1}')
\end{align*}
Applying the above inequality recursively gives
\begin{align*}
 v^\top\E_A[x_t' ]
=&  v^\top(1 - \alpha \mu)^t x_0 + \alpha \mu'  \Big(\sum_{t'=0}^{t-1} (1 - \alpha \mu)^{t - 1 - t'} \E_A(b^\top \lambda_{t'+1}')\Big) \\
\geq &  v^\top(1 - \alpha \mu)^t x_0 + \alpha \mu'  \frac{1 - (1 - \alpha \mu)^t}{\alpha \mu} 
\tag*{\text{$b^\top \lambda \geq 1$ for all $\lambdain$}}\\
=&  (1 - \alpha \mu)^t \big(v^\top x_0 - \mu'\mu^{-1} \big) +  \mu'\mu^{-1}   .
\end{align*}
Since $x_0 = 7v$, $\mu'\mu^{-1} \leq 1$, it holds that
\begin{align*}
 v^\top \E_A[x_t' ]
 =& (1 - \alpha \mu)^t \big(7 - \mu'\mu^{-1}  \big) +  \mu'\mu^{-1}  
 \geq  
 6(1 - \alpha \mu)^t +  \mu'\mu^{-1}  .
\end{align*}

Then it follows that
\begin{align*}
v^\top \E_A [A x_t' - z_{t,2} b^\top \lambda_t'] 
=& \mu\E_A [ v^\top x_t' -  b^\top \lambda_t'] 
\geq 
\mu \Big( 6(1 - \alpha \mu)^t + \mu'\mu^{-1} - (1+\sqrt{2}) \Big) \\
\geq &
\mu \Big( 6(1 -\alpha \mu t )   - (1+\sqrt{2}) \Big)  
\tag*{\text{$\mu'\mu^{-1}\geq 0$}} \\
\geq &
\mu \Big( 6(1 - \frac{1}{4})   - (1+\sqrt{2}) \Big) 
\geq 0
\tag*{\text{$\alpha = \frac{1}{4\mu T}$}}
\end{align*} 
By the update of $\lambda_t'$ from \eqref{eq:lam_update_lower_bound_P}, 
\begin{align*}
 b^\top \E_A[\lambda_{t+1}'-\lambda_t'] 
 {=}& b^\top 
 \gamma b_P \E_A[z_{t,1}'^\top  \big( A x_t' - z_{t,2}' b^\top \lambda_{t}' \big) ]\\
 =& \gamma \E_A[ \mu'v^\top \big( A x_t' - z_{t,2}' b^\top \lambda_{t}' \big) ] 
 \tag*{\quad\quad\text{$\E_A[z_{t,1}'] = \mu' v$, $z_{t,1}'$ independent of $\lambda_{t}'$ and $x_{t}'$}} \\
 =& \gamma \mu'v^\top  \E_A[   A x_t' - \mu'v b^\top \lambda_{t}'  ] 
 \tag*{\text{$\E_A[z_{t,2}'] = \mu' v$, $z_{t,2}'$ independent of $\lambda_{t}'$}}\\
 \geq & \gamma \mu'v^\top  \E_A[   A x_t' - \mu v b^\top \lambda_{t}'  ] 
 \tag*{\text{$\mu'\leq \mu $, $b^\top \lambda \geq 1$ for all $\lambdain$}} \\
 =& \gamma \mu'v^\top  \E_A[   A x_t' - z_{t,2} b^\top \lambda_{t}'  ] \geq 0
 \tag*{\text{$\E_A[z_{t,2}] = \mu v$, $z_{t,2}$ independent of $\lambda_{t}'$}}
 .
\end{align*}
The proof is complete.
\end{proof}

\begin{thm}\label{thm:lower_bound}
Suppose Assumptions~\ref{assmp:lip_cont_grad_f} and \ref{assmp:sconvex} hold.
Under Example~\ref{exmp:lower_bound_quadratic_SC} with $M = 2$, choose $\lambda_0 = \frac{1}{M} \mathbf{1}$, $x_0 = x_0' = 7v$, $\alpha = \frac{1}{4\mu T}$,  $0< \gamma \leq \frac{1}{2M T \ell_F \ell_f}$, $\rho=0$, and $T\leq 4 n^\frac{2}{3}$ for the MoDo algorithm. 
Denote $\{x_t\}$, $\{\lambda_t\}$ and $\{x_t'\}$, $\{\lambda_t'\}$ as the sequences  generated by the MoDo algorithm with dataset $S$ and  $S'$, respectively.
Then it holds that  
\begin{align}
\E[\|x_T - x_T'\|] \geq \frac{ \gamma T}{2n^2} + \frac{1}{16 n} .
\end{align}   
\end{thm}

\begin{proof}
Denote $\delta_t = \|x_t - x_t'\|$, $\zeta_t = \|\lambda_t - \lambda_t'\|$.
From Lemma~\ref{lemma:lower_bound_v_bp}, it holds that
\begin{align}
&\E[\delta_{t+1}] = \E[ |\varphi_{x,t+1}| \|v\|] = \E[ |\varphi_{x,t+1}| ] \geq \E[ \varphi_{x,t+1}] \\
&\E[\zeta_{t+1}] = \E[ |\varphi_{\lambda,t+1}| \|v\|] = \E[ |\varphi_{\lambda,t+1}| ]
\geq \E[\varphi_{\lambda,t+1}]
\end{align}
where $\varphi_{x,t} $, $\varphi_{\lambda,t} $ satisfy
\begin{align*}
& \varphi_{x,t+1} = (1  - \alpha \mu + \alpha  \gamma c_{t,3} c_{t,1}  \mu ) \varphi_{x,t} 
 +\alpha c_{t,3} (1 -\gamma c_{t,1} c_{t,2} ) \varphi_{\lambda,t} 
 + \mathds{1}(i_{t,3} = j) \alpha 
 (b^\top \lambda_{t+1}') \\
 & \qquad \qquad  
 + \mathds{1}(i_{t,1} = j) 
 \gamma \alpha  c_{t,3} (\mu v^\top x_t' - c_{t,2} b^\top \lambda_t')
 - \mathds{1}(i_{t,2} = j) \gamma \alpha  c_{t,3} c_{t,1}' 
  b^\top \lambda_t' 
\numberthis\\ 
& \varphi_{\lambda,t+1} = ( 1  -\gamma c_{t,1} c_{t,2} ) \varphi_{\lambda,t} 
 +\gamma  c_{t,1} \mu \varphi_{x,t}\\
&\qquad \qquad + \mathds{1}(i_{t,1} = j) 
 \gamma \big(\mu ( v^\top x_t') - c_{t,2} (b^\top \lambda_t') \big)
 -\mathds{1}(i_{t,2} = j) \gamma
 c_{t,1}' (b^\top \lambda_t'). 
\numberthis
\end{align*}

The expectation of $\varphi_{x,t+1}$ can be further bounded as
\begin{align*}
\E[\varphi_{x,t+1}] = &
\E [(1  - \alpha \mu + \alpha  \gamma c_{t,3} c_{t,1}  \mu ) \varphi_{x,t} 
 +\alpha c_{t,3} (1 -\gamma c_{t,1} c_{t,2} ) \varphi_{\lambda,t} 
 + \frac{1}{n} \alpha 
 (b^\top \lambda_{t+1}') \\
 & 
 + \frac{1}{n}
 \gamma \alpha  c_{t,3} (\mu v^\top x_t' - c_{t,2} b^\top \lambda_t')
 - \frac{1}{n} \gamma \alpha  c_{t,3} c_{t,1}' 
  b^\top \lambda_t'] \\
\stackrel{(a)}{\geq} & 
(1 - \alpha \mu (1 - \gamma \mu^2) ) \E[\varphi_{x,t} ]
 +\alpha \mu (1 - \gamma \mu^2) \E[\varphi_{\lambda,t}]
 + \frac{1}{n} \alpha b^\top \E[\lambda_{t+1}']
 - \frac{1}{n} \gamma \alpha \mu \mu' b^\top \E[\lambda_t']\\
 \geq & 
 (1 - \alpha \mu (1 - \gamma \mu^2) ) \E[\varphi_{x,t} ]
 +\alpha \mu (1 - \gamma \mu^2) \E[\varphi_{\lambda,t}]
 + \frac{1}{n} \alpha b^\top \E[\lambda_t'] (1 - \gamma \mu^2)
\numberthis
\end{align*}
where $(a)$ follows from $\E_A[c_{t,s}] = \mu$,  $\E_A[c_{t,s}'] = \mu' \leq \mu$, $\E[\lambda_{t+1}'] \geq \E[\lambda_{t}']$ by Lemma~\ref{lemma:grad_geq_0}, and the fact that $c_{t,s}$ is independent of $\varphi_{x,t}$.

Similarly, the expectation of $\varphi_{\lambda,t+1}$ can be further bounded as
\begin{align*}
\E[\varphi_{\lambda,t+1} ] =&
\E[ ( 1 -\gamma c_{t,1} c_{t,2} ) \varphi_{\lambda,t} 
 +\gamma  c_{t,1} \mu \varphi_{x,t}
 + \frac{1}{n} \gamma \big(\mu ( v^\top x_t') - c_{t,2} (b^\top \lambda_t') \big)
 -\frac{1}{n} \gamma c_{t,1}' (b^\top \lambda_t') ] \\
\stackrel{(b)}{\geq} & 
( 1 - \gamma \mu^2 ) \E[\varphi_{\lambda,t}  ]
 + \gamma \mu^2 \E[\varphi_{x,t}] 
 - \frac{1}{n}  \gamma \mu' b^\top \E[\lambda_t']
\numberthis
\end{align*}
where $(b)$ follows from $\E_A[c_{t,s}] = \mu$,  $\E_A[c_{t,s}'] = \mu' \leq \mu$, and Lemma~\ref{lemma:grad_geq_0}.

The above arguments prove that
\begin{equation}
 \!\!\begin{bmatrix}
    \E[\delta_{t+1}] \\
    \E[\zeta_{t+1}]
  \end{bmatrix} 
  \geq \begin{bmatrix}
    \E[\varphi_{x,t+1}] \\
    \E[\varphi_{\lambda,t+1}]
  \end{bmatrix} 
  \geq  \underbrace{\begin{bmatrix}
    (1 - \alpha\mu (1 - \gamma \mu^2)) &\alpha\mu (1 - \gamma \mu^2) \\
    \gamma \mu^2 & (1 - \gamma \mu^2)
  \end{bmatrix}}_{B} 
  \begin{bmatrix}
    \E[\varphi_{x,t}] \\
    \E[\varphi_{\lambda,t}]
  \end{bmatrix} 
  +  \frac{1}{n}
  \begin{bmatrix}
    \alpha  (1 - \gamma \mu^2)\\
    -\mu' \gamma
  \end{bmatrix} 
 \label{eq:delta_zeta_iter_B}
 \end{equation} 
where the inequality for vectors denotes the inequality of each corresponding element in the vectors, and matrix $B$ has $v_{B,1} = [1, 1]^\top$ as an eigenvector associated with the eigenvalue $1$ because
\begin{equation}  
B v_{B,1}
= \begin{bmatrix}
(1 - \alpha\mu (1 - \gamma \mu^2)) &\alpha\mu (1 - \gamma \mu^2) \\
\gamma \mu^2 & (1 - \gamma \mu^2)
\end{bmatrix}
\begin{bmatrix}
1 \\
1
\end{bmatrix}
= \begin{bmatrix}
1 \\
1
\end{bmatrix} .
\end{equation}  
Similarly, since
\begin{align*}  
B \begin{bmatrix}
\alpha (1 - \gamma \mu^2) \\
- \gamma \mu
\end{bmatrix}
=& \begin{bmatrix}
    (1 - \alpha\mu (1 - \gamma \mu^2)) &\alpha\mu (1 - \gamma \mu^2) \\
    \gamma \mu^2 & (1 - \gamma \mu^2)
  \end{bmatrix}
  \begin{bmatrix}
\alpha  (1 - \gamma \mu^2) \\
- \gamma \mu
\end{bmatrix} \\
=& \begin{bmatrix}
(1 - \alpha\mu (1 - \gamma \mu^2)) \alpha (1 - \gamma \mu^2)
- \gamma \mu \alpha\mu (1 - \gamma \mu^2) \\
\gamma \mu^2 \alpha (1 - \gamma \mu^2)
- \gamma \mu (1 - \gamma \mu^2)
\end{bmatrix} \\
=& \begin{bmatrix}
(1- \alpha\mu) (1 - \gamma \mu^2) \alpha  (1 - \gamma \mu^2) \\
-(1- \alpha\mu) (1 - \gamma \mu^2) \gamma \mu
\end{bmatrix}
= (1- \alpha\mu) (1 - \gamma \mu^2)
\begin{bmatrix}
 \alpha  (1 - \gamma \mu^2) \\
-\gamma \mu 
\end{bmatrix},
\numberthis
\end{align*}  
then $v_{B,2} = [ \alpha (1 - \gamma \mu^2), 
-\gamma \mu]^\top $ is another eigenvector of $B$ with a positive eigenvalue $ (1- \alpha\mu) (1 - \gamma \mu^2) < 1$.
Let $Q_B = [v_{B,1}, v_{B,2} ]$, which can be expressed as
\begin{align}
Q_B = [v_{B,1}, v_{B,2} ]
= \begin{bmatrix}
1 & \alpha  (1 - \gamma \mu^2)\\
1 & -\gamma \mu
\end{bmatrix}.
\end{align}
Then $B$ has eigenvalue decomposition $B = Q_B \Lambda_B Q_B^{-1}$, where $\Lambda_B  = \mathrm{diag}([1, (1- \alpha\mu) (1 - \gamma \mu^2)])$, and thus $B^t = Q_B \Lambda_B^t Q_B^{-1}$ for $t \in [T]$.

Let $[\alpha(1 - \gamma \mu^2), -\mu'\gamma]^\top = Q_B [c_{B,1}, c_{B,2}]^\top$, where $[c_{B,1}, c_{B,2}]^\top = Q_B^{-1} [\alpha(1 - \gamma \mu^2), -\mu'\gamma]^\top$ can be computed by
\begin{align*}
\begin{bmatrix}
c_{B,1} \\
c_{B,2}
\end{bmatrix} 
=&
Q_B^{-1} 
\begin{bmatrix}
\alpha(1 - \gamma \mu^2) \\
-\mu'\gamma
\end{bmatrix} 
=  -\frac{1}{\alpha  (1 - \gamma \mu^2) + \gamma \mu } 
\begin{bmatrix}
-\gamma \mu & -\alpha  (1 - \gamma \mu^2)  \\
-1  & 1
\end{bmatrix} 
\begin{bmatrix}
\alpha \\
-\mu'\gamma
\end{bmatrix} \\
=& \frac{1}{\alpha  (1 - \gamma \mu^2) + \gamma \mu }  
\begin{bmatrix}
\alpha \gamma (\mu - \mu' )(1 - \gamma \mu^2)   \\
  \alpha + \mu' \gamma
\end{bmatrix}
\geq 
\begin{bmatrix}
\frac{1}{2n} \gamma    \\
1
\end{bmatrix}
\numberthis
\end{align*}
where the last inequality follows from $\mu - \mu' = \frac{1}{n} $ and $\alpha (1 - \gamma \mu^2)\geq \frac{1}{2}\alpha = 1/(8\mu T) \geq \gamma\mu$ for $c_{B,1} \geq \frac{1}{2n} \gamma$, and $\alpha \mu^2 = \mu/(4T) = 4 n^{-\frac{1}{3}} T^{-1} \geq n^{-1} = \mu - \mu'$, so that $\alpha + \mu'\gamma = \alpha + \gamma(\mu'-\mu + \mu) \geq 
\alpha + \gamma(\mu - \alpha\mu^2) = \alpha(1-\gamma\mu^2) + \gamma\mu$ for $c_{B,2} \geq 1$.

Since all elements in $B$ are positive, multiplying $B$ on both sides preserves inequality. 
Applying \eqref{eq:delta_zeta_iter_B} recursively yields
\begin{align*}
\begin{bmatrix}
\E[\delta_{T}] \\
\E[\zeta_{T}]
\end{bmatrix} 
\geq &
\sum_{t=0}^{T-1} B^{T-1 - t}  
\frac{1}{n}
\begin{bmatrix}
\alpha  (1 - \gamma \mu^2)\\
-\mu' \gamma
\end{bmatrix} 
= \sum_{t=0}^{T-1} B^{T-1 - t}  
\frac{1}{n}
\begin{bmatrix}
\alpha  (1 - \gamma \mu^2)\\
-\mu' \gamma
\end{bmatrix} 
= \sum_{t=0}^{T-1} B^{T-1 - t}  
\frac{1}{n} Q_B 
\begin{bmatrix}
c_{B,1}  \\
c_{B,2}
\end{bmatrix}  \\
= &
\frac{1}{n}
\sum_{t=0}^{T-1} 1^{T-1-t} c_{B,1} v_{B,1}
+  \frac{1}{n} \sum_{t=0}^{T-1} \big((1 - \alpha \mu)(1 - \gamma \mu^2) \big)^{T-1-t} c_{B,2} {v_{B,2}}  \\
\geq &
\frac{T}{n } c_{B,1} v_{B,1}
+ \frac{1}{8 n \alpha} c_{B,2} {v_{B,2}} 
\geq \frac{\gamma T}{2n^2} v_{B,1}
+ \frac{1}{8 n \alpha} {v_{B,2}} 
\numberthis
\end{align*} 
where the last inequality follows from $c_{B,1} \geq \frac{1}{2n} \gamma$, and $c_{B,2} \geq 1$.
Plugging in
$v_{B,1} = [1,1]^\top$ and $v_{B,2} = [ \alpha (1 - \gamma \mu^2), -\gamma \mu]^\top $, and since $(1 - \gamma \mu^2) \geq \frac{1}{2}$,
it follows that $\E[\delta_{T}] \geq \frac{ \gamma T}{2n^2} + \frac{1}{16 n}$.
\end{proof}

\subsubsection{Proof of Theorem~\ref{thm:grad_stability_modo_sc}}
\label{app_subs:proof_stab_SC_modo}

\begin{proof}[Theorem~\ref{thm:grad_stability_modo_sc}]
Combining the argument stability in Theorem~\ref{thm:bound_E_diff_x_lam_warm_modo}, and Assumption~\ref{assmp:lip_cont_grad_f}, the MOL uniform stability can be bounded by
\begin{align*}
& \sup_z \E_A[\|\nabla F_z(A(S)) - \nabla F_z(A(S'))\|_{\rm F}^2] \\
\leq &
\E_A[ \ell_{F,1}^2 \|A(S) - A(S')\|^2] 
\tag*{by Assumption~\ref{assmp:lip_cont_grad_f}}\\
\leq &
\frac{48}{\mu n} \ell_f^2 \ell_{F,1}^2  
\Big(\alpha + \frac{12 + 4M \ell_f^2}{\mu n} + \frac{10 M\ell_f^4 \gamma}{\mu} \Big) .
\numberthis
\label{eq:bound_MOL_stab}
\end{align*}
Then based on Propositions~\ref{prop:gen_err_minnorm_bound_grad}-\ref{prop:stability_gen_grad_F}, we have
\begin{align*}
 \E_{A,S}[R_{\rm gen}(A(S))]
 {\leq} &
 \E_{A,S} [ \|\nabla F(A(S))- \nabla F_S(A(S)) \|_{\rm F}] 
 \tag*{by Proposition~\ref{prop:gen_err_minnorm_bound_grad}}\\
 {\leq} &
 4\epsilon_{\rm F} 
+ \sqrt{n^{-1} \mathbb{E}_S\left[\mathbb{V}_{z\sim \cal D}(\nabla F_z(A(S) ))\right]} 
\tag*{by Proposition~\ref{prop:stability_gen_grad_F}}\\
{=}& 
\mathcal{O}( n^{-\frac{1}{2}}) .
\tag*{by \eqref{eq:bound_MOL_stab}}
\end{align*}
The proof of the upper bound is complete. We then prove the MOL uniform stability lower bound based on  the argument uniform stability lower bound in Theorem~\ref{thm:lower_bound}.
By the strong convexity of the function $f_{z,m}(x)$, for all $m \in [M]$
\begin{align*}
\sup_z \E_A[\|\nabla F_z(A(S)) - \nabla F_z(A(S'))\|_{\rm F}^2] 
\geq &
\E_A[ M \mu^2 \|A(S) - A(S')\|^2]
\tag*{by Assumption~\ref{assmp:sconvex}}\\
\geq &
\frac{M \mu^2}{256 n^2} .
\tag*{by Theorem~\ref{thm:lower_bound} and Jensen's inequality}
\end{align*}
The proof of the lower bound is complete.
\end{proof}

\section{Bounding the CA Distance}
\label{sec:bound_CA_dist}

\subsection{Auxiliary lemmas}
\label{sub_app:opt_err_aux_lemma_app}

Section~\ref{ssub:properties_of_the_subproblem} summarizes properties of the generalized subproblem $\min_{\lambdain}\|Q \lambda\|^2 + \rho \|\lambda\|^2$ along with their proofs, where $\rho \geq 0$.
Section~\ref{ssub:other_supporting_lemmas} summarizes properties of the update function of the MoDo algorithm.
These properties are then used to derive the main theorems.

\subsubsection{Properties of the subproblem} 
\label{ssub:properties_of_the_subproblem}

Given $Q \in \mathbb{R}^{d\times M}$, $\rho \geq 0$, 
define the subproblem as
\begin{align}
  \min_{\lambdain}& ~~\|Q \lambda\|^2 + \rho \|\lambda\|^2
\end{align}
which is a constrained quadratic programming problem, and $Q$ can be the full-batch gradient $\nabla F_S(x)$ or its stochastic estimate. In this section, we provide basic properties of this problem.
Before proceeding, we first define a few notations we will use repeatedly in this section.

\begin{subequations}
\begin{align}
\text{The CA weight }\qquad  &\lambda_{Q,\rho}^* \in \mathop{\arg\min}_{\lambdain} ~\|Q \lambda\|^2 + \rho \|\lambda\|^2 
\label{eq:def_CA_weight} \\
\text{The CA direction }\qquad  &d_{Q,\rho} \coloneqq Q \lambda_{Q,\rho}^*
\label{eq:def_CA_direction}
\end{align}
\end{subequations}

\begin{lmm}[Uniqueness of CA direction]
\label{lemma:uniqueness_CA}
  Given $Q \in \mathbb{R}^{d\times M}$, 
  then $d_{Q,\rho} \coloneqq Q\lambda^*_{Q,\rho}$ with $\lambda^*_{Q,\rho} \in \arg\min_{\lambda \in \Delta^{M}}\|Q\lambda\|^2  + \rho \|\lambda\|^2$ exists, and $d_{Q,\rho}$ is unique.
\end{lmm}
\begin{proof}
\label{proof:uniqueness_CA}
When $\rho = 0$,
the proof is given in~\citep[Section~2]{Desideri2012mgda}. 
When $\rho > 0$, it is a standard result for strictly convex problem with a unique $\lambda^*_{Q,\rho}$, thus unique $d_{Q,\rho}$.
\end{proof}

\begin{lmm}\label{lemma:descent direction}
Given $Q \in \mathbb{R}^{d\times M}$, recall $\lambda^*_{Q,\rho}$ with $\rho \geq 0$ is defined as
\begin{equation}
   \lambda^*_{Q,\rho} \in \mathop{\arg\min}_{\lambda \in \Delta^{M} }\|Q \lambda \|^2 + \rho \|\lambda\|^2.  \label{eq:descent-dir-lemma} 
\end{equation}
Then, for any   $\lambda \in \Delta^{M}$, it holds that
\begin{subequations}
\begin{align}
&\langle Q \lambda^*_{Q,\rho},Q \lambda \rangle \geq \|Q \lambda^*_{Q,\rho}\|^2 - \rho, \label{eq:convex_inner_prod_property}\\
\text{and}~~& \|Q\lambda - Q \lambda^*_{Q,\rho} \|^2
\leq \|Q\lambda\|^2 - \|Q \lambda^*_{Q,\rho}\|^2
+ 2 \rho.
\end{align}
\end{subequations}
\end{lmm}
\begin{proof}
By the first order optimality condition for \eqref{eq:descent-dir-lemma} , for any   $\lambda \in \Delta^{M}$, we have
\begin{align}
\langle Q^\top Q \lambda^*_{Q,\rho}, \lambda - \lambda^*_{Q,\rho} \rangle \geq - \rho.
\end{align}
By rearranging the above inequality, we obtain
\begin{align}
\langle Q \lambda^*_{Q,\rho}, Q \lambda\rangle \geq \|Q \lambda^*_{Q,\rho} \|^2 - \rho,
\end{align}
which is precisely the first inequality in the claim. Furthermore, we can also have
\begin{align*}
\|Q \lambda - Q \lambda^*_{Q,\rho} \|^2 
&= \|Q \lambda\|^2  + \| Q \lambda^*_{Q,\rho} \|^2 - 2 \langle Q \lambda^*_{Q,\rho}, Q \lambda\rangle \\
&\leq \|Q \lambda\|^2  + \| Q \lambda^*_{Q,\rho} \|^2 - 2 \|Q \lambda^*_{Q,\rho}\|^2 + 2 \rho \\
&= \| Q \lambda \|^2 - \|Q \lambda^*_{Q,\rho}\|^2 + 2 \rho,
\numberthis
\end{align*}
which is the desired second inequality in the claim. Hence, the proof is complete.
\end{proof}

\begin{lmm}[Continuity of $\lambda^*_{Q,\rho}$ with $\rho >0$]\label{lemma:lambda* lip}
Given $Q \in \mathbb{R}^{d\times M}$, $\rho>0$ and $x\in\mathbb{R}^d$, 
recall 
$$\lambda^*_{Q,\rho} = \mathop{\arg\min}_{\lambdain} g(\lambda;Q,\rho) =  \mathop{\arg\min}_{\lambdain} \frac{1}{2}\|Q \lambda\|^2 +  \frac{1}{2} \rho \|\lambda\|^2,$$  
then the following inequality holds
\begin{align}\label{eq:lambda* lip}
\|\lambda^*_{Q,\rho}-\lambda^*_{Q',\rho}\| \leq \rho^{-1} \|Q^\top Q - Q'^{\top} Q' \|.
\end{align}
Furthermore, suppose either 1) Assumptions~\ref{assmp:lip_cont_grad_f},~\ref{assmp:lip_cont_f} hold, or 2) Assumptions~\ref{assmp:lip_cont_grad_f},~\ref{assmp:sconvex} hold, with $\ell_F$ defined in Lemma~\ref{lemma:x_t_bounded_sc_smooth}. Then 
for $x \in \{x_t\}_{t=1}^T$, $x' \in \{x_t'\}_{t=1}^T$ generated by MoDo algorithm on training dataset $S$ and $S'$, respectively, let $\lambda^*_\rho(x) = \lambda^*_{\nabla F_S(x),\rho}$, $\lambda^*_\rho(x') = \lambda^*_{\nabla F_S(x'),\rho}$,
it implies 
\begin{align}
 \|\lambda^*_\rho(x)-\lambda^*_\rho(x')\| \leq 2 \rho^{-1} \ell_{F,1} \ell_F \|x - x' \| .   
\end{align}
\end{lmm}

\begin{proof}
We provide proof leveraging the convergence properties of the projected gradient descent algorithm on strongly convex objectives below.
Consider the problem $\min_{\lambdain} g(\lambda;Q,\rho) $, which is $\rho$-strongly convex.
Let $\{\lambda_{Q,\rho,k}\}$ for $k = 0,1,\dots, K$ denote the sequence obtained from applying projected  gradient descent (PGD) on the objective $g(\lambda;Q ,\rho) $, i.e., 
\begin{align*}
  \lambda_{Q,\rho, k+1}  = &
  \Pi_{\Delta^{M}} \Big(\lambda_{Q,\rho, k} - \eta Q^\top Q \lambda_{Q,\rho, k} - \eta \rho \lambda_{Q,\rho, k} \Big) \\
  = & \Pi_{\Delta^{M}} \Big(\big((1 - \eta \rho) \mathrm{I} - \eta Q^\top Q \big) \lambda_{Q,\rho, k} \Big)
\numberthis
\end{align*}
where $\eta $ is the step size that $\eta < 1/ (\|Q^\top Q\|+\rho)$.
Note that both $\rho,\eta$ are independent of $K$.
By the convergence result of PGD on strongly convex objective, we know that $\lambda_{Q,\rho}^* $ is the limit point of $\{\lambda_{Q,\rho,k} \}_{k=0}^\infty$.
By the non-expansiveness of projection, we have
\begin{align*}
\|\lambda_{Q,\rho, k+1} - \lambda_{Q', \rho, k+1}\| 
 \leq &
  \|\big((1 - \eta \rho) \mathrm{I} - \eta Q^\top Q \big) \lambda_{Q,\rho, k} - 
  \big((1 - \eta \rho) \mathrm{I} - \eta {Q'}^\top {Q'} \big) \lambda_{Q',\rho, k} \| \\
 \leq &
   \|(1 - \eta\rho) \mathrm{I} - \eta Q^\top Q \| \|\lambda_{Q,\rho, k} - \lambda_{Q',\rho, k}\| + \eta \| ( Q^\top Q - {Q'}^\top {Q'} )\lambda_{Q',\rho, k}\| \\
 \!\!\!  \leq &
    (1 - \eta\rho)  \|\lambda_{Q,\rho, k} - \lambda_{Q',\rho, k}\| 
   + \eta \|(Q^\top Q - {Q'}^\top {Q'} )\lambda_{Q',\rho, k}\|. 
\numberthis
\end{align*}
Since $\lambda_{Q, \rho, 0} = \lambda_{Q',\rho, 0} $, applying the above inequality recursively from $k=0,\dots, K-1$ gives
\begin{align*}
  \|\lambda_{Q,\rho, K}  - \lambda_{Q',\rho, K} \|  
  \leq & \eta \|(Q^\top Q - {Q'}^\top {Q'} )\lambda_{Q',\rho, k}\| 
  \Big(\sum_{k=0}^{K-1} (1 - \eta\rho)^k \Big) \\
  =&\eta \|(Q^\top Q - {Q'}^\top {Q'} ) \lambda_{Q',\rho, k}\| 
  \frac{1 - (\eta\rho)^K}{\eta\rho} \\
  \leq & \rho^{-1} (1 - (\eta\rho)^K) \|Q^\top Q - {Q'}^\top {Q'}\| .
  \numberthis
\end{align*}
Then it follows that
\begin{align*}
 \|\lambda^*_{Q,\rho}-\lambda^*_{Q',\rho}\| 
 \leq & 
  \lim_{K \to \infty} \big( \|\lambda^*_{Q,\rho}-\lambda_{Q,\rho,K}\| + \|\lambda^*_{Q',\rho}-\lambda_{Q',\rho,K}\|  + \|\lambda_{Q,\rho,K}-\lambda_{Q',\rho,K}\| \big) \\
  \leq & 
  \lim_{K \to \infty} \big( \|\lambda^*_{Q,\rho}-\lambda_{Q,\rho,K}\| + \|\lambda^*_{Q',\rho}-\lambda_{Q',\rho,K}\|\big) \\
  &+\lim_{K \to \infty} 
  \rho^{-1} (1 - (\eta\rho)^K) \|Q^\top Q - {Q'}^\top {Q'}\|  \\
 \stackrel{(a)}{\leq} & \rho^{-1}  \|Q^\top Q - {Q'}^\top {Q'}\|  + \lim_{K \to \infty} 2\sqrt{\frac{4}{\rho \eta K}} 
 \leq \rho^{-1}  \|Q^\top Q - {Q'}^\top {Q'}\|
 \numberthis\label{eq:lam_rho_star_cont}
\end{align*}
where $(a)$ follows from $\lim_{K\to \infty}1 - (\eta \rho)^K = 1$, and from the convergence of PGD~\citep[Theorem~1.1]{Beck2010GradientbasedAW} on $\rho$-strongly convex objectives that
\begin{align*}
 \|\lambda^*_{Q,\rho}-\lambda_{Q,\rho,K}\|^2 \leq \frac{2}{\rho}
 \Big(g(\lambda_{Q,\rho,K}; Q,\rho)
- g(\lambda^*_{Q,\rho}; Q,\rho) \Big) 
\leq  \frac{2}{\rho} 
\frac{\|\lambda_{Q,\rho,0} - \lambda^*_{Q,\rho}\|^2}{2\eta K}
\leq \frac{4}{\rho\eta K}.
\end{align*}
This proves \eqref{eq:lambda* lip}.
In addition, under Assumptions~\ref{assmp:lip_cont_grad_f},~\ref{assmp:lip_cont_f}, the above result directly implies that
\begin{align*}
 \|\lambda^*_\rho(x)-\lambda^*_\rho(x')\| \leq  & \rho^{-1}  \|\nabla F_S(x)^\top \nabla F_S(x) - \nabla F_S(x')^\top \nabla F_S(x')\|   \\
 \leq & \rho^{-1} \|\nabla F_S(x)+ \nabla F_S(x')\| \| \nabla F_S(x) - \nabla F_S(x')\|  \\
 \leq & 2 \rho^{-1} \ell_{F,1} \ell_F \|x - x'\|  .
 \numberthis
\end{align*}
 While under Assumptions~\ref{assmp:lip_cont_grad_f} and \ref{assmp:sconvex}, and for $\ell_F$ defined in Lemma~\ref{lemma:x_t_bounded_sc_smooth},
 and for $x \in \{x_t\}_{t=1}^T$, $x' \in \{x_t'\}_{t=1}^T$ generated by MoDo algorithm on training dataset $S$ and $S'$, respectively,
 $\|\nabla F_S(x)\|\leq \ell_F$, $\|\nabla F_S(x')\|\leq \ell_F$, which along with~\eqref{eq:lam_rho_star_cont} implies that
\begin{align*}
 \|\lambda^*_\rho(x)-\lambda^*_\rho(x')\| 
 \leq & \rho^{-1} \|\nabla F_S(x)+ \nabla F_S(x')\| \| \nabla F_S(x) - \nabla F_S(x')\|
 \leq 2 \rho^{-1} \ell_{F,1} \ell_F \|x - x'\| .
\end{align*}
The proof is complete.
\end{proof}

\begin{lmm}\label{lemma:frho subopt}
Given $Q \in \mathbb{R}^{d\times M}$, $\rho\geq 0, \bar{\rho} > 0$, 
with $\lambda^*_{Q,\rho}$ defined in \eqref{eq:def_CA_weight},
then we have
\begin{align}
  - \bar{\rho} \Big(1 - \frac{1}{M} \Big) \leq \|Q \lambda^*_{Q,\rho}\|^2 - \|Q\lambda^*_{Q,\bar{\rho}}\|^2 \leq {\rho}\left(1-\frac{1}{M}\right).
\end{align}
\end{lmm}

\begin{proof}\label{proof:frho subopt}
 Since
 $\lambda^*_{Q,\bar{\rho}} = \arg\min_{\lambdain} \|Q \lambda\|^2 + \bar{\rho} \|\lambda\|^2$, therefore
 \begin{align*}
   &\|Q \lambda^*_{Q,\rho}\|^2 - \|Q\lambda^*_{Q,\bar{\rho}}\|^2 \\
   =& \|Q \lambda^*_{Q,\rho}\|^2 + \bar{\rho} \|\lambda^*_{Q,{\rho}}\|^2
   - \min_{\lambdain} (\|Q \lambda\|^2 + \bar{\rho} \|\lambda\|^2)
   + \bar{\rho} (\|\lambda^*_{Q,\bar{\rho}}\|^2
   - \|\lambda^*_{Q,{\rho}}\|^2)
   \geq - \bar{\rho} \Big(1 - \frac{1}{M} \Big) .
   \numberthis
 \end{align*}
 Similarly, since $\lambda^*_{Q,\rho} = \arg\min_{\lambdain} \|Q \lambda\|^2 +\rho  \|\lambda\|^2$, therefore
 \begin{align}
 \|Q\lambda^*_{Q,\bar{\rho}}\|^2 +\rho  \|\lambda^*_{Q,\bar{\rho}}\|^2
 - \|Q \lambda^*_{Q,\rho}\|^2 
 - \rho \|\lambda^*_{Q,\rho}\|^2 \geq 0.
 \end{align}
 Rearranging the above inequality gives
 \begin{align}
  \|Q \lambda^*_{Q,\rho}\|^2 - \|Q\lambda^*_{Q,\bar{\rho}}\|^2
  \leq &
  \rho \|\lambda^*_{Q,\bar{\rho}}\|^2 - \rho \|\lambda^*_{Q,\rho}\|^2 
  \leq \rho \Big(1 - \frac{1}{M} \Big) .
 \end{align}
\end{proof}

\subsubsection{Properties of the update} 
\label{ssub:other_supporting_lemmas}

\begin{lmm}[Properties of MoDo update of $\lambda_t$]
\label{lemma:grad_lam_t_lam}
Consider $\{x_t\}$, $\{\lambda_t\}$ generated by the MoDo algorithm. For all $\lambdain$, $\rho \geq 0$, it holds that
\begin{align*}
& 2\gamma_t \E_A \langle \lambda_{t} - \lambda, (\nabla F_S(x_t)^\top \nabla F_S(x_t) +\rho \mathrm{I} )\lambda_t \rangle \\
\leq   & 
\E_A [\|\lambda_{t} - \lambda  \|^2]
- \E_A [\|\lambda_{t+1} - \lambda \|^2]
+ \gamma_t^2 \E_A [\|(\nabla F_{z_{t,1}} (x_t)^\top \nabla F_{z_{t,2}} (x_t) +\rho \mathrm{I} )\lambda_t\|^2],
\numberthis
\label{eq:all_lam_bound1}\\
\text{and} \quad 
&\gamma_t \E_A (\|\nabla F_S(x_t) \lambda_t\|^2  
- \|\nabla F_S(x_t) \lambda \|^2 
+ \rho\| \lambda_t\|^2 - \rho\| \lambda\|^2 
+ \rho\| \lambda_t - \lambda\|^2 ) \\
\leq & 
\E_A [\|\lambda_{t} - \lambda \|^2]
- \E_A [\|\lambda_{t+1} - \lambda \|^2] 
+ \gamma_t^2 \E_A [\|(\nabla F_{z_{t,1}} (x_t)^\top \nabla F_{z_{t,2}} (x_t) +\rho \mathrm{I} )\lambda_t\|^2] .  
\numberthis
\label{eq:all_lam_bound2}
\end{align*}
\end{lmm}

\begin{proof}
By the update of $\lambda$, for all $\lambdain$, we have
\begin{align*}
  & \|\lambda_{t+1} - \lambda \|^2 \\
  =&
  \|\Pi_{\Delta^{M}}(\lambda_{t} - \gamma_t(\nabla F_{z_{t,1}} (x_t)^\top \nabla F_{z_{t,2}} (x_t) +\rho \mathrm{I} )\lambda_t) - \lambda \|^2 \\
  \leq &
  \|\lambda_{t} - \gamma_t(\nabla F_{z_{t,1}} (x_t)^\top \nabla F_{z_{t,2}} (x_t) +\rho \mathrm{I} )\lambda_t - \lambda \|^2 \\
  = &
  \|\lambda_{t} - \lambda  \|^2
  -2\gamma_t \langle \lambda_{t} - \lambda , (\nabla F_{z_{t,1}} (x_t)^\top \nabla F_{z_{t,2}} (x_t) +\rho \mathrm{I} )\lambda_t \rangle 
  + \gamma_t^2 \|(\nabla F_{z_{t,1}} (x_t)^\top \nabla F_{z_{t,2}} (x_t) +\rho \mathrm{I} )\lambda_t\|^2 .
\end{align*}    
Taking expectation over $z_{t,1}, z_{t,2}$ on both sides and rearranging   proves~\eqref{eq:all_lam_bound1}.\\
By the $\rho$-strong convexity of the subproblem, $\min_{\lambdain} \frac{1}{2}\|\nabla F_S(x_t) \lambda\|^2 + \frac{1}{2} \rho \|\lambda\|^2 $, we have
\begin{align*}
& \gamma_t \E_A (\|\nabla F_S(x_t) \lambda_t\|^2 + \rho\| \lambda_t\|^2
- \|\nabla F_S(x_t) \lambda \|^2 -  \rho\| \lambda\|^2 + {\rho} \|\lambda_t - \lambda\|^2) \\
\leq & 
2\gamma_t \E_A\langle \lambda_{t} - \lambda , (\nabla F_S(x_t)^\top \nabla F_S(x_t) +\rho \mathrm{I} )\lambda_t \rangle \\
\stackrel{\eqref{eq:all_lam_bound1}}{\leq}   & 
\E_A [\|\lambda_{t} - \lambda  \|^2]
- \E_A [\|\lambda_{t+1} - \lambda \|^2] 
+ \gamma_t^2 \E_A [\|(\nabla F_{z_{t,1}} (x_t)^\top \nabla F_{z_{t,2}} (x_t) +\rho \mathrm{I} )\lambda_t\|^2] .
\numberthis
\end{align*}
Rearranging the above inequality  proves~\eqref{eq:all_lam_bound2}.
\end{proof}

\subsection{Proof of Theorem~\ref{lemma:converge_MGDA_bounded_grad} -- CA direction distance of MoDo}
\label{sub_app:proof_CA_dist_MoDo}

\paragraph{Organization of proof.}
In Lemma~\ref{lemma:converge_mgda}, we prove the upper bound of the CA direction distance, $\frac{1}{T} \sum_{t=1}^T  \E_A [\|\nabla F_S(x_t) \lambda_t 
- \nabla F_S(x_t) \lambda^*_\rho(x_t)\|^2 ] $, in terms of two average of sequences, $S_{1,T}$, and $S_{2,T}$. Then   under either Assumptions~\ref{assmp:lip_cont_grad_f},~\ref{assmp:lip_cont_f}, or Assumptions~\ref{assmp:lip_cont_grad_f},~\ref{assmp:sconvex},
we prove the upper bound of $S_{1,T}$, and $S_{2,T}$, and thus the CA direction distance in Theorem~\ref{lemma:converge_MGDA_bounded_grad}.

\begin{lmm}
\label{lemma:converge_mgda}
Suppose Assumption~\ref{assmp:lip_cont_grad_f}
holds. 
Let $\{x_t\}, \{\lambda_t\}$ be the sequences produced by the MoDo algorithm with step sizes $\alpha_t = \alpha > 0$, $\gamma_t = \gamma > 0$, and regularization $\rho \geq 0$.
With a positive constant $\bar{\rho} >0$, 
define
\begin{subequations}
\begin{align}
  S_{1,T} =& \frac{1}{T} \sum_{t=1}^T
 \E_A [\|(\nabla F_{z_{t,1}}(x_t)^\top \nabla F_{z_{t,2}}(x_t) + \rho \mathrm{I} )\lambda_t\|^2] \\
  S_{2,T} =& \frac{1}{T} \sum_{t=1}^T
  \E_A [\|\nabla F_S(x_{t+1}) + \nabla F_S(x_{t})\|
     \|\nabla F_{Z_{t+1}}(x_t)\lambda_{t+1}\|] .
\end{align}
\end{subequations}
Then it holds that
\begin{align}
\frac{1}{T} \sum_{t=1}^T  \E_A [\|\nabla F_S(x_t) \lambda_t\|^2 
- \|\nabla F_S(x_t) \lambda_{\rho}^*(x_t)\|^2 ] 
\leq &  
\bar{\rho} +
\frac{4}{\gamma T} ( 1 + \bar{\rho}^{-1} \alpha \ell_{F,1} T S_{2,T} ) + \frac{\rho}{\gamma} 
+ \gamma S_{1,T}  .
\end{align}  
\end{lmm}

\begin{proof}
Define $\lambda^*_{\bar{\rho}}(x_t) = \arg\min_{\lambdain} \frac{1}{2}\|\nabla F_S(x_t) \lambda\|^2 +\frac{\bar{\rho}}{2} \|\lambda\|^2$ with $\bar{\rho} > 0$.
Note that different from $\rho \geq 0$, $\bar{\rho} > 0$ is strictly positive, and  used as an intermediate parameter only for analysis but not for algorithm update.

Substituting $\lambda = \lambda^*_{\bar{\rho}}(x_t)$ in Lemma~\ref{lemma:grad_lam_t_lam}, \eqref{eq:all_lam_bound2}, we have
\begin{align*}
& \gamma_t \E_A(\|\nabla F_S(x_t) \lambda_t\|^2  
- \|\nabla F_S(x_t) \lambda_{\bar{\rho}}^*(x_t) \|^2 
+ \rho\| \lambda_t\|^2 - \rho\| \lambda_{\bar{\rho}}^*(x_t)\|^2 
+ \rho\| \lambda_t - \lambda_{\bar{\rho}}^*(x_t)\|^2 ) \\
\leq & 
\E_A[\|\lambda_{t} - \lambda^*_{\bar{\rho}}(x_t) \|^2]
- \E_A[\|\lambda_{t+1} - \lambda^*_{\bar{\rho}}(x_t)\|^2]
+ \gamma_t^2 \E_A[\|(\nabla F_{z_{t,1}} (x_t)^\top \nabla F_{z_{t,2}} (x_t) + \rho \mathrm{I} )\lambda_t\|^2].
\numberthis
\end{align*}
Setting $\gamma_t = \gamma > 0$,  and telescoping  the above inequality gives
\begin{align*}
 &\frac{1}{T} \sum_{t=1}^T  
 \E_A [\|\nabla F_S(x_t) \lambda_t\|^2  
- \|\nabla F_S(x_t) \lambda_{\bar{\rho}}^*(x_t)\|^2 ] \\
\leq &
\frac{1}{T} \sum_{t=1}^T
\frac{1}{\gamma} 
\E_A [\|\lambda_{t} - \lambda^*_{\bar{\rho}}(x_t) \|^2
- \|\lambda_{t+1} - \lambda^*_{\bar{\rho}}(x_t)\|^2 ]
+ \frac{1}{T} \sum_{t=1}^T
\gamma \E_A [\|(\nabla F_{z_{t,1}} (x_t)^\top \nabla F_{z_{t,2}} (x_t) + \rho \mathrm{I} )\lambda_t\|^2] \\
&- \frac{\rho}{\gamma } 
\E_A [\| \lambda_t\|^2 - \| \lambda_{\bar{\rho}}^*(x_t)\|^2 
+ \| \lambda_t - \lambda_{\bar{\rho}}^*(x_t)\|^2] \\
= &
\frac{1}{\gamma T} \underbrace{\Big(\sum_{t=1}^T 
\E_A [\|\lambda_{t} - \lambda^*_{\bar{\rho}} (x_t)\|^2
- \|\lambda_{t+1} - \lambda^*_{\bar{\rho}}(x_t)\|^2 ] \Big)}_{I_1}
+ \frac{1}{T} \sum_{t=1}^T
\gamma \E_A [\|(\nabla F_{z_{t,1}} (x_t)^\top \nabla F_{z_{t,2}} (x_t) + \rho \mathrm{I} )\lambda_t\|^2] \\
&- \frac{\rho}{\gamma } 
\E_A [\| \lambda_t\|^2 - \| \lambda_{\bar{\rho}}^*(x_t)\|^2 
+ \| \lambda_t - \lambda_{\bar{\rho}}^*(x_t)\|^2]
\numberthis\label{eq:bound_FS_lam_rho_iter_I1}
\end{align*}
where $I_1$ can be further derived as
\begin{align*}
  I_1 =&  \sum_{t=1}^T 
\E_A  [\|\lambda_{t} - \lambda^*_{\bar{\rho}}(x_t) \|^2]
- \E_A [\|\lambda_{t+1} - \lambda^*_{\bar{\rho}}(x_t)\|^2] \\
=& \E_A [\|\lambda_{1} - \lambda^*_{\bar{\rho}}(x_1) \|^2 ]
- \E_A [\|\lambda_{T+1} - \lambda^*_{\bar{\rho}}(x_T)\|^2 ]
+ \sum_{t=1}^{T-1} 
\E_A [\|\lambda_{t+1} - \lambda^*_{\bar{\rho}}(x_{t+1}) \|^2
- \|\lambda_{t+1} - \lambda^*_{\bar{\rho}}(x_t)\|^2 ]\\
\leq &\E_A [\|\lambda_{1} - \lambda^*_{\bar{\rho}}(x_1) \|^2 ]
- \E_A [\|\lambda_{T+1} - \lambda^*_{\bar{\rho}}(x_T)\|^2] \\
&+ \sum_{t=1}^{T-1} 
\E_A [\|2\lambda_{t+1} - \lambda^*_{\bar{\rho}}(x_{t+1}) - \lambda^*_{\bar{\rho}}(x_t)\| 
\|\lambda_{\bar{\rho}}^*(x_{t+1}) - \lambda^*_{\bar{\rho}}(x_t)\| ] \\
\leq & 4
+ 4\sum_{t=1}^{T-1} 
\E_A [\|\lambda_{\bar{\rho}}^*(x_{t+1}) - \lambda^*_{\bar{\rho}}(x_t)\|]
\numberthis
\end{align*}
where $\|\lambda_{\bar{\rho},t+1}^*(x_{t+1}) - \lambda^*_{\bar{\rho}}(x_t)\|$, by Lemma~\ref{lemma:lambda* lip}, can be bounded by
\begin{align*}
  \|\lambda_{\bar{\rho},t+1}^*(x_{t+1}) - \lambda^*_{\bar{\rho}}(x_t)\|
  \leq & \bar{\rho}^{-1} 
  \|\nabla F_S(x_{t+1}) + \nabla F_S(x_{t})\|
  \|\nabla F_S(x_{t+1}) - \nabla F_S(x_{t})\| \\
  \leq & \bar{\rho}^{-1} \ell_{F,1}
  \|\nabla F_S(x_{t+1}) + \nabla F_S(x_{t})\|
   \|x_{t+1} - x_{t}\| \\
  \leq & \bar{\rho}^{-1} \alpha \ell_{F,1}
  \|\nabla F_S(x_{t+1})
  + \nabla F_S(x_{t})\|
   \|\nabla F_{Z_{t+1}}\lambda_{t+1}\|.
   \numberthis
\end{align*}
Hence, it follows that
\begin{align*}
  I_1 \leq & 
  4 + 4\bar{\rho}^{-1} \alpha \ell_{F,1}
  \sum_{t=1}^T
  \E_A [\|\nabla F_S(x_{t+1})
    + \nabla F_S(x_{t})\|
     \|\nabla F_{Z_{t+1}}\lambda_{t+1}\|] \\
  = &
  4 + 4\bar{\rho}^{-1} \alpha \ell_{F,1}
  T S_{2,T} 
\numberthis
\end{align*}
plugging which into \eqref{eq:bound_FS_lam_rho_iter_I1}  gives
\begin{align}
&\frac{1}{T} \sum_{t=1}^T  \E_A [\|\nabla F_S(x_t) \lambda_t\|^2 
- \|\nabla F_S(x_t) \lambda_{\bar{\rho}}^*(x_t)\|^2 ] 
\leq 
\frac{4}{\gamma T} ( 1 + \bar{\rho}^{-1} \alpha \ell_{F,1} T S_{2,T})
+ \gamma S_{1,T} + \frac{\rho}{\gamma} .
\label{eq:bound_diff_func_lam_rho_S1S2}
\end{align}
Recall $\lambda^*_\rho(x_t) = \arg\min_{\lambdain} \|\nabla F_S(x_t) \lambda\|^2 + \rho \|\lambda\|^2 $.
Then
\begin{align*}
  &\frac{1}{T} \sum_{t=1}^T  \E_A [\|\nabla F_S(x_t) \lambda_t\|^2 
- \|\nabla F_S(x_t) \lambda^*_\rho(x_t)\|^2 ]\\
= & \frac{1}{T} \sum_{t=1}^T  \E_A [\|\nabla F_S(x_t) \lambda_t\|^2 
- \|\nabla F_S(x_t) \lambda_{\bar{\rho}}^*(x_t)\|^2 
+ \|\nabla F_S(x_t) \lambda_{\bar{\rho}}^*(x_t)\|^2 
- \|\nabla F_S(x_t) \lambda^*_\rho(x_t)\|^2] \\
\stackrel{\eqref{eq:bound_diff_func_lam_rho_S1S2}}{\leq}  & 
\frac{4}{\gamma T} ( 1 + \bar{\rho}^{-1} \alpha \ell_{F,1} T S_{2,T})
+ \gamma S_{1,T} + \frac{\rho}{\gamma} 
+ \frac{1}{T} \sum_{t=1}^T \E_A [
 \|\nabla F_S(x_t) \lambda_{\bar{\rho}}^*(x_t)\|^2 
- \|\nabla F_S(x_t) \lambda^*_\rho(x_t)\|^2] \\
\leq  &
\frac{4}{\gamma T} ( 1 + \bar{\rho}^{-1} \alpha \ell_{F,1} T S_{2,T})
+ \gamma S_{1,T} + \frac{\rho}{\gamma} 
+ {\bar\rho}
\numberthis
\end{align*}
where the last inequality follows from Lemma~\ref{lemma:frho subopt}.
The proof is complete.
\end{proof}

\begin{proof}[Theorem~\ref{lemma:converge_MGDA_bounded_grad}]
\label{proof:converge_MGDA_bounded_grad}
Building on the result in Lemma~\ref{lemma:converge_mgda},  and by the convexity of the subproblem, $\min_{\lambdain} \frac{1}{2}\|\nabla F_S(x_t) \lambda\|^2 + \rho \|\lambda\|^2$, and Lemma~\ref{lemma:descent direction}, we have
\begin{align*}
&\frac{1}{T} \sum_{t=1}^T  \E_A [\|\nabla F_S(x_t) \lambda_t 
- \nabla F_S(x_t) \lambda^*_\rho(x_t)\|^2 ] \\
\leq &
\frac{1}{T} \sum_{t=1}^T  \E_A [\|\nabla F_S(x_t) \lambda_t\|^2 
- \|\nabla F_S(x_t) \lambda^*_\rho(x_t)\|^2 ] + 2\rho \\
\leq &   \bar{\rho} + 2\rho  +
\frac{4}{\gamma T} ( 1 + \bar{\rho}^{-1} \alpha T \ell_{F,1} S_{2,T} ) + \frac{\rho}{\gamma} 
+ \gamma S_{1,T} .
\numberthis
\label{eq:bound_converge_CA_dist_avg_S1S2}
\end{align*}
By Assumptions~\ref{assmp:lip_cont_grad_f},~\ref{assmp:lip_cont_f} or Assumptions~\ref{assmp:lip_cont_grad_f},~\ref{assmp:sconvex} and Lemma~\ref{lemma:x_t_bounded_sc_smooth}, we have
\begin{align}
S_{1,T} =& \frac{1}{T} \sum_{t=1}^T
\E_A [\|(\nabla F_{z_{t,1}} (x_t)^\top \nabla F_{z_{t,2}} (x_t) + \rho \mathrm{I} )\lambda_t\|^2] 
\leq (\ell_f \ell_F + \rho )^2 = (M^{\frac{1}{2}} \ell_f^2 + \rho )^2 \\
S_{2,T} =& \frac{1}{T} \sum_{t=1}^T
\E_A [\|\nabla F_S(x_{t+1})
+ \nabla F_S(x_{t})\|
\|\nabla F_{Z_{t+1}}\lambda_{t+1}\| ]
\leq 2 \ell_f \ell_F = 2 M^{\frac{1}{2}} \ell_f^2.
\end{align}
Substituting $S_{1,T}$, $S_{2,T}$ in ~\eqref{eq:bound_converge_CA_dist_avg_S1S2} with the above bound yields
\begin{align}
 &\frac{1}{T} \sum_{t=1}^T  \E_A [\|\nabla F_S(x_t) \lambda_t 
- \nabla F_S(x_t) \lambda^*_\rho (x_t)\|^2 ] \nonumber \\
\leq &  
  \bar{\rho} + 2\rho + 
\frac{4}{\gamma T} ( 1 + 2 \bar{\rho}^{-1} \alpha  T \ell_{F,1} \ell_f \ell_F ) + \frac{\rho}{\gamma} 
+ \gamma (M^{\frac{1}{2}} \ell_f^2 + \rho )^2 .   
\end{align}
Based on the definition of the CA direction distance, we have
\begin{align*}
  &\mathcal{E}_{\rm ca}(x_t, \lambda_{t+1}) = 
 \|\E_A[\nabla F_{Z_{t+1}}(x_t)\lambda_{t+1} - d(x_t)]\| ^2
 = \|\E_A[\nabla F_S(x_t)\lambda_{t+1} 
 - \nabla F_S(x_t) \lambda^*_\rho(x_t)]\| ^2 \\
 \leq & \E_A [\|\nabla F_S(x_t) \lambda_{t+1} 
- \nabla F_S(x_t) \lambda^*_\rho(x_t)\|^2 ] \\
\leq & 2\E_A [\|\nabla F_S(x_t) \lambda_t 
- \nabla F_S(x_t) \lambda^*_\rho(x_t)\|^2 ] 
+ 2\E_A [\|\nabla F_S(x_t)(\lambda_{t+1} - \lambda_t)\|^2 ] \\
\leq & 2\E_A [\|\nabla F_S(x_t) \lambda_t 
- \nabla F_S(x_t) \lambda^*_\rho(x_t)\|^2 ] 
+ 2\gamma^2 \ell_F^2\E_A [\|(\nabla F_{z_{t,1}} (x_t)^\top \nabla F_{z_{t,2}} (x_t) + \rho \mathrm{I} )\lambda_t\|^2].
\numberthis
\label{eq:CA_direct_dist_bound_modo}
\end{align*}
Because $\ell_{F,1}\ell_F \leq M \ell_{f,1}\ell_f$,
choosing $\bar{\rho} = 2 (\alpha  M \ell_{f,1}\ell_f^2 /\gamma)^{\frac{1}{2}}$  yields
\begin{align*}
\frac{1}{T} \sum_{t=1}^T \mathcal{E}_{\rm ca}(x_t, \lambda_{t+1})
 \leq & \frac{1}{T} \sum_{t=1}^T  2\E_A [\|\nabla F_S(x_t) \lambda_t 
- \nabla F_S(x_t) \lambda^*_\rho(x_t)\|^2 ] 
+ 2 M \gamma^2 \ell_f^2 S_{1,T}\\
{\leq} &  2\bar{\rho} + 4\rho + 
\frac{8}{\gamma T} ( 1 + 2 \bar{\rho}^{-1} \alpha  T M\ell_{f,1} \ell_f^2)
+ \frac{2\rho}{\gamma} 
+ 2\gamma(1+M\gamma) (M^{\frac{1}{2}} \ell_f^2 + \rho )^2 \\
=& \frac{8}{\gamma T} + 12 \sqrt{M \ell_{f,1}\ell_f^2 \frac{\alpha}{\gamma} } + \frac{2\rho}{\gamma}  + 4\rho + 2\gamma(1+M\gamma) (M^{\frac{1}{2}} \ell_f^2 + \rho )^2 .
\numberthis
\end{align*}
This proves the result.
\end{proof}

\subsection{Proof of Theorem~\ref{thm:dist_CA_weight} -- CA weight distance of MoDo}
\label{sub_app:proof_CA_weight_dist_MoDo}

In this section, we consider the regularization $\rho> 0$, and prove Theorem~\ref{thm:dist_CA_weight}, the guarantee of CA weight distance, which is stronger than the guarantee of CA direction distance.

\begin{proof}[Theorem~\ref{thm:dist_CA_weight}]
\label{proof:distance_CA_weight}  
Consider the function $g(\lambda; \nabla F_S(x), \rho) \coloneqq \frac{1}{2} \|\nabla F_S(x) \lambda \|^2 + \frac{1}{2} \rho \|\lambda\|^2 \geq 0$, which is $\rho$-strongly convex.
Based on Lemma~\ref{lemma:grad_lam_t_lam}, \eqref{eq:all_lam_bound2}, the property of the update of $\lambda$, we have
\begin{align*}
 & \gamma_t \E_A [  g(\lambda_t; \nabla F_S(x_t), \rho)
 - g(\lambda; \nabla F_S(x_t), \rho)
+ \rho\| \lambda_t - \lambda\|^2 ] \\
\leq & 
\E_A[\|\lambda_{t} - \lambda  \|^2]
- \E_A [\|\lambda_{t+1} - \lambda \|^2] 
+ \gamma_t^2 \E_A [\|(\nabla F_{z_{t,1}} (x_t)^\top \nabla F_{z_{t,2}} (x_t) +\rho \mathrm{I} )\lambda_t\|^2]
\end{align*}
where setting $\gamma_t = \gamma > 0$ and rearranging yields
\begin{align*}
\E_A [\|\lambda_{t+1} - \lambda \|^2] 
\leq &
(1 - \rho \gamma) \E_A[\|\lambda_{t} - \lambda \|^2] 
+ \gamma^2 \E_A [\|(\nabla F_{z_{t,1}} (x_t)^\top \nabla F_{z_{t,2}} (x_t) +\rho \mathrm{I} )\lambda_t\|^2] \\
& - \gamma \E_A [  g(\lambda_t; \nabla F_S(x_t), \rho)
 - g(\lambda; \nabla F_S(x_t), \rho) ] 
 \numberthis .
\end{align*}
Substituting $\lambda = \lambda^*_{\rho}(x_t) = \arg\min_{\lambdain} g(\lambda; \nabla F_S(x_t), \rho)$, we have
\begin{align*}
\E_A [\|\lambda_{t+1} - \lambda^*_{\rho}(x_t) \|^2 ]
\leq &
(1 - \rho \gamma) \E_A[\|\lambda_{t} - \lambda^*_{\rho}(x_t) \|^2 ]
+ \gamma^2 \E_A [\|(\nabla F_{z_{t,1}} (x_t)^\top \nabla F_{z_{t,2}} (x_t) +\rho \mathrm{I} )\lambda_t\|^2] \\
& - \gamma \E_A [  g(\lambda_t; \nabla F_S(x_t), \rho)
 - g(\lambda^*_{\rho}(x_t); \nabla F_S(x_t), \rho) ] \\
\leq & 
(1 - \rho \gamma) \E_A[\|\lambda_{t} - \lambda^*_{\rho}(x_t) \|^2 ]
+ \gamma^2 (M^{\frac{1}{2}} \ell_f^2 + \rho)^2 \numberthis
\label{eq:property_lam_t_lam_rho_star}
\end{align*}
where the last inequality holds because $g(\lambda^*_{\rho}(x_t); \nabla F_S(x_t), \rho) = \min_{\lambdain} g(\lambda; \nabla F_S(x_t), \rho)$, and $\|(\nabla F_{z_{t,1}} (x_t)^\top \nabla F_{z_{t,2}} (x_t) +\rho \mathrm{I} )\lambda_t\| \leq M^{\frac{1}{2}} \ell_f^2 + \rho$. \\
Then $\E_A [\|\lambda_t - \lambda^*_{\rho}(x_t) \|^2]$ can be further bounded by
\begin{align*}
  \E_A [\|\lambda_t - \lambda^*_{\rho}(x_t) \|^2]
  \leq & 
\E_A [\|\lambda_t - \lambda^*_{\rho}(x_{t-1}) \|^2]
+ \E_A (\|\lambda_{t} - \lambda^*_{\rho}(x_{t}) \|^2
- \|\lambda_{t} - \lambda^*_{\rho}(x_{t-1}) \|^2) \\
\leq &\E_A \|\lambda_{t} - \lambda^*_{\rho}(x_{t-1}) \|^2
+ 4 \E_A \|\lambda^*_{\rho}(x_{t}) - \lambda^*_{\rho}(x_{t-1})\| 
\numberthis
\label{eq:lam_rho_approx}
\end{align*}
where $\|\lambda_{\rho}^*(x_{t}) - \lambda^*_{\rho}(x_{t-1})\|$, by Lemma~\ref{lemma:lambda* lip}, can be bounded by
\begin{align*}
  \|\lambda_{\rho}^*(x_{t}) - \lambda^*_{\rho}(x_{t-1})\|
  \leq & \rho^{-1} 
  \|\nabla F_S(x_{t-1}) + \nabla F_S(x_{t})\|
  \|\nabla F_S(x_{t-1}) - \nabla F_S(x_{t})\| \\
  \leq & \rho^{-1} \ell_{F,1}
  \|\nabla F_S(x_{t-1}) + \nabla F_S(x_{t})\|
   \|x_{t-1} - x_{t}\| \\
  \leq & 2 \rho^{-1} \alpha \ell_{F,1}
  \|\nabla F_S(x_{t-1}) + \nabla F_S(x_{t})\|
   \|\nabla F_{Z_{t}}\lambda_{t}\|
   \leq 2 \rho^{-1} \alpha \ell_{F,1}
  \ell_F \ell_f
   \numberthis \label{eq:lam_rho_iter_bound}
\end{align*}
where the last inequality follows from either: 1) Assumptions~\ref{assmp:lip_cont_grad_f},~\ref{assmp:lip_cont_f}; or 2) Assumptions~\ref{assmp:lip_cont_grad_f}, \ref{assmp:sconvex}, with $\ell_f $ and $\ell_F$ defined in  Lemma~\ref{lemma:x_t_bounded_sc_smooth}. 
\\
Combining \eqref{eq:property_lam_t_lam_rho_star}, \eqref{eq:lam_rho_approx} and \eqref{eq:lam_rho_iter_bound} gives
\begin{align*}
  \E_A [\|\lambda_{t+1} - \lambda^*_{\rho}(x_{t}) \|^2]
  \leq & 
  (1 - \rho \gamma) \E_A[\|\lambda_{t} - \lambda^*_{\rho}(x_{t-1}) \|^2 ]
+ \gamma^2 (M^{\frac{1}{2}} \ell_f^2 + \rho)^2
+ 8 \rho^{-1} \alpha M \ell_{f,1} \ell_f^2.
\end{align*}
Applying the above inequality recursively yields
\begin{align*}
 \E_A [\|\lambda_{T+1} - \lambda^*_{\rho}(x_{T}) \|^2]
  \leq & 
  (1 - \rho \gamma)^{T} \E_A[\|\lambda_1 - \lambda^*_{\rho}(x_0) \|^2]
  + \rho^{-1}\gamma (M^{\frac{1}{2}} \ell_f^2 + \rho)^2
+ 8 \rho^{-2} \gamma^{-1} \alpha M \ell_{f,1} \ell_f^2 \\
\leq & 
  4 (1 - \rho \gamma)^{T} 
  + \rho^{-1}\gamma (M^{\frac{1}{2}} \ell_f^2 + \rho)^2
+ 8 \rho^{-2} \gamma^{-1} \alpha M \ell_{f,1} \ell_f^2.
\end{align*}
The proof is complete.
\end{proof}

\subsection{Proof of Theorem~\ref{crlr:bias_CA_mitigate} -- CA direction distances of SMG and MoCo} 
\label{sub_app:proof_CA_direction_SMG_MoCo}

\begin{proof}[Theorem~\ref{crlr:bias_CA_mitigate}]
\label{proof:bias_CA_mitigate}
As a direct consequence of Lemma~\ref{lemma:bound_diff_update_holder} by plugging in $Q' = \nabla F_S(x)$,
given $x\in \mathbb{R}^d$ and $Q\in \mathbb{R}^{d\times M}$, define $\lambda^*(x) \in \arg\min_{\lambdain} \|\nabla F_S(x)\lambda\|^2$, and $\lambda^*_Q \in \arg\min_{\lambdain} \|Q\lambda\|^2$, then it holds that
\begin{align}\label{eq:bound_Q_update}
\| \nabla F_S(x) \lambda^*(x) -Q  \lambda^*_Q \|^2& 
\leq 4 \max \Big\{\sup_{\lambdain}\|Q\lambda\|, \sup_{\lambdain}\|\nabla F_S(x)\lambda\| \Big\} \cdot \|Q-\nabla F_S(x)\|.
\end{align}
If  $\max\{\sup_{\lambdain}\|Q\lambda\|, \sup_{\lambdain}\|\nabla F_S(x)\lambda\|\}  \leq \ell_f$, then the CA direction distance can be further bounded by
\begin{align*}
    \|\nabla F_S(x) \lambda^*(x) - \E_A[Q \lambda^*_{Q}]\|^2 
    & \leq 
    \E_A [\|\nabla F_S(x) \lambda^*(x) - Q \lambda^*_{Q}\|^2] \\
    & \stackrel{\eqref{eq:bound_Q_update}}{\leq} 4 \ell_f \E_A [\| Q - \nabla F_S(x) \|] .
    \numberthis
    \label{eq:bias_F_Z_update_bound}
  \end{align*}

  For the \emph{SMG} algorithm, plugging in $Q = \nabla F_Z(x)$,
  with $Z$ denoting a subset or mini-batch randomly sampled from $S$.
  Then it holds that
  \begin{align}
    \|\nabla F_S(x) \lambda^*(x) - \E_A[\nabla F_Z(x) \lambda^*_{\nabla F_Z(x)}]\|^2 
    \leq 4 \ell_f \E_A [\| \nabla F_Z(x) - \E_A[\nabla F_Z(x)] \|] 
     = \mathcal{O} \Bigg(\frac{1}{\sqrt{|Z|}} \Bigg).
  \end{align}
  This suggests when the size $|Z|$  increases, $\E_Z [\| \nabla F_Z(x) - \nabla F_S(x) \|]$ decreases, then the upper bound of $\|\nabla F_S(x) \lambda^*(x) - \E_Z[\nabla F_Z(x) \lambda^*_{\nabla F_Z(x)}]\|^2 $ also decreases.
  This proves the bias to the CA direction is mitigated by increasing the batch size.
  With $\{x_t\}$, $\{Z_t\}$ denoting the sequence of models and the stochastic mini-batch of data generated by the SMG algorithm, and $|Z_t| = \mathcal{O}(t)$,  it holds that
  \begin{align}
    \frac{1}{T}\sum_{t=1}^T \E_A[\|\nabla F_S(x_t) \lambda^*(x_t) - \E_A[\nabla F_{Z_t}(x_t) \lambda^*_{\nabla F_{Z_t}(x_t)}]\|^2] 
    = \mathcal{O}(T^{-\frac{1}{2}}) .
  \end{align}

Similarly, for the \emph{MoCo} algorithm, $Q=Y_t = (1 - \beta_{t-1})Y_{t-1} + \beta_{t-1} \nabla F_{z_{t-1}}(x_{t-1})$, denotes its moving average gradient at iteration $t$.
Let $\beta_t=\beta >0$ be a constant given $T$, then by (46) in~\citep{fernando2022mitigating}, set $\alpha=\Theta(T^{-\frac{3}{4}})$, and $\beta=\Theta(T^{-\frac{1}{2}})$, we have
\begin{align}
  \frac{1}{T} \sum_{t=1}^T 
\E_A[\|Y_t-\nabla F_S (x_t)\|^2] = \mathcal{O}( \beta^{-1} T^{-1} + \beta + \alpha^2 \beta^{-2}) = \mathcal{O}(T^{-\frac{1}{2}}).
\end{align} 
This states that $\frac{1}{T}\sum_{t=1}^T \E_A [\| Y_t - \nabla F_S(x_t) \|]$  is converging, so is the CA direction distance of MoCo, given by
\begin{align*}
   &\frac{1}{T}\sum_{t=1}^T \E_A[\|\nabla F_S(x_t) \lambda^*(x_t) - \E_A[Y_t \lambda^*_{Y_t}]\|^2] 
  \leq \frac{1}{T}\sum_{t=1}^T  4\ell_f \E_A[\|\nabla F_S(x_t)  - \E_A[Y_t]\| ] \\
  \leq & \frac{1}{T}\sum_{t=1}^T  4\ell_f \E_A[\|\nabla F_S(x_t) - Y_t \|]
  \leq \Bigg(\frac{1}{T}\sum_{t=1}^T  4\ell_f \E_A[\|\nabla F_S(x_t) - Y_t \|^2 ] \Bigg)^{\frac{1}{2}}
  = \mathcal{O}(T^{-\frac{1}{4}}) .
\end{align*}
The proof is complete.
\end{proof}

\section{Bounding the PS Optimization Error}
\label{sec_app:bound_opt_err}

\subsection{Proof of Theorems~\ref{thm:opt_err_modo_nonconvex_bounded_grad} -- Optimization of MoDo}
\label{sub_app:bound_opt_err_MoDo}

\paragraph{Technical contributions.} The optimization error bound in Theorem~\ref{thm:opt_err_modo_nonconvex_bounded_grad} is improved with either relaxed assumption or improved convergence rate compared to prior stochastic MOL algorithms~\citep{zhou2022_SMOO,fernando2022mitigating,liu2021stochastic} (see Table~\ref{tab:opt_err_compare_detail}).   This is achieved by 1) instead of bounding the approximation error to $\lambda^*(x_t)$, we bound that to the CA direction $d(x_t) = -\nabla F_S(x_t) \lambda^*(x_t)$
 as a whole, and 2) instead of using the descent lemma of $F_S(x_t) \lambda^*(x_t)$ with a dynamic weight, we use that of $F_S(x_t) \lambda$ with a fixed weight (see Lemma~\ref{lemma:opt_err_PS_min_norm},~\eqref{eq:descent_lemma_all_lam}), thereby improving the tightness of the bound. 

\paragraph{Organization of proof.}
In Lemma~\ref{lemma:opt_err_PS_min_norm}, we prove the upper bound of the PS optimization error, $\frac{1}{T}\sum_{t=1}^T 
\E_A \| \nabla F_S(x_t) \lambda_t^*(x_t) \|^2 $, in terms of three average of sequences, $S_{1,T}$, $S_{3,T}$, and $S_{4,T}$. Then we prove the upper bound of $S_{1,T}$, $S_{3,T}$, and $S_{4,T}$, and thus the PS optimization error either in the NC case under Assumptions~\ref{assmp:lip_cont_grad_f},~\ref{assmp:lip_cont_f} or in the SC case under Assumptions~\ref{assmp:lip_cont_grad_f},~\ref{assmp:sconvex}.
In Lemma~\ref{lemma:PS_opt_SC_MoDo_gammaT}, we prove the last-iterate convergence in the SC case, which can be tighter than Lemma~\ref{lemma:opt_err_PS_min_norm} in the SC case with $\gamma = \mathcal{O}(T^{-\frac{3}{2}})$.
Combining the results leads to Theorem~\ref{thm:opt_err_modo_nonconvex_bounded_grad}.

\subsubsection{Auxiliary Lemmas} 
\label{ssub_app:auxiliary_lemmas_opt_MoDo}

\begin{lmm}
\label{lemma:opt_err_PS_min_norm}
Suppose Assumption~\ref{assmp:lip_cont_grad_f} holds.
Consider the sequence $\{x_t\}, \{\lambda_t\}$ generated by MoDo in unbounded domain for $x$. 
Define
\begin{subequations}\label{eq.def-S134}
 \begin{align}
 S_{1,T} =& \frac{1}{T} \sum_{t=1}^T
 \E_A [\|(\nabla F_{z_{t,1}} (x_t)^\top \nabla F_{z_{t,2}} (x_t) +\rho \mathrm{I} ) \lambda_t\|^2]   \\
 S_{3,T} =& \frac{1}{T}\sum_{t=1}^T \E_{A}[\|(\nabla F_{z_{t,1}} (x_t)^\top \nabla F_{z_{t,2}} (x_t) +\rho \mathrm{I}) \lambda_t\| \|\nabla F_S(x_t)^\top \nabla F_S(x_t)\lambda_1 \|] \\
 S_{4,T} =& \frac{1}{T}\sum_{t=1}^T \E_{A}[\| \nabla F_{Z_{t+1}}(x_t)\lambda_{t+1}\|^2] .
\end{align}
\end{subequations}
Then it holds that
\begin{align*}
 \frac{1}{T}\sum_{t=1}^T 
\E_A [\| \nabla F_S(x_t) \lambda_t^*(x_t) \|^2]  
\leq    
 \frac{1}{\alpha T} \E_A [F_S (x_{1}) - F_S (x_{T+1}) ] \lambda_1   + \frac{1}{2} \gamma S_{1,T}
+ \gamma S_{3,T}  + \frac{1}{2} \alpha \ell_{f,1} S_{4,T} +\rho .
\end{align*}
\end{lmm}

\begin{proof}
By the $\ell_{f,1}$-smoothness of $F_S (x)\lambda$ for all $\lambdain$, we have
\begin{align*}
F_S (x_{t+1}) \lambda - F_S (x_{t}) \lambda  
\leq &
\langle \nabla F_S(x_{t}) \lambda, x_{t+1} - x_t \rangle
+ \frac{\ell_{f,1}}{2} \|x_{t+1} - x_t\|^2 \\
= &
-\alpha_t \langle \nabla F_S(x_{t}) \lambda, \nabla F_{Z_{t+1}}(x_t)\lambda_{t+1} \rangle
+ \frac{\ell_{f,1}}{2} \alpha_t^2 \| \nabla F_{Z_{t+1}}(x_t)\lambda_{t+1}\|^2.
\numberthis\label{eq:descent_lemma_all_lam}
\end{align*}
Taking expectation over $Z_{t+1}$ on both sides of the above inequality gives
\begin{align}\label{eq:bound_descent_FS_lam}
&  \E_{Z_{t+1}} [F_S (x_{t+1})] \lambda - F_S (x_{t}) \lambda 
\leq  
-\alpha_t \langle \nabla F_S(x_{t}) \lambda, \nabla F_S(x_t)\lambda_{t+1} \rangle
+ \frac{\ell_{f,1}}{2} \alpha_t^2 \E_{Z_{t+1}}[\| \nabla F_{Z_{t+1}}(x_t)\lambda_{t+1}\|^2] .
\end{align}
By Lemma~\ref{lemma:grad_lam_t_lam}, \eqref{eq:all_lam_bound1}, we have
\begin{align*}
  & 2\gamma_t \E_A \langle \lambda_{t} - \lambda, (\nabla F_S(x_t)^\top \nabla F_S(x_t) +\rho \mathrm{I} )\lambda_t \rangle \\
\leq   & 
\E_A [\|\lambda_{t} - \lambda  \|^2]
- \E_A [\|\lambda_{t+1} - \lambda \|^2]
+ \gamma_t^2 \E_A [\|(\nabla F_{z_{t,1}} (x_t)^\top \nabla F_{z_{t,2}} (x_t) +\rho \mathrm{I} )\lambda_t\|^2] .
\numberthis
\end{align*}
Rearranging the above inequality and letting $\gamma_t = \gamma > 0$ gives
\begin{align*}
  &- \E_A\langle \lambda, (\nabla F_S(x_t)^\top \nabla F_S(x_t))  \lambda_t \rangle 
  \leq 
  - \E_A\langle \lambda_t, (\nabla F_S(x_t)^\top \nabla F_S(x_t) +\rho \mathrm{I} )\lambda_t \rangle + \rho \E_A [\lambda^\top \lambda_t] \\
  &\qquad\qquad + \frac{1}{2\gamma} 
  \E_A [\|\lambda_{t} - \lambda \|^2 - \|\lambda_{t+1} - \lambda\|^2 ]
  + \frac{1}{2} 
  \gamma \E_A[\|(\nabla F_{z_{t,1}} (x_t)^\top \nabla F_{z_{t,2}} (x_t) +\rho \mathrm{I} )\lambda_t\|^2] \\
  \leq &
  - \E_A[\| \nabla F_S(x_t) \lambda_t \|^2]
  + \rho \E_A[\lambda^\top \lambda_t - \| \lambda_t \|^2]
  + \frac{1}{2\gamma} 
  \E_A [\|\lambda_{t} - \lambda \|^2 - \|\lambda_{t+1} - \lambda\|^2 ] \\
  & + \frac{1}{2} 
  \gamma \E_A[\|(\nabla F_{z_{t,1}} (x_t)^\top \nabla F_{z_{t,2}} (x_t) +\rho \mathrm{I} )\lambda_t\|^2] .
\numberthis
\end{align*}
Plugging the above inequality into \eqref{eq:bound_descent_FS_lam}, and setting $\alpha_t = \alpha > 0$, we have
\begin{align*}
   & \E_A [F_S (x_{t+1}) \lambda - F_S (x_{t}) \lambda ]
\leq  
-\alpha \E_A \langle \nabla F_S(x_{t}) \lambda, \nabla F_S(x_t)\lambda_{t+1} \rangle
+ \frac{\ell_{f,1}}{2} \alpha^2 \E_A[\| \nabla F_{Z_{t+1}}(x_t)\lambda_{t+1}\|^2] \\
\leq &
- \alpha \E_A[\| \nabla F_S(x_t) \lambda_t \|^2]
  + \frac{\alpha}{2\gamma} 
  \E_A[ \|\lambda_{t} - \lambda \|^2 - \|\lambda_{t+1} - \lambda\|^2 ]  + \alpha \rho \\
&  +\alpha \E_A\langle \nabla F_S(x_{t}) \lambda, \nabla F_S(x_t)(\lambda_t - \lambda_{t+1}) \rangle   + \frac{1}{2} \alpha^2 \ell_{f,1} \E_{A}[\| \nabla F_{Z_{t+1}}(x_t)\lambda_{t+1}\|^2] \\
 & + \frac{1}{2} 
  \alpha \gamma \E_A[\|(\nabla F_{z_{t,1}} (x_t)^\top \nabla F_{z_{t,2}} (x_t) +\rho \mathrm{I})\lambda_t\|^2].
\numberthis
\end{align*}
Taking telescope sum and rearranging yields, for all $\lambdain$,
\begin{align*}
& \frac{1}{T}\sum_{t=1}^T 
\E_A[\| \nabla F_S(x_t) \lambda_t \|^2]  \\
\leq & \rho +
\frac{1}{2\gamma T}\sum_{t=1}^T \E_A[ \|\lambda_{t} - \lambda \|^2 - \|\lambda_{t+1} - \lambda\|^2] 
+ \frac{1}{\alpha T}\sum_{t=1}^T
\E_A [F_S (x_{t}) 
- F_S (x_{t+1})] \lambda \\
& + \frac{1}{2T} \sum_{t=1}^T 
\Big(\gamma  \E_A[\|(\nabla F_{z_{t,1}} (x_t)^\top \nabla F_{z_{t,2}} (x_t) +\rho \mathrm{I})\lambda_t\|^2] + \alpha \ell_{f,1} \E_A[\| \nabla F_{Z_{t+1}}(x_t)\lambda_{t+1}\|^2] \\
&\qquad\qquad + 2 \E_A\langle \nabla F_S(x_{t}) \lambda, \nabla F_S(x_t)(\lambda_t - \lambda_{t+1}) \rangle 
 \Big)   \\
\leq & \rho + 
\frac{1}{2\gamma T} \E_A [\|\lambda_{1} - \lambda \|^2 -  \|\lambda_{T+1} - \lambda\|^2 ]
+ \frac{1}{\alpha T} 
\E_A [F_S (x_{1}) - F_S (x_{T+1}) ] \lambda  
+ \frac{1}{2} \gamma S_{1,T}
+ \gamma S_{3,T}  + \frac{1}{2} \alpha \ell_{f,1} S_{4,T} .
\numberthis
\end{align*}
Setting $\lambda = \lambda_1$ in the above inequality yields
\begin{align*}
 \frac{1}{T}\sum_{t=1}^T 
\E_A[\| \nabla F_S(x_t) \lambda_t \|^2]
\leq &
\frac{1}{\alpha T} \E_A [F_S (x_{1}) - F_S (x_{T+1}) ] \lambda_1  + \frac{1}{2} \gamma S_{1,T}
+ \gamma S_{3,T}  + \frac{1}{2} \alpha \ell_{f,1} S_{4,T} + \rho .
\end{align*}
Finally, the results follow from the definition of $\lambda_t^*(x_t)$ that $\frac{1}{T}\sum_{t=1}^T 
\E_A [\| \nabla F_S(x_t) \lambda_t^*(x_t) \|^2]  
\leq    
\frac{1}{T}\sum_{t=1}^T 
\E_A [\| \nabla F_S(x_t) \lambda_t  \|^2]$.
\end{proof}

\begin{lmm}
\label{lemma:PS_opt_SC_MoDo_gammaT}
Suppose  Assumptions~\ref{assmp:lip_cont_grad_f}, \ref{assmp:sconvex}
hold, with $\ell_f $   defined in  Lemma~\ref{lemma:x_t_bounded_sc_smooth}.
Define $c_F$ such that $\E_A [F_S (x_{1}) \lambda_1] - \min_{x\in \mathbb{R}^d} \E_A [F_S (x) \lambda_1] \leq c_F$.
Considering $\{x_t\}$ generated by MoDo, with $\alpha_t = \alpha \leq 1/(2\ell_{f,1})$, $\gamma_t = \gamma$, then  it holds that
\begin{align*}
  \E_A \Big[\min_{\lambdain} \|\nabla F_S(x_T)\lambda\|^2\Big]
  =& \mathcal{O}\Big( (1 - \alpha \mu)^{T} 
  + \alpha + M (\gamma T )^2 \Big) .
\end{align*}
\end{lmm}

\begin{proof}
By the $\ell_{f,1}$-smoothness of $F_S (x)\lambda$ for all $\lambdain$, we have
\begin{align*}
F_S (x_{t+1}) \lambda - F_S (x_{t}) \lambda  
\leq &
\langle \nabla F_S(x_{t}) \lambda, x_{t+1} - x_t \rangle
+ \frac{\ell_{f,1}}{2} \|x_{t+1} - x_t\|^2 \\
= &
-\alpha_t \langle \nabla F_S(x_{t}) \lambda, \nabla F_{Z_{t+1}}(x_t)\lambda_{t+1} \rangle
+ \frac{\ell_{f,1}}{2} \alpha_t^2 \| \nabla F_{Z_{t+1}}(x_t)\lambda_{t+1}\|^2.
\numberthis
\end{align*}
Taking expectation over $Z_{t+1}$ on both sides of the above inequality gives
\begin{align*}
&  \E_{Z_{t+1}} [F_S (x_{t+1})] \lambda - F_S (x_{t}) \lambda 
\leq  
-\alpha_t \langle \nabla F_S(x_{t}) \lambda, \nabla F_S(x_t)\lambda_{t+1} \rangle
+ \frac{\ell_{f,1}}{2} \alpha_t^2 \E_{Z_{t+1}}[\| \nabla F_{Z_{t+1}}(x_t)\lambda_{t+1}\|^2] \\
= & 
-\frac{1}{2}\alpha_t \big(\| \nabla F_S(x_{t}) \lambda\|^2
+ \| \nabla F_S(x_{t}) \lambda_{t+1}\|^2
- \| \nabla F_S(x_{t}) (\lambda -\lambda_{t+1}) \|^2 \big)
+ \frac{\ell_{f,1}}{2} \alpha_t^2 \E_{Z_{t+1}}[\| \nabla F_{Z_{t+1}}(x_t)\lambda_{t+1}\|^2] \\
\leq & -\frac{1}{2}\alpha_t 2\mu (F_S(x_t) \lambda -\inf_x F_S(x)\lambda ) 
+ \frac{1}{2}\alpha_t \| \nabla F_S(x_{t}) (\lambda -\lambda_{t+1}) \|^2
+ \frac{\ell_{f,1}}{2} \alpha_t^2 \E_{Z_{t+1}}[\| \nabla F_{Z_{t+1}}(x_t)\lambda_{t+1}\|^2].
\numberthis
\end{align*}
Let $\alpha_t = \alpha $ and rearranging the above inequality yields
\begin{align*}
&  \E_{Z_{t+1}} [F_S (x_{t+1})] \lambda - \inf_x F_S(x)\lambda \\
\leq & (1-\alpha \mu) (F_S(x_t) \lambda -\inf_x F_S(x)\lambda ) 
+ \frac{1}{2}\alpha \| \nabla F_S(x_{t}) (\lambda -\lambda_{t+1}) \|^2
+ \frac{\ell_{f,1}}{2} \alpha^2 \E_{Z_{t+1}}[\| \nabla F_{Z_{t+1}}(x_t)\lambda_{t+1}\|^2] \\
\leq & (1-\alpha \mu) (F_S(x_t) \lambda -\inf_x F_S(x)\lambda ) 
+ \frac{1}{2}\alpha \| \nabla F_S(x_{t}) (\lambda -\lambda_{t+1}) \|^2
+ \frac{1}{2} \ell_{f,1} \alpha^2 \ell_f^2
\\
=& (1-\alpha \mu) (F_S(x_t) \lambda -\inf_x F_S(x)\lambda ) 
+ \frac{1}{2}\alpha s_t^2
+ \frac{1}{2} \alpha^2 \ell_{f,1} \ell_f^2 
\numberthis
\end{align*}
where we let $s_t = \| \nabla F_S(x_{t}) (\lambda -\lambda_{t+1}) \|^2$.
Apply the above inequality recursively, we get
\begin{align*}
  &\E_A [F_S (x_T) \lambda - \inf_x F_S(x)\lambda] \\
  \leq & (1 - \alpha \mu)^{T-1}
  \E_A [F_S (x_1) \lambda - \inf_x F_S(x)\lambda]
  + \frac{1}{2} \alpha^2 \ell_{f,1} \ell_f^2 \sum_{t=1}^{T} (1 - \alpha \mu)^{t-1}
  + \frac{1}{2} \alpha \sum_{t=1}^{T} (1 - \alpha \mu)^{T-t} s_t \\
  \leq &
  (1 - \alpha \mu)^{T-1}
  \E_A [F_S (x_1) \lambda - \inf_x F_S(x)\lambda]
  + \frac{1}{2} \alpha \mu^{-1} \ell_{f,1} \ell_f^2 
  + \frac{1}{2} \alpha \sum_{t=1}^{T} (1 - \alpha \mu)^{T-t} s_t
  \numberthis
\end{align*}
where let $\lambda = \lambda_1$, then $s_t \leq (\gamma t \ell_F)^2$, then
\begin{align*}
  & \E_A [F_S (x_T) \lambda_1 - \inf_x F_S(x)\lambda_1]
  \leq 
  (1 - \alpha \mu)^{T-1} c_F
  + \frac{1}{2} \alpha \mu^{-1} \ell_{f,1} \ell_f^2
  + \frac{1}{2} \alpha \sum_{t=1}^{T} (1 - \alpha \mu)^{T-t} 
  (\gamma t \ell_F )^2 \\
  \leq &
  (1 - \alpha \mu)^{T-1} c_F
  + \frac{1}{2} \alpha \mu^{-1} \ell_{f,1} \ell_f^2
  + \frac{1}{2} \alpha (\gamma T \ell_F )^2 
  \sum_{t=1}^{T} (1 - \alpha \mu)^{T-t} \\
  \leq &
  (1 - \alpha \mu)^{T-1} c_F
  + \frac{1}{2} \alpha \mu^{-1} \ell_{f,1} \ell_f^2
  + \frac{1}{2} \mu^{-1} M(\gamma T \ell_f )^2 .
  \numberthis
\end{align*}
And by the smoothness of the functions $F_S(x)\lambda_1$, it holds that
\begin{align}
  \E_A \Big[\min_{\lambdain} \|\nabla F_S(x_T)\lambda\|^2\Big]
  \leq & \E_A [\|\nabla F_S(x_T)\lambda_1\|^2] 
  \leq  2\ell_{f,1} \E_A [F_S (x_T) \lambda_1 - \inf_x F_S(x)\lambda_1] \nonumber \\
  =& \mathcal{O}\Big( (1 - \alpha \mu)^{T} 
  + \alpha + M (\gamma T )^2 \Big).
\end{align}
The proof is complete.
\end{proof}

\subsubsection{Proof of Theorem~\ref{thm:opt_err_modo_nonconvex_bounded_grad}} 
\label{ssub_app:proof_opt_MoDo}

\begin{proof}[Theorem~\ref{thm:opt_err_modo_nonconvex_bounded_grad}]
\label{proof:opt_error_sc_smooth}
Lemma~\ref{lemma:opt_err_PS_min_norm} states that, under Assumption~\ref{assmp:lip_cont_grad_f}, we have
\begin{align*}
 \frac{1}{T}\sum_{t=1}^T 
\E_A [\| \nabla F_S(x_t) \lambda_t^*(x_t) \|^2 ] 
\leq &
 \frac{1}{\alpha T} \E_A [F_S (x_{1}) - F_S (x_{T+1}) ] \lambda_1 + \rho + \frac{1}{2} \gamma S_{1,T}
+ \gamma S_{3,T}  + \frac{1}{2} \alpha \ell_{f,1} S_{4,T}  .
\end{align*}  
Then we proceed to bound
$S_{1,T}, S_{3,T}, S_{4,T}$.
Under either Assumptions~\ref{assmp:lip_cont_grad_f}, \ref{assmp:lip_cont_f}, or Assumptions~\ref{assmp:lip_cont_grad_f}, \ref{assmp:sconvex} with $\ell_f$, $\ell_F$ defined in Lemma~\ref{lemma:x_t_bounded_sc_smooth}, we have that 
for all $z \in S$ and $\lambdain$, $\|\nabla F_z(x_t) \lambda \| \leq  \ell_{f}$, 
and $\|\nabla F_z(x_t)\| \leq \ell_F $.
Then $S_{1,T}, S_{3,T}, S_{4,T}$ can be bounded  below 
\begin{subequations} 
\begin{align}
S_{1,T} =& \frac{1}{T} \sum_{t=1}^T
\E_A [\|(\nabla F_{z_{t,1}} (x_t)^\top \nabla F_{z_{t,2}} (x_t) +\rho \mathrm{I})\lambda_t\|^2] 
 \leq (M^{\frac{1}{2}} \ell_f^2 + \rho)^2 \\
 S_{3,T} =& \frac{1}{T}\sum_{t=1}^T\E_{A}[\|(\nabla F_{z_{t,1}} (x_t)^\top \nabla F_{z_{t,2}} (x_t) +\rho \mathrm{I}) \lambda_t\| \|\nabla F_S(x_t)^\top \nabla F_S(x_t)\lambda_1 \|]
 \leq (M^{\frac{1}{2}} \ell_f^2) (M^{\frac{1}{2}} \ell_f^2 + \rho)  \\
 S_{4,T} =& \frac{1}{T}\sum_{t=1}^T \E_{A}[\| \nabla F_{Z_{t+1}}(x_t)\lambda_{t+1}\|^2] \leq \ell_f^2 
\end{align}
\end{subequations}
which proves that
\begin{align*}
\frac{1}{T}\sum_{t=1}^T 
\E_A [\| \nabla F_S(x_t) \lambda_t^*(x_t) \|^2 ] 
\leq &
 \frac{1}{\alpha T} c_F  + \rho  + \frac{1}{2} \gamma (M^{\frac{1}{2}} \ell_f^2 + \rho)(3M^{\frac{1}{2}} \ell_f^2 + \rho)  + \frac{1}{2} \alpha \ell_{f,1} \ell_f^2. \numberthis
\end{align*}
Then by $\frac{1}{T}\sum_{t=1}^T 
\E_A [\| \nabla F_S(x_t) \lambda_t^*(x_t) \|] \leq \Big( \frac{1}{T}\sum_{t=1}^T  \E_A [\| \nabla F_S(x_t) \lambda_t^*(x_t) \|^2]  \Big)^{\frac{1}{2}}$ from the Jensen's inequality and the convexity of the square function, as well as the subadditivity of the square root function, we have that under either Assumptions~\ref{assmp:lip_cont_grad_f} and~\ref{assmp:lip_cont_f} or Assumptions~\ref{assmp:lip_cont_grad_f} and~\ref{assmp:sconvex}, it holds that
\begin{align}
 & \E_A \Big[\min_{t\in [T]}~R_{\rm opt}(x_t) \Big] \leq
\min_{t\in [T]}~\E_A [\| \nabla F_S(x_t) \lambda_t^*(x_t) \|] \nonumber \\
\leq & \frac{1}{T}\sum_{t=1}^T 
\E_A [\| \nabla F_S(x_t) \lambda_t^*(x_t) \|] = 
\mathcal{O}\big( \alpha^{-\frac{1}{2}} T^{-\frac{1}{2}}  + {\gamma^{\frac{1}{2}} M^{\frac{1}{2}} }  + { \alpha^{\frac{1}{2}} } + {\rho^{\frac{1}{2}}} 
\big).
\label{eq:MoDo_opt_NC_result}
\end{align}
This proves the first part of Theorem~\ref{thm:opt_err_modo_nonconvex_bounded_grad}.
And by Lemma~\ref{lemma:PS_opt_SC_MoDo_gammaT}, in the SC case, under Assumptions~\ref{assmp:lip_cont_grad_f} and~\ref{assmp:sconvex}, it additionally holds that
\begin{align}
& \E_A \Big[\min_{t\in [T]}~R_{\rm opt}(x_t) \Big] \leq
\min_{t\in [T]}~\E_A \big[\| \nabla F_S(x_t) \lambda_t^*(x_t) \| \big] \nonumber \\
\leq &
\E_A \Big[\min_{\lambdain} \|\nabla F_S(x_T)\lambda\|\Big]
= \mathcal{O}\Big( (1 - \alpha \mu)^{\frac{T}{2}}
  + \alpha^{\frac{1}{2}} + M^{\frac{1}{2}}\gamma T \Big) .
  \label{eq:MoDo_opt_SC_lastiter_result}
\end{align}
Combining~\eqref{eq:MoDo_opt_NC_result} and~\eqref{eq:MoDo_opt_SC_lastiter_result} proves the second  part of Theorem~\ref{thm:opt_err_modo_nonconvex_bounded_grad}.
\end{proof}

\subsection{Proof of Theorem~\ref{thm:opt_err_smg_moco} -- Optimization of SMG and MoCo} 
\label{sub_app:proof_PS_opt_SMG_MoCo}

\begin{proof}[Theorem~\ref{thm:opt_err_smg_moco}]
We use the general notation $Q_t \in \mathbb{R}^{d\times M}$ to represent the gradient estimate for SMG or MoCo at iteration $t$.
Recall that $Q_t = \nabla F_{Z_t}(x_t)$ for SMG update, and $Q_t = Y_t$ for MoCo update. 
We first derive the results with the general $Q_t$ which holds for both SMG and MoCo. Then we derive the bounds for SMG and MoCo separately by substituting $Q_t$ with their actual gradient estimate, i.e., $\nabla F_{Z_t}(x_t)$ or $Y_t$.

By the $\ell_{f, 1}$-smoothness of $F_S(x)\lambda$ for all $\lambda \in \Delta^M$, we have 
\begin{align}
F_S (x_{t+1}) \lambda - F_S (x_{t}) \lambda  
\leq &
\langle \nabla F_S(x_{t}) \lambda, x_{t+1} - x_t \rangle
+ \frac{\ell_{f,1}}{2} \|x_{t+1} - x_t\|^2 
\label{eq:proof_ps_opt_smg_smooth}
\end{align}
where $x_{t+1} - x_t = \alpha_t d_{Q_t}$, with $d_{Q_t} \coloneqq Q_t\lambda^*_{Q_t}, ~\mathrm{s.t.}~\lambda^*_{Q_t}\in \arg\min_{\lambdain}\|Q_t\lambda\|^2$.
Then,
\begin{align*}
F_S (x_{t+1}) \lambda - F_S (x_{t}) \lambda  
\leq &
-\alpha_t \langle \nabla F_S(x_{t}) \lambda, Q_t \lambda^*_{Q_t} \rangle
+ \frac{\ell_{f,1}}{2} \alpha_t^2 \| Q_t \lambda^*_{Q_t} \|^2.
\numberthis
\label{eq:descent_YlamYstar}
\end{align*}
The inner product term can be bounded as
\begin{align*}
 -\langle\nabla F_S (x_t)\lambda, Q_t \lambda_{Q_t}^* \rangle 
=& \langle\nabla F_S (x_t) \lambda, \nabla F_S (x_t) \lambda_t^*(x_t) - Q_t \lambda^*_{Q_t}\rangle - \langle\nabla F_S (x_t) \lambda , \nabla F_S (x_t) \lambda_t^*(x_t)\rangle \\
\stackrel{(a)}{\leq} & 
\langle\nabla F_S (x_t) \lambda, \nabla F_S (x_t) \lambda_t^*(x_t) - Q_t \lambda^*_{Q_t}\rangle - \|\nabla F_S (x_t) \lambda_t^*(x_t)\|^2 \\
\leq &
\ell_f\|\nabla F_S (x_t) \lambda_t^*(x_t) - Q_t \lambda^*_{Q_t}\|
- \|\nabla F_S (x_t) \lambda_t^*(x_t)\|^2 \\
\stackrel{(b)}{\leq} & 
2 \ell_f^{\frac{3}{2}}\|Q_t-\nabla F_S (x_t)\|^{\frac{1}{2}}
-\|\nabla F_S (x_t) \lambda_t^*(x_t)\|^2
\numberthis
\label{eq:bound_cross_descent_YlamYstar}
\end{align*}
where $(a)$ follows from Lemma~\ref{lemma:descent direction}, \eqref{eq:convex_inner_prod_property}, $(b)$ follows from Lemma~\ref{lemma:bound_diff_update_holder}.
Plugging \eqref{eq:bound_cross_descent_YlamYstar} into \eqref{eq:descent_YlamYstar},
taking expectations on both sides and rearranging yield
\begin{align*}
\alpha_t \E_A[\|\nabla F_S(x_t) \lambda_t^*(x_t) \|^2] 
\leq & 
\E_A[F_S(x_t)-F_S(x_{t+1})]\lambda
+2 \ell_f^{\frac{3}{2}} 
\alpha_t \E_A [\|Q_t-\nabla F_S (x_t)\|^{\frac{1}{2}} ]
+\frac{\ell_{f, 1}}{2} \ell_f^2 \alpha_t^2.
\end{align*}
For all $t \in [T]$, plugging in $\alpha_t=\alpha$, and taking the telescope sum on both sides of the last inequality yield
\begin{align*}
&\frac{1}{T} \sum_{t=1}^T \E_A
[\|\nabla F_S (x_t) \lambda_t^*(x_t)\|^2]  \\
\leq &
 \frac{1}{\alpha T} \E_A[F_S(x_t)-F_S(x_{t+1})]\lambda 
+2  \ell_f^{\frac{3}{2}} \frac{1}{T} \sum_{t=1}^T 
\E_A[\|Q_t-\nabla F_S(x_t)\|^{\frac{1}{2}}]
+\frac{\ell_{f, 1}}{2} \ell_f^2 \alpha \\
\leq &
 \frac{1}{\alpha T} \E_A[F_S(x_t)-F_S(x_{t+1})]\lambda  
+2  \ell_f^{\frac{3}{2}}  
\Big(\frac{1}{T} \sum_{t=1}^T 
\E_A[\|Q_t-\nabla F_S (x_t)\|^2]\Big)^{\frac{1}{4}}
+\frac{\ell_{f, 1}}{2} \ell_f^2 \alpha.
\numberthis
\label{eq:genera_opt_SMG_MoCo}
\end{align*}
For SMG, by increasing the batch size during optimization with $|Z_t| = {\cal O}(t)$, it holds that
\begin{align}
  \frac{1}{T} \sum_{t=1}^T 
\E_A[\|\nabla F_{Z_t}(x_t)-\nabla F_S (x_t)\|^2 ]
= \mathcal{O}\Bigg( \frac{1}{T} \sum_{t=1}^T t^{-1} \Bigg)
= \mathcal{O}(T^{-1}\ln T).
\label{eq:SMG_diff_grad}
\end{align}
Therefore, for SMG, plugging~\eqref{eq:SMG_diff_grad} back into~\eqref{eq:genera_opt_SMG_MoCo}, its PS optimization error is
\begin{align}
\E_A \Big[\min_{t\in [T], \lambdain} \|\nabla F_S (x_t) \lambda\|^2 \Big] \leq
  \frac{1}{T} \sum_{t=1}^T \E_A
[\|\nabla F_S (x_t) \lambda_t^*(x_t)\|^2]
= {\cal O}\Big( \alpha^{-1} T^{-1} + \alpha + (T^{-1}\ln T)^{\frac{1}{4}}\Big)
\end{align}
where by applying Jensen's inequality, subadditivity of the square root function, and choosing $\alpha = \Theta(T^{-\frac{1}{2}})$, it holds that
\begin{align*}
  \E_A\Big[\min_{t\in [T]} R_{\rm opt}(x_t) \Big]
  = \tilde{\cal O}(T^{-\frac{1}{8}}).
\end{align*}
For MoCo, $Q_t = Y_t = (1 - \beta_{t-1})Y_{t-1} + \beta_{t-1} \nabla F_{z_{t-1}}(x_{t-1})$.
Let $\beta_t=\beta >0$ be a constant given $T$, then by (46) in~\citep{fernando2022mitigating}, we have
\begin{align}
  \frac{1}{T} \sum_{t=1}^T 
\E_A[\|Y_t-\nabla F_S (x_t)\|^2] = \mathcal{O}( \beta^{-1} T^{-1} + \beta + \alpha^2 \beta^{-2})
\end{align}
where by setting 
$\alpha=\Theta(T^{-\frac{3}{4}})$, and $\beta=\Theta(T^{-\frac{1}{2}})$, and plugging back into~\eqref{eq:genera_opt_SMG_MoCo},
we obtain
\begin{align}
\E_A\Big[\min_{t\in [T]} R_{\rm opt}(x_t) \Big] \leq
  \Bigg(\frac{1}{T} \sum_{t=1}^T \mathbb{E}_A
  [\|\nabla F_S (x_t) \lambda_t^*(x_t)\|^2 ] \Bigg)^{\frac{1}{2}}
= \mathcal{O}( T^{-\frac{1}{16}}).
\end{align} 
The proof is complete.
\end{proof}

\section{Additional Experiments and Implementation Details}

\paragraph{Compute.}
Experiments are done on a machine with
 GPU NVIDIA RTX A5000. We use MATLAB R2021a for the synthetic experiments in strongly convex case, and  Python 3.8, CUDA 11.7, Pytorch 1.8.0 for other experiments.
Unless otherwise specified, all experiments are repeated with 5 random seeds. And their average performance with standard deviations are reported.

\subsection{Synthetic experiments}
\label{sub_app:synthetic_experiment}

\subsubsection{Experiments on non-convex objectives}

The toy example used in Figure \ref{fig:toy-comp} is modified from~\citep{liu2021conflict} to consider stochastic data.
Denote the model parameter as
$x=[x_1, x_2]^\top \in \mathbb{R}^2$, stochastic data as  $z = [z_1, z_2]^\top \in \mathbb{R}^2$ sampled from the standard multi-variate Gaussian distribution.
The individual empirical objectives are defined as:
\begin{align*}
f_{z,1}(x) & =c_1(x) h_1(x)+c_2(x) g_{z,1}(x) \text { and } 
f_{z,2}(x)=c_1(x) h_2(x)+c_2(x) g_{z,2}(x), \text { where } \\
h_1(x) & =\ln  (\max  ( |0.5 (-x_1-7 )-\tanh  (-x_2 ) |, 0.000005 ) )+6, \\
h_2(x) & =\ln  (\max  ( |0.5 (-x_1+3 )-\tanh  (-x_2 )+2 |,  0.000005 ) )+6, \\
g_{z,1}(x) & = ( (-x_1+ 3.5 )^2+0.1 * (-x_2-1 )^2 ) / 10-20 - 2* z_1 x_1 - 5.5* z_2 x_2, \\
g_{z,2}(x) & = ( (-x_1- 3.5 )^2+0.1 * (-x_2-1 )^2 ) / 10-20 + 2* z_1 x_1 - 5.5* z_2 x_2, \\
c_1(x) & =\max  (\tanh  (0.5 * x_2 ), 0 ) \text { and } c_2(x)=\max  (\tanh  (-0.5 * x_2 ), 0 ) .
\numberthis
\end{align*}
Since $z$ is zero-mean, the individual population objectives are correspondingly:
\begin{align*}
f_1(x) & =c_1(x) h_1(x)+c_2(x) g_1(x) \text { and } 
f_2(x)=c_1(x) h_2(x)+c_2(x) g_2(x), \text { where } \\   
g_{1}(x) & = ( (-x_1+ 3.5 )^2+0.1 * (-x_2-1 )^2 ) / 10-20 , \\
g_{2}(x) & = ( (-x_1- 3.5 )^2+0.1 * (-x_2-1 )^2 ) / 10-20 .
\numberthis
\end{align*}

The training dataset size is $n = |S| = 20$.
For all methods, i.e., MGDA, static weighting, and MoDo, the number of iterations is $T=50000$.
The initialization of $\lambda$ is $\lambda_0 = [0.5, 0.5]^\top$.
The hyperparameters for this experiment are summarized in Table~\ref{tab:hp-toy_nc}.

\begin{table}[ht]
\vspace{-0.2cm}
\caption{Summary of hyper-parameter choices for nonconvex synthetic example.}
\label{tab:hp-toy_nc}
\fontsize{8}{9}\selectfont
\centering
\setlength{\tabcolsep}{1.0em} %
{\renewcommand{\arraystretch}{1.3}%
\begin{tabular}{|c| c | c | c|}
\hline
 & Static & MGDA & MoDo \\
 \hline
 optimizer of $x_t$ & Adam & Adam & Adam \\
 $x_t$ step size ($\alpha_t$) & $5\times 10^{-3}$ & $5\times 10^{-3}$ & $5\times 10^{-3}$ \\ 
 $\lambda_t$ step size ($\gamma_t$) & - & - & $10^{-4}$\\
 batch size  & $16$ & full & $16$\\
 \hline
\end{tabular}
}
\end{table} 

\subsubsection{Experiments on strongly convex objectives}

Below we provide the details of experiments that generate Figure~\ref{fig:alpha_gamma_T_SC}.
We use the following synthetic example  for the experiments in the strongly convex case. The $m$-th objective function with stochastic data sample $z$ is specified as
\begin{align}
 f_{z,m}(x) = \frac{1}{2} b_{1,m} x^\top A x - b_{2,m} z^\top x  
\end{align}
where $b_{1,m}> 0 $ for all $m\in [M]$, and $b_{2,m}$ is another scalar. 
We set $M=3$, $b_1 = [b_{1,1}; b_{1,2}; b_{1,3}] = [1;2;1]$, and 
$b_2 = [b_{2,1}; b_{2,2}; b_{2,3}] = [1;3;2]$.
The default  parameters are $T = 100$, $\alpha = 0.01$, $\gamma=0.001$.
In other words, in Figure~\ref{fig:T_opt_gen}, we fix $ \alpha = 0.01, \gamma=0.001$, and vary $T$;
in Figure~\ref{fig:alpha_opt_gen}, we fix $ T=100, \gamma=0.001$, and vary $\alpha$;
and in Figure~\ref{fig:gamma_opt_gen}, we fix $ T=100,  \alpha=0.01$, and vary $\gamma$.

\subsubsection{MNIST dataset experiments}

Below are the details to generate Figure~\ref{fig:T-alpha-gamma-ablation}.
The model architecture is a  two-layer multi-layer perceptron (MLP), with each hidden layer size 512, and no hidden layer activation. Input size is 784, the size of an MNIST image in the vector form, and the output size  is 10, the number of digit classes. The training, validation, and testing data sizes are 50k, 10k, and 10k, respectively.  
Hyper-parameters such as step sizes are chosen based on each algorithm's validation accuracy performance, as given in Table~\ref{tab:hp-mnist}. 

\begin{table}[ht]
\vspace{-0.2cm}
\caption{Summary of hyper-parameter choices for MNIST image classification}
\label{tab:hp-mnist}
\fontsize{8}{9}\selectfont
\centering
\setlength{\tabcolsep}{1.0em} 
{\renewcommand{\arraystretch}{1.3}
\begin{tabular}{|c| c | c | c|}
\hline
 & Static & MGDA & MoDo \\
 \hline
 optimizer of $x_t$ & SGD & SGD & SGD \\
 $x_t$ step size ($\alpha_t$) & 0.1 & 5.0 & 1.0 \\ 
 $\lambda_t$ step size ($\gamma_t$) & - & - & 1.0\\
 batch size  & 64 & 64 & 64 \\
 \hline
\end{tabular}
\vspace{-0.2cm}}
\end{table}


\subsection{Multi-task supervised learning experiments}
\label{sub_app:image_classification_experiment}

In this section, we present experiment details and additional results for comparing recent MOL baselines with MoDo, under real-world multi-task supervised learning problems using the opensource library LibMTL~\citep{LibMTL}.

\subsubsection{Office-31 and Office-home dataset experiments}
 Office-31 and Office-home datasets are built for multi-domain image classification tasks.  Office-31 and Office-home consist of 31 and 65 image classes, respectively. The image domains for Office-31 are: ``Amazon'', with object images from Amazon, ``DSLR'', with high-resolution images of objects, and ``Webcam'', with low-resolution images from objects. Similarly, the image domains for Office-31 are: ``Art'', ``Clipart'', ``Product'', and ``Real-world''. For both datasets, we tune the step sizes and weight decay parameters for all algorithms. We use batch size 64 to update Static and MGDA, and use 2 independent batches with batch size 32 to update MoDo. A summary of hyper-parameters used for Office-31 and Office-home for each algorithm are given in Table~\ref{tab:hp-office-31} and Table~\ref{tab:hp-office-home}, respectively. Other experiment setups are the same for all algorithms, and the default LibMTL configuration is used.
 
 The results on Office-31 and Office-home are given in Tables~\ref{tab:office-31-results} and~\ref{tab:office-home-results}, respectively (average over 5 seeds, with standard deviation reported). 
 For Office-31, $S_{\mathcal{B}, m}$ values are $87.50\%$ for ``Amazon'', $98.88\%$ for ``DSLR'', and $97.32\%$ for ``Webcam''. For Office-home, $S_{\mathcal{B}, m}$ values are $66.98\%$ for ``Art'', $82.02\%$ for ``Clipart'', $91.53\%$ for ``Product'', and $80.97\%$ for ``Real-world''. 
\begin{table}[ht]
\vspace{-0.2cm}
\caption{Summary of hyper-parameter choices for Office-31 task}
\label{tab:hp-office-31}
\fontsize{8}{9}\selectfont
\centering
\setlength{\tabcolsep}{1.0em} 
{\renewcommand{\arraystretch}{1.3}
\begin{tabular}{|c| c | c | c|}
\hline
 & Static & MGDA & MoDo \\
 \hline
 optimizer of $x_t$ & Adam & Adam & Adam \\
 $x_t$ step size ($\alpha_t$) & $10^{-4}$ & $10^{-4}$ & $10^{-4}$ \\ 
 $\lambda_t$ step size ($\gamma_t$) & - & - & $10^{-3}$\\
 weight decay & $10^{-3}$ & $10^{-7}$ & $10^{-5}$ \\
 batch size  & 64 & 64 & 64\\
 \hline
\end{tabular}
}
\end{table} 

\begin{table}[ht]
\caption{Summary of hyper-parameter choices for Office-home task}
\label{tab:hp-office-home}
\fontsize{8}{9}\selectfont
\centering
\setlength{\tabcolsep}{1.0em} 
{\renewcommand{\arraystretch}{1.3}
\begin{tabular}{|c| c | c | c|}
\hline
 & Static & MGDA & MoDo \\
 \hline
 optimizer of $x_t$ & Adam & Adam & Adam \\
 $x_t$ step size ($\alpha_t$) & $10^{-4}$ & $10^{-4}$ & $10^{-4}$ \\ 
 $\lambda_t$ step size ($\gamma_t$) & - & - & $10^{-3}$\\
 weight decay & $10^{-3}$ & $10^{-6}$ & $10^{-5}$ \\
 batch size  & 64 & 64 & 64\\
 \hline
\end{tabular}
\vspace{-0.4cm}}
\end{table}

\subsubsection{NYU-v2 dataset experiments}
NYU-v2 dataset consists of image segmentation, depth estimation, and surface normal estimation tasks. Unlike Office-31 and Office-home datasets, this is a single-input multi-task learning problem. The dataset consists of images from indoor video sequences. We tune step size of $x$ and weight decay parameters for static weighting and MGDA algorithms, and tune step sizes of $x$ and $\lambda$ for MoDo algorithm. We use batch sizes of 4 to update static weighting and MGDA, and use 2 independent batches with batch size 2 to update MoDo. We used 50 epochs for all methods. 
A summary of hyper-parameters used for each algorithm is in Table~\ref{tab:hp-nyu}. All other experiment setup is shared for all algorithms, and the LibMTL configuration is used.

\begin{table}[ht]
\vspace{-0.2cm}
\caption{Summary of hyper-parameter choices for NYU-v2 task}
\label{tab:hp-nyu}
\fontsize{8}{9}\selectfont
\centering
\setlength{\tabcolsep}{1.0em} 
{\renewcommand{\arraystretch}{1.3}
\begin{tabular}{|c| c | c | c|}
\hline
 & Static & MGDA & MoDo \\
 \hline
 optimizer of $x_t$ & Adam & Adam & Adam \\
 $x_t$ step size ($\alpha_t$) & $10^{-4}$ & $10^{-4}$ & $10^{-4}$ \\ 
 $\lambda_t$ step size ($\gamma_t$) & - & - & $10^{-3}$\\
 weight decay & $10^{-4}$ & $10^{-6}$ & $10^{-5}$ \\
 batch size  & 4 & 4 & 4\\
 \hline
\end{tabular}
\vspace{-0.2cm}}
\end{table}

The results of NYU-v2 experiments are in Tables \ref{tab:nyu-v2-results} (average over 3 seeds, the error indicates standard deviation). Here we use the average per-task performance drop of metrics $S_{\mathcal{A}, m}$ for method $\mathcal{A}$ with respect to corresponding baseline measures $S_{\mathcal{B}, m}$ as a measure of the overall performance of a given method. We use the best results for each task obtained by dedicated independent task learners of each task as $S_{\mathcal{B}, m}$. The independent task learners are tuned for the learning rate and weight decay parameter. For segmentation task, $S_{\mathcal{B}, m}$ values are $53.94\%$ for ``mIoU'', and $75.67\%$ for ``Pix Acc''. For depth estimation task, $S_{\mathcal{B}, m}$ values are $0.3949$ for ``Abs Err'', and $0.1634$ for ``Rel Err''. For surface normal estimation task, $S_{\mathcal{B}, m}$ values are $22.12$ for ``Angle Distance - Mean'', $15.49$ for ``Angle Distance - Median'', $38.35\%$ for ``Within $11.25^\circ$'', $64.30\%$ for ``Within $22.5^\circ$'', and $74.70\%$ for ``Within $30^\circ$''.

\subsubsection{Additional experiments for comparison with other MOL baselines}
\label{sub_app:additional_experiment}

In this section, we provide a comparison between MoDo and other popular MOL baselines. For this purpose, we use the same benchmark datasets as the previous section and use the experiment setup provided in~\citep{lin2022reasonable_RGW} to run experiments with MoDo. Hence, we use experiment results provided by \citep{lin2022reasonable_RGW} for other baselines for comparison. Additionally, we implement MoCo \citep{fernando2022mitigating}, which is not included in \citep{lin2022reasonable_RGW}. In addition to the overall performance measure $\Delta \mathcal{A}_{\rm id} \%$ used in previous experiment results, in this section we also report $\Delta \mathcal{A}_{\rm st} \%$, which is the average performance degradation of a given method compared to static weighting (similar to the definition in~\citep{lin2022reasonable_RGW}, but lower the better).

The results on Office-31, Office-home, and NYU-v2  are given in Tables~\ref{tab:benchmark-office-31-results-1},~\ref{tab:benchmark-office-home-results-1}, and~\ref{tab:bench-nyu-1}, respectively, where MoDo outperforms all the baselines for most tasks, and has a better overall performance in $\Delta \mathcal{A}_{\rm st} \%$ and $\Delta \mathcal{A}_{\rm id} \%$.
The hyper-parameters of MoDo for the above experiments are given in Table \ref{tab:hp-modo}.

\begin{table}[ht]
\caption{Comparison with other methods on Office-31 dataset. }
\label{tab:benchmark-office-31-results-1}
\fontsize{9}{9}\selectfont
\centering
{\renewcommand{\arraystretch}{1.3}
\begin{tabular}{c c c c c c c}
\toprule
 Method & Amazon & DSLR & Webcam
 & $\Delta \mathcal{A}_{\text{st}}\% \downarrow$ & $\Delta \mathcal{A}_{\text{id}}\% \downarrow$\\
 \midrule
  Static (EW) & 81.02  & 96.72  & 96.11  & 0.00 & 2.96 \\ 
  MGDA-UB~\citep{sener2018multi} & 81.02  & 95.90  & \textbf{97.77}  & 0.40 & 3.32 \\
  GradNorm~\citep{chen2018gradnorm} & 83.93  & 97.54  & 94.44  & -0.19 & 2.80 \\
  PCGrad~\citep{yu2020gradient} & 82.22  & 96.72  & 95.55  & 0.40 & 3.35 \\
  CAGrad~\citep{liu2021conflict} & 82.22  & 96.72  & 96.67  & 0.01  & 2.96 \\
  RGW~\citep{lin2022reasonable_RGW} & 84.27  & 96.72  & 96.67  & -0.81 & 2.18 \\
  MoCo~\citep{fernando2022mitigating} & 85.30  & 97.54  & 97.22  & -1.70 & 1.32 \\
  \rowcolor{RoyalBlue!10}
  MoDo (ours) & \textbf{85.47}  & \textbf{98.36}  & 96.67  & \textbf{-1.86} & \textbf{1.17} \\
 \bottomrule
\end{tabular}}
\end{table}

\begin{table}[ht]
\caption{Comparison with other methods on Office-home dataset. }
\label{tab:benchmark-office-home-results-1}
\fontsize{9}{9}\selectfont
\centering
{\renewcommand{\arraystretch}{1.3}
\begin{tabular}{c c c c c c c c}
\toprule
 Method & Art & Clipart & Product & Real-world
 & $\Delta \mathcal{A}_{\text{st}}\% \downarrow$ & $\Delta \mathcal{A}_{\text{id}}\% \downarrow$\\ 
 \midrule
  Static (EW) & 62.99  & 76.48  & 88.45  & 77.72 & 0.00 & 5.02 \\
  MGDA-UB~\citep{lin2022reasonable_RGW} & 64.32  & 75.29  & 89.72  & 79.35 & -1.02 & 4.04 \\
  GradNorm~\citep{chen2018gradnorm} & 65.46  & 75.29  & 88.66  & 78.91 & -1.03 & 4.04 \\
  PCGrad~\citep{yu2020gradient} & 63.94  & 76.05  & 88.87  & 78.27 & -0.53 & 4.51 \\
  CAGrad~\citep{liu2021conflict} & 63.75  & 75.94  & 89.08  & 78.27  & -0.48  & 4.56 \\
  RGW~\citep{lin2022reasonable_RGW} & 65.08  & 78.65  & 88.66  & 79.89 & -2.30 & 2.85 \\
  MoCo~\citep{fernando2022mitigating} & 64.14  & \textbf{79.85}  & 89.62  & 79.57 & -2.48 & 2.68 \\
  \rowcolor{RoyalBlue!10}
  MoDo (ours) & \textbf{66.22}  & 78.22  & \textbf{89.83}  & \textbf{80.32} & \textbf{-3.08} & \textbf{2.11} \\
 \bottomrule
\end{tabular}}
\end{table}

\begin{table}[ht]
\tiny
\caption{Comparison with other methods on NYU-v2 dataset.}
\label{tab:bench-nyu-1}
\centering
\begin{tabular}{c c c c c c c c c c c c}
\toprule
 \multirow{3}{*}{Method} &\multicolumn{2}{c}{Segmentation} &\multicolumn{2}{c}{Depth} &\multicolumn{5}{c}{Surface Normal} &\multirow{3}{*}{\makecell[c]{$\Delta \mathcal{A}_{\text{st}}\%$ \\ $\downarrow$}} &\multirow{3}{*}{\makecell[c]{$\Delta \mathcal{A}_{\text{id}}\%$ \\ $\downarrow$}} \\ \cline{2-3}\cline{4-5}\cline{6-10}
 &\multicolumn{2}{c}{(Higher Better)} &\multicolumn{2}{c}{(Lower Better)} &\multicolumn{2}{c}{\makecell{Angle Distance \\ (Lower Better)}} &\multicolumn{3}{c}{\makecell{Within $t^\circ$ \\ (Higher better)}}\\
 &mIoU &Pix Acc &Abs Err &Rel Err &Mean &Median &11.25 &22.5 &30\\ \hline
Static (EW) &53.77  &75.45  &0.3845 &0.1605  &23.57  &17.04 &35.04  &60.93  &72.07  &0.00 &1.63 \\
\makecell[c]{MGDA-UB} &50.42  &73.46  &0.3834 &0.1555  &22.78  &16.14 &36.90  &62.88  &73.61  &-0.38 &1.26 \\
\makecell[c]{GradNorm} &53.58  &75.06  &0.3931 &0.1663  &23.44  &16.98 &35.11  &61.11  &72.24  &0.99 &2.62 \\
\makecell[c]{PCGrad} &53.70  &75.41  &0.3903 &0.1607  &23.43  &16.97 &35.16  &61.19  &72.28  &0.16 &1.79 \\
\makecell[c]{CAGrad} &53.12  &75.19  &0.3871 &0.1599  &\textbf{22.53}  &\textbf{15.88} &\textbf{37.42}  &\textbf{63.50}  &\textbf{74.17}  &-1.36  &0.26 \\
RGW &53.85  &\textbf{75.87}  &0.3772 &0.1562  &23.67 &17.24 &34.62  &60.49  &71.75  &-0.62 &1.03 \\
\makecell[c]{MoCo} &\textbf{54.05}  &75.58  &0.3812 &\textbf{0.1530}  &23.39  &16.69 &35.65  &61.68  &72.60  &-1.47 &0.18 \\
\rowcolor{RoyalBlue!10}
MoDo (ours) &53.37  &75.25  &\textbf{0.3739} &0.1531  &23.22  &16.65 &35.62  &61.84  &72.76  &\textbf{-1.59} &\textbf{0.07} \\
 \bottomrule
\end{tabular}
\vspace{0.5cm}
\end{table}

\begin{table}[ht]
\caption{Summary of hyper-parameter choices for MoDo}
\label{tab:hp-modo}
\fontsize{8}{9}\selectfont
\centering
\setlength{\tabcolsep}{1.0em} 
{\renewcommand{\arraystretch}{1.3}
\begin{tabular}{|c| c | c | c|}
\hline
 & Office-31 & Office-home & NYU-v2 \\
 \hline
 optimizer of $x_t$ & Adam & Adam & Adam \\
 $x_t$ step size ($\alpha_t$) & $1\times 10^{-4}$ & $5 \times 10^{-4}$ & $2.5 \times 10^{-4}$ \\ 
 $\lambda_t$ step size ($\gamma_t$) & $10^{-6}$ & $10^{-3}$ & $10^{-3}$\\
 weight decay & $10^{-5}$ & $10^{-5}$ & $10^{-5}$ \\
 batch size  & 128 & 128 & 8\\
 \hline
\end{tabular}
\vspace{-0.4cm}}
\end{table}

\clearpage
\bibliography{myabrv,MOO_stability,MOO_opt,MTL}

\end{document}